\documentclass[10pt]{article} 

\usepackage[utf8]{inputenc}
\usepackage{amsmath}
\usepackage{amsfonts}
\usepackage{amssymb}
\usepackage{mathtools}
\usepackage{listings}
\usepackage{mathrsfs}
\usepackage{times}
\usepackage{float}
\usepackage{color}
\usepackage{setspace}
\usepackage{color,soul}

\usepackage{amsmath,amsfonts,amsthm,eucal}
\usepackage{enumerate}
\usepackage[letterpaper,margin=3cm]{geometry}
\usepackage{float}
\usepackage[title]{appendix}
\usepackage{color}
\usepackage{tabularx}
\usepackage{siunitx}
\usepackage[tableposition=below]{caption}
\usepackage{array}

\usepackage[
pdfstartview=XYZ,
bookmarks=true,
colorlinks=true,
linkcolor=blue,
urlcolor=blue,
citecolor=blue,
pdftex,
bookmarks=true,
linktocpage=true,   
hyperindex=true
]{hyperref}

\usepackage{lipsum}
\usepackage{lineno}

\usepackage[FIGBOTCAP,TABTOPCAP,bf,tight]{subfigure}

\usepackage{natbib}
\setcitestyle{authoryear}

\usepackage[pdftex]{graphicx}
\usepackage{epstopdf}
\usepackage{booktabs} 
\usepackage{tikz,mathpazo}
\usetikzlibrary{shapes.geometric, arrows}
\usepackage{caption}

\newcommand{\tensor}[1]{\ensuremath{\boldsymbol{#1}}}

\DeclareMathOperator*{\argmin}{argmin}

\usepackage{algorithm}
\usepackage{algorithmicx}
\usepackage[noend]{algpseudocode}

\theoremstyle{remark}

\newtheorem{remark}{Remark}

\renewcommand{\vec}[1]{\ensuremath{\boldsymbol{#1}}}
\theoremstyle{definition}

\usepackage{algorithm}
\usepackage[noend]{algpseudocode}
\newcolumntype{M}[1]{>{\centering\arraybackslash}m{#1}}

\title{Sobolev training of thermodynamic-informed neural networks for smoothed elasto-plasticity models with level set hardening} 

\begin{document}


\author{Nikolaos N. Vlassis\thanks{Department of Civil Engineering and Engineering Mechanics, 
 Columbia University, 
 New York, NY 10027.     \textit{nnv2102@columbia.edu}  } 
\and
        WaiChing Sun\thanks{Department of Civil Engineering and Engineering Mechanics, 
 Columbia University, 
 New York, NY 10027.
  \textit{wsun@columbia.edu}  (corresponding author)   }
}

\maketitle

\begin{abstract}
We introduce a deep learning framework 
designed to train smoothed elastoplasticity models 
with interpretable components, such as a smoothed 
stored elastic energy function, 
a yield surface, and a plastic flow that are evolved 
based on a set of deep neural network predictions. 
By recasting the yield function as an evolving level set, 
we introduce a machine learning approach  to 
predict the solutions of the 
Hamilton-Jacobi equation that governs the hardening mechanism. 
This machine learning hardening law  
may recover classical hardening models and discover new 
mechanisms that are otherwise very difficult to anticipate 
and hand-craft. 
 This treatment enables us to use supervised machine learning to generate models that are thermodynamically consistent, interpretable, but also 
exhibit excellent learning capacity. 
Using a 3D FFT solver to create a polycrystal database, numerical experiments are conducted and the implementations of each component of the models are individually verified. Our numerical experiments reveal that this new approach provides more robust and accurate forward predictions of cyclic stress paths than these obtained from black-box deep neural network models such as a recurrent GRU neural network, a 1D convolutional neural network, and a multi-step feedforward model.  
\end{abstract}


\section{Introduction}
\label{intro}

Plastic deformation of materials is a history-dependent process manifested by irreversible and permanent changes of microstructures, such as dislocation, pore collapses, growth of defects and phase transition. Macroscopic consitutive models designed to capture the history-dependent constitutive responses 
can be categorized into multiple families. For example, hypoplasticity models often do not distinguish the reversible and irreversible strain \citep{dafalias1986bounding,  kolymbas1991outline, wang_identifying_2016}. 
Unlike the classical elastoplasticity models where the plastic flow is normal to the stress gradient of the plastic potential and the evolution of it is governed by a set of hardening rules \citep{rice1971inelastic, 
hill1998mathematical, sun_unified_2013, bryant2019micromorphically}, 
hypoplasticity models do not employ a yield function to characterize the initial yielding. Instead, the relationship between the strain rate and the stress rate
is captured by a set of evolution laws originated from 
a combination of phenomenological observations and physics constraints. 
Interestingly, the early design of neural network models such as 
 \citet{ghaboussi_knowledge-based_1991, furukawa1998implicit, pernot1999application, lefik_artificial_2009}, would often adopt this approach with a 
purely supervised learning strategy to adjust the weights of neurons to minimize the errors.
Using the strain from current and previous time steps to 
estimate the current stress, these models would essentially 
predict the stress rate without utilizing a yield function and, hence, 
can be viewed as hypoplasticity models with machine learning derived evolution laws. 
The major issue of these machine learning generated evolution laws that, in retrospect, limits the adaptations of neural network constitutive model is the lack of interpretability and the vulnerability to over-fitting. 
While there are existing regularization techniques such as dropout layers \citep{wang_multiscale_2018}, cross-validation \citep{heider2020so, vlassis2020geometric},  adversarial attack \citep{wang2020non}
and/or increasing the size of the database could be helpful, it remains difficult to assess the credibility without the interpretability of the underlying laws deduced from the neural network. 
Another approach could involve symbolic regression through reinforcement learning \citep{wang2019cooperative} or genetic algorithms \citep{versino2017data} 
that may lead to explicitly written evolution laws, however, the fitness of these equations is often at the expense of readability. 

\begin{figure}[h!]
\newcommand\siz{.35\textwidth}
\centering

\begin{tabular}{M{.35\textwidth}M{.50\textwidth}}
\includegraphics[width=.25\textwidth, angle=0]{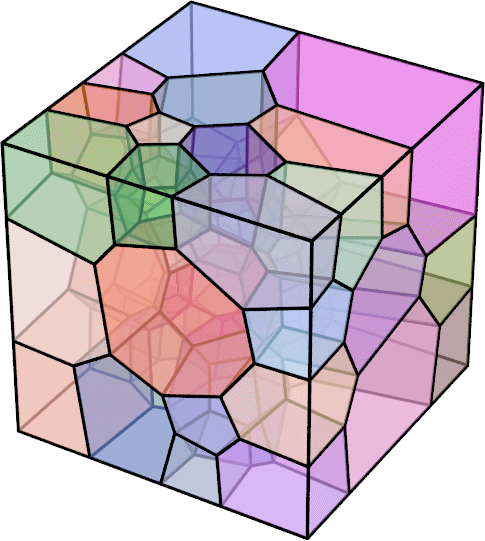} &
\includegraphics[width=.4\textwidth ,angle=0]{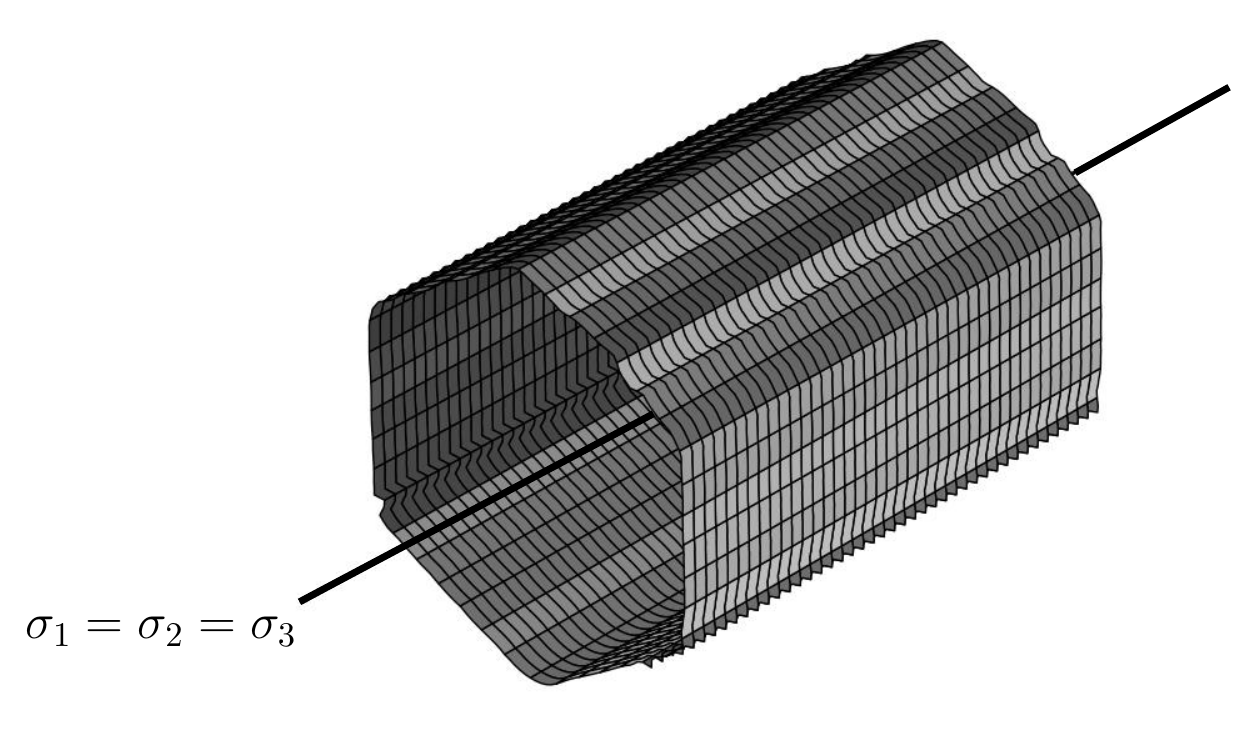} \\
 
\end{tabular}

\caption{Neural network discovery of isotropic pressure-independent yield surface for polycrystal microstructure.}
\label{fig:function_discovery}
\end{figure}

Another common approach to predict plastic deformation is the classical elasto-plasticity model where an elasticity model is coupled with a yield function that evolves with a set of internal variables that represents the history 
of the materials. 
Within the framework of the classical elasto-plasticity model -- where constitutive models are driven by the evolution of yield surface and the underlying elastic model, there has been a significant number of works dedicated to 
refining the initial shapes and forms of the yield functions in the stress space (e.g. \citet{mises1913mechanik, prager1955theory, william1974constitutive}) and the corresponding hardening laws that govern the evolution of these yield functions with the plastic strain (e.g. \citet{drucker1950some, borja1994multiaxial, taiebat2008sanisand, nielsen2010ductile, foster2005implicit, sun2014modeling}). There are several key upshots brought by the existence of the yield function. For instance, the existence of a yield function facilitates the geometric interpretation of plasticity and, therefore, enables us to connect mechanics concepts, such as the thermodynamic law, with geometric concepts, such as convexity in the principal stress space \citep{miehe_anisotropic_2002, borja_plasticity_2013, vlassis2020geometric}. 
Furthermore, the existence of a distinctive elastic region in the stress or strain space also allows to introduce a multi-step transfer learning strategy. In this case,  
the machine learning of the elastic responses can be viewed as a pre-training step for the plasticity machine learning where the predicted elastic responses can help determine the underlying split of the elastic and plastic strain upon the initial yielding and, hence, allows for a more accurate hardening law and plastic flow to be discovered.

\subsection{Why Sobolev training for plasticity}
Recently, there have been attempts to rectify the limitations of machine learning models that do not distinguish or partition the elastic and plastic strain. 
\citet{xu2020learning}, for instance, introduce a differentiable transition function to create a smooth transition between the elastic and plastic range for an incremental constitutive law generated from supervised learning. \citet{mozaffar2019deep} and later \citet{zhang2020using} inroduce the machine learning to deduce the yield function and enable linear and distortion hardening using loss functions that minimizing the $L_{2}$ norm of the yield function discrepancy. 

While these machine learning exercises are effective in identifying the yield locus, calibrating and even selecting the existing hardening mechanisms ((isotropic, kinematic, rotation....etc)) for predictions, 
more flexibility and control over capturing the evolution of the yield surface with respect to strain history is needed for more general applications 
where the dominated hardening mechanisms are not known in advance. 
Furthermore, since the constitutive laws are updated via a system of linearized constrains, the \textit{solvability} of the constitutive laws and the resultant global system of equations all depends not only on the accuracy of the predictions of the yield function, but also the gradient and Hessian.  

This research is specifically designed to fill this knowledge gap in order to make the machine learning models more robust and practical when incorporating into PDE solvers. In particular, we introduce a set of new supervised learning problems where a family of higher-order norms is used to regularize the predictions of scalar functionals that lead to the elastic and elasto-plastic responses of isotropic path-dependent materials. By adopting the Haigh-Westergaard coordinate system to simplify the parametrization, we introduce a simple training program that can generate accurate and robust stress predictions but also yield the elastic energy and elasto-plastic tangent operator that is sufficiently smooth for numerical predictions -- one of the technical barriers that prevent the adoption of neural network models since their inception in the 90s \citep{hashash2004numerical}.

\subsection{Organization of the content and notations}  
The organization of the rest of the paper is as follows. We first 
provide a detailed account of the different designs of 
Sobolev higher-order training introduced 
to generate the elastic stored energy functional, 
yield function, flow rules and hardening models 
in their corresponding parametric space. 
The setup of control experiments with other common alternative
black-box models is then described. 
We then demonstrate how to leverage this new design 
of an interpretable machine learning framework to 
analyze the thermodynamic behavior of the machine learning derived 
constitutive laws, while illustrating the geometrical interpretation
of the proposed modeling framework.
A brief highlight for the adopted return mapping algorithm implementation that leverages automatic differentiation 
is provided in Section \ref{sec:return_mapping_algorithm}, followed by the 
numerical experiments and the conclusions that outline
this work's major findings. 

As for notations and symbols in this current work, bold-faced letters
denote tensors (including vectors which are rank-one tensors); 
the symbol '$\cdot$' denotes a single contraction of adjacent indices of two tensors (e.g. $\vec{a} \cdot \vec{b} = a_{i}b_{i}$ or $\tensor{c}
\cdot \vec{d} = c_{ij}d_{jk}$ ); the symbol `:' denotes a double
contraction of adjacent indices of tensor of rank two or higher (
e.g. $\tensor{C} : \vec{\epsilon^{e}}$ = $C_{ijkl} \epsilon_{kl}^{e}$
); the symbol `$\otimes$' denotes a juxtaposition of two vectors
(e.g. $\vec{a} \otimes \vec{b} = a_{i}b_{j}$) or two symmetric second
order tensors (e.g. $(\vec{\alpha} \otimes \vec{\beta})_{ijkl} =
\alpha_{ij}\beta_{kl}$). Moreover, $(\tensor{\alpha}\oplus\tensor{\beta})_{ijkl} = \alpha_{jl} \beta_{ik}$ and $(\tensor{\alpha}\ominus\tensor{\beta})_{ijkl} = \alpha_{il} \beta_{jk}$. We also define identity tensors $(\tensor{I})_{ij} = \delta_{ij}$, $(\tensor{I}^4)_{ijkl} = \delta_{ik}\delta_{jl}$, and $(\tensor{I}^4_{\text{sym}})_{ijkl} = \frac{1}{2} (\delta_{ik}\delta_{jl} + \delta_{il}\delta_{kj})$, where $\delta_{ij}$ is the Kronecker delta. As for sign conventions, unless specified otherwise,
we consider the direction of the tensile stress and dilative pressure as positive.

\section{Framework for Sobolev training of elastoplasticity models} 
\label{sec:framework}

This section presents the framework to train the multiple 
deep neural networks 
to predict the three constitutive laws studied in this work -- the elastic stored energy 
functional, the yield function, and the plastic flow. 

We discuss the neural network architecture that will allow for smoother predictions and higher-order Sobolev optimization. 
We introduce how this higher-order training can benefit the data-driven approximation of a hyperelastic energy functional.
Continuing with plasticity, we describe how we achieve the dimensionality reduction of our training problem by adopting the $\pi$-plane interpretation of the stress space.
The yield function prediction and evolution is presented as a Hamilton-Jacobi extension problem to facilitate the data pre-processing and neural network training.
Finally, the training objectives for associative and non-associative plasticity laws that obey thermodynamic constraints are described.

To simplify the proposed constitutive laws, we assume that the deformation is infinitesimal and the plastic deformation is rate independent. 
 Meanwhile, the elastic response and the initial yield function of the materials we considered can both be approximated as isotropic. 
Extensions that relax these assumptions will be considered in future studies but is out of the scope of this current paper.

\subsection{Neural network design for Sobolev training of a smooth scalar functional with physical constraints}
Here we will provide a brief account on the design of the neural network that aims to generate a
scalar functional 
 with sufficient smoothness and continuity for a variety of mechanics problems where both the functional itself and its derivatives are both of interest (e.g. elasticity energy functional, yield function). 
The specific data preparation
and the loss function required to complete the Sobolev training will be 
discussed in the next Sections \ref{sec:higher_order_sobolev} and \ref{sec:yield}.  

\begin{figure}[h!]
\newcommand\siz{.23\textwidth}
\newcommand\sizb{.23\textwidth}
\centering
\begin{tabular}{M{.23\textwidth}M{.23\textwidth}M{.23\textwidth}M{.23\textwidth}}
\hspace{-1cm}Architecture: ddd & \hspace{-1cm}Architecture: dmdd & \hspace{-1cm}Architecture: dmdmd & \hspace{-1cm}Architecture: dmmdmd \\
\hspace{-1cm}\includegraphics[width=.23\textwidth ,angle=0]{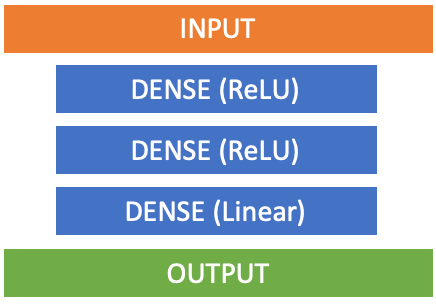} &
\hspace{-1cm}\includegraphics[width=.23\textwidth ,angle=0]{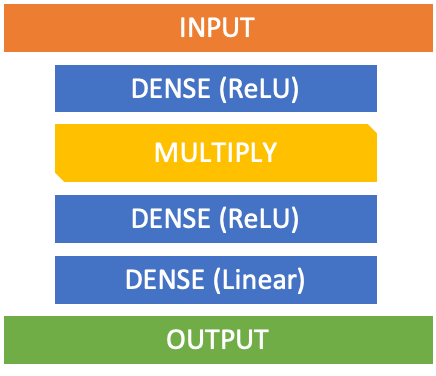} &
\hspace{-1cm}\includegraphics[width=.23\textwidth ,angle=0]{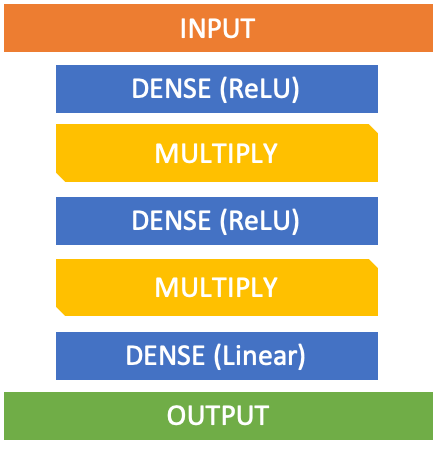} &
\hspace{-1cm}\includegraphics[width=.23\textwidth ,angle=0]{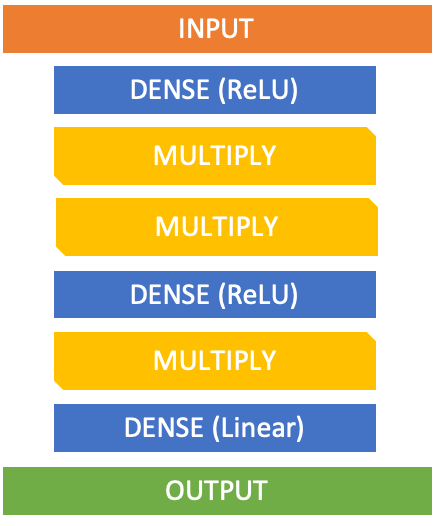}   
\end{tabular}
\caption{Modification of standard two-layer feed-forward architecture with the introduction of Multiply layers that increase local non-linearity. For brevity, the letters d and m represent the Dense and Multiply layers respectively that form an architecture (e.g. architecture dmdd has the layer structure Dense $\rightarrow$ Multiply $\rightarrow$ Dense $\rightarrow$ Dense).}
\label{fig:multiply_layer_examples}
\end{figure}

While many common tasks that employ supervised learning
such as classification of images and game playing, rarely mandate 
 accurate predictions of the approximated function's derivatives, while
computational mechanics problems that are based on variational principle often require their determination. 

To facilitate the Sobolev training, we must employ a setup to ensure 
that the learned function is an element of a Sobolev space. 
First, we must define the norm and loss function to measure 
the distance between the approximated function and the benchmark. 
Second, we must introduce the types of activities such that the basis 
of the learned function spans a Sobolev space. 
Most feed-forward neural network architectures used for regression utilize a combination of linear or piece-wise linear activation functions. For example, a common architecture for regression may employ the ReLU activation function, $\text{ReLU}(\bullet) = \max(0,\bullet)$, for the intermediate hidden layers of the network and the linear activation function, $\text{Linear}(\bullet) = \bullet$, for the output layer. The second-order derivative of such architecture with respect to the input would be constant and equal to zero and, hence, not suitable for our purpose where the second derivatives of the potentials are essential. 

To ensure the degree of continuity of the learned function to gain control of the errors of the higher-order derivatives, 
 we introduce a simple technique of adding Multiply layers. These layers are placed 
in between two hidden dense layers of the network to modify the output of the preceding layer. The Multiply layer receives as input the output $\tensor{h}^{n-1}$ of the preceding layer and outputs $\tensor{h}^n $, such that, 

\begin{equation}
\tensor{h}^n = \text{Multiply}(\tensor{h}^{n-1}) = \tensor{h}^{n-1} \circ \tensor{h}^{n-1},
\label{eq:multiply_layer}
\end{equation} 
where $\circ$ is the element-wise product of two-vectors. Conceptually,  introducing a Multiply layer is equivalent to modifying the activation function of the preceding layer. By placing the Multiply layers
in between dense layers, 
this simple modification enables us to control the order of continuity of the learned multivariable function. 
Without the introduction of any additional weights or handcrafting custom activation functions, this simple technique provides a simple solution to overcome 
the issue of vanishing 2nd-order derivatives 
that would impede the training and the deployment of machine learning elastoplasticity models that heavily rely on the network's differentiability.

The placement and the number of intermediate Multiply layers 
are hyper-parameters that can be fine-tuned along with the rest of the hyperparameters of the neural network (e.g. dropout rate, number of neurons per layer, number of layers). 
The tuning of these hyperparameters can be performed manually or through automatic hyperparameter tuning algorithms (cf. \citet{bergstra2015hyperopt, komer2014hyperopt}). Several variations to the standard two-layer architecture we tested for shown in Fig.~\ref{fig:multiply_layer_examples}. The performance of these different neural networks that complete
the higher-order Sobolev training is demonstrated in the numerical experiments showcased in Section~\ref{sec:small_strain_sobolev}.

\subsection{Sobolev training of hyperelastic energy functional}
\label{sec:higher_order_sobolev}

The first component of the elastoplastic framework we train is the elastic stored energy $\psi^{e}$. The elasticity energy functional is not only useful for predicting the stress from the elastic strain, but can also be used to re-interpret the experimental data to identify the accumulated plastic strain and the plastic flow directions that are crucial for the training of the yield function and plastic flow. 

 In the infinitesimal strain regime, the hyperelastic energy functional $\psi^{\mathrm{e}}(\tensor{\epsilon}^{\mathrm{e}}) \in \mathbb{R}^{+}$ can be define as a non-negative valued function of elastic infinitesimal strain of which the first derivative is the Cauchy stress tensor 
 $\tensor{\sigma} \in \mathbb{S} $ and the Hessian is the tangential elasticity tensor 
 $\tensor{c}^{e} \in \mathbb{M}$:

\begin{equation}
\tensor{\sigma}=\frac{\partial \psi^{\mathrm{e}}\left(\tensor{\epsilon}^{\mathrm{e}}\right)}{\partial \tensor{\epsilon}^{\mathrm{e}}} \; , \;
\tensor{c}^{\mathrm{e}}=\frac{\partial \tensor{\sigma}}{\partial \tensor{\epsilon}^{\mathrm{e}}}=\frac{\partial^{2} \psi^{\mathrm{e}}\left(\tensor{\epsilon}^{\mathrm{e}}\right)}{\partial \tensor{\epsilon}^{\mathrm{e}} \otimes \partial \tensor{\epsilon}^{\mathrm{e}}}, 
\label{eq:hyperelastic_stiffness}
\end{equation}
where $\mathbb{S}$ is the space of the second-order symmetric tensors and $\mathbb{M}$ is the space of the fourth-order tensors that possess major and minor symmetries \citep{heider2020so}. 
The true hyperelastic energy functional $\psi^{\mathrm{e}}$ of the material is approximated by the neural network learned function $ \widehat{\psi}^{\mathrm{e}} (\tensor{\epsilon}^{\mathrm{e}} |\tensor{W},\tensor{b})$ with the elastic strain tensor $\tensor{\epsilon}^{\mathrm{e}}$ as the input, parametrized by weights $\tensor{W}$ and biases $\tensor{b}$ obtained from a supervised learning procedure. 

In a conventional setting, 
 the training often employs the mean square error or the $L_2$ norm as the loss function. The $L_2$ norm training objective for the training samples $i \in [1,...,N]$ takes the following form: 

\begin{equation}
\tensor{W}',\tensor{b}' = \argmin_{\tensor{W},\tensor{b}}\left( \frac{1}{N} \sum_{i=1}^{N} \gamma_1 \left\lVert \psi^{\mathrm{e}} - \widehat{\psi}^{\mathrm{e}}_{i}\right\rVert^2_2\right),
\label{eq:energy_l2_loss}
\end{equation} 
where $\gamma_1$ is a scaling coefficient discussed further in Remark~\ref{rem:scaling}.

However, the issue is that the convergence of the stored energy 
measured by the $L_{2}$ norm does not \textit{guarantee} the quality and even the existence of the stress and elastic tangent operators stemmed from the learned energy function. To rectify this issue, we introduce the use of an $H^{1}$ norm as the loss function to train a generic anisotropic energy functional that predict the polycrystal elasticity in finite deformation regime \citet{vlassis2020geometric}. In the infinitesimal regime, the supervised learning procedure is to find the weights and biases for the neural network such that,

\begin{equation}
W^{\prime}, \boldsymbol{b}^{\prime}=\underset{\boldsymbol{W}, \boldsymbol{b}}{\operatorname{argmin}}\left(\frac{1}{N} \sum_{i=1}^{N}  \left( \gamma_1\left\|\psi^{\mathrm{e}}_{i}-\widehat{\psi}^{\mathrm{e}}_{i}\right\|_{2}^{2}+ \gamma_2 \left\|\frac{\partial \psi^{\mathrm{e}}_{i}}{\partial \tensor{\epsilon}^{\mathrm{e}} _{i}}-\frac{\partial \widehat{\psi}^{\mathrm{e}}_{i}}{\partial \tensor{\epsilon}^{\mathrm{e}} _{i}}\right\|_{2}^{2}\right)\right), 
\label{eq:energy_h1_loss}
\end{equation} 
where we assume that both the energy and the stress measures are sampled together at each data point and the total number of sample is 
$N$. 
In this work, our goal is to generate elasto-plastic model that is practical 
for implicit solvers. 
This, however, is considered a difficult task in the earlier attempts on using neural network as a replacement for constitutive laws (cf. \citet{hashash2004numerical}) and, hence, the proposed solution 
is either to bypass the calculation of tangent with an explicit time integrator or to introduce finite differences on the stress predictions. 

By leveraging the differentiability achieved 
by the Multiply layer and adopting an $H^{2}$ norm as the training objective, we introduce an alternative that renders the neural network model applicable for implicit solver while eliminating the potential 
spurious oscillations of the tangent operators.  
The new training objective for the hyperelastic energy functional approximator $\widehat{\psi}^{\mathrm{e}}$ includes constraints for the predicted energy, stress, and stiffness values. This training objective, modeled after an $H^2$ norm, for the training samples $i \in [1,...,N]$ would have the following form: 

\begin{equation}
W^{\prime}, \boldsymbol{b}^{\prime}=\underset{\boldsymbol{W}, \boldsymbol{b}}{\operatorname{argmin}}\left(\frac{1}{N} \sum_{i=1}^{N} \left( \gamma_1 \left\|\psi^{\mathrm{e}}_{i}-\widehat{\psi}^{\mathrm{e}}_{i}\right\|_{2}^{2}+
\gamma_2 \left\|\frac{\partial \psi^{\mathrm{e}}_{i}}{\partial \tensor{\epsilon}^{\mathrm{e}} _{i}}-\frac{\partial \widehat{\psi}^{\mathrm{e}}_{i}}{\partial \tensor{\epsilon}^{\mathrm{e}} _{i}}\right\|_{2}^{2}
+\gamma_3 \left\| \frac{\partial^{2} \psi^{\mathrm{e}}_i}{\partial \tensor{\epsilon}^{\mathrm{e}}_i \otimes \partial \tensor{\epsilon}^{\mathrm{e}}_i} 
-\frac{\partial^{2} \widehat{\psi}^{\mathrm{e}}_i}{\partial \tensor{\epsilon}^{\mathrm{e}}_i \otimes \partial \tensor{\epsilon}^{\mathrm{e}}_i}\right\|_{2}^{2}\right)\right).
\label{eq:energy_h2_loss}
\end{equation}

\subsubsection{Simplified training for isotropic elasticity}
In this work, our primary focus is on small strain isotropic hyperelasticity which can completely be described in spectral form by the principal strain and stress values (without the principal directions). Thus, for isotropic infinitesimal hyperelasticity, the $H_2$ training objective of Eq.~\eqref{eq:energy_h2_loss} for the training samples $i \in [1,...,N]$ can be rewritten in terms of principal values as:

\begin{eqnarray}
W^{\prime}, \boldsymbol{b}^{\prime}=\underset{\boldsymbol{W}, \boldsymbol{b}}{\operatorname{argmin}}\Bigl(\frac{1}{N} \sum_{i=1}^{N} \Bigl( \gamma_1 \left\|\psi^{\mathrm{e}}_{i}-\widehat{\psi}^{\mathrm{e}}_{i}\right\|_{2}^{2}+\sum_{A=1}^{3} \gamma_2 \left\|\frac{\partial \psi^{\mathrm{e}}_{i}}{\partial \epsilon^{\mathrm{e}} _{A,i}}-\frac{\partial \widehat{\psi}^{\mathrm{e}}_{i}}{\partial \epsilon^{\mathrm{e}}_{A,i}}\right\|_{2}^{2} \notag \\
+\sum_{A=1}^{3}\sum_{B=1}^{3} \gamma_3 \left\| \frac{\partial^{2} \psi^{\mathrm{e}}_i}{\partial \epsilon^{\mathrm{e}} _{A,i}  \partial \epsilon^{\mathrm{e}} _{B,i}} 
-\frac{\partial^{2} \widehat{\psi}^{\mathrm{e}}_i}{\partial \epsilon^{\mathrm{e}} _{A,i}  \partial \epsilon^{\mathrm{e}} _{B,i}}\right\|_{2}^{2}\Bigr)\Bigr),
\label{eq:energy_h2_loss_principal}
\end{eqnarray} 
where $\epsilon^{\mathrm{e}} _{A}$ for $A = 1,2,3$ are the principal values of the elastic strain tensor $\tensor{\epsilon}^{\mathrm{e}}$. The approximated energy functional for this training objective is a function of the three input principal strains and not of a full second-order tensor of 6 input components (reduced from 9 by assuming symmetry). This effectively reduces the input parametric space of the learned function and facilitates learning by minimizing complexity. 

The training objective can further be simplified to two input variables by adopting an invariant space, commonly used in geotechnical studies when the intermediate principal stress does not exhibit a dominating effect on the elastic response 
or when the intermediate principal stress is not measured at all due to the limitation of the experiment apparatus \citep{wawersik1997new, haimson2010effect}. 
In this case, a small-strain isotropic hyperelastic law can equivalently be described with two strain invariants (volumetric strain $\epsilon _v^{\mathrm{e}}$, deviatoric strain $\epsilon_s^{\mathrm{e}}$). The strain invariants are defined as:

\begin{equation}
\epsilon_{v}^{\mathrm{e}} =\operatorname{tr}\left(\tensor{\epsilon}^{\mathrm{e}}\right), \quad \epsilon_{s}^{\mathrm{e}}=\sqrt{\frac{2}{3}}\left\|\tensor{e}^{\mathrm{e}}\right\|, \quad \tensor{e}^{\mathrm{e}}=\tensor{\epsilon}^{\mathrm{e}}-\frac{1}{3} \epsilon_{v}^{\mathrm{e}} \tensor{1},
\end{equation}
where $\tensor{\epsilon}^{\mathrm{e}}$ is the small strain tensor and $\tensor{e}^{\mathrm{e}}$ the deviatoric part of the small strain tensor. Using the chain rule, the Cauchy stress tensor can be described in the invariant space as follows:

\begin{equation}
\boldsymbol{\sigma}=\frac{\partial \psi^{\mathrm{e}}}{\partial \epsilon_{v}^{\mathrm{e}}} \frac{\partial \epsilon_{v}^{\mathrm{e}}}{\partial \tensor{\epsilon}^{\mathrm{e}}}+\frac{\partial \psi^{\mathrm{e}}}{\partial \epsilon_{s}^{\mathrm{e}}} \frac{\partial \epsilon_{s}^{\mathrm{e}}}{\partial \tensor{\epsilon}^{\mathrm{e}}}.
\end{equation}

In the above, the mean pressure $p$ and deviatoric (von Mises) stress $q$ can be defined as:

\begin{equation}
p=\frac{\partial \psi^{\mathrm{e}}}{\partial \epsilon_{v}^{\mathrm{e}}} \equiv \frac{1}{3} \operatorname{tr}(\boldsymbol{\sigma}), \quad q=\frac{\partial \psi^{\mathrm{e}}}{\partial \epsilon_{s}^{\mathrm{e}}} \equiv \sqrt{\frac{3}{2}}\|\boldsymbol{s}\|,
\end{equation}
where $\tensor{s}$ is the deviatoric part of the Cauchy stress tensor. Thus, the Cauchy stress tensor can be expressed by the stress invariants as:

\begin{equation}
\boldsymbol{\sigma}=p \mathbf{1}+\sqrt{\frac{2}{3}} q \widehat{\boldsymbol{n}},
\end{equation}

\begin{equation}
\text{where} \: \:  \widehat{\boldsymbol{n}}=\boldsymbol{e}^{\mathrm{e}} /\left\|\boldsymbol{e}^{\mathrm{e}}\right\|=\sqrt{2 / 3} e^{\mathrm{e}} / \epsilon_{s}^{\mathrm{e}}.
\end{equation}

The $H_2$ training objective of Eq.~\eqref{eq:energy_h2_loss_principal} for the training samples $i \in [1,...,N]$ can now be rewritten in terms of the two strain invariants:

\begin{eqnarray}
\tensor{W}^{\prime}, \tensor{b}^{\prime}=\underset{\tensor{W}, \tensor{b}}{\operatorname{argmin}}\Bigl( \frac{1}{N} \sum_{i=1}^{N} \Bigl( \gamma_1 \left\|\psi^{\mathrm{e}}_{i}-\widehat{\psi}^{\mathrm{e}}_{i}\right\|_{2}^{2}+ 
\gamma_4 \left\| p_i - \widehat{p}_i \right\|_{2}^{2}  \notag \\ 
+ \gamma_5 \left\| q_i - \widehat{q}_i \right\|_{2}^{2}
+\sum_{\alpha=1}^{2}\sum_{\beta=1}^{2} \gamma_6 \left\| D_{\alpha\beta,i}^{\mathrm{e}}  - \widehat{D}_{\alpha\beta,i}^{\mathrm{e}}  \right\|_{2}^{2}\Bigr)\Bigr),
\label{eq:energy_h2_loss_invariant}
\end{eqnarray}
where:
\begin{equation}
 D_{11}^{\mathrm{e}} =\frac{\partial ^2 \psi }{\partial \epsilon_{v}^{\mathrm{e} \, 2}}, \quad D_{22}^{\mathrm{e}} =\frac{\partial ^2 \psi }{\partial \epsilon_{s}^{\mathrm{e} \, 2}}, \quad \text{and} \quad D_{12}^{\mathrm{e}} =D_{21}^{\mathrm{e}}=\frac{\partial ^2 \psi }{\partial \epsilon_{v}^{\mathrm{e}}\partial \epsilon_{s}^{\mathrm{e}}}.
 \end{equation}

Finally, it should be noted that while the loss function listed in Eqs. \eqref{eq:energy_h2_loss_principal} and \eqref{eq:energy_h2_loss_invariant}
is sufficient to control all degree of freedoms for the stored energy, the stress and the tangent simultaneously for the isotropic and two-invariant cases, these two loss functions can also be used for the general anisotropic case if only partial control on the stress and tangent is needed. 

\remark{\label{rem:scaling}\textbf{Rescaling of the training data}. The terms of the loss functions  mentioned in this section constrain measures of different units (energy, stress, stiffness). In every loss function in this work, we have introduced scaling coefficients $\gamma_\alpha$ for consistency of the units in the formulation. In practice, as pre-processing step all input and output measures used in neural network training have been scaled to a unitless feature range from 0 to 1. Thus, the use of other scaling unit coefficients was not deemed necessary during training.  }

\subsection{Training of evolving yield function as level set}
\label{sec:yield}
This section introduces the theoretical framework that regards 
the evolution of the yield surface as a level set evolution problem. 
To illustrate the key ideas with a visual geometrical interpretation and without the burden of generating a large database, we 
restrict our attentions to construct a yield function that remains pressure-insensitive (Fig.~\ref{fig:function_discovery}) but may otherwise evolve in any arbitrary way on the $pi$-plane, including moving, expanding, contracting and deforming the 
elastic region. The goal of the supervised learning task is to determine the optimal way the yield function should evolve such that it is consistent 
with the observed experimental data collected after the plastic yielding 
and obeying the thermodynamic constraints that can be interpreted geometrically in the stress space. 

\subsubsection{Reducing the dimension of data by leveraging symmetries}
\label{sec:dimensionalreduction}
Here we provide a brief review of the geometrical interpretation 
of the stress space and how it can be used to reduce the dimensions 
of the data and reduce the difficulty of the machine learning tasks. 
In this work, we consider a convex elastic domain $\mathbb{E}$ defined by a yield surface $f$. This yield function is a function of Cauchy stress 
 $\tensor{\sigma}$ and the internal variable $\xi$ that represent the history-dependent behavior of the material, i.e.,  (cf. \citet{borja_plasticity_2013}), 
 \begin{equation}
 \xi = \int_{0}^{t} \dot{\lambda} dt, 
 \end{equation}
where $\xi$ is an monotonically increasing function of time and $\dot{\lambda}$ is the rate of change of the plastic multiplier where 
$\dot{\tensor{\epsilon}}^{p} = \dot{\lambda} \partial g / \partial \tensor{\sigma}$ and $g$ is the plastic potential.  
The yield function returns a negative value in the elastic region and equals to zero when the material is yielding. 
 The stress on the boundary $f(\tensor{\sigma}, \xi)=0$ is therefore the yielding stress and all admissible stress belong to 
the closure of the elastic domain, i.e., 
\begin{equation}
\mathbb{E} := \{ (\tensor{\sigma}, \xi) \in \mathbb{S}\times \mathbb{R}^1| f(\tensor{\sigma}, \xi) \leq 0 \}.
\label{eq:elastic_domain}
\end{equation}

First, we assume that the yield function depends only on the principal stress. This treatment reduces the dimension of the stress from six to three. Then, we assume that the plastic yielding is not sensitive to the mean pressure. As such, the shape of the yield surface in the principal stress space can be sufficiently 
described by a projection on the $\pi$-plane and, hence, further reduce the dimensions of the independent stress input from three to two. 
To further simplify the interpolation of the yield surface, we introduce 
a polar coordinate system on the $\pi$-plane such that different monotonic stress paths commonly obtained from triaxial tests can be easily described via the Lode's angle. 
 
Recall that the $\pi$-plane refers to a projection of the principal stress space based on the space diagonal defined by $\sigma_1 = \sigma_2 = \sigma_3$. More specifically, the $\pi$-plane is defined by the equation:
\begin{equation}
\sigma_1 + \sigma_2 + \sigma_3 = 0.
\label{eq:pi_plane}
\end{equation}

 \begin{figure}[h!]
\newcommand\siz{.49\textwidth}
\centering
\begin{tabular}{M{.49\textwidth}M{.49\textwidth}}
\includegraphics[width=.33\textwidth ,angle=0]{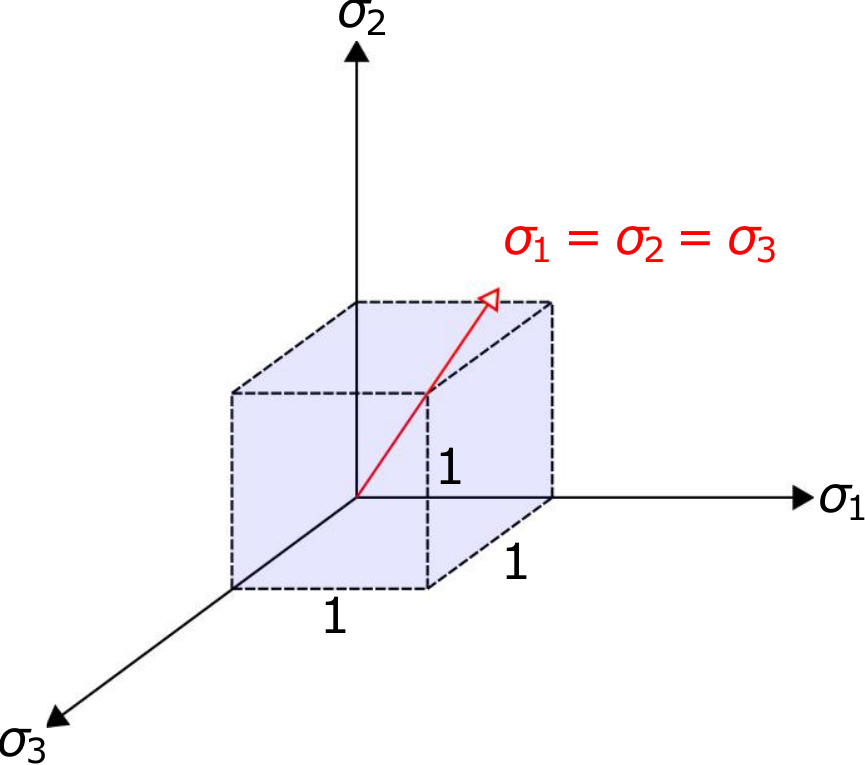} &
\includegraphics[width=.33\textwidth ,angle=0]{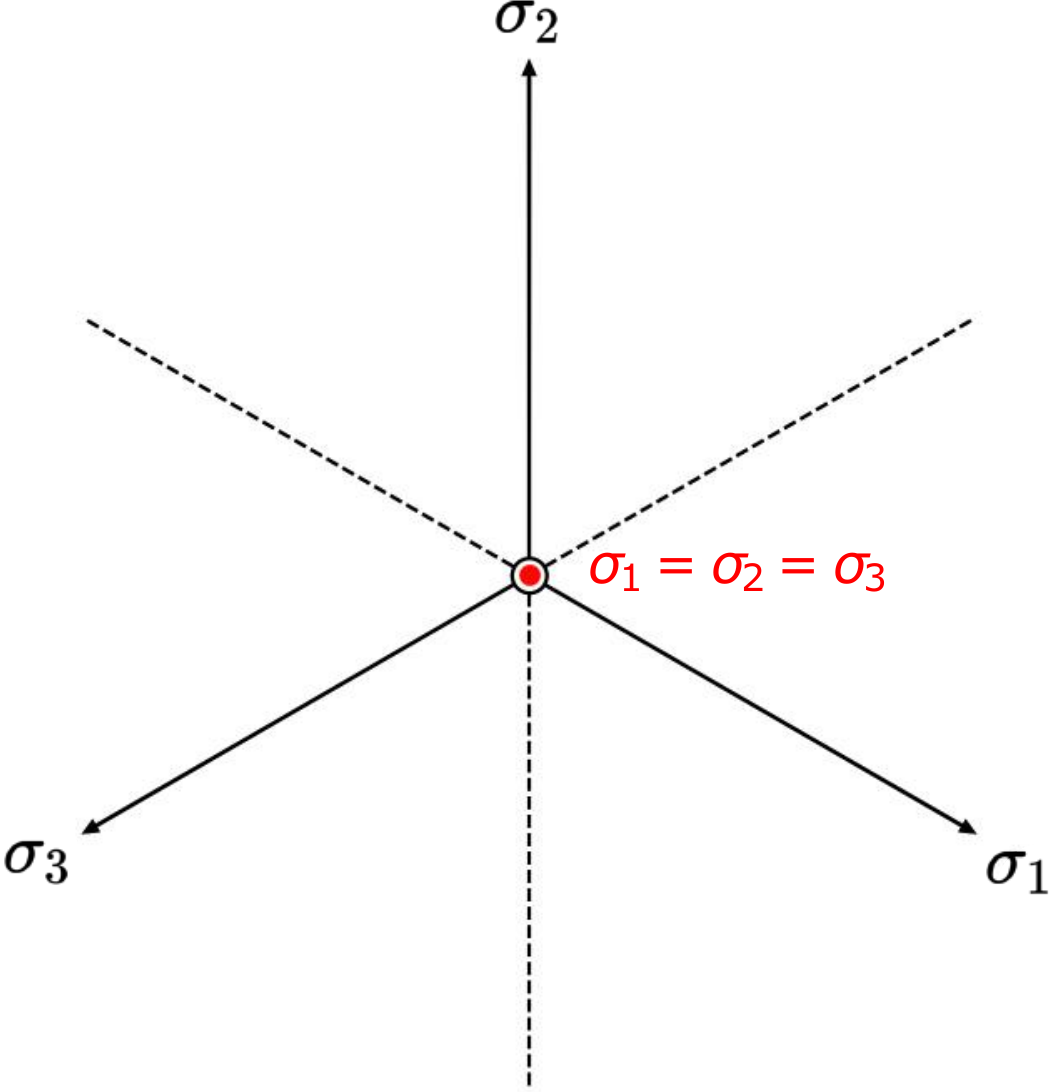} \\     
\end{tabular}
\caption{The $\pi$-plane is perpendicular to the space diagonal and is passing through the origin of the principal stress axes. Figure reproduced from \citet{borja_plasticity_2013}.}
\label{fig:pi_plane_axes}
\end{figure}

 The transformation from the original stress space coordinate system $(\sigma_1,\sigma_2,\sigma_3)$ to the $\pi$-plane can be decomposed into two specific rotations $\tensor{R}$ and $\tensor{R''}$ of the coordinate system (cf. \citep{borja_plasticity_2013}) such that, 

\begin{equation}
\left\{\begin{array}{l}
\sigma_{1} \\
\sigma_{2} \\
\sigma_{3}
\end{array}\right\}=\tensor{R}' \tensor{R}^{\prime \prime} \left\{\begin{array}{c}
\sigma_{1}^{\prime \prime} \\
\sigma_{2}^{\prime \prime} \\
\sigma_{3}^{\prime \prime}
\end{array}\right\} =\left[\begin{array}{ccc}
\sqrt{2} / 2 & 0 & \sqrt{2} / 2 \\
0 & 1 & 0 \\
-\sqrt{2} / 2 & 0 & \sqrt{2} / 2
\end{array}\right]\left[\begin{array}{ccc}
1 & 0 & 0 \\
0 & \sqrt{2 / 3} & 1 / \sqrt{3} \\
0 & -1 / \sqrt{3} & \sqrt{2 / 3}
\end{array}\right]\left\{\begin{array}{c}
\sigma_{1}^{\prime \prime} \\
\sigma_{2}^{\prime \prime} \\
\sigma_{3}^{\prime \prime}
\end{array}\right\}
\label{eq:transform_to_sigma_dprim_coords}
\end{equation} 

For pressure-insensitive plasticity, $\sigma_{3}''$ is not needed, as the principal stress differences are function of $\sigma_{1}''$ and $\sigma_{2}''$ and  are independent of $\sigma_{3}''$.  

 We opt to describe the stress states of the material on the $\pi$-plane using two stress invariants, the polar radius $r$ and the Lode's angle $\theta$ \citep{lode1926versuche}. These invariants are derived by solving the characteristic equation of the deviatoric component $\tensor{s} \in \mathbb{S}$ of the Cauchy stress tensor, following \citep{borja_plasticity_2013}:
 
 \begin{equation}
s^3-J_2 s -J_3 = 0,
\label{eq:dev_stress_characteristic_eq}
\end{equation}
where $s$ is a principal value of $\tensor{s}$, and 
 \begin{equation}
J_2 = \frac{1}{2}\text{tr} (\tensor{s}^2), \qquad J_3 = \frac{1}{3}\text{tr} (\tensor{s}^3),
\label{eq:j2_j3}
\end{equation}
are respectively the second and third invariants of the tensor $\tensor{s}$. Utilizing the identity:
 \begin{equation}
\cos^3 \theta - 3/4 \cos \theta - 1/4 \cos 3 \theta = 0,
\label{eq:trig_identity}
\end{equation}
and writing $s$ in polar coordinates such that:
 \begin{equation}
s = \rho \cos \theta,
\label{eq:in_polar_coords}
\end{equation}
and substituting in \eqref{eq:dev_stress_characteristic_eq}, the polar radius and the Lode's angle, can be retrieved as:
 \begin{equation}
\rho = 2 \sqrt{J_2 / 3}, \qquad \text{and} \qquad \cos 3 \theta = \frac{3 \sqrt{3} J_3}{2 J_2^{3/2}}.
\label{eq:radius_angle_definition}
\end{equation}
In terms of the $\pi$-plane coordinates $\sigma_{1}^{\prime \prime}$ and $\sigma_{2}^{\prime \prime}$, the Lode's coordinates $\rho$ and $\theta$ can be respectively written as:
 \begin{equation}
\rho =  \sqrt{\sigma_{1}^{\prime \prime \, 2} + \sigma_{2}^{\prime \prime \, 2} }, \qquad \text{and} \qquad \tan \theta = \frac{\sigma_{2}^{\prime \prime }}{\sigma_{1}^{\prime \prime }}.
\label{eq:radius_angle_definition_prime_coords}
\end{equation}

Thus, for an isotropic pressure-independent plasticity model, the yield surface can equivalently be described by an approximator using either the principal stresses $\sigma_1$, $\sigma_2$, and $\sigma_3$ or the stress invariant $\rho$, and, $\theta$ such that:

 \begin{equation}
\overline{f}(\sigma_1,\sigma_2,\sigma_3,\xi) = \widehat{f}(\rho, \theta, \xi) = 0.
\label{eq:yield_level_set}
\end{equation}

 \begin{figure}[h!]
\centering
\includegraphics[width=0.95\textwidth, angle=-0]{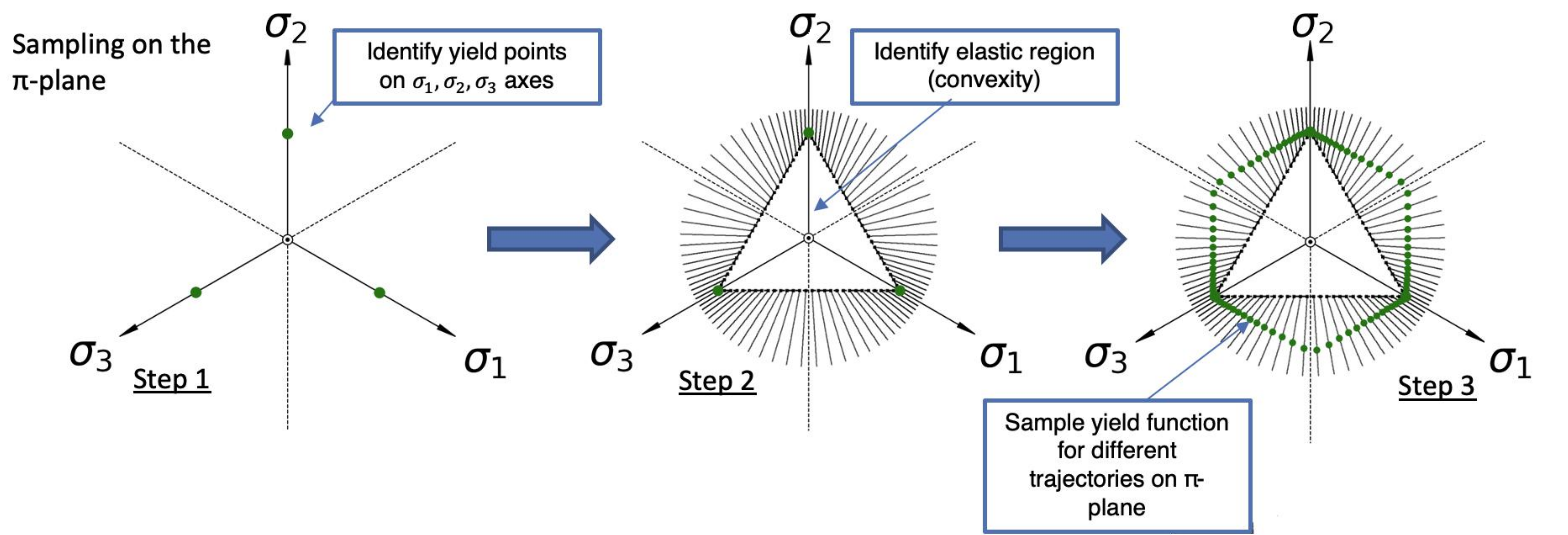}
\caption{The $\pi$-plane is used as the basis for the polycrystal plasticty dataset generation. The plane is explored radially for different Lode angle inputs. Every Lode angle constitutes a different simulation. The material is first loaded in the direction of $\sigma_1$, $\sigma_2$, and $\sigma_3$ to find the yielding points in these directions. Knowing the elastic domain is convex, the exploration can be reduced to outside of the area delimited by the three yield points in the principal directions.} 
\label{fig:data_generation_scheme}
\end{figure}

In this case, both the isotropy of the yield function and the symmetry along the hydrostatic axis offer great opportunity to simplify the 
training process. First, the reduction of the dimensions may greatly simplify the supervised learning \citep{heider2020so, wang_multiscale_2018}. 
Second, the geometrical interpretation 
of yield function on the $\pi$-plane provided a visual guidance for more effective data exploration (see Fig. \ref{fig:data_generation_scheme} for instance). In our numerical examples, our database consists of results of direct numerical simulations of polycrystals undergoing isochoric plastic deformation. As a result, we simply design experiments that covers stress paths for different Lode's angle and that will provide a mean for us to determine the initial yield function and the subsequent evolution.

\subsubsection{Detecting initial yielding from direct numerical simulations}
In the plasticity literature, the accurate prediction of the initial yielding 
point often does not receive sufficient attention. 
This may be attributed to the fact that the nonlinearity of the stress-strain curves make it difficult to pinpoint the actual yielding. While predicting the initial yield surface is not necessary crucial for \textit{curve-fitting} stress-strain curves (as the imprecise elasticity and yield surface 
can be masked by a overfitting hardening curves), such a practice may significantly reduce the predictive capacity of the model \citep{wang2016identifying}. 
In principle, it is possible to accurately locate the actual initial yielding surface by applying elastic unloading to extrapolate
the stress at which plastic strain begins accumulating for different stress paths. 
However, this technique is not practical due to the cost of experiments and the impossible task of preparing multiple identical specimens. 
Hence, the alternative is to assume an elasticity model and locate the point at which a departure between the Cauchy stress and the stress assuming no plastic deformation for a given stress path. 

In our numerical experiments, we use a FFT solver to generate 
3D polycrystal simulations and use these simulated data to constitute 
the material database. As such, we simply detect the initial yielding by monitoring the RVE and record the stress when at least one location of the RVE develops plastic strain.

\subsubsection{Data preparation for training the yield function as a level set}
\label{sec:level_set}
Identifying the set of stress at which the initial plastic yielding occurs 
is a necessary but not sufficient condition to generate a yield surface. 
In fact, a yield surface $f(\tensor{\sigma}, \xi)$ must be well-defined not just at $f=0$ but also anywhere in the product space $\mathbb{S} \times R^{1}$. 
Another key observation is that, in order for the yield surface to function properly, the value of $f(\tensor{\sigma}, \xi)$ inside and outside the yield surface may vary, provided that the orientation of the stress gradient remains consistent. For instance, consider two classical $J_{2}$ yield functions, 
\begin{equation}
f_1(\tensor{\sigma}, \xi) = \sqrt{2 J_{2}} - \kappa \leq 0 \; \; ; \; \;
f_2(\tensor{\sigma}, \xi) = \sqrt{J_{2}} - \kappa/\sqrt{2} \leq 0.
\end{equation}

These two models will yield identical constitutive responses except that, in each incremental step, the plastic multiplier deduced from $f_{1}$ is $\sqrt{2}$ times smaller than that of $f_{2}$, as the stress gradient of $f_{1}$ is $\sqrt{2}$ times larger than that of $f_{2}$. 
With these observations in mind, we introduce a level set approach 
where the yield surface is postulated to be a signed distance function
and the evolution of the yield function is governed by a Hamilton-Jacobi equation that is not solved but generated from a supervised learning with the following steps. 

\begin{enumerate}
\item \textbf{Generate auxiliary data points to train the signed distance yield function}. In this first step, we first attempt to construct a signed distance function $\phi$ in the stress space when the internal variable is fixed on a given value, i.e. $\xi = \overline{\xi}$ where $\rho=0$ when yielding. 
Let $\Omega$ be the solution domain of the stress space of which the signed distance function is defined. Assume that the yield function can be sufficiently described in $\pi-$plane. For simplicity, we will adopt the polar coordinate system to parametrize the signed distance function $\phi$ that is used to train the yield surface , i.e., 

\begin{equation}
\tensor{x}(\sigma_{11},\sigma_{22},\sigma_{33},\sigma_{12},\sigma_{23},\sigma_{13}) = \overline{\tensor{x}}(\sigma_1,\sigma_2,\sigma_3) =  \widehat{\tensor{x}}(\rho,\theta).
\label{eq:position_vectors}
\end{equation}

\begin{figure}[h!]
\centering
\includegraphics[width=.70\textwidth,angle=-0]{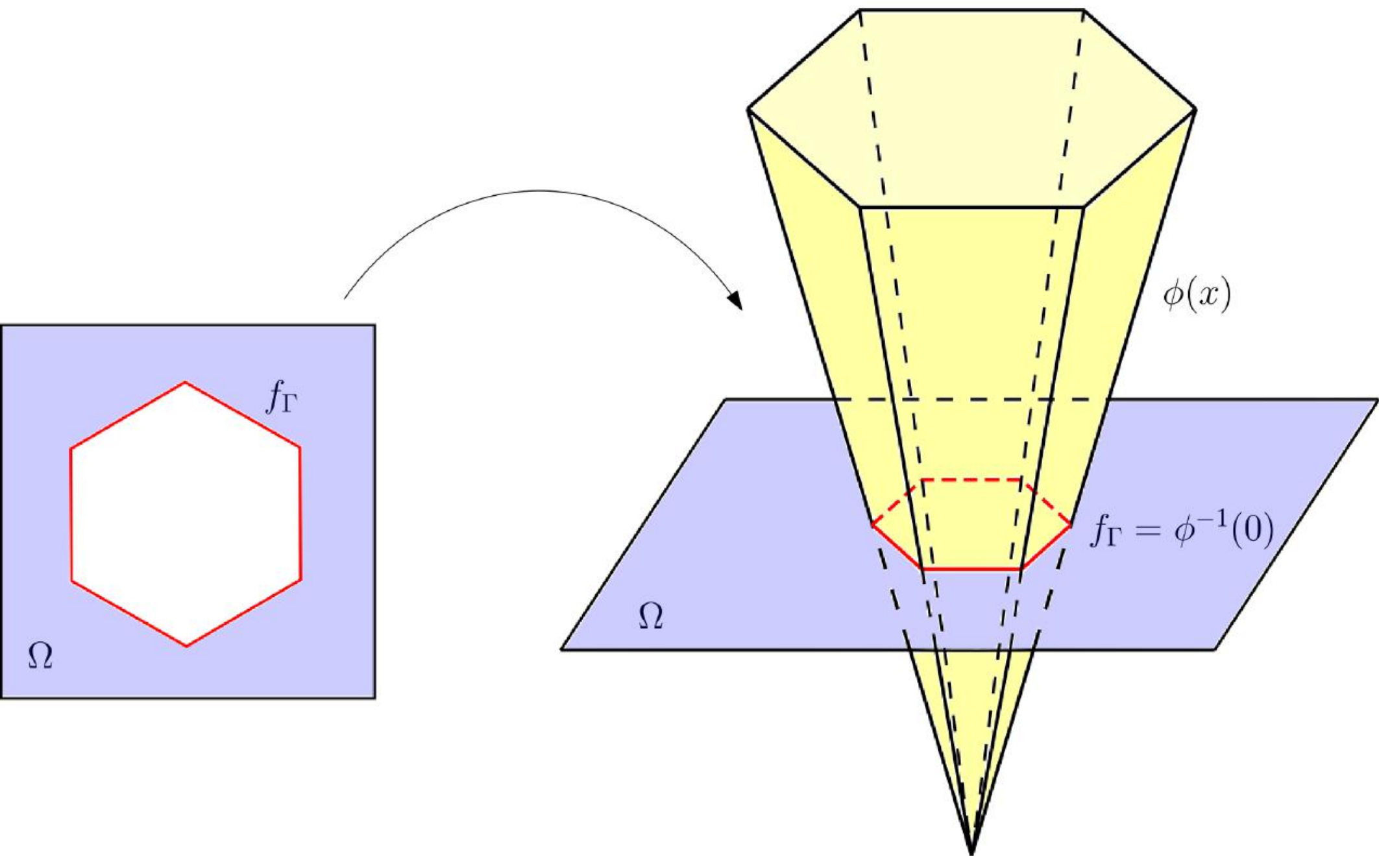}
\caption{Based on the level set method, the yield surface interface $f_{\Gamma}$ can be represented as the the zero level set of some higher-dimensional function $\phi(x)$.} 
\label{fig:levelset_sketch}
\end{figure}

The signed distance function (see, for instance Figure~\ref{fig:levelset_sketch}) is defined as 
\begin{equation}
\phi(\widehat{\mathbf{x}}, t)=\left\{\begin{array}{cl}
d(\widehat{\mathbf{x}}) & \text{ outside } f_{\Gamma} (\text{inadmissible stress}) \\
0 & \text{ on } f_{\Gamma} (\text{yielding}) \\
-d(\widehat{\mathbf{x}}) & \text{ inside } f_{\Gamma} \text{ (elastic region)}
\end{array}\right. ,
\label{eq:signed_distance_function}
\end{equation}
where $d(\widehat{\tensor{x})}$ is the minimum Euclidean distance between any point $\tensor{x}$ of $\Omega$ and the interface $f_\Gamma = \{ \widehat{\vec{x}} \in \mathbb{R}^{2} | f(\widehat{\vec{x}}) = 0 \}$, defined as:

\begin{equation}
d(\widehat{\tensor{x}})=\min \left(\left|\widehat{\vec{x}}-\widehat{\vec{x}}_{\Gamma}\right|\right).
\label{eq:distance_function}
\end{equation}
where $\vec{x}_{\Gamma}$ is the yielding stress for a given $\xi$. 
The signed distance function is obtained by solving the 
 Eikonal equation $|\nabla^{\widehat{\vec{x}}} \phi|=1$ while prescribing the signed distance function as 0 at $\vec{x} \in f_{\Gamma}$ In the polar coordinate system, 
 the Eikonal equation reads, 
 \begin{equation}
(\frac{\partial{\phi}}{\partial {\rho}} )^{2} + \frac{1}{\rho^{2}} ((\frac{\partial{\phi}}{\partial {\theta}} )^{2} = 1.
 \end{equation}
Note that the is a singularity at the polar coordinate of the $\pi-$ plane at $\rho =0$ and, hence, the origin point is not used as an auxiliary point to train the yield function. The Eikonal solution can be simply solved by a fast marching solver in the 2D polar coordinate.  
 \begin{figure}[h!]
\newcommand\siz{.49\textwidth}
\centering
\begin{tabular}{M{.49\textwidth}M{.49\textwidth}}
\includegraphics[width=.43\textwidth ,angle=0]{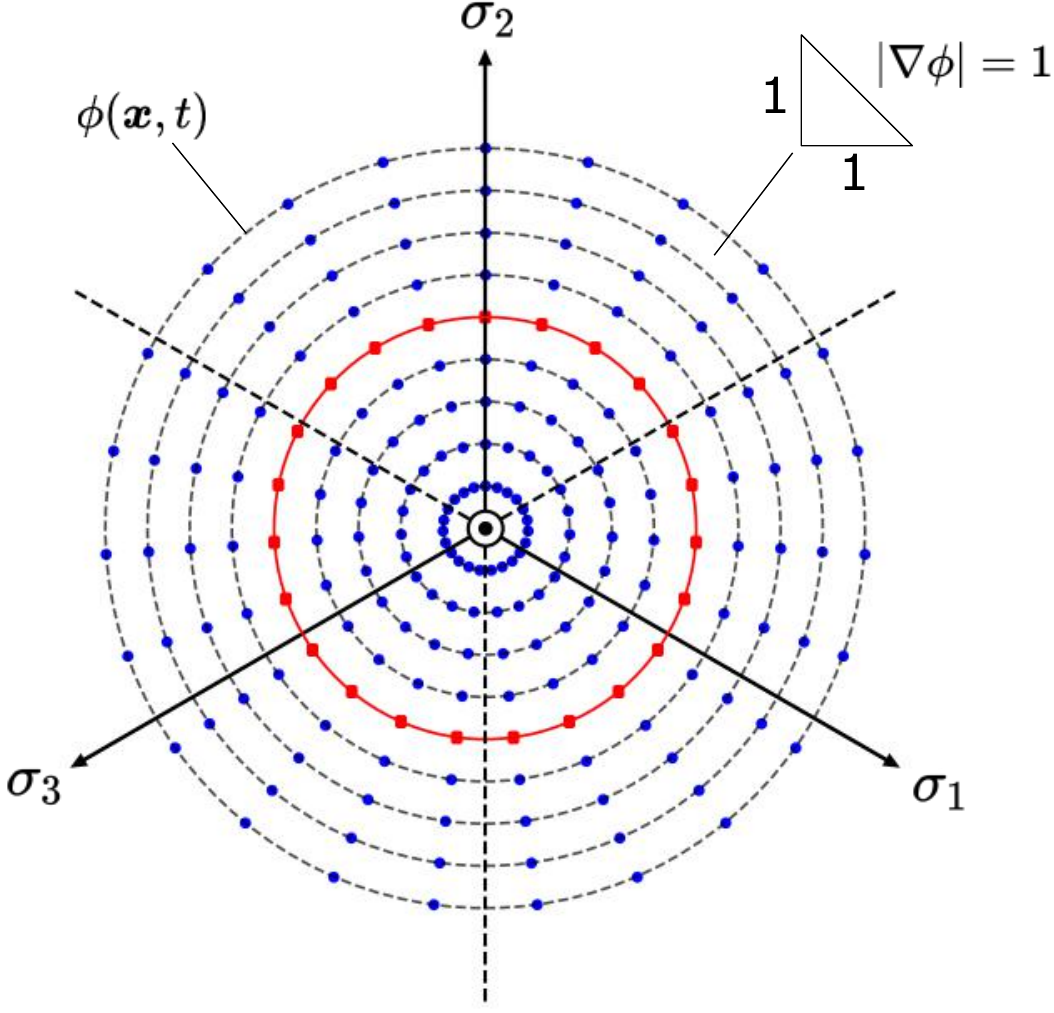} &
\includegraphics[width=.43\textwidth ,angle=0]{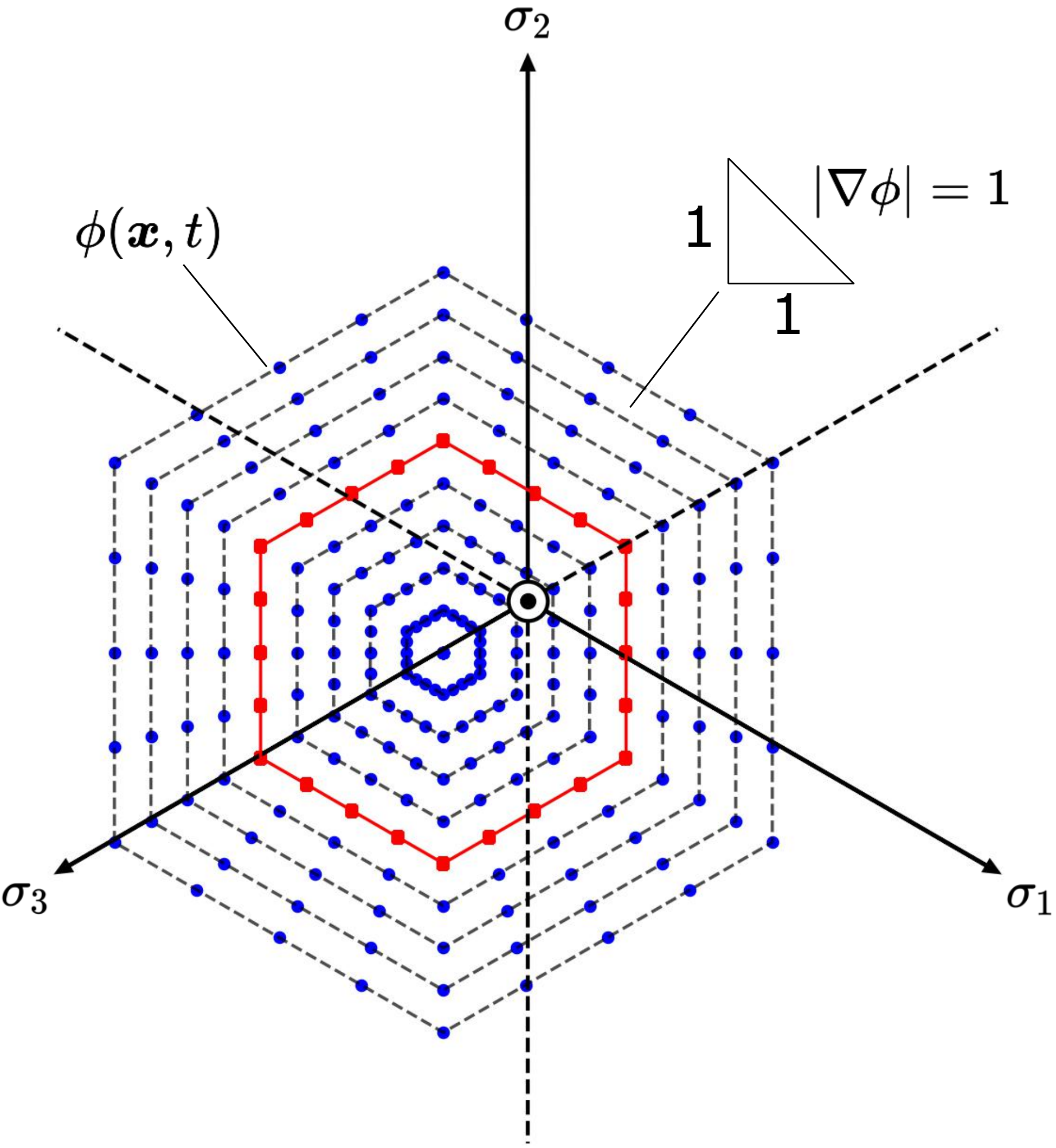} \\     
\includegraphics[width=.41\textwidth ,angle=0]{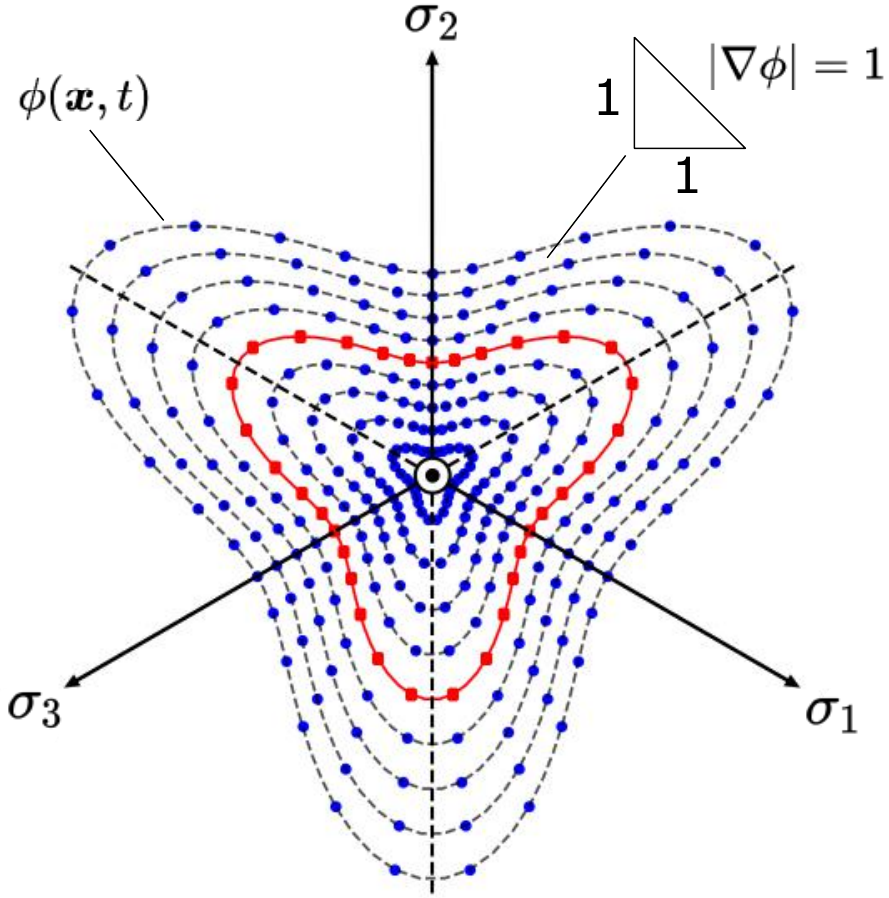} &
\includegraphics[width=.43\textwidth ,angle=0]{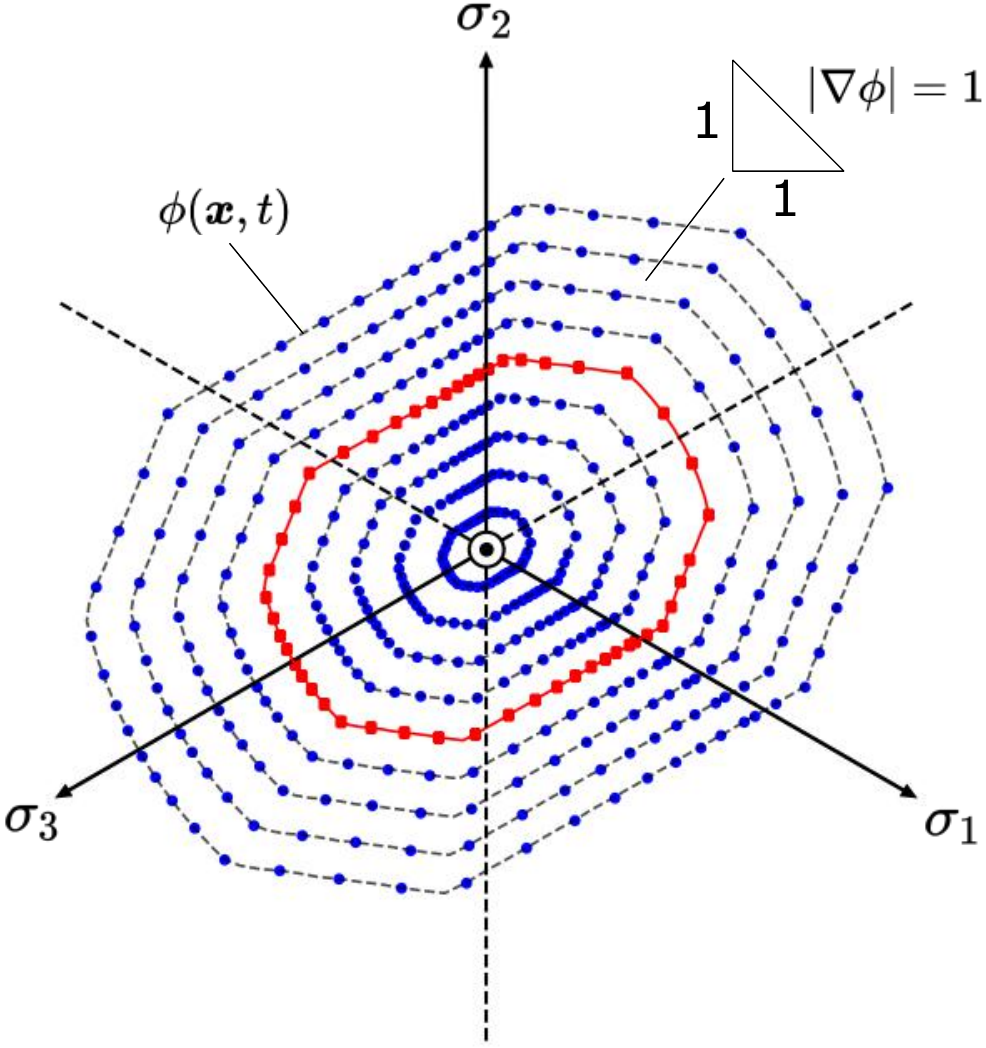} \\    
\end{tabular}
\includegraphics[width=.75\textwidth]{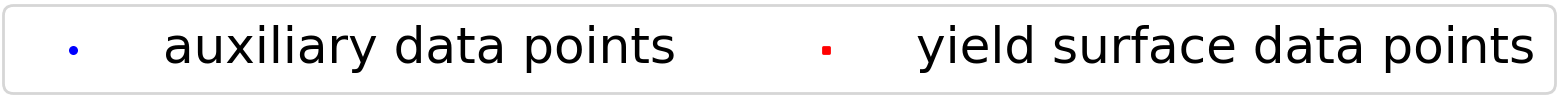} 
\caption{ Generation of auxiliary data points through level set re-initialization.The yield function level set $\phi(\tensor{x},t)$ is created using a signed distance function. The initial yield surface points are given by the experimental results -- a level set is constructed for every accumulated plastic strain $\bar{\epsilon}_p$ value in the data set. The isocontour curves represent the projection of the signed distance function level set on the $\pi$-plane.}
\label{fig:signed_distance_function}
\end{figure}
Figure \ref{fig:signed_distance_function} shows an example of solution 
a number of signed distance function converted from classical yield surfaces or deduced from direct numerical simulations.  
\\

 \item \textbf{Obtain the speed function to constitute Hamilton-Jacobi hardening of the yield function}
After we generate a sequence of signed distance function for different $\xi$, we might introduce an inverse problem to obtain the velocity function for the Hamilton-Jacobi equation that evolves the signed distance function. Recall that the Hamilton-Jacobi equation may take the following forms: 

\begin{equation}
\frac{\partial \phi}{\partial t}+\tensor{v} \cdot \nabla^{\widehat{\vec{x}}} \phi=0,
\label{eq:phi_convection}
\end{equation}

where $\tensor{v}$ is the normal velocity field that defines the geometric evolution of the boundary and, in the case of plasticity, is chosen to describe the observed hardening mechanism. The velocity field is given by:
\begin{equation}
\tensor{v} = F \tensor{n},
\label{eq:velocity_field}
\end{equation}

where $F$ is a scalar function describing the magnitude of the boundary change and $\tensor{n}=\nabla^{\widehat{\vec{x}}} \phi /|\nabla^{\widehat{\vec{x}}}  \phi| $. Using $\nabla \phi \cdot \nabla \phi=|\nabla \phi|^{2}$ in Eq.~\eqref{eq:phi_convection}, the level set Hamilton-Jacobi equation for stationary yield function can be simplified as, 
\begin{equation}
\frac{\partial \phi}{\partial t}+F|\nabla^{\widehat{\vec{x}}} \phi|=0.
\label{eq:hamilton_jacobi}
\end{equation}
Note that $t$ is a pseudo-time and since the snapshot of $\phi$ we obtained from Step 1 remains a signed distance function, then $|\nabla^{\widehat{\vec{x}}} \phi| = 1$. 
Next, we replace the pseudo-time $t$ with $\xi$. 
Assuming that the experimental data collected from different stress paths are collected data points $N$ times beyond the initial yielding point, each time with the same incremental plastic strain $\Delta \lambda$, then Step 1  will provide us a collection 
of signed distance function $\{ \phi_{0}, \phi_{1},  ....,\phi_{n+1} \}$ corresponding to $\{ \xi_{0}, \xi_{1}, ...., \xi_{n+1} \}$.
Then, the corresponding velocity function can be obtained via finite difference, i.e., 
\begin{equation}
F_{i} \approx \frac{\phi_{i} - \phi_{i+1}}{\xi_{i+1} - \xi_{i}},  
\end{equation}
where $F_{i}(\rho, \theta) = F(\rho, \theta, \xi_{i})$ and $i=0, 1, 2,..., n+1$. 
By setting the signed distance function that fulfills Eq. \ref{eq:hamilton_jacobi} as the yield function, i.e., $f(\rho, \theta, \xi) = \phi(\rho, \theta, \xi)$ ,
we may use experimental data generated from the loading paths demonstrated in Fig. \ref{fig:data_generation_scheme} to train 
a neural network to predict a new yield function or velocity function for an arbitrary $\xi$ that represent the history of the strain (see Figure \ref{fig:velocity_field}). 

 \begin{figure}[h!]
\centering
\includegraphics[width=.5\textwidth]{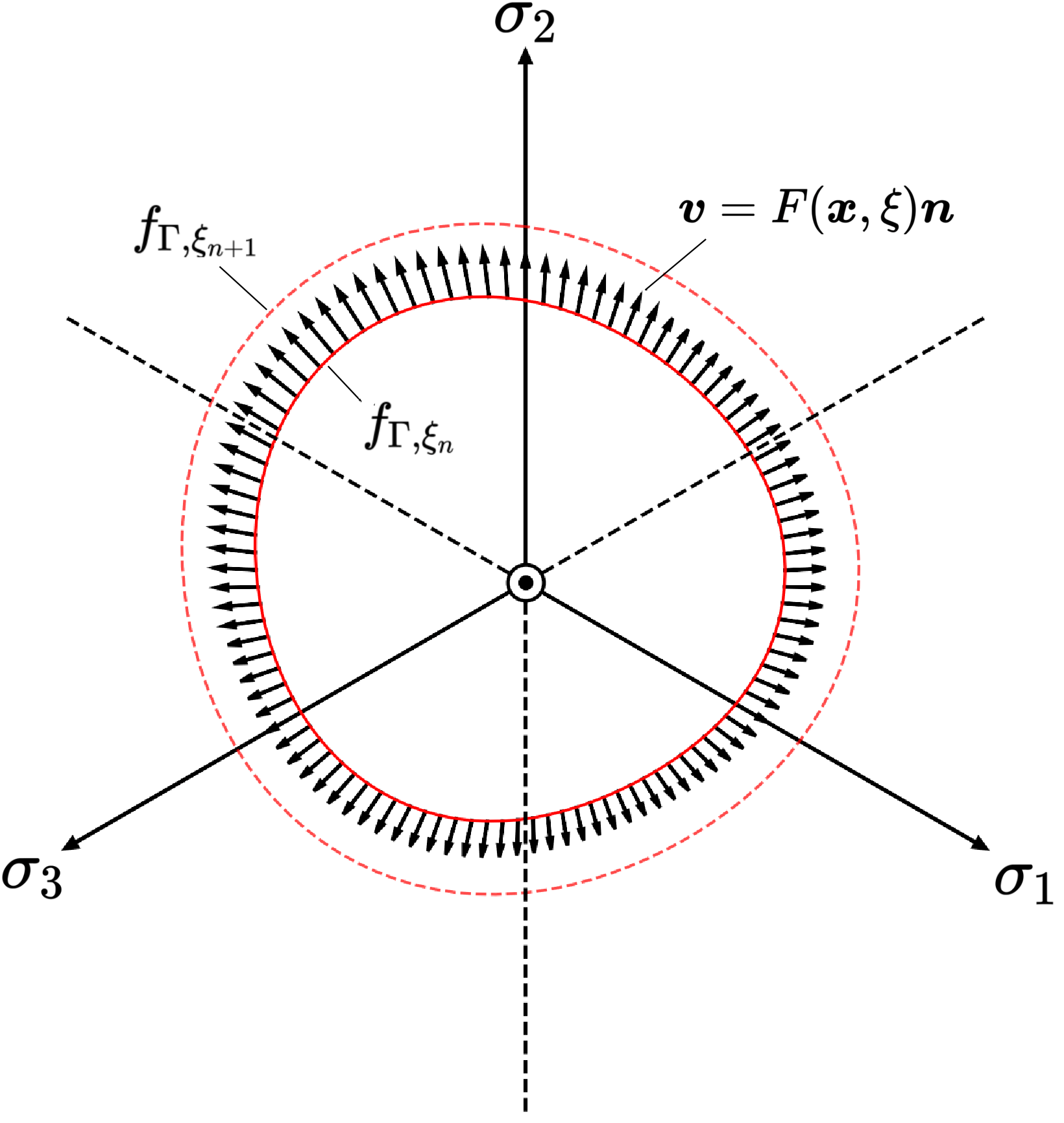} 
\caption{The evolution of the yield surface $f_{\Gamma}$ is connected to a level set $\phi(\tensor{x},\xi)$ extension problem. The velocity field of the Hamilton-Jacobi equation~\eqref{eq:hamilton_jacobi} emulates the material hardening law. The yield surface evolution and the velocity field are inferred from the data through the neural network training.}
\label{fig:velocity_field}
\end{figure}

\end{enumerate}

More importantly, we show that the evolution of the yield function can be modeled as a level set evolving according to 
a Hamilton-Jacobi equation. This knowledge may open up many new possibilities to capture hardening without any hand-crafted treatment. 
To overcome the potential cost to solve the Hamilton-Jacobi equation, 
we will introduce a supervised learning procedure to obtain the updated yield function for a given strain history represented by the internal variable $\xi$ without explicitly solving the Hamilton-Jacobi equation (see the next sub-sections). 
Consequently, this treatment will enable us to create a generic elasto-plasticity framework that can replace the hard-crafted yield functions and hardening laws without the high computational costs and the burden of repeated modeling trial-and-errors.

\subsubsection{Training yield function with associative plastic flow}
Assuming an associative flow rule, the generalized Hooke's law utilizing a yield function neural network approximator $\widehat{f}$ can be written in rate form as:

\begin{equation}
\dot{\tensor{\sigma}}=\tensor{c}^{\mathrm{e}}:\left(\dot{\tensor{\epsilon}}-\dot{\lambda} \frac{\partial \widehat{f}}{\partial \tensor{\sigma}}\right).
\label{eq:hookes_law_rate_form}
\end{equation}

And in incremental form, the predictor-corrector scheme is written as:

\begin{equation}
\tensor{\sigma}_{n+1}=\tensor{\sigma}_{n+1}^{\mathrm{tr}}-\Delta \lambda \tensor{c}_{n+1}^{\mathrm{e}}:\left.\frac{\partial \widehat{f}}{\partial \tensor{\sigma}}\right|_{n+1}, 
\label{eq:hookes_law_incremental_form}
\end{equation}

where

\begin{equation}
\tensor{\sigma}_{n+1}^{\mathrm{tr}}=\tensor{\sigma}_{n}+\tensor{c}_{n+1}^{\mathrm{e}}: \Delta \tensor{\epsilon}=\tensor{c}_{n+1}^{\mathrm{e}}: \tensor{\epsilon}_{n+1}^{\mathrm{e} \operatorname{tr}}, \quad \tensor{\epsilon}_{n+1}^{\mathrm{etr}}=\tensor{\epsilon}_{n}^{\mathrm{e}}+\Delta \tensor{\epsilon}.
\label{eq:stress_strain_increments}
\end{equation}

The strain and stress tensor predictors can be written in spectral form as follows:

\begin{equation}
\tensor{\sigma}_{n+1}^{\mathrm{tr}}=\sum_{A=1}^{3} \sigma_{A, n+1}^{\mathrm{tr}} \tensor{n}_{n+1}^{\operatorname{tr}(A)} \otimes \tensor{n}_{n+1}^{\operatorname{tr}(A)}, \qquad
\tensor{\epsilon}_{n+1}^{\mathrm{etr}}=\sum_{A=1}^{3} \epsilon_{A, n+1}^{\mathrm{etr}} \tensor{n}_{n+1}^{\operatorname{tr}(A)} \otimes \tensor{n}_{n+1}^{\operatorname{tr}(A)}.
\label{eq:stress_strain_increments_spectral}
\end{equation}

The predictor-corrector scheme can be rewritten in spectral form, omitting the subscript $(n+1)$

\begin{equation}
\sum_{A=1}^{3} \sigma_{A} \tensor{n}^{(A)} \otimes \tensor{n}^{(A)}=\sum_{A=1}^{3} \sigma_{A}^{\operatorname{tr}} \tensor{n}^{\operatorname{tr}(A)} \otimes \tensor{n}^{\operatorname{tr}(A)}-\Delta \lambda \sum_{A=1}^{3}\left(\sum_{B=1}^{3} a_{A B}^{\mathrm{e}} \widehat{f}_{B}\right) \tensor{n}^{(A)} \otimes \tensor{n}^{(A)}
\label{eq:predictor_corrector_spectral}
\end{equation}

\begin{equation}
\tensor{n}^{(A)} \otimes \tensor{n}^{(A)}=\tensor{n}^{\operatorname{tr}(A)} \otimes \tensor{n}^{\operatorname{tr}(A)}
\label{eq:isotropy_spectral directions}
\end{equation}

By assuming that the plastic flow obeys the normality rule, we may 
use the observed plastic flow from the data to regularize the shape 
of the evolving yield function. To do so, we will leverage the fact that 
we have already obtained an elastic energy functional from the previous training. The plastic deformation mode can then be obtained by the difference between the trial and the true Cauchy stress at each incremental step where the data are recorded in an experiment or 
directed numerical simulations, i.e., 

\begin{equation}
\begin{aligned}
\sigma_{A} &=\sigma_{A}^{\mathrm{tr}}-\Delta \lambda \sum_{B=1}^{3} a_{A B}^{\mathrm{e}} f_{B}, \\
f_{A} &=\partial f / \partial \sigma_{A} = \frac{\epsilon_{A}^{\mathrm{etr}} - \epsilon_{A}^{\mathrm{e}} }{\Delta \lambda} \text { for } A=1,2,3.\\
\end{aligned}
\label{eq:predictor_corrector_scheme}
\end{equation}

At every incremental step of the data generating simulations, we have information on the total strain and total stress of the material. Having the knowledge of the underlying hyperelastic model, we can utilize an inverse mapping to estimate the elastic strain that would correspond to the current total stress, if there was no plasticity. Thus, we can post-process the available data to gather the plastic flow information necessary for the network training.

The quantities $f_{1},f_2,f_{3}$ correspond to the amount of plastic flow in the principal directions $A=1,2,3$. A neural network approximator of the yield function should have adequately accurate stress derivatives that are necessary for the implementation of the return mapping algorithm, discussed in Section~\ref{sec:return_mapping_algorithm}, and so as to provide an accurate plastic flow, in the case of associative plasticity. The normalized plastic flow direction vector $\tensor{\overline{f}}_{\text{norm}}$ can be defined as

\begin{equation}
\tensor{\overline{f}}_{\text{norm}}=\langle \widehat{f}_{1}, \widehat{f}_{2}, \widehat{f}_{3} \rangle / \| \langle \widehat{f}_{1},\widehat{f}_{2},\widehat{f}_{3} \rangle \|,
\label{eq:normalized_flow_vector}
\end{equation}
and holds information about the yield function shape in the $\pi$-plane.

In the case of the simple MLP feed-forward network, the network can be seen as an approximator function $\widehat{f} = \widehat{f}(\rho, \theta, \xi |\tensor{W},\tensor{b})$ of the true yield function level set $f$ with input the Lode's coordinates $\rho$, $\theta$, and the hardening parameter $\xi$, parametrized by weights $\tensor{W}$ and biases $\tensor{b}$. A classical training objective, following an $L_2$ norm, would only constrain the predicted yield function values. The corresponding training objective is to minimize the discrepancy measured at 
$N$  number of sample points $(\widehat{\vec{x}}, \xi) 
\in \mathbb{S} \times \mathbb{R}^{1}$ reads,
\begin{equation}
\tensor{W}',\tensor{b}' = \argmin_{\tensor{W},\tensor{b}}\left( \frac{1}{N} \sum_{i=1}^{N} \gamma_7 \left\lVert f_{i} - \widehat{f}_{i}\right\rVert^2_2\right),
\label{eq:yield_l2_loss}
\end{equation} 
where $f_{i} = f((\widehat{\vec{x}}_{i}, \xi_{i})$ and  $\widehat{f}_{i} = f((\widehat{\vec{x}}_{i}, \xi_{i})$.
A second training objective can be modeled after an $H_1$ norm, constraining both $f$ and its first derivative with respect to the stress state $\sigma_{1},\sigma_{2},\sigma_{3}$. For a neural network aprroximator parametrized as $\overline{f} = \overline{f}(\sigma_{1},\sigma_{2},\sigma_{3},\xi |\tensor{W},\tensor{b})$ using the principal stresses as inputs, this training objective for the training samples $i \in [1,...,N]$ would have the following form: 

\begin{equation}
\tensor{W}',\tensor{b}' = \argmin_{\tensor{W},\tensor{b}}\left( \frac{1}{N} \sum_{i=1}^{N} \left(  \gamma_7 \left\lVert f_{i} - \overline{f}_{i}\right\rVert^2_2 + \sum_{A=1}^{3}  \gamma_8 \left\lVert \frac{\partial f_{i}}{\partial \tensor{\sigma_{A}}_i} - \frac{\partial \overline{f}_{i}}{\partial \tensor{\sigma_{A}}_i}\right\rVert^2_2  \right)\right).
\label{eq:yield_h1_loss_principal}
\end{equation} 

Utilizing an equivalent representation of the stress state with Lode's coordinates in the $\pi$-plane, the above training objective can further be simplified. The normalized flow direction vector $\tensor{\overline{f}}_{\text{norm}}$ in Lode's coordinates can solely be described using an angle $\theta_{f}$ since the vector has a magnitude equal to unity. To constrain the flow direction angle, we modify the loss function of this higher order training objective by adding a distance function metric between two rotation tensors $\tensor{R}_{\theta ,i}$, $\tensor{R}_{\widehat{\theta} ,i}$, corresponding to $\theta_{f,i}$ and $\widehat{\theta}_{f,i}$ -- the flow vector directions in the $\pi$-plane for the data and approximated yield function respectively for the $i$-th sample. The two rotation tensors belong to the Special Orthogonal Group, SO(3) and the metric is based on the distance from the identity matrix. For the $i$-th sample, the rotation related term can be calculated as:

\begin{equation}
\overline{\Phi}_{i}=\left\|\tensor{I}-\tensor{R}_{\theta ,i}\left(\tensor{R}_{\widehat{\theta} ,i}\right)^{T}\right\|_{F}=\sqrt{2\left[3-\operatorname{tr}\left[\tensor{R}_{\theta ,i}\left(\tensor{R}_{\widehat{\theta} ,i}\right)^{T}\right]\right]},
\label{eq:yield_h1_loss_principal}
\end{equation}
where $\|\cdot\|_{F}$ is the Frobenius norm.
For a neural network aprroximator parametrized via 
the Lode's coordinates as input, i.e.
$\widehat{f} = \widehat{f}(\rho,\theta,\xi |\tensor{W},\tensor{b})$, 
the Sobolev training objective for the training samples $i \in [1,...,N]$ reads, 

\begin{equation}
\tensor{W}',\tensor{b}' = \argmin_{\tensor{W},\tensor{b}}\left( \frac{1}{N} \sum_{i=1}^{N} \left( \gamma_7  \left\lVert f_{i} - \widehat{f}_{i}\right\rVert^2_2 +  \gamma_9 \overline{\Phi}_{i} \right)\right), 
\label{eq:yield_h1_loss_lode}
\end{equation} 
where we minimize both the discrepancy of the yield function and the direction 
of the gradient in the stress space. 

\remark{\textbf{Discrete data points for yield function}. Note that the training of the yield function involves not just the points at $f(\rho, \theta)=0$ but also the new auxiliary data generated from the re-initialization of the level set/yield function. Strictly speaking, the accuracy of the elasto-plastic responses only depend on how well boundary of the admissible stress range $f(\rho, \theta) = \neq 0$ is kept track of. However, the additional yield function values inside and outside the admission range is helpful for evolving the yield function with sufficient smoothness. To emphasize the importance of data across $f(\rho, \theta) = \neq 0$, we may introduce a higher weighting factor of these data points for Eq. \eqref{eq:yield_h1_loss_lode}.
}

\subsubsection{Training yield function and non-associative plastic flow}
\label{sec:nonassociative}
Here, we present the training for the plastic flow without assuming 
that the plastic flow follows the normality rule. As such, the yield function and the plastic flow must be trained separately. 
We adopt the idea of generalized plasticity in which the plastic flow direction is directly deduced from Sobolev training of a neural network with  the experimental data  \citep{zienkiewicz1999computational}. 

Firstly, the yield function training is similar to the associative flow cases, except that the terms that control the stress gradient of the yield function cannot be directly obtained from the plastic flow due to the 
non-associative flow rule. Nevertheless, the stress gradient of the yield function may still be constrained by the convexity (if there is no intended phase transition that requires non-convexity yield function).
Recall that the convexity requires 
\begin{equation}
(\tensor{\tensor{\sigma}}^{*} - \tensor{\sigma}) : \frac{\partial \widehat{f}}{\partial \tensor{\sigma}} \leq 0 ,
\label{eq:convexity}
\end{equation}
where $\tensor{\sigma}^{*}$ is an arbitrary stress. One necessary condition we can incorporate as a thermodynamic constraint is a special case where we simply set  $\tensor{\sigma}^{*} =0$, as such we obtain,  
\begin{equation}
 \sum_{A = 1}^{3} \sigma_{A} \widehat{f}_{A} \geq 0 .
 \label{eq:convexity_penalty_yield}
\end{equation}
One way to enforce that constraint is to apply a penalty term for the loss function in Eq. \eqref{eq:yield_l2_loss}, .e.g, 
\begin{equation}
w_{\text{nnp}} \; \text{sign}(  -\sum_{A = 1}^{3} \sigma_{A} \widehat{f}_{A}),
\end{equation}
where this term will not be activated if the learned yield function is obeying the convexity. However, the sign operator may lead to a jump of the loss function, which is not desirable for training. As a result, a regularized Heaviside step function can be used to replace the sign operator, for instance,
\begin{equation}
\text{sign}^{\text{approx}}(-\sigma_{A} \widehat{f}_{A} ) = \frac{1}{2} + \frac{1}{2} \tanh (-k \sigma_{A} \widehat{f}_{A}  ),
\end{equation}
where $k$ controls how sharp the transition is at $\sigma_{A} \widehat{f}_{A} =0$. 
As shown in our numerical experiments, this additional term may not be required if the raw experimental data itself does not violate the thermodynamic restriction. 
To obtain a plastic flow, we again obtain the flow information incrementally from the experimental data via the following equations, i.e.,

\begin{equation}
\begin{aligned}
\sigma_{A} &=\sigma_{A}^{\mathrm{tr}}-\Delta \lambda \sum_{B=1}^{3} a_{A B}^{\mathrm{e}} g_{B} \\
g_{A} &=\partial g / \partial \sigma_{A} = \frac{\epsilon_{A}^{\mathrm{etr}} - \epsilon_{A}^{\mathrm{e}} }{\Delta \lambda} \text { for } A=1,2,3.\\
\end{aligned}
\label{eq:predictor_corrector_scheme_flow}
\end{equation}

We can gather the plastic flow information by post-processing the simulation data, similar to Equation \eqref{eq:predictor_corrector_scheme}. Once the plastic flow $g_{A}$ is determined incrementally for different $\xi$, we then introduce another supervised learning that reads, 
\begin{equation}
\tensor{W}',\tensor{b}' = \argmin_{\tensor{W},\tensor{b}}\left( \frac{1}{N} \sum_{i=1}^{N} \left(  \gamma_{10} \left\lVert g_{A,i} - \widehat{g}_{A,i}\right\rVert^2_2 \right)\right). 
\label{eq:plasticflow}
\end{equation}

The non-negative plastic work is the thermodynamic constraint that requires $\dot{W}^{p} = \tensor{\sigma}: \dot{\tensor{\epsilon}}^{p} \geq 0$. The corresponding incremental form for isotropic material reads, 
\begin{equation}
\Delta W = \sigma_{n+1} : \Delta \tensor{\epsilon}^{p}  = \Delta \lambda \sum_{A = 1}^{3} \sigma_{A} \widehat{g}_{A} \geq 0 .
\end{equation}
Notice that the stress beyond the initial yielding point satisfies the yield function $f=0$. As a result, this inequality can be recast as an additional term for the loss function that trains the yield function (Eq. \eqref{eq:yield_h1_loss_lode}) such that 
\begin{equation}
w_{\text{nnp}} \; \text{sign}( -\Delta \lambda \sum_{A = 1}^{3} \sigma_{A} \widehat{g}_{A}),
 \label{eq:convexity_penalty_flow}
\end{equation}
where $w_{\text{nnp}}$ is the penalty parameter. Notice that when the non-negative plastic work is fulfilled during the training of neural network, the penalty term would not be activated and will not affect the back-propagation step. Furthermore, if the yield function is convex and the flow rule is associative, this constraint is always fulfilled and, hence, not necessary. This constraint, however, should be helpful to regulate the relationships of the yield function and plastic flow when we intend to train the plastic flow direction independent of the stress gradient of the yield function.

\section{Implementation highlights: return mapping algorithm with automatic differentiation}
\label{sec:return_mapping_algorithm}
Here, we provide a review of the implementation of a fully implicit 
stress integration algorithm used for the proposed Hamilton-Jacobi hardening framework. For isotropic materials where the elastic strain and stress are co-axial, the stress integration can be done via spectral decomposition as shown in Alg. \ref{return_map_algorithm}. An upshot of the proposed method is that there is only a small modification necessary
to incorporate the Hamilton-Jacobi hardening and the generalized plasticity. 

\begin{algorithm}
	\caption{Return mapping algorithm in strain-space in principal axes for an isotropic hyperelastic-plastic model}\label{return_map_algorithm}
	\begin{algorithmic}[1]
		\State Compute $\tensor{\epsilon}_{n+1}^{\mathrm{e \, tr}}=\tensor{\epsilon}_{n}^{\mathrm{e}}+\Delta \tensor{\epsilon}$.
		\State Spectrally decompose $\tensor{\epsilon}_{n+1}^{\mathrm{e \, tr}}=\sum_{A=1}^{3} \epsilon_{A}^{\mathrm{e \, tr}} \tensor{n}^{\operatorname{tr}(A)} \otimes \tensor{n}^{\operatorname{tr}(A)}$.
		\State Compute $\sigma_{A}^{\mathrm{tr}}=\partial \widehat{\psi}^{\mathrm{e}} / \partial \epsilon_{A}^{\mathrm{e}}$ at  $\epsilon_{n+1}^{\mathrm{e \, tr}}$.
		\If{$\widehat{f}\left(\sigma_{1}^{\operatorname{tr}}, \sigma_{2}^{\operatorname{tr}}, \sigma_{3}^{\operatorname{tr}}, \xi_{n}\right) \leq 0 $}
			\State Set $\tensor{\sigma}_{n+1}=\sum_{A=1}^{3} \sigma_{A}^{\operatorname{tr}} \tensor{n}^{\operatorname{tr}(A)} \otimes \tensor{n}^{\operatorname{tr}(A)}$ and exit.
		\Else
			\State Solve for $\epsilon_{1}^{\mathrm{e}}, \epsilon_{2}^{\mathrm{e}}, \epsilon_{3}^{\mathrm{e}}$, and $\xi_{n+1}$ such that $\widehat{f}\left(\sigma_{1}^{\operatorname{tr}}, \sigma_{2}^{\operatorname{tr}}, \sigma_{3}^{\operatorname{tr}}, \xi_{n+1}\right) = 0 $.
			\State Compute $ \tensor{\sigma}_{n+1}=\sum_{A=1}^{3}\left(\partial \widehat{\psi}^{\mathrm{e}} / \partial \epsilon_{A}^{\mathrm{e}}\right) \tensor{n}^{\operatorname{tr}(A)} \otimes \tensor{n}^{\operatorname{tr}(A)}$ and exit.
		\EndIf	
	\end{algorithmic}
\caption{Return mapping algorithm with machine learning Hamilton-Jacobi hardening and generalized plasticity. }
\end{algorithm}

In this current work -- unless otherwise stated, all the necessary information for the return mapping algorithm about the elastic and plastic and constitutive responses is derived from the trained neural networks of the hyperelastic energy functional and yield function respectively using the Keras \citep{chollet2015keras} and Tensorflow \citep{abadi2016tensorflow} libraries. No additional explicit forms of constitutive laws are defined. Furthermore, the algorithm requires that all the strain and stress variables are in the principal axes. However, as it was stated in Section~\ref{sec:framework}, in order to facilitate the machine learning algorithms, we have opted to train with the strain invariants $\epsilon _v^{\mathrm{e}}$ and $\epsilon _s^{\mathrm{e}}$, and the stress invariants $\rho$ and $\theta$. Integrating the machine learning algorithms with the return mapping algorithm requires a set of coordinate system transformations, which, in turn, require the calculation of the partial derivatives of said transformations to use in the chain rule formulation. The partial derivative calculation is performed using the Autograd library \citep{maclaurin2015autograd} for automatic differentiation. 

Autograd enables the automatic calculation of the partial derivatives of explicitly defined functions. Thus, we can easily define the transformation of any input parameter space for our neural networks to the principal space and readily have the necessary partial derivatives for the chain rule implementation. This allows to use equivalent expressions of our neural network approximators $\widehat{\psi}^{\mathrm{e}}(\epsilon _v^{\mathrm{e}},\epsilon _s^{\mathrm{e}})$ and $\widehat{f}\left(\rho, \theta, \xi \right)$ in the principal space, such that:

\begin{equation}
\widehat{\psi}^{\mathrm{e}}(\epsilon _v^{\mathrm{e}},\epsilon _s^{\mathrm{e}})=\widehat{\psi}_{\text{principal}}^{\tensor{\epsilon}}\left(\epsilon_{1}^{\mathrm{e}}, \epsilon_{2}^{\mathrm{e}}, \epsilon_{3}^{\mathrm{e}}\right) \qquad \text{and } \qquad
\widehat{f}\left(\rho, \theta, \xi \right)=\widehat{f}_{\text{principal}}\left(\epsilon_{1}^{\mathrm{e}}, \epsilon_{2}^{\mathrm{e}}, \epsilon_{3}^{\mathrm{e}}, \xi\right).
\label{eq:approx_in_strain_space}
\end{equation}

In this work, integrating the neural network approximators in the return mapping requires the following coordinate system transformations $(\epsilon _v^{\mathrm{e}},\epsilon _s^{\mathrm{e}}) \longleftrightarrow (\epsilon_{1}^{\mathrm{e}}, \epsilon_{2}^{\mathrm{e}}, \epsilon_{3}^{\mathrm{e}})$,  $\left(\rho, \theta \right)  \longleftrightarrow (\sigma_1, \sigma_2, \sigma_3)$, $\left(\sigma_{1}^{\prime \prime} , \sigma_{2}^{\prime \prime}\right)  \longleftrightarrow (\sigma_1, \sigma_2, \sigma_3)$, and $\left(\sigma_{1}^{\prime \prime} , \sigma_{2}^{\prime \prime}\right)  \longleftrightarrow (\rho, \theta )$. These transformations require a large number of chain rules increasing the possibility of formulation errors, as well as rendering replacing the networks' input space less flexible. Thus, we opt for the automation of this process using Autograd.

Due to the fact that the machine learning training has created 
a mapping that automatically generates an updated yield function whenever the internal variables $\xi$ are updated, there is no need to add additional constraints for the linearized hardening rules. The return mapping algorithm can be described with a system of four equations that are solved iteratively. 
For a local iteration $k$, we solve for the solution vector $\tensor{x}$ such that $\boldsymbol{A}^{k} \cdot \Delta \boldsymbol{x}=\boldsymbol{r}\left(\boldsymbol{x}^{k}\right), \quad \boldsymbol{x}^{k+1} \leftarrow \boldsymbol{x}^{k}-\Delta \boldsymbol{x}, \quad k \leftarrow k+1$ until the residual norm $\|\boldsymbol{r}\|$ is below a set error threshold. The residual vector $\tensor{r}$ and the local tangent $\tensor{A}^{k}$ can be assembled for the calculation of $\tensor{x}$ by a series of neural network evaluations and automatic differentiations, such that:

\begin{align}
\tensor{r}(\tensor{x})&=\left\{\begin{array}{c}
\epsilon_{1}^{\mathrm{e}}-\epsilon_{1}^{\mathrm{etr}}+\Delta \lambda \widehat{g}_{1} \\
\epsilon_{2}^{\mathrm{e}}-\epsilon_{2}^{\mathrm{etr}}+\Delta \lambda \widehat{g}_{2} \\
\epsilon_{3}^{\mathrm{e}}-\epsilon_{3}^{\mathrm{etr}}+\Delta \lambda \widehat{g}_{3} \\
\widehat{f}\left(\epsilon_{1}^{\mathrm{e}}, \epsilon_{2}^{\mathrm{e}}, \epsilon_{3}^{\mathrm{e}}, \xi\right)
\end{array}\right\} , \quad 
\tensor{A}^{k}=\boldsymbol{r}^{\prime}\left(\tensor{x}^{k}\right)=\left[\begin{array}{cccc}
c_{11} & c_{12} & c_{13} & \widehat{g}_{1} \\
c_{21} & c_{22} & c_{23} & \widehat{g}_{2} \\
c_{31} & c_{32} & c_{33} & \widehat{g}_{3} \\
\partial \widehat{f} / \partial \epsilon_{1}^{\mathrm{e}} & \partial \widehat{f} / \partial \epsilon_{2}^{\mathrm{e}} &  \partial \widehat{f} / \partial \epsilon_{3}^{\mathrm{e}} & \partial \widehat{f} / \partial \xi
\end{array}\right], \notag \\
\text{ and }
\tensor{x}&=\left\{\begin{array}{c}
\epsilon_{1}^{\mathrm{e}} \\
\epsilon_{2}^{\mathrm{e}} \\
\epsilon_{3}^{\mathrm{e}} \\
\Delta \lambda
\end{array}\right\},
\label{eq:equation_system}
\end{align}
where $\epsilon_{I}^{\mathrm{e \, tr}}$ is the trial state principal strain, $\widehat{g}_I = \partial \widehat{f} / \partial \sigma_{I}$ for an associative flow rule and:
\begin{equation}
c_{I J}=\delta_{I J}+\Delta \lambda \frac{\partial \widehat{g}_{I}}{\partial \epsilon_{J}^{\mathrm{e}}}, \quad I, J=1,2,3 .
\label{eq:local_cij}
\end{equation}

This framework is also readily available to implement in finite element simulations (Section~\ref{sec:computational_examples}). We can assemble the algorithmic consistent tangent $\tensor{c}_{n+1}$ in principal axes for a global Newton iteration $n$:

\begin{equation}
\tensor{c}_{n+1} =\sum_{A=1}^{3} \sum_{B=1}^{3} a_{A B} \tensor{m}^{(A)} \otimes \tensor{m}^{(B)} 
+\frac{1}{2} \sum_{A=1}^{3} \sum_{B \neq A}\left(\frac{\sigma_{B}-\sigma_{A}}{\epsilon_{B}^{\mathrm{etr}}-\epsilon_{A}^{\mathrm{etr}}}\right)\left(\boldsymbol{m}^{(A B)} \otimes \tensor{m}^{(A B)}+\tensor{m}^{(A B)} \otimes \tensor{m}^{(B A)}\right), 
\label{eq:tangent_spectral}
\end{equation}
where $\tensor{m}^{(A B)} = \tensor{n}^{(A)} \otimes \tensor{n}^{(B)}$t he matrix of elastic moduli in principal axes is given as:

\begin{equation}
a_{A B}:=\frac{\partial \sigma_{A}}{\partial \epsilon_{B}^{\mathrm{etr}}}=\sum_{C=1}^{3}\left(\frac{\partial^{2} \widehat{\psi}^{\mathrm{e}}}{\partial \epsilon_{A}^{\mathrm{e}} \partial \epsilon_{C}^{\mathrm{e}}}\right) \frac{\partial \epsilon_{C}^{\mathrm{e}}}{\partial \epsilon_{B}^{\mathrm{etr}}}.
\label{eq:local_alpha_matrix}
\end{equation}

Utilizing Tensorflow and Autograd, the return mapping algorithm is fully generalized for any isotropic hyperelastic and yield function data-driven constitutive laws. It also allows for quick implementation of any parametrization of the neural network architectures. In future work, the framework can be extended to accommodate anisotropic responses, as well as architectures with complex internal variables and higher descriptive power.

\section{Alternative comparison models for control experiments} 
\label{sec:comparison_models}

In this section, we will briefly review some simple black-box neural network architectures that can be employed to predict the path-dependent plasticity behaviors. The predictive capabilities of these behaviors will be compared to our neural network elastoplasticity framework in Section~\ref{sec:comparison_black_box}. Three different architectures will be designed for comparison with our framework: a multi-step feed-forward network, a recurrent GRU network, and a 1-D convolutional network. All of these networks demonstrate the ability to capture path-dependent behavior utilizing different memory mechanisms.

\begin{table}[h]
{\small
\centering
{
\begin{tabular}{p{.1\textwidth}p{.8\textwidth}}
\hline\noalign{\smallskip}
Model & Description\\
\noalign{\smallskip}\hline\\[-3mm]
${\cal M}_{\text{stepDense}}$ & \textbf{Dense} (100 neurons / ReLU) $\rightarrow$ \textbf{Dense} (100 neurons / ReLU) $\rightarrow$ \textbf{Dense} (100 neurons / ReLU) $\rightarrow$ Output \textbf{Dense} (Linear) \\[4mm]
 & \\
${\cal M}_{\text{GRU}}$ & \textbf{GRU} (32 units / tanh) $\rightarrow$ \textbf{GRU} (32 units / tanh) $\rightarrow$ \textbf{Dense} (100 neurons / ReLU) $\rightarrow$ \textbf{Dense} (100 neurons / ReLU) $\rightarrow$ Output \textbf{Dense} (Linear)\\[4mm]
 & \\
${\cal M}_{\text{Conv1D}}$&  \textbf{Conv1D} (32 filters / ReLU) $\rightarrow$ \textbf{Conv1D} (64 filters / ReLU) $\rightarrow$ \textbf{Conv1D} (128 filters / ReLU) $\rightarrow$ \textbf{Flatten} $\rightarrow$ \textbf{Dense} (100 neurons / ReLU) $\rightarrow$ \textbf{Dense} (100 neurons / ReLU) $\rightarrow$ Output \textbf{Dense} (Linear)\\
\noalign{\smallskip}\hline
\end{tabular}
}}
\caption{Summary of black-box neural network architectures used for control experiments.}
\label{tab:model_architecture}       
\end{table}

The first architecture is a feed-forward network that learns from information of a previous time-step to predict the stress behavior of the current one. The feed-forward architecture consists of fully-connected Dense layers that have the following formulation in matrix form:

\begin{equation}
\tensor{h}_{\text {dense }}^{(l+1)}=a\left(\tensor{h}^{(l)} \tensor{W}^{(l)}+\tensor{b}^{(l)}\right),
\label{eq:dense_layer}
\end{equation}
where $\tensor{h}_{\text {dense }}^{(l+1)}$ is the output of the Dense layer, $\tensor{h}^{(l)}$ is the output of the previous layer $l$, $a$ is an activation function, $\tensor{W}^{(l)}$, $\tensor{b}^{(l)}$ are the trainable weight matrix  and bias vector of the layer respectively. It is noted that the layer formulation itself cannot hold any memory information. The memory of path-dependence in this architecture is derived from the input of the neural network. The input is the full strain tensor $\tensor{\epsilon}_n$ at time step $n$ and the full stress tensor $\tensor{\sigma}_{n-1}$ at the previous time step $(n-1)$, both in Voigt notation. The output prediction of the network is the stress tensor $\tensor{\sigma}_{n}$ at time step $n$. The network attempts to infer the path-dependent behavior by associating the previous stress state with the current one. The architecture consists of three Dense hidden layers (100 neurons each) with ReLU activation functions and the output Dense layer with a Linear activation function. 

The second architecture is a recurrent network that learns the path-dependent behavior in the form of time series. The architecture utilizes the Gated Recurrent Unit (GRU) layer formulation, a recurrent architecture introduced in \citep{cho2014learning} -- a variation of the popular Long Short Term Memory (LSTM) recurrent architecture \citep{gers1999learning}. The GRU cell controls memory information by utilizing three gates (an update gate, a reset gate, and a current memory gate), the formulation of which is omitted for brevity. The architecture input is a formatted as a timeseries of the input strain -- a training sample input is a time series of the strain tensors in Voigt notation with a history length of $\ell$. The variable $\ell$ is a network hyperparameter that is fine-tuned to give optimal results and it signifies the amount of information from the previous time steps that are taken into consideration to make a prediction for the current time step. Thus, a GRU network sample for time step $n$ has input the series of strain tensors $[ \tensor{\epsilon}_{n-\ell},...,\tensor{\epsilon}_{n-1},\tensor{\epsilon}_{n}]$ and output the stress tensor for the current step in Voigt notation $\tensor{\sigma}_n$. The architecture used in this work consists of two GRU hidden layers (32 recurrent units each) with a ReLU activation function, followed by two Dense layers (100 neurons) with ReLU activations and a Dense output layer with a linear activation function. The history variable was set to $\ell = 20$.

The last architecture we compare our framework learn the path-dependent information from time series by extracting features through a 1-D convolution filter. The convolution filter extracts higher-order from time series of fixed length and has be used for time-series predictions \citep{lecun1995convolutional} and audio processing \citep{oord2016wavenet}. The input of this architecture is he series of strain tensors $[ \tensor{\epsilon}_{n-\ell},...,\tensor{\epsilon}_{n-1},\tensor{\epsilon}_{n}]$ and output the stress tensor for the current step in Voigt notation $\tensor{\sigma}_n$. The 1D convolutional filter processes segments of the path-dependent time series data in a rolling window manner and has length equal to $\ell$. The architecture consists of three 1D convolution networks (32, 64, and 128 filters respectively) with ReLU activation functions. The output features of the last convolutional layer are flattened and then fed into two consecutive Dense layers (100 neurons) with ReLU activations, followed by a Dense output layer with a Linear activation function.

All the architectures were trained for 500 epochs with the Nadam optimizer with a batch size of 64. They were trained on different data sets to illustrate the comparisons with our elastoplasticity framework -- the data sets are described in the context of the numerical experiments in Section~\ref{sec:comparison_black_box}. The hyperparameters of these architectures were fine-tuned through trial and error in an effort to provide optimal results and a fair comparison with our elastoplasticity framework to the best of our knowledge. The three black-box architectures are summarized in Table~\ref{tab:model_architecture}.

\section{Numerical Experiments} 
\label{sec:numerical_experiments}

In this section, we report the results of numerical experiments we conducted to verify the implementation and evaluate the predictive capacity of the presented elastoplasticity ANN framework. 
For brevity, some background materials and simple verification exercise are placed in the Appendices. 
In Section \ref{sec:small_strain_sobolev}, we demonstrate how the training of the hyperelastic energy functional approximator can benefit by the use of higher-order activations function and higher-order Sobolev constraints. In Section~\ref{sec:yield_function_training}, we demonstrate the training of the yield function level set neural networks and their approximation of the evolving yield functions. In Section~\ref{sec:comparison_black_box}, we are comparing the three recurrent architectures of Section~\ref{sec:comparison_models} and our elastoplasticity framework as surrogate models for a polycrystal microstructure. Finally, in Section~\ref{sec:computational_examples}, we demonstrate the ability of our framework to integrate into a finite element simulation by fully replacing the elastic and plastic constitutive models with their data-driven counterparts.

\subsection{Benchmark study 1: Higher-order Sobolev training of hyperelastic energy functional}
\label{sec:small_strain_sobolev}

In this numerical experiment, we demonstrate the benefits of training of a neural network on hyperelastic energy functional data utilizing higher-order activation functions with higher-order Sobolev training objectives. The generation for the hyperelastic energy functional data sets are discussed in Appendix~\ref{sec:dataset_hyperelastic}.

The neural network models in this work are trained on two datasets of the energy functionals for linear elasticity and non-linear elasticity (Eq.~\eqref{eq:borja_hyperelastic_psi}) of 2500 sample points each. The points are sampled in a uniform grid, since there is no path-dependence in the elastic behavior and no strain history will need to be taken into consideration.

\begin{figure}[h!]
\newcommand\siz{.32\textwidth}
\centering

\begin{tabular}{M{.01\textwidth}M{.33\textwidth}M{.33\textwidth}M{.33\textwidth}}
\hspace{-2.1cm}(a) &
\hspace{-2.5cm}\includegraphics[width=.32\textwidth ,angle=0]{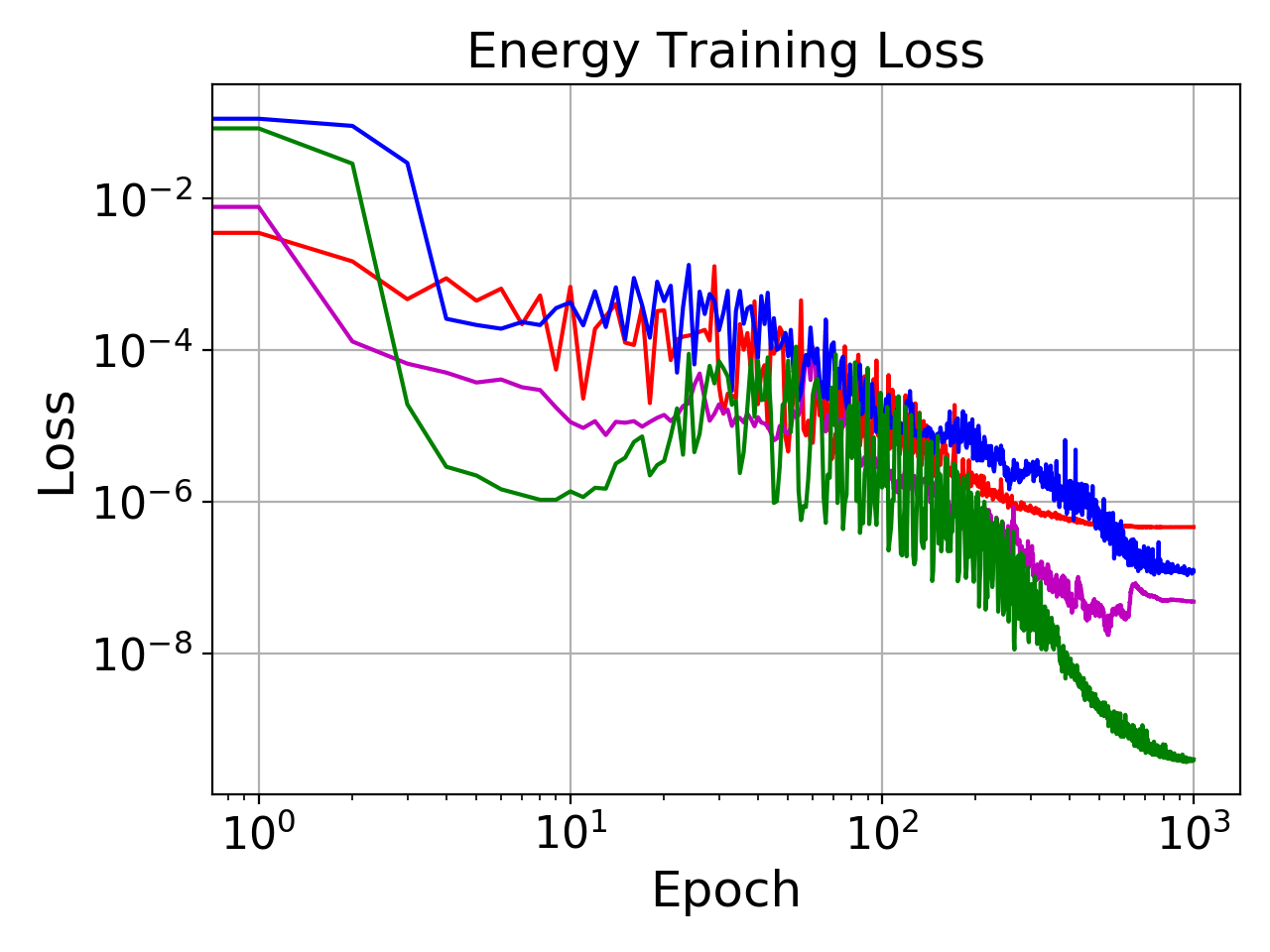} &
\hspace{-2.5cm}\includegraphics[width=.32\textwidth ,angle=0]{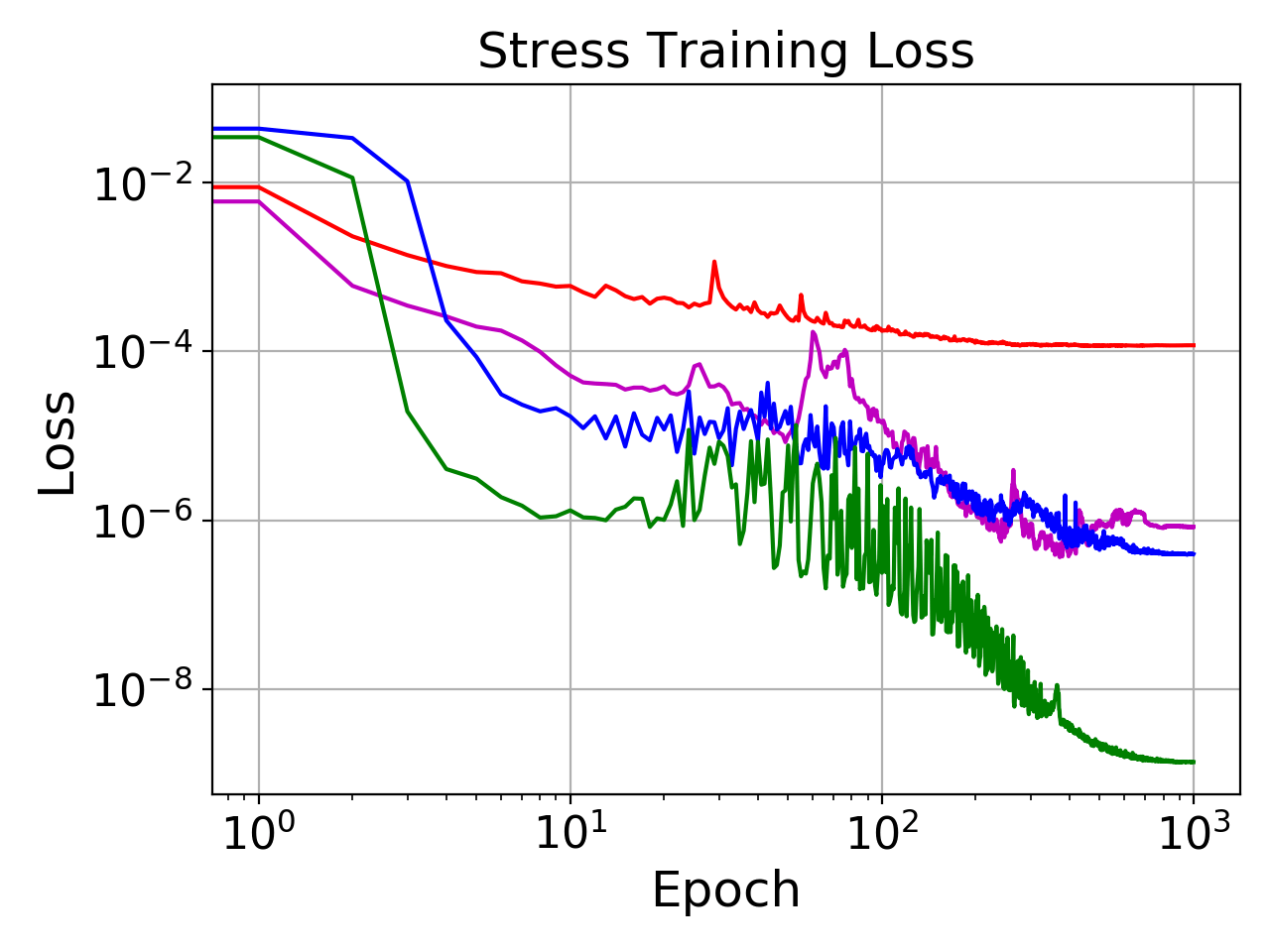} &
\hspace{-2.5cm}\includegraphics[width=.32\textwidth ,angle=0]{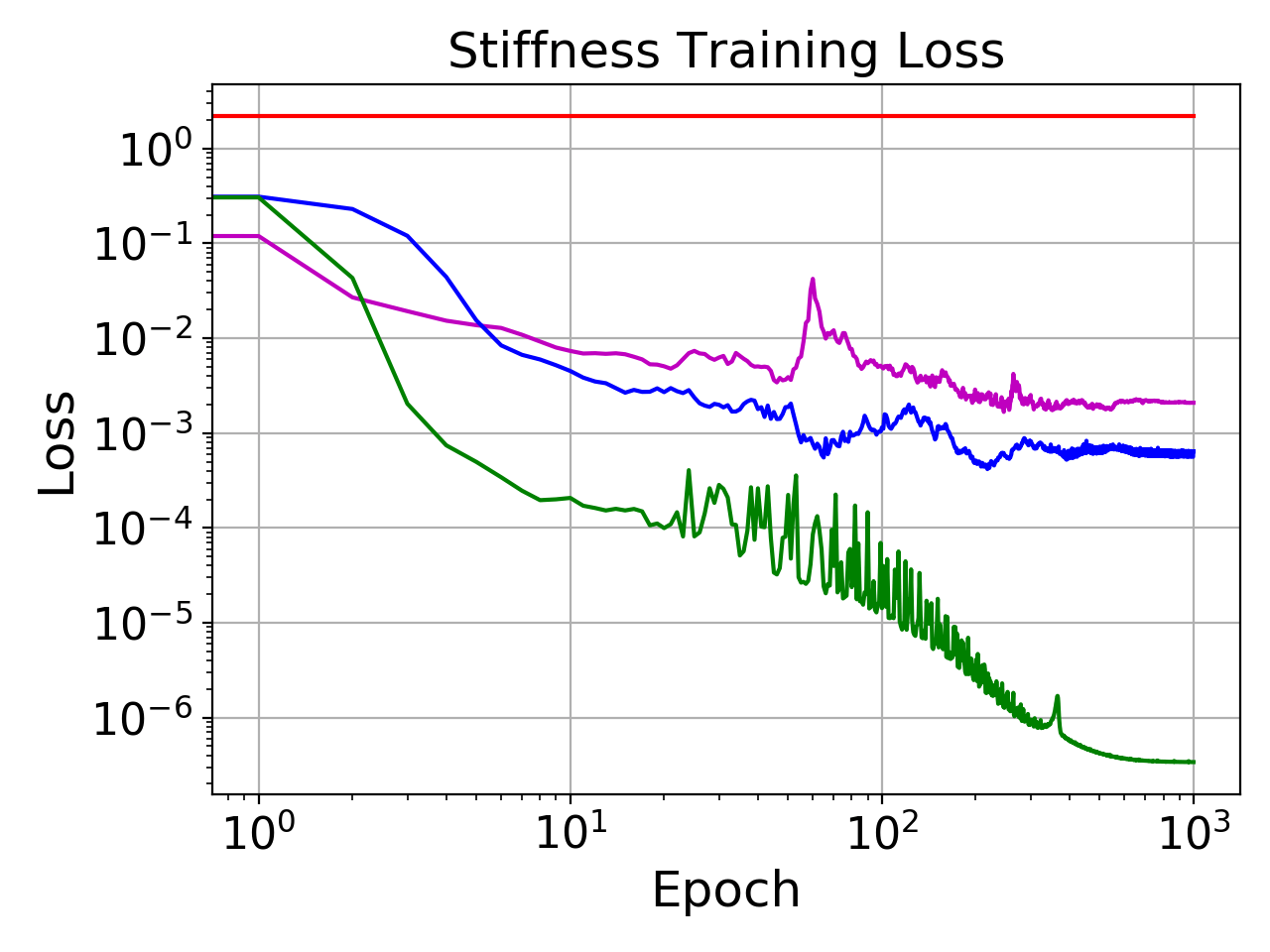} \\     

\hspace{-2.1cm}(b) &
\hspace{-2.5cm}\includegraphics[width=.32\textwidth ,angle=0]{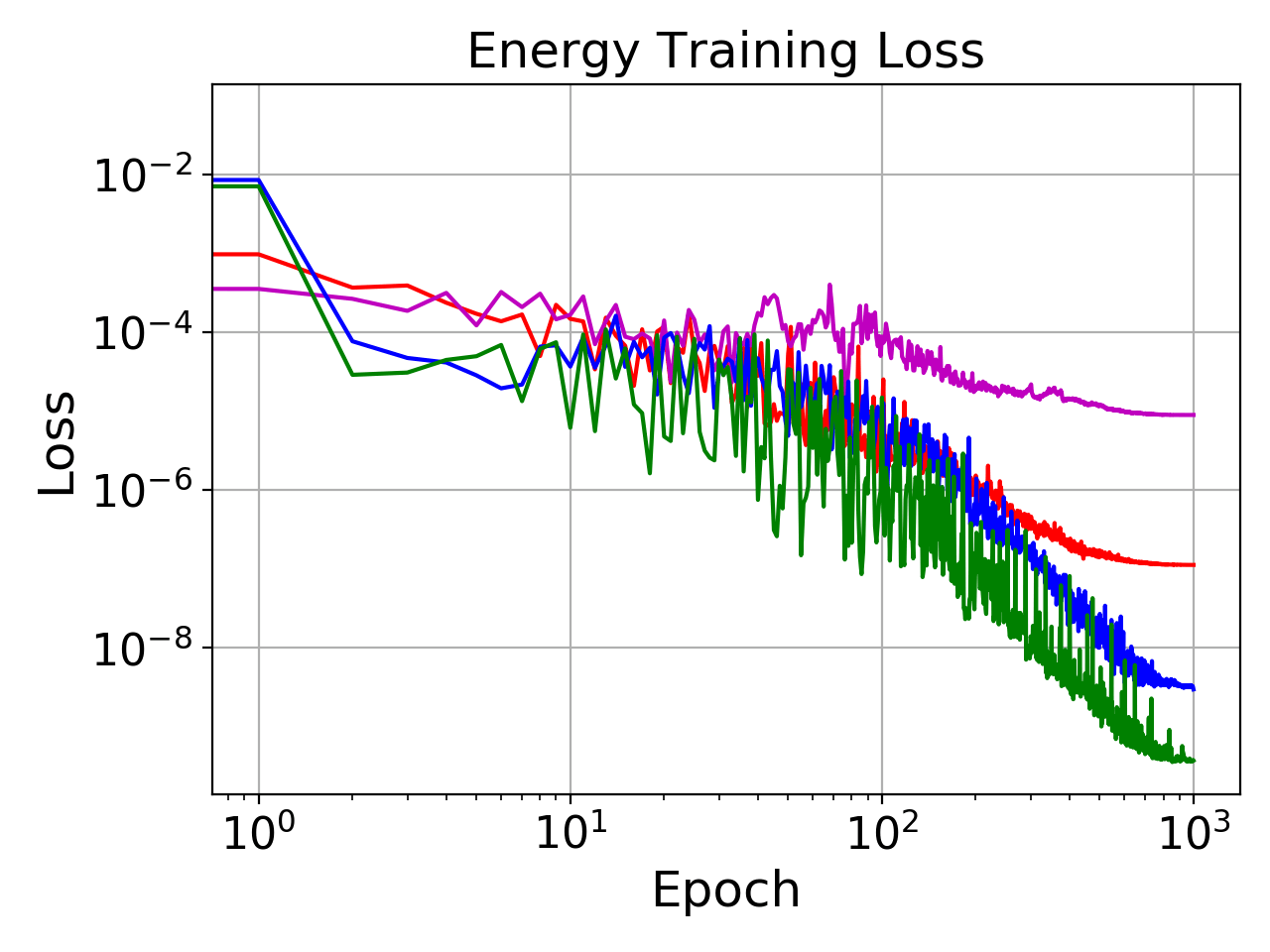} &
\hspace{-2.5cm}\includegraphics[width=.32\textwidth ,angle=0]{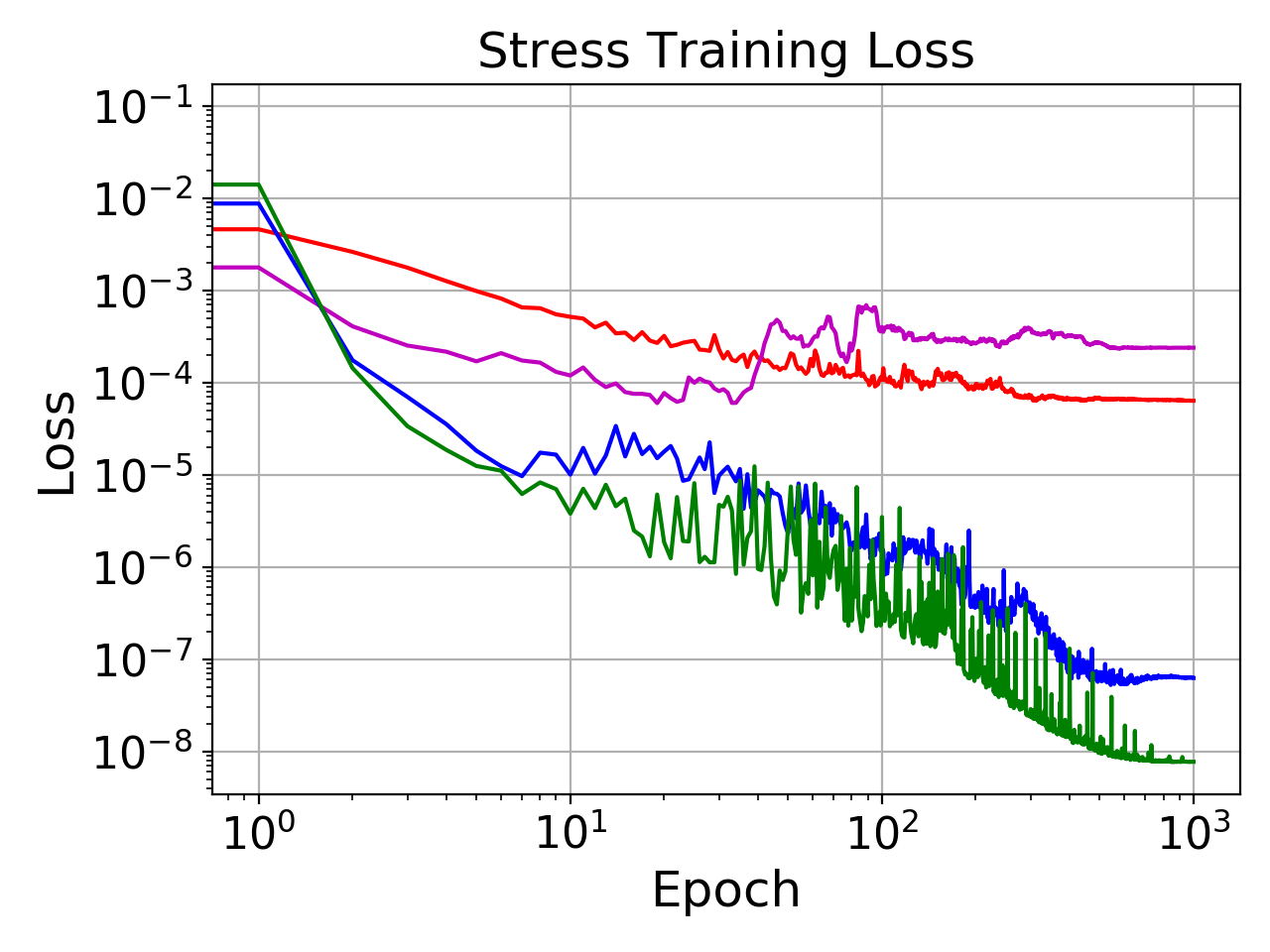} &
\hspace{-2.5cm}\includegraphics[width=.32\textwidth ,angle=0]{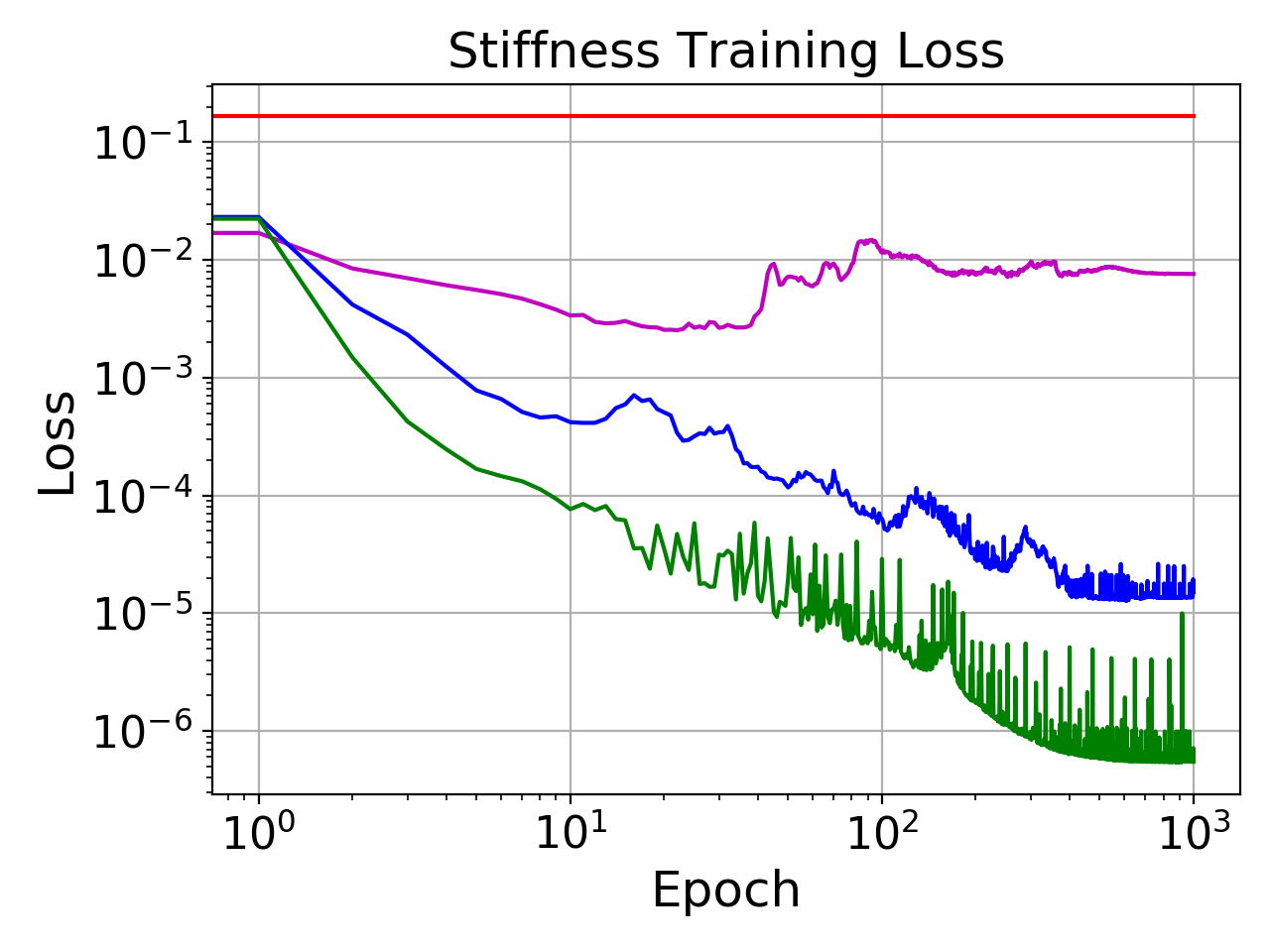} 

\end{tabular}
\includegraphics[width=.8\textwidth]{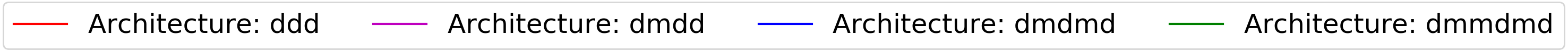} 

\caption{Training loss comparison of feed-forward architectures with a progressively larger number of Multiply layers with an $H^2$ training objective for (a) linear elasticity and (b) Modified Cam-Clay hyperelastic law \citep{borja2001cam}. As more non-linearity is introduced in the network architecture, the stiffness accuracy prediction increases - more control is allowed for the $H^2$ terms of the training objective.}
\label{fig:activation_function_comparison}
\end{figure}

In the first part of this numerical experiment, we investigate the ability of the feed-forward architecture to fulfill the higher-order Sobolev constraints. It is expected that architectures with only piece-wise linear activation functions will not be able to handle the higher-order $H^2$ constraints. We will be increasing the ability of the feed-forward architecture to capture higher-order non-linearity by progressively introducing Multiply layers. The different architectures tested are shown in Fig.~\ref{fig:multiply_layer_examples}. For brevity, the letters d and m in the names of architecture represent the Dense and Multiply layers respectively that form the architecture (e.g. architecture dmdd has the layer structure Dense $\rightarrow$ Multiply $\rightarrow$ Dense $\rightarrow$ Dense). The initial architecture has two hidden feed-forward Dense layers (100 neurons each) with ReLU activation functions and an output Dense layer with a Linear activation function (architecture: ddd in Fig.~\ref{fig:activation_function_comparison}). We progressively introduce more intermediate Multiply layers in the architecture. Specifically, we test for architectures with one, two, and three Multiply layers (arhitectures dmdd, dmdmd, and dmmdmd in Fig.~\ref{fig:multiply_layer_examples} respectively). Other than the number of intermediate Multiply layers, all the other hyperparameters are identical among all the architectures. The training objective used is the one resembling an $H^2$ norm, similar to Eq~\eqref{eq:energy_h2_loss}. All the models were trained for 1000 epochs with a batch size of 32 using the Nadam optimizer, set with default values \citep{dozat2016incorporating}.

The results of this numerical experiment can be seen in Fig.~\ref{fig:activation_function_comparison}. Increasing the non-linearity of the architecture progressively increases the accuracy of the captured $H^2$ constrained stiffness measure. Without any Multiply layers, the piece-wise linear architecture ddd cannot capture and improve the stiffness measure during training at all, as it was expected. The more non-linearity introduced the more control there is over the $H^2$ terms. An improvement in the energy and stress predictions is also observed with more allowed non-linearity which can be interpreted as the architectures ability to capture the non-linear energy response surfaces better and the $H^2$ norm training objective being fulfilled. It is highlighted that the number of Multiply layers is considered as another architecture hyperparameter to be tune as to improve accuracy of the predictions. The combination of different number of layers, activation functions, and Multiply layers should be specific for the data set approximated. It was also observed that excessively increasing the non-linearity of the architecture would lead the training procedure to diverge.

\begin{figure}[h!]
\newcommand\siz{.32\textwidth}
\centering
\begin{tabular}{M{.01\textwidth}M{.33\textwidth}M{.33\textwidth}M{.33\textwidth}}
\hspace{-2.1cm}(a) &
\hspace{-2.5cm}\includegraphics[width=.32\textwidth ,angle=0]{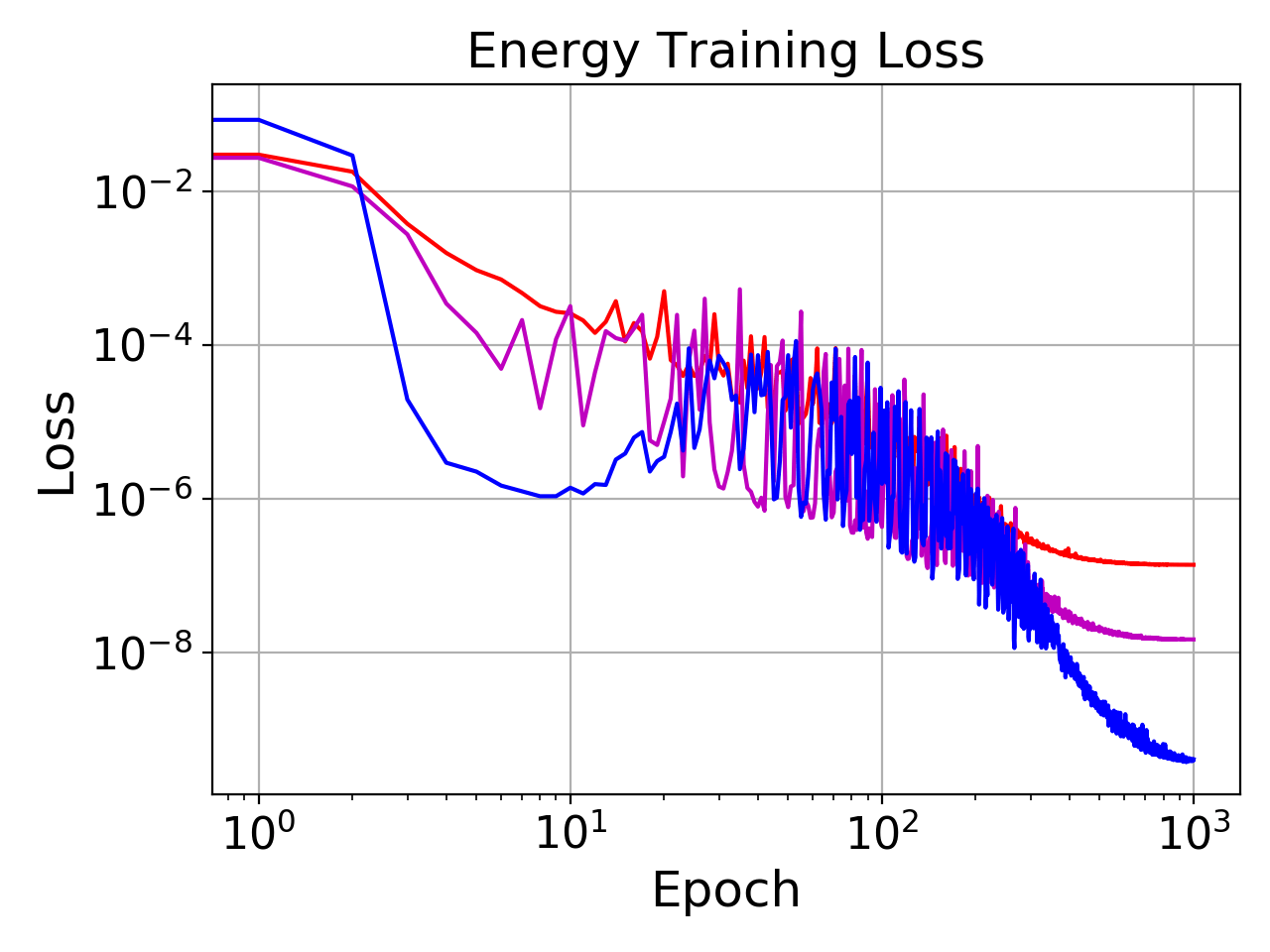} &
\hspace{-2.5cm}\includegraphics[width=.32\textwidth ,angle=0]{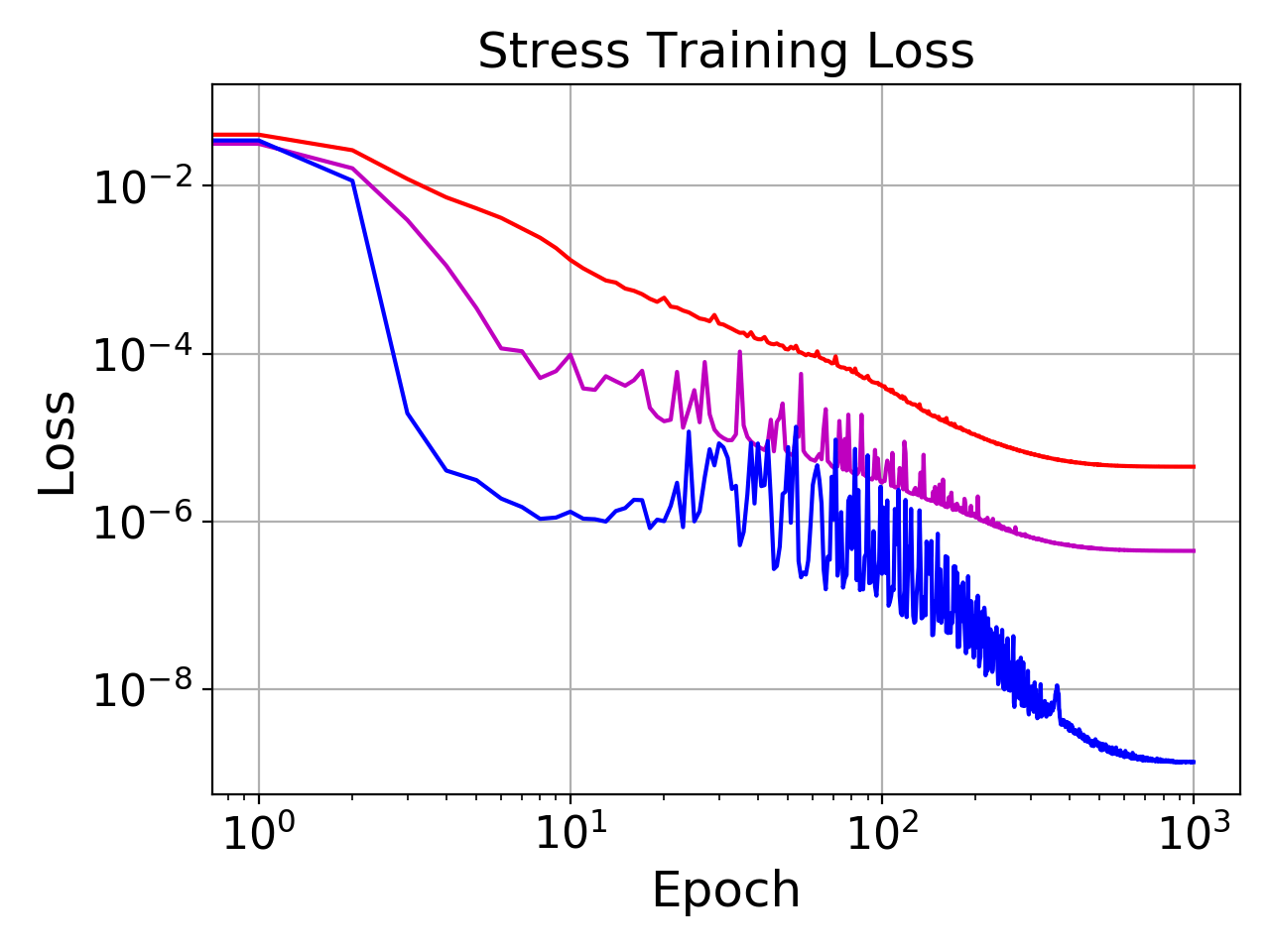} &
\hspace{-2.5cm}\includegraphics[width=.32\textwidth ,angle=0]{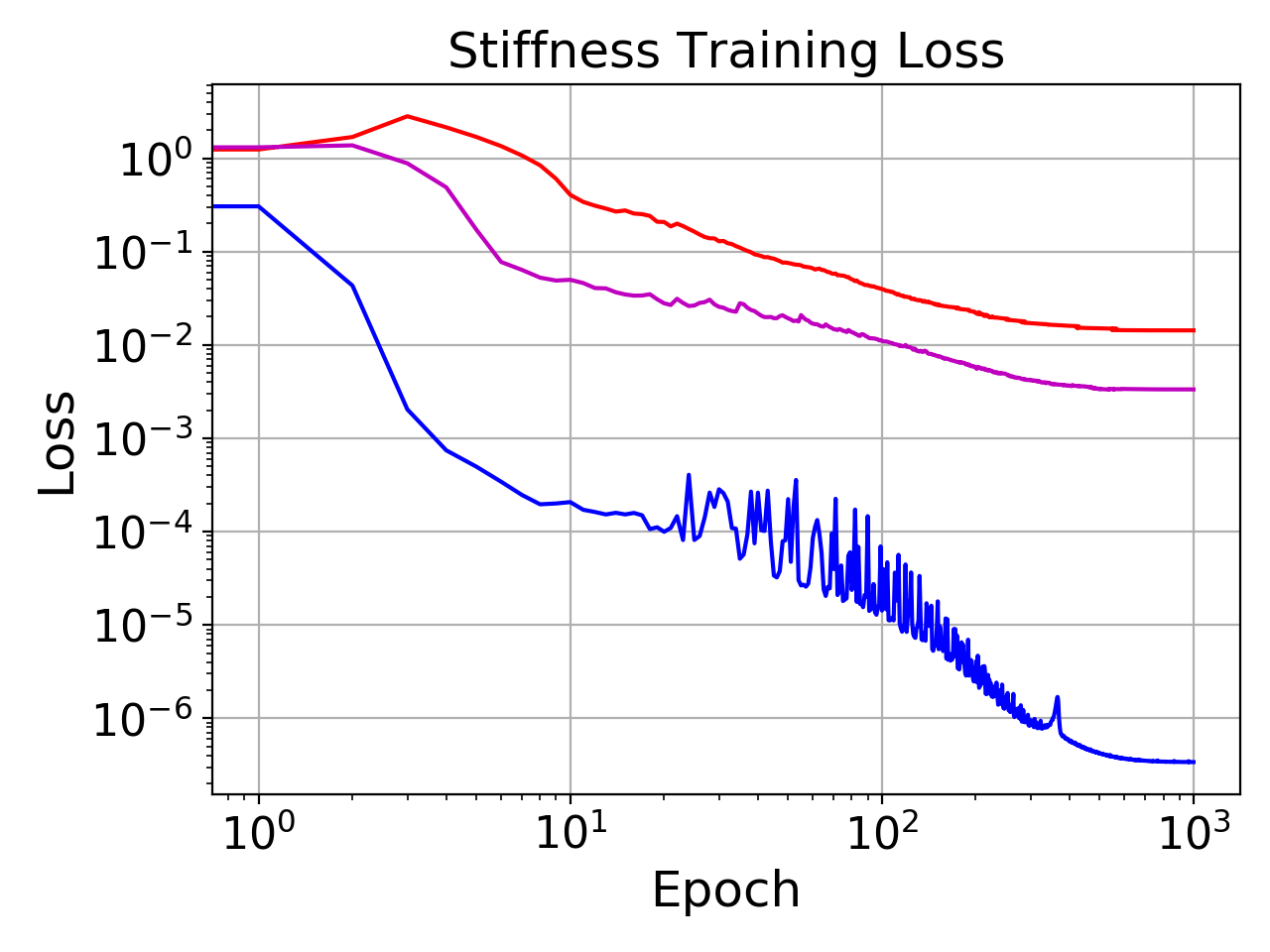} \\     

\hspace{-2.1cm}(b) &
\hspace{-2.5cm}\includegraphics[width=.32\textwidth ,angle=0]{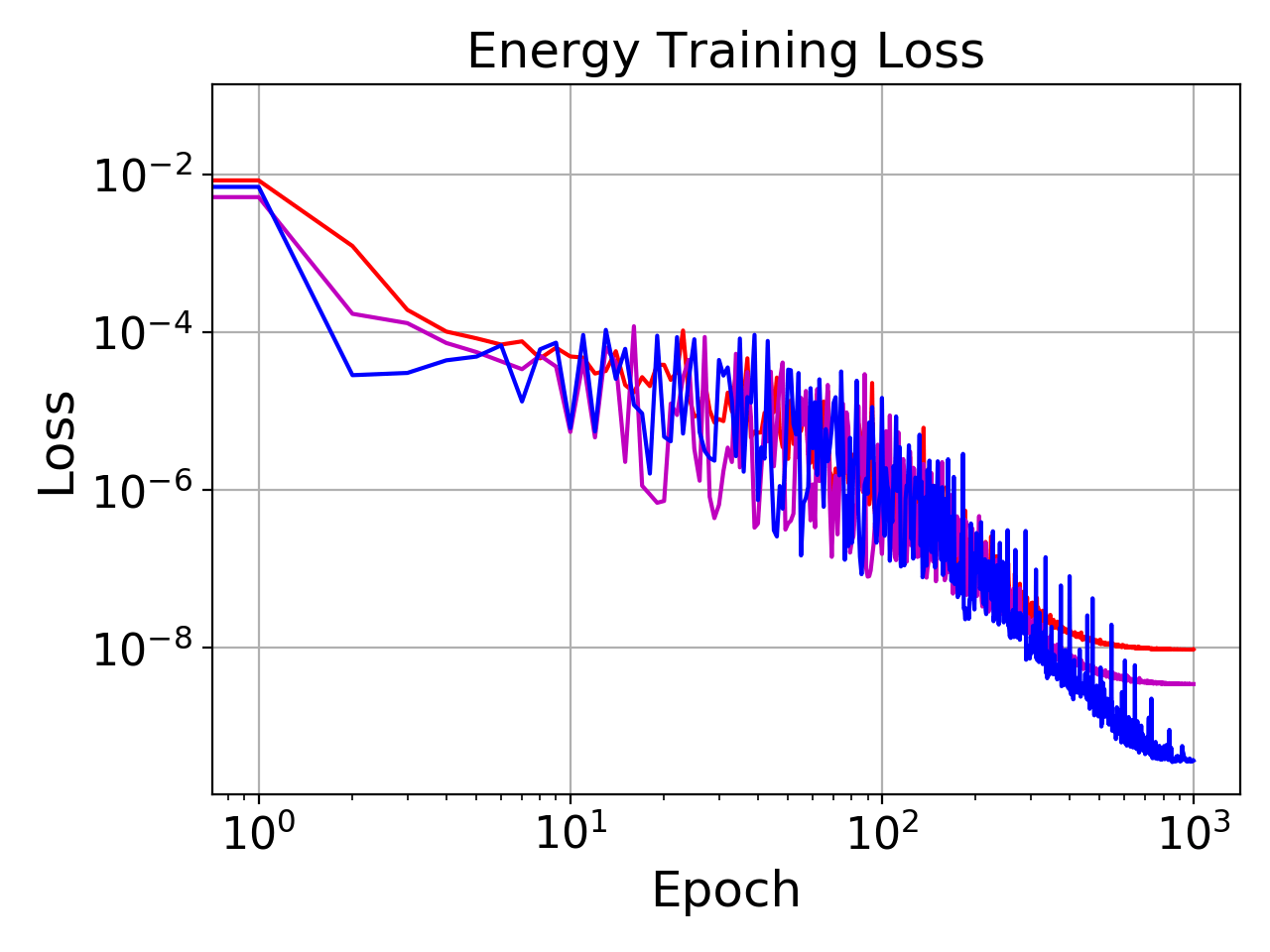} &
\hspace{-2.5cm}\includegraphics[width=.32\textwidth ,angle=0]{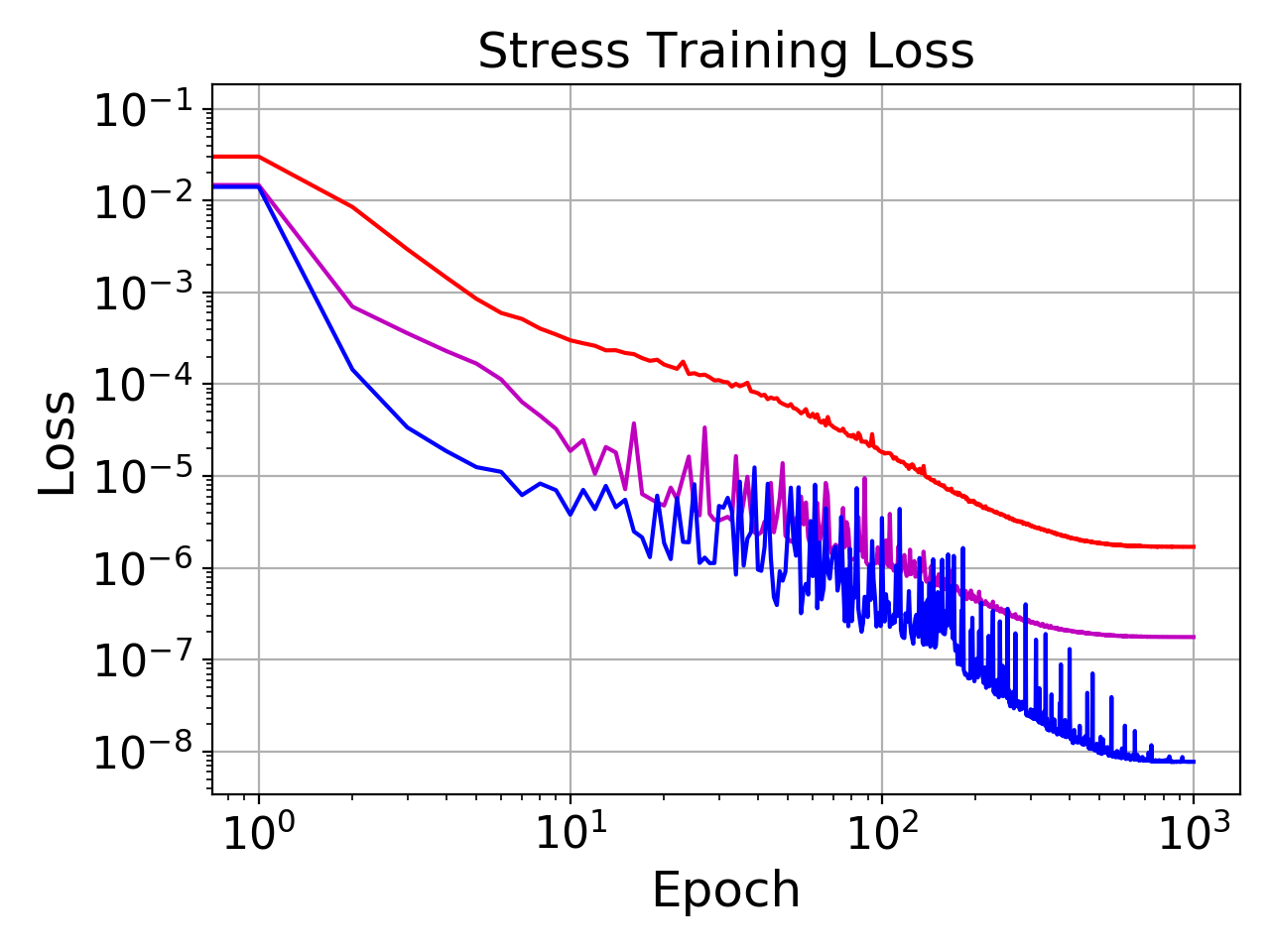} &
\hspace{-2.5cm}\includegraphics[width=.32\textwidth ,angle=0]{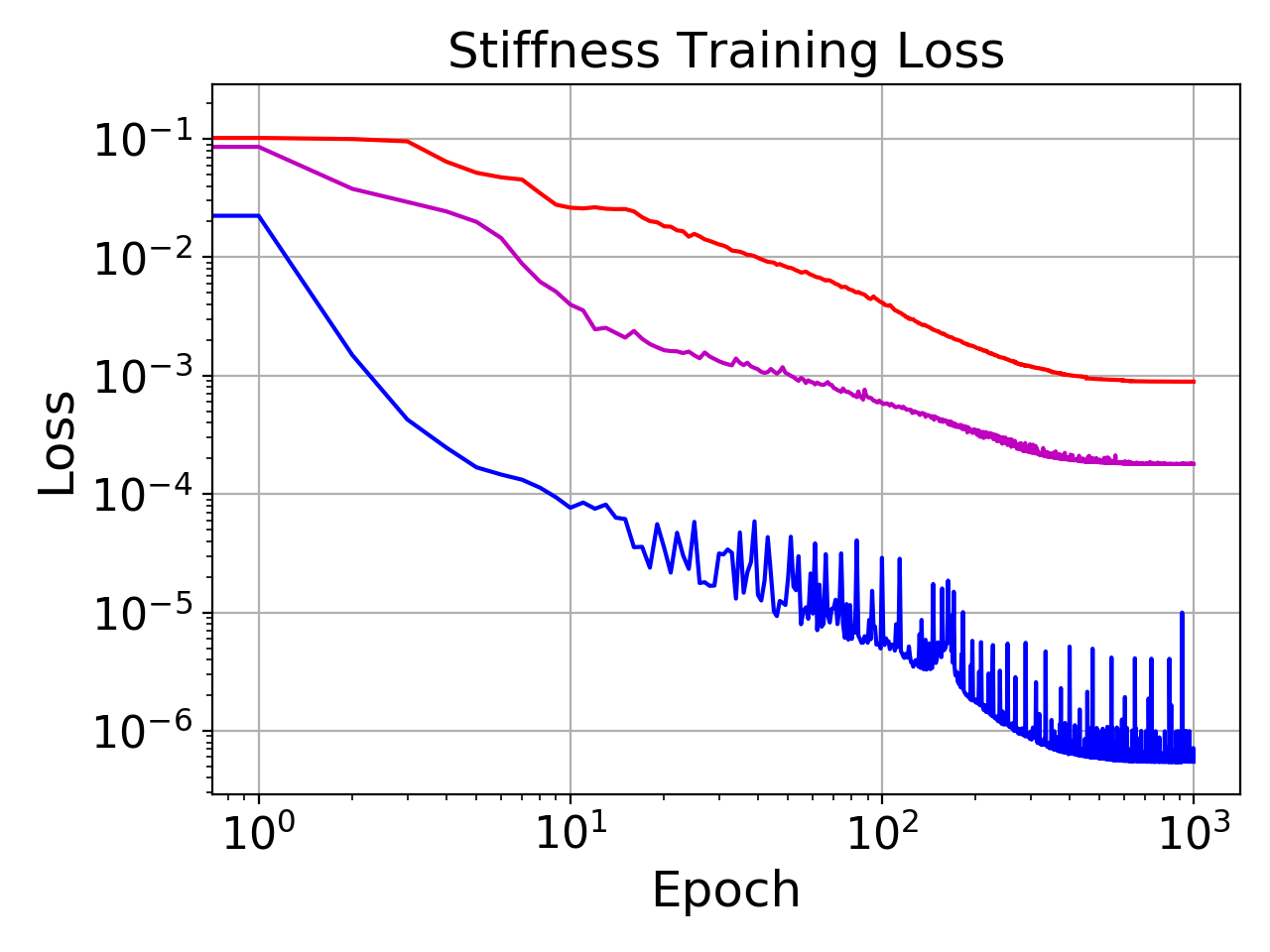} 

\end{tabular}
\includegraphics[width=.9\textwidth]{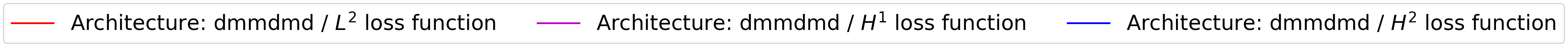} 

\caption{Training loss comparison for $L^2$, $H^1$,  and $H^2$ training objectives of an architecture with three multiply layers (dmmdmd) for (a) linear elasticity and (b) Modified Cam-Clay hyperelastic law \citep{borja2001cam}. The $H^2$ training objective procures more accurate results than the $L^2$ and $H^1$ objectives for all of the energy, stress, and stiffness fields.}
\label{fig:loss_function_comparison}
\end{figure}

In the second part of this numerical experiment, we investigate the predictive accuracy of the dmmdmd architecture (as shown in Fig.~\ref{fig:multiply_layer_examples}) trained using an $L^2$, an $H^1$, and $H^2$ norm-based training objective. For all the training procedures, all the other the architecture and training hyperparameters are identical to the ones used in the first part of the experiment. The results of these three training experiments can be seen in Fig.~\ref{fig:loss_function_comparison}. The predictive capability of the model increases when higher-order Sobolev training is utilized with the best overall scores procured for $H^2$ norm-based training. \cite{czarnecki_sobolev_2017} had observed that constraining the $H^1$ terms in the loss function improves the function value prediction accuracy. We are showing that by constraining the $H^2$ terms, we are improving both the prediction of the function values and the first-order derivatives along with the second-order derivatives of the function.

\begin{figure}[h!]
\newcommand\siz{.30\textwidth}
\centering

\begin{tabular}{M{.3\textwidth}M{.3\textwidth}M{.3\textwidth}}

\hspace{-1cm}\includegraphics[width=.30\textwidth ,angle=0]{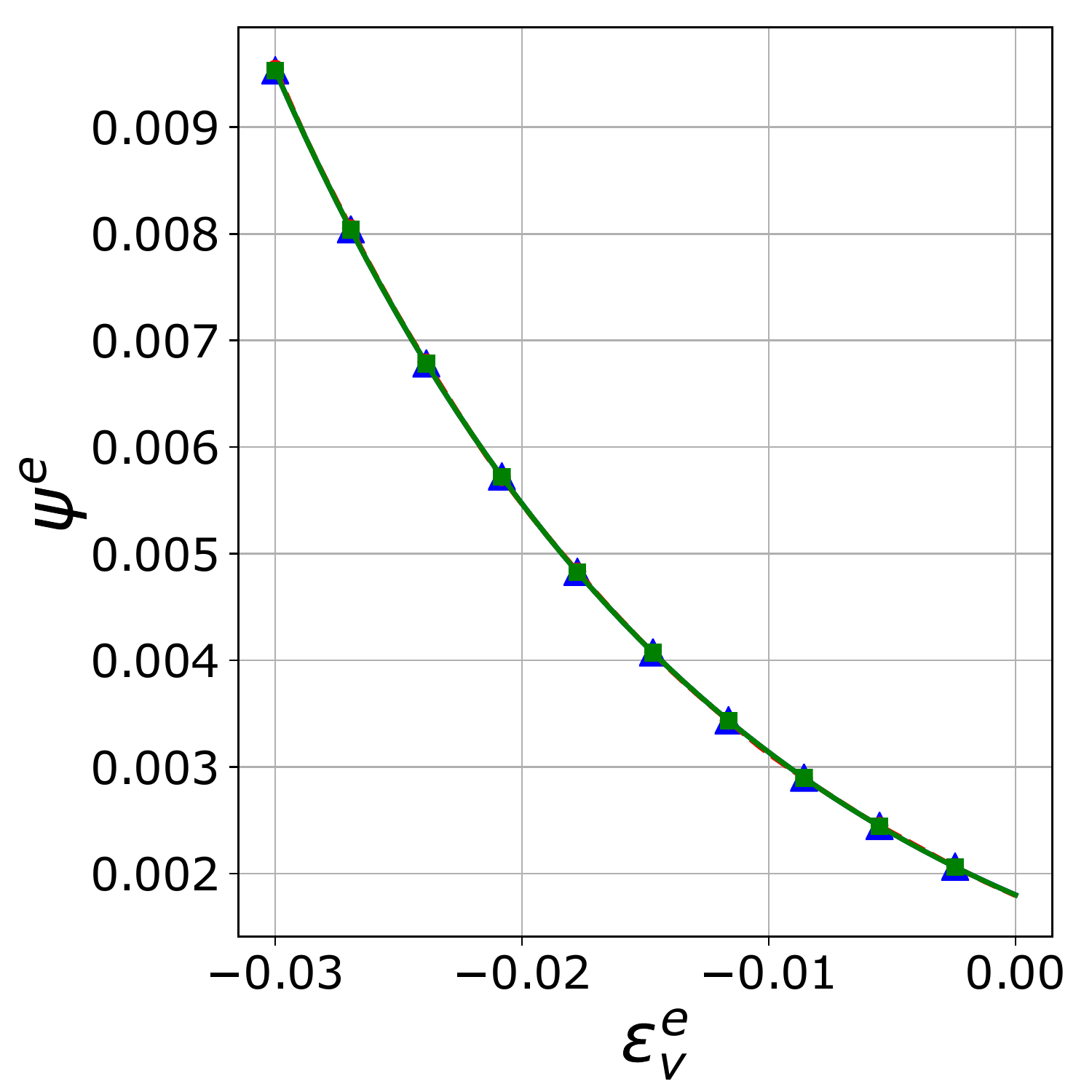} &
\hspace{-1cm}\includegraphics[width=.30\textwidth ,angle=0]{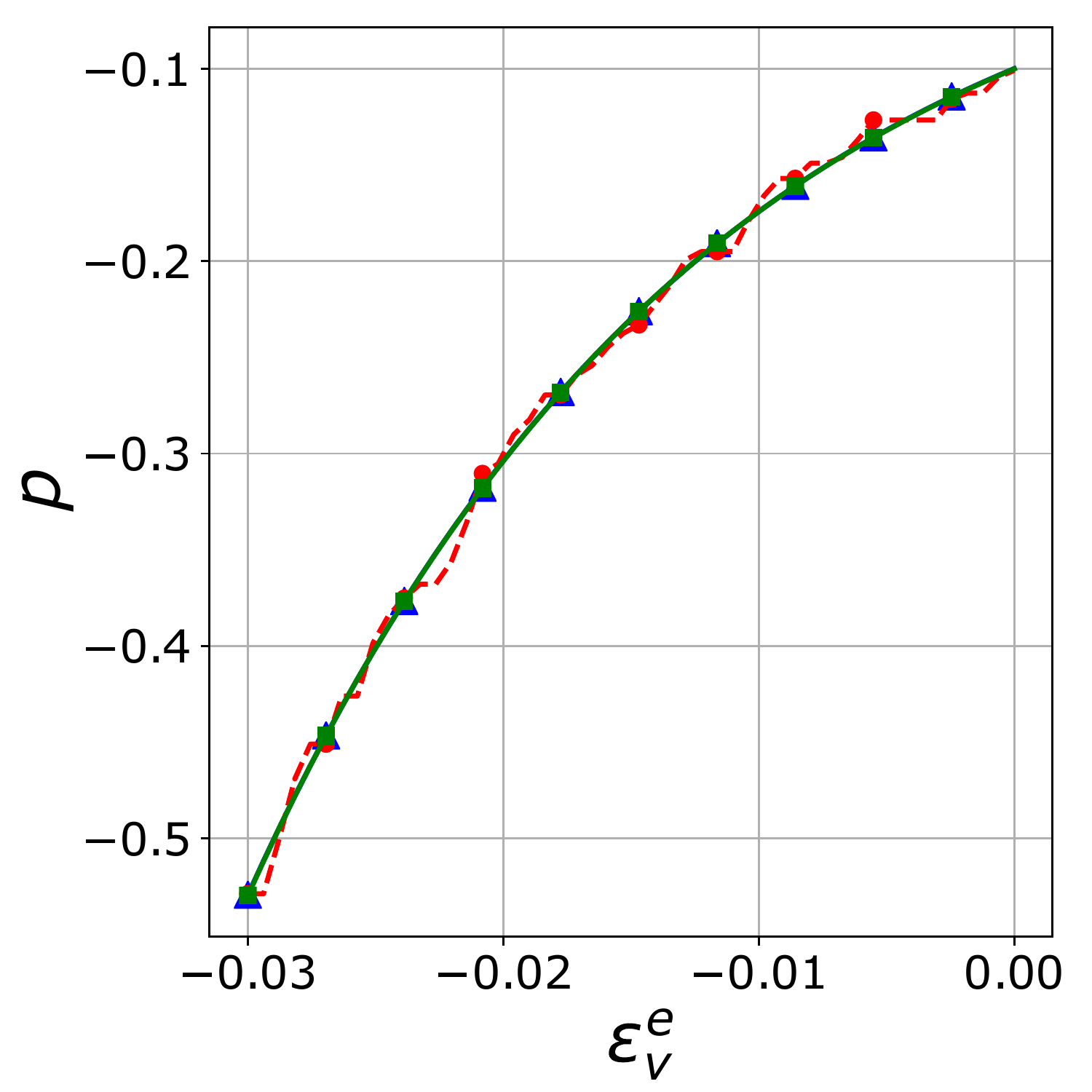} &
\hspace{-1cm}\includegraphics[width=.30\textwidth ,angle=0]{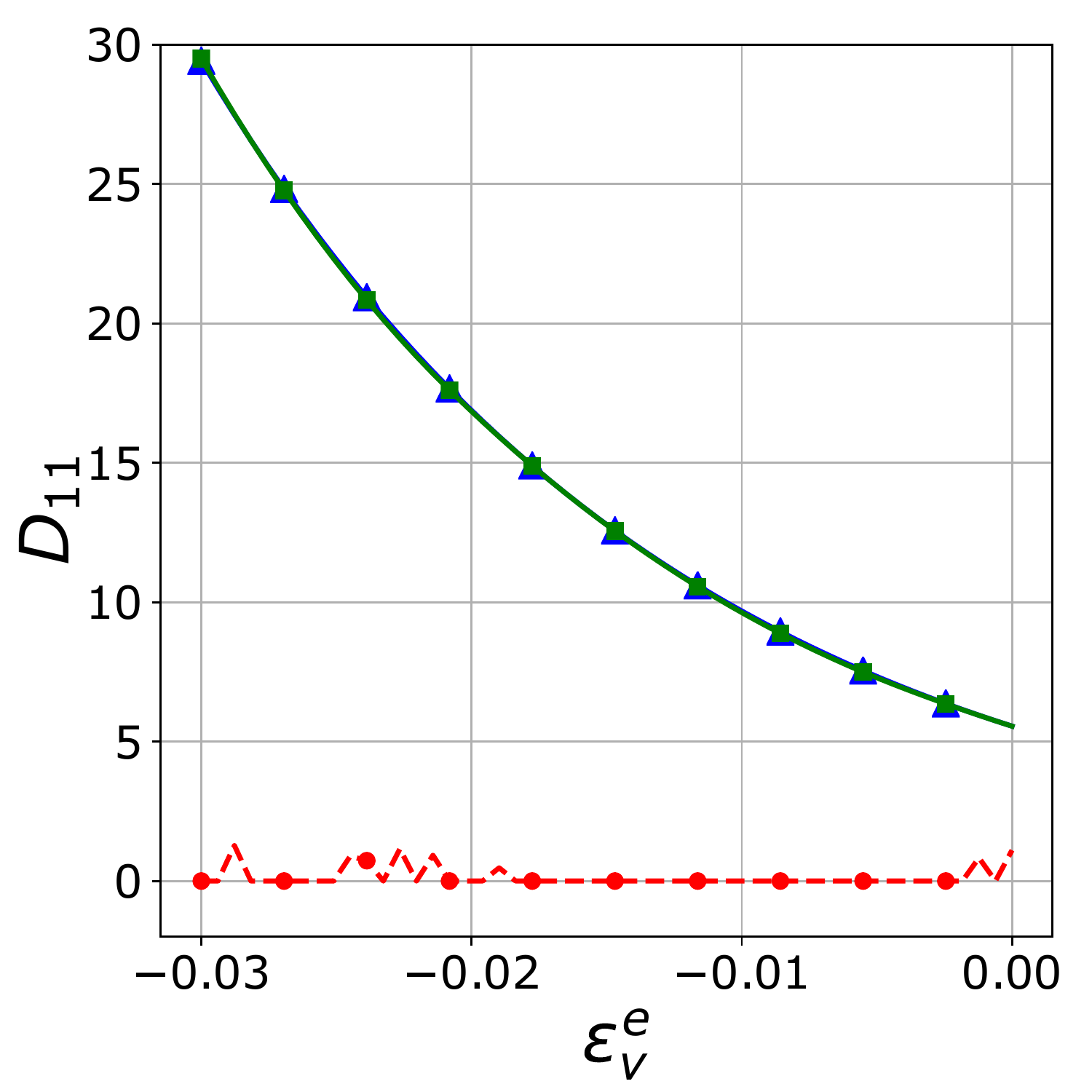} \\

\hspace{-1cm}\includegraphics[width=.30\textwidth ,angle=0]{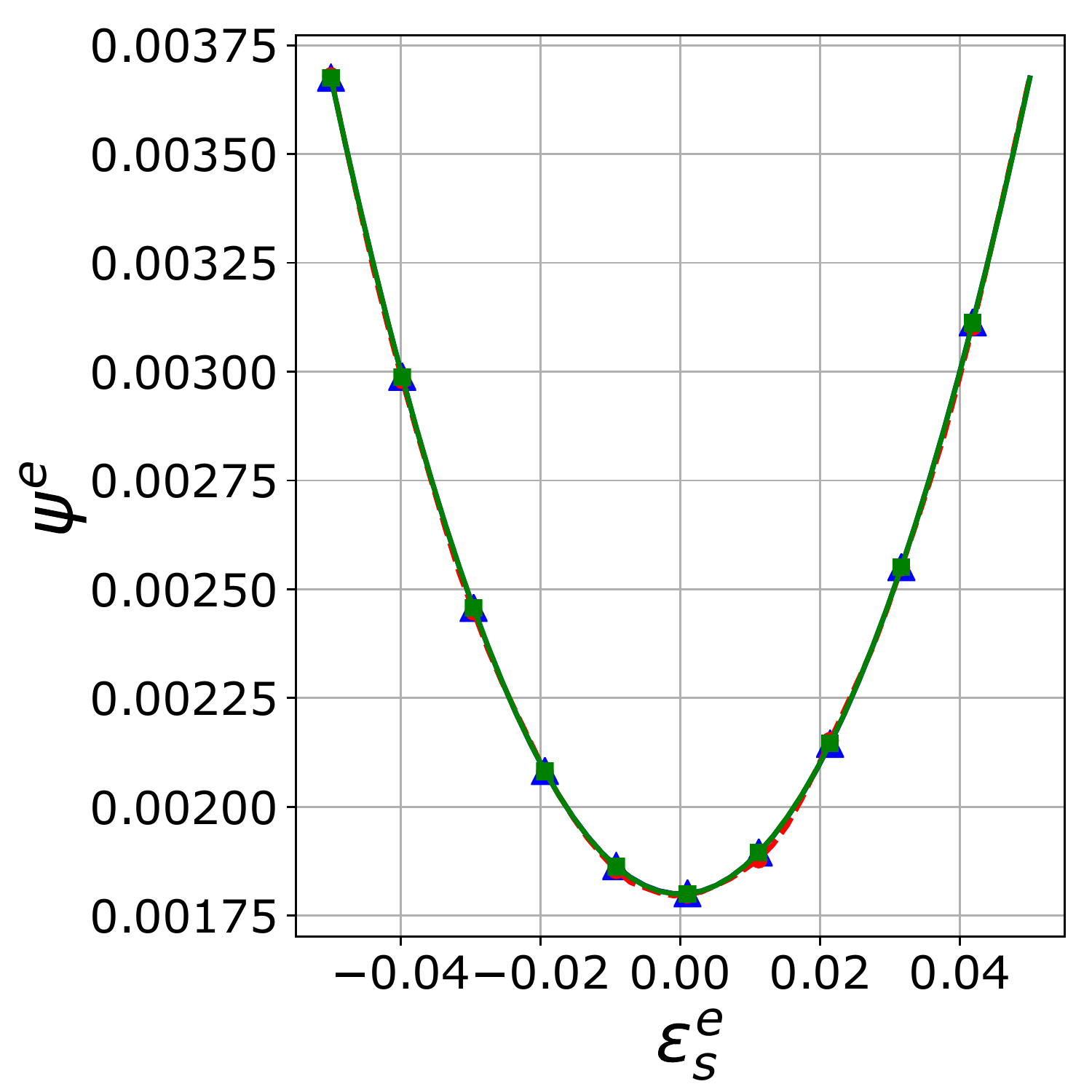} &
\hspace{-1cm}\includegraphics[width=.30\textwidth ,angle=0]{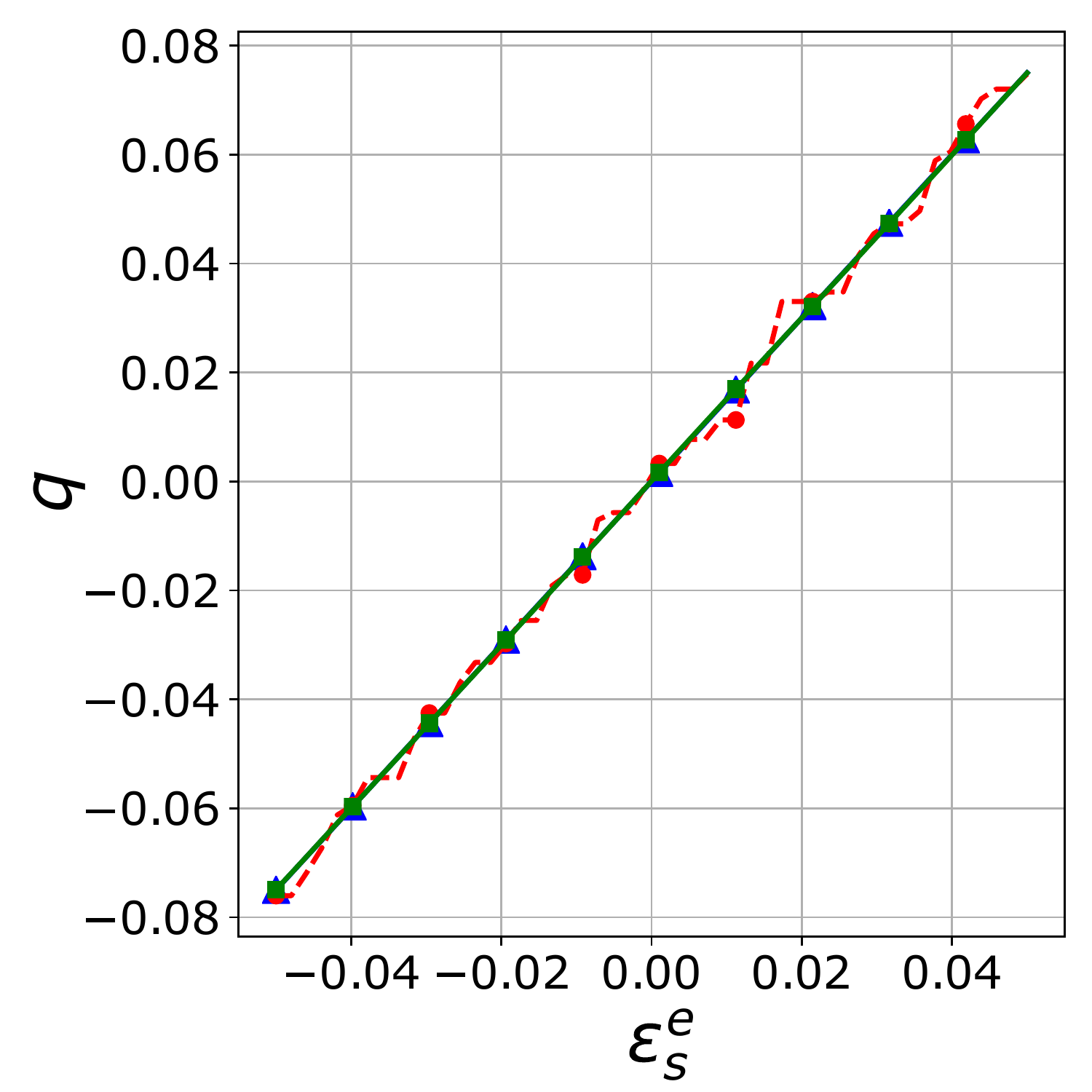} &
\vspace{0.5cm}\hspace{-1cm}\includegraphics[width=.30\textwidth ,angle=0]{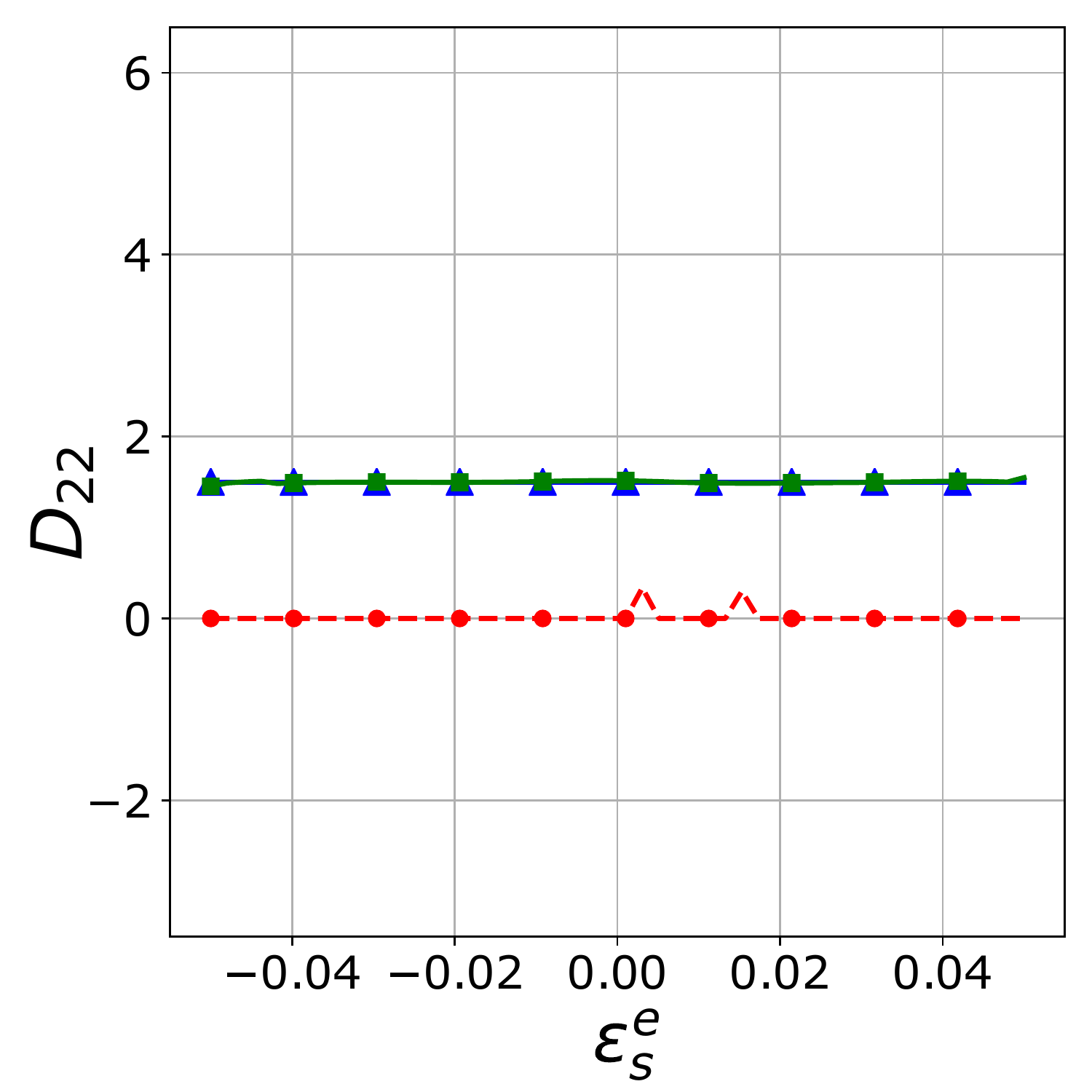} 

\end{tabular}
\includegraphics[width=.9\textwidth]{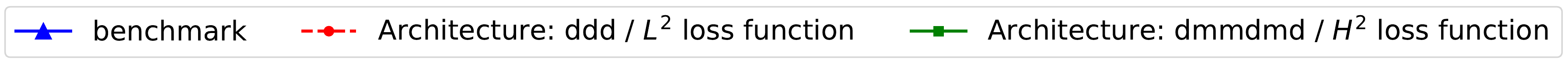} 

\caption{Comparison of the predictions of an $L^2$ trained ddd network with and an $H^2$ trained dmmdmd network for the energy functional, stress, and stiffness measures of the Modified Cam-Clay hyperelastic law \citep{borja2001cam}. The ddd architecture (piece-wise linear activation functions) can only predict local second-order derivatives ($D_{11}, D_{22}$) to be equal to 0. The dmmdmd architecture, modified with Multiply layers, can capture these higher-order derivatives.}
\label{fig:hyper_stress_curve_comp}
\end{figure}

A comparison of the predictive capabilities of an architecture without no added Multiply layers (ddd) and an $L^2$ norm training objective with a dmmdmd $H^2$ trained architecture can be seen in Fig.~\ref{fig:hyper_stress_curve_comp} for the Modified Cam-Clay hyperelastic law \citep{borja2001cam}. Without any Multiply layers, the ddd architecture cannot predict the stiffness measure properly -- all the predictions are 0.

\subsection{Benchmark Study 2: Training of yield function as a level set} 
\label{sec:yield_function_training}

In this section, we demonstrate the training process of the neural network level set yield functions utilized in this paper. We demonstrate the neural networks ability to recover yield surfaces and their evolution completely from the data. The purpose of the yield function neural networks is twofold: to automate the discovery of complex yield surfaces, and to facilitate the expression of non-linear hardening laws. 

\begin{figure}[h!]
\newcommand\siz{.32\textwidth}
\centering

\begin{tabular}{M{.45\textwidth}M{.45\textwidth}}
\includegraphics[width=.45\textwidth ,angle=0]{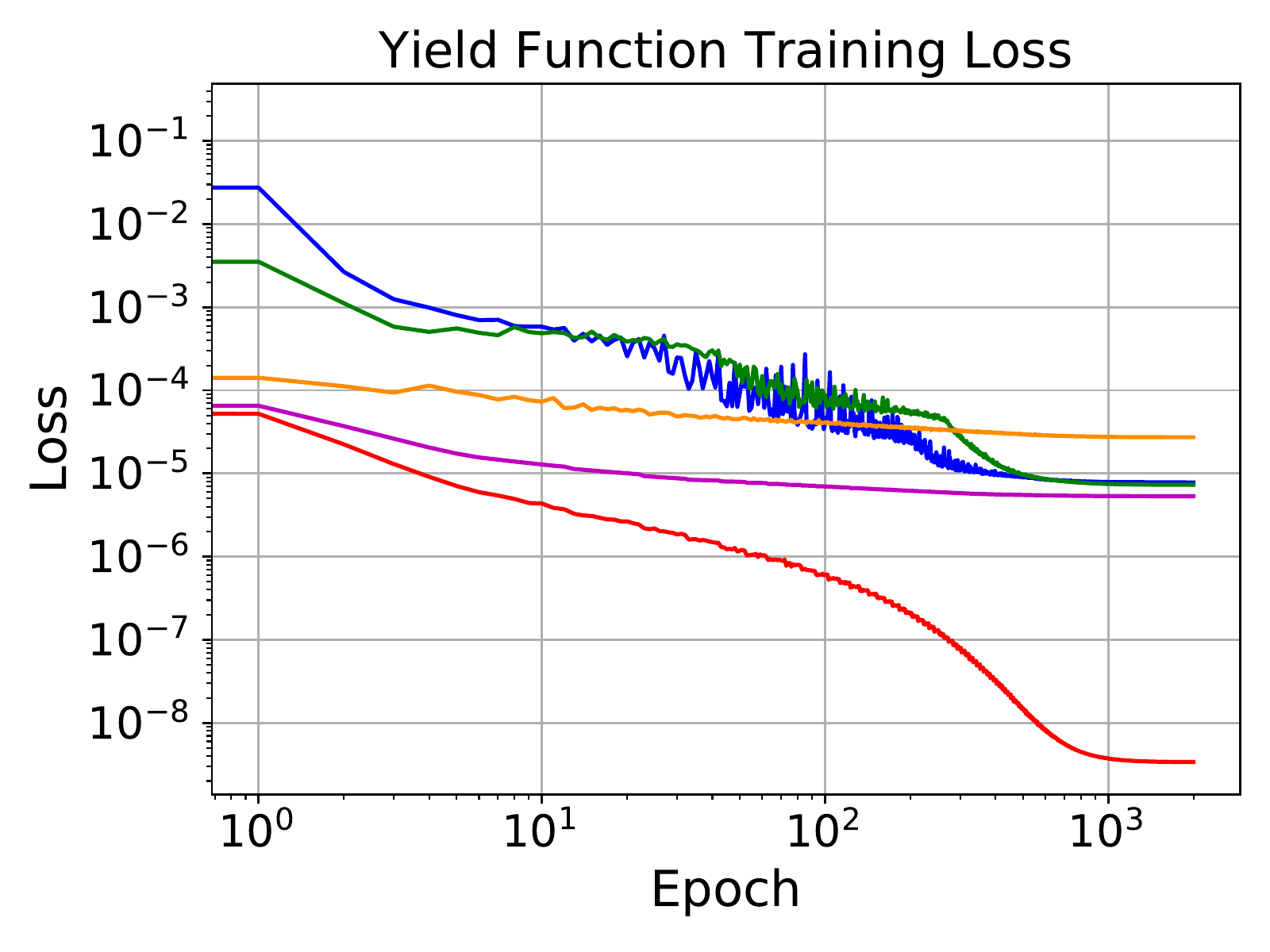} &
\includegraphics[width=.3\textwidth ,angle=0]{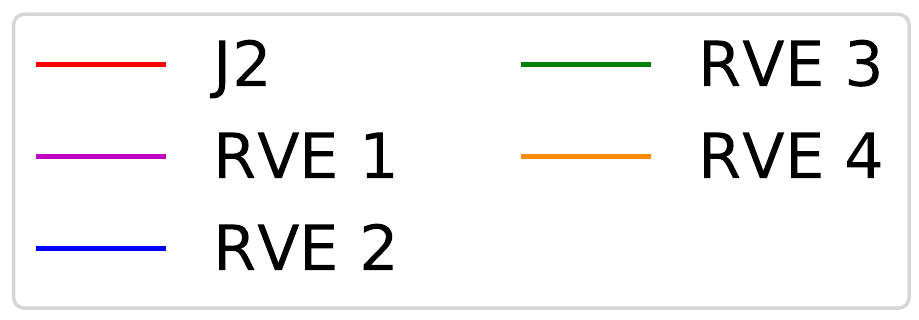} \\
 
\end{tabular}

\caption{Training loss curves for the J2 plasticity and 4 different polycrystal RVEs' yield function level sets.}
\label{fig:yield_function_loss}
\end{figure}

In a first numerical experiment, we test the ability of the neural networks to learn from a yield function level set data set and capture varying yield surface shapes. 
The yield function neural networks have a feed-forward architecture of a hidden Dense layer (100 neurons / ReLU), followed by two Multiply layers, then another hidden Dense layer (100 neurons / ReLU) and an output Dense layer (Linear). All the models were trained for 2000 epochs with a batch size of 128 using the Nadam optimizer, set with default values.
The neural networks were trained on a data set of J2 plasticity as well as data sets for 4 different polycrystal RVEs as described in the Appendix~\ref{sec:dataset_yield}. The training loss curves for this experiment with an $L^2$ training objective are show in Fig~\ref{fig:yield_function_loss}.

\begin{figure}[h!]
\newcommand\siz{.33\textwidth}
\centering

\begin{tabular}{M{.33\textwidth}M{.33\textwidth}M{.33\textwidth}}
\hspace{-2cm}\includegraphics[width=.33\textwidth ,angle=0]{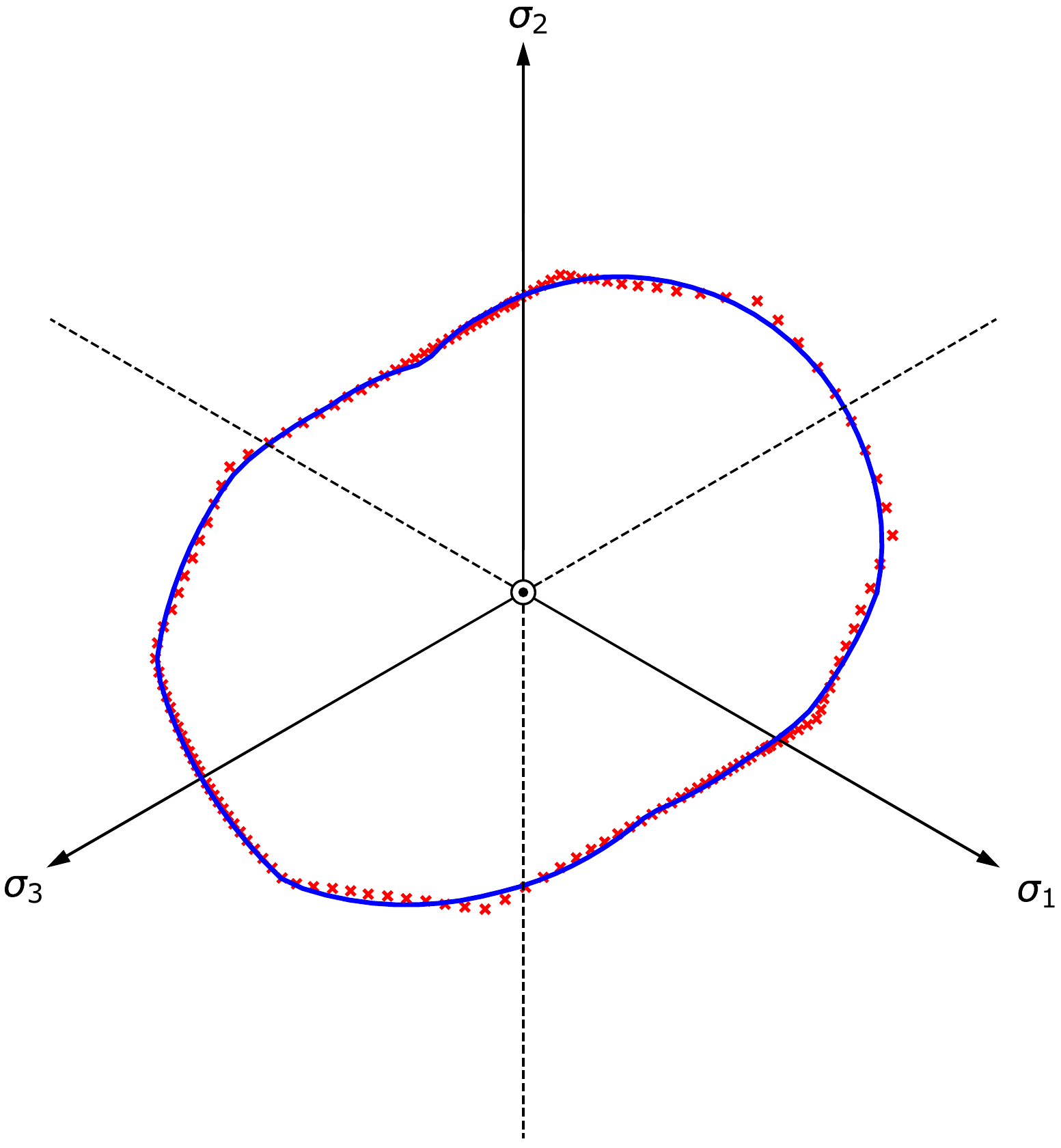} &
\hspace{-2cm}\includegraphics[width=.33\textwidth ,angle=0]{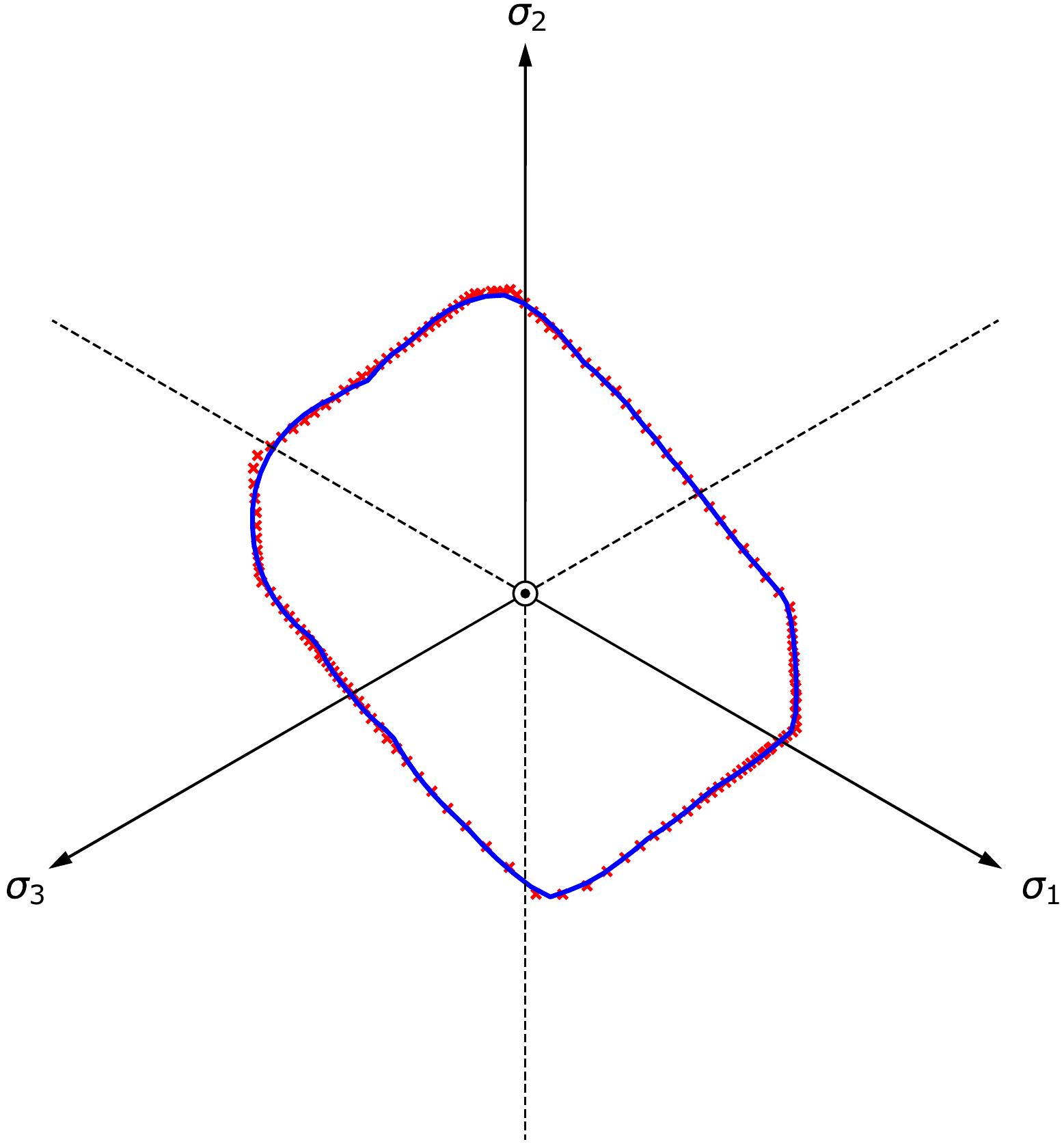} &
\hspace{-2cm}\includegraphics[width=.33\textwidth ,angle=0]{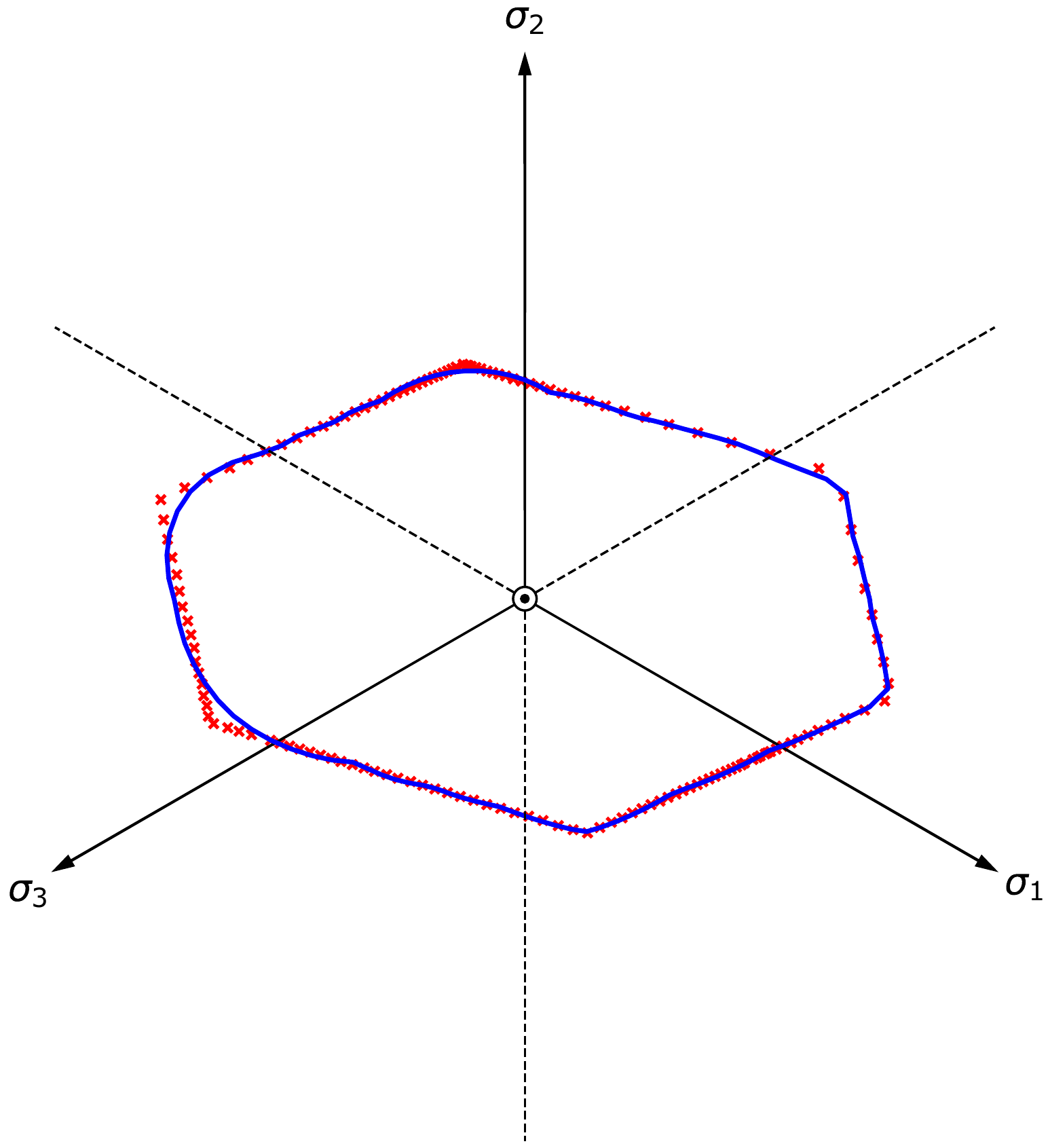} \\
 
\hspace{-1cm} RVE 2 & \hspace{-1cm}RVE 3 & \hspace{-1cm}RVE 4\\
\end{tabular}
\includegraphics[width=.33\textwidth ,angle=0]{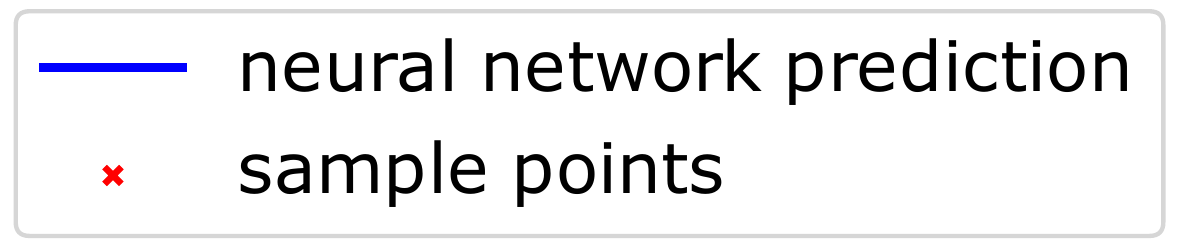} \\

\caption{Yield surface neural network predictions for three polycrystal RVEs with different crystal orientations.}
\label{fig:three_yield_functions}
\end{figure}

The ability to capture a yield surface directly from the data becomes crucial in materials such as the polycrystal microstructures -- where complex constitutive responses may manifest from spatial heterogeneity and grain boundary interactions. In Fig.~\ref{fig:three_yield_functions}, it is shown that a polycrystal RVE of the same size with different crystal orientations can have distinctive initial yield surfaces. Anticipating 
the geometry of the yield surface in the stress space and then handcrafting these them with with mathematical expressions would be a great undertaking and possibly futile since a change in the crystal properties would require deducing geometric shape design from scratch. Our framework automates the discovery of these yield surfaces. This will also be the basis to describe the plastic behavior of anisotropic materials where the yield surface changes for orientations. This will be considered in future work by expanding the stress invariant input space to include orientations and possibly more descriptive plastic internal variables that are derived from the topology of the microstructures.

\subsubsection{Smoothing non-smooth yield surfaces}
Another key feature of the proposed machine learning approach is that 
the Sobolev traning with the right activation functions may 
generate smoothed yield surfaces on the $\pi$-plane. 
Classical non-smooth and multi-yield surface models often lead to 
 sharp tips and corners on the yield surface that makes the stress gradient 
 of the yield function bifurcated. This is not only an issue for stability 
 but also requires specialized algorithmic designs for the return mapping 
 algorithm to function (cf. \citet{de_souza_neto_computational_2011}).
 As a result, there have been decades of 
 efforts to hand-craft derive implicit functions that are smoothed approximations of well-known multi-yield surface models 
 \citet{matsuoka1985relationship, abbo1995smooth}).
Our numerical experiments indicate that such a treatment can be 
automated with the proposed Sobolev training. As shown 
 in Fig.~\ref{fig:three_yield_functions}, our machine learning framework 
 may generate a smoothed yield surface that can be easily incorporated into 
an existing generalized return mapping algorithm without significant modifications.

\begin{figure}[h!]
\newcommand\siz{.33\textwidth}
\centering

\begin{tabular}{M{.33\textwidth}M{.33\textwidth}M{.33\textwidth}}
\hspace{-2cm}\includegraphics[width=.33\textwidth ,angle=0]{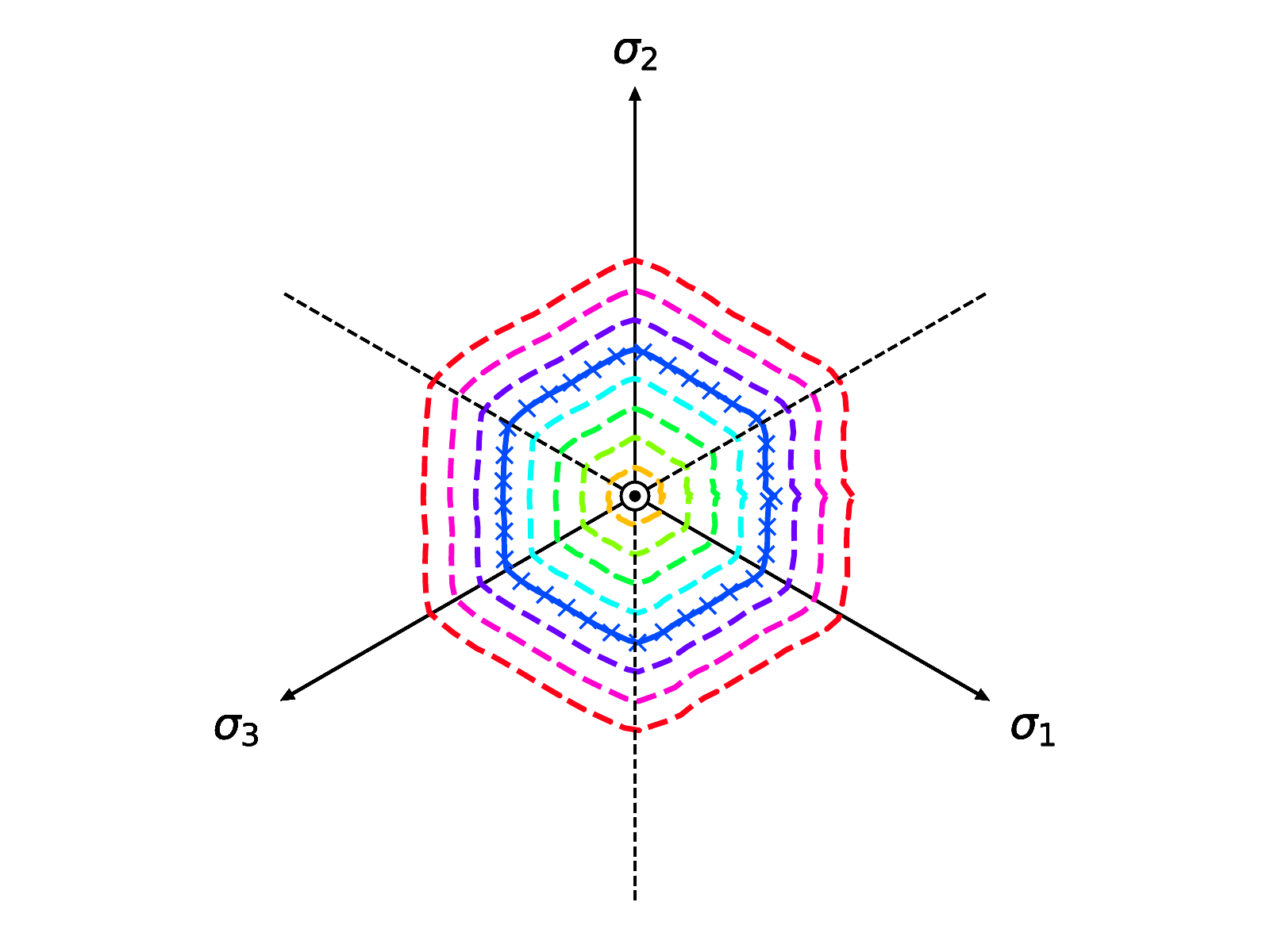} &
\hspace{-2cm}\includegraphics[width=.33\textwidth ,angle=0]{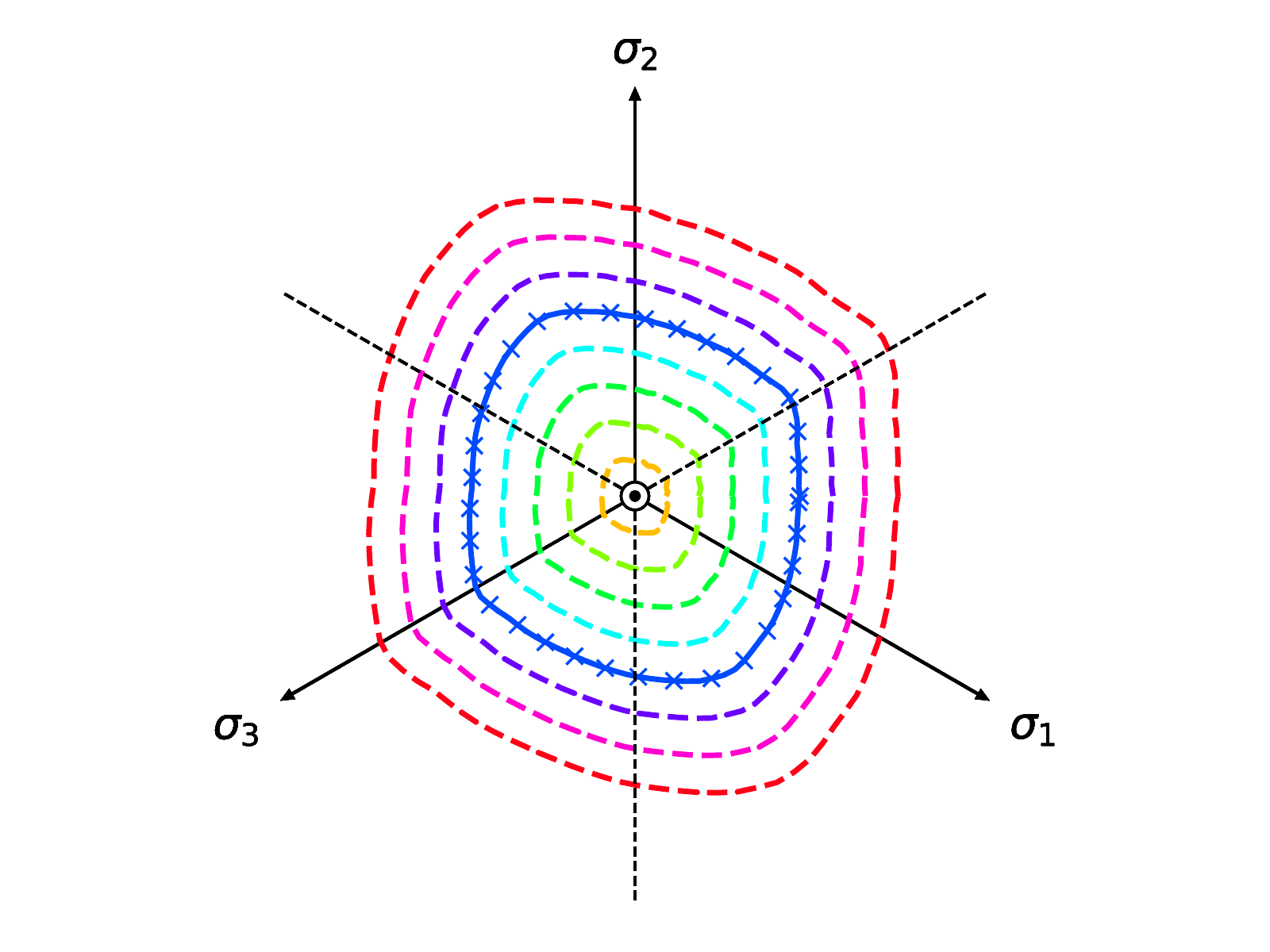} &
\hspace{-2cm}\includegraphics[width=.33\textwidth ,angle=0]{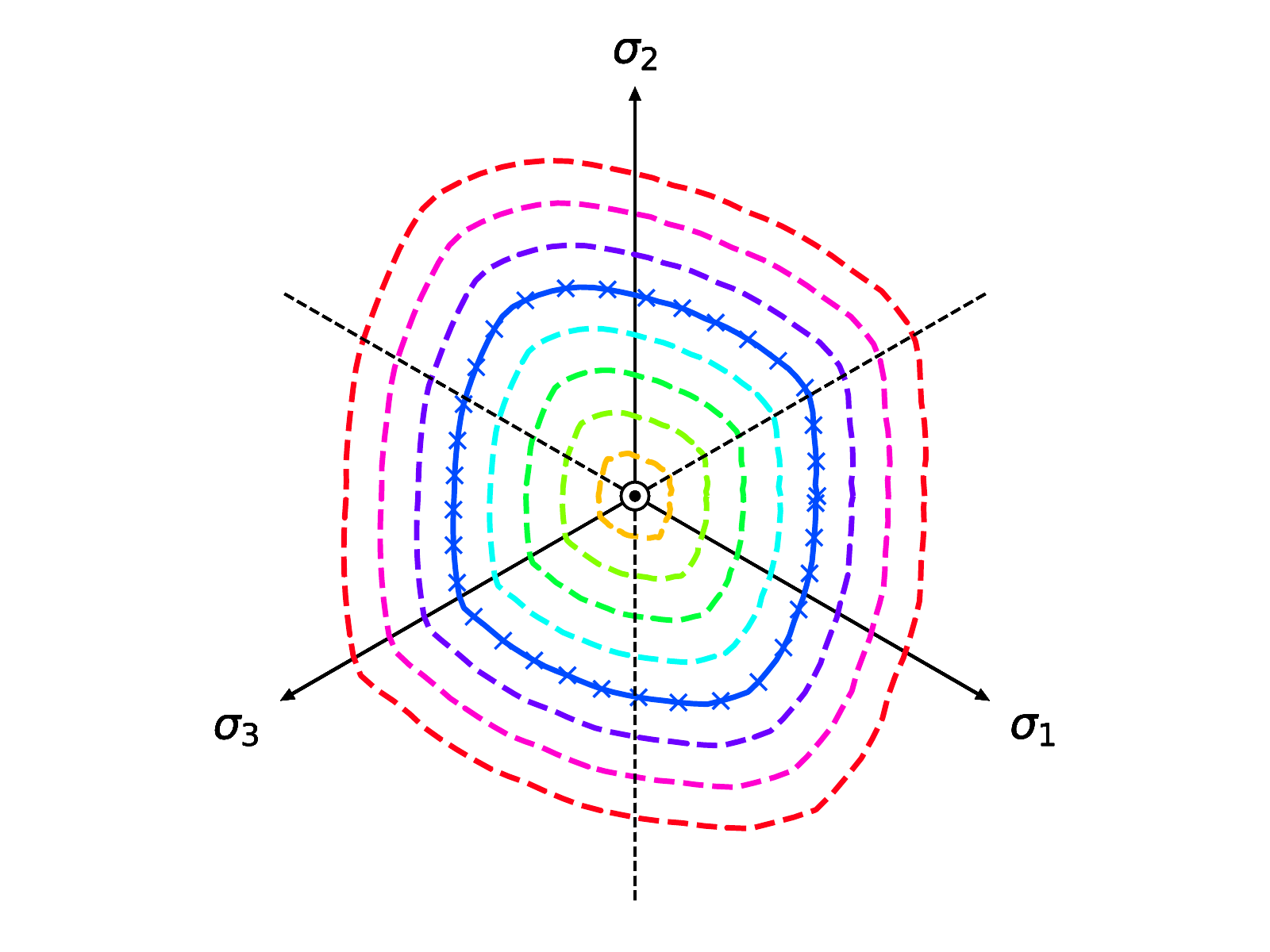} \\
 
\end{tabular}
\includegraphics[width=.4\textwidth ,angle=0]{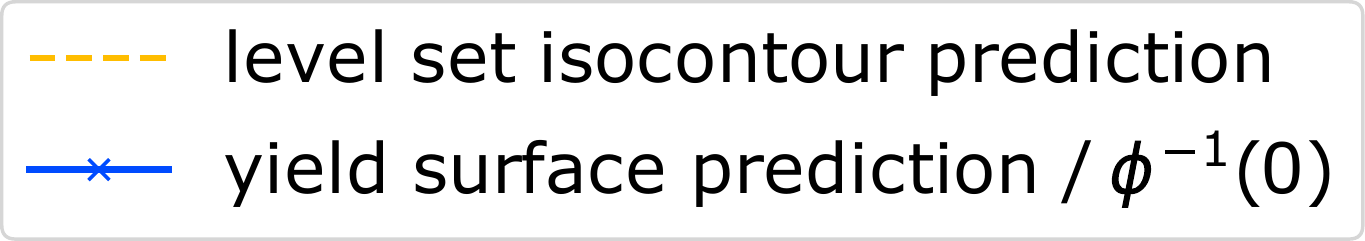} \\

\caption{Predicted level set isocontours for three evolving yield surfaces of RVE 1 of a specimen undergoing increasing axial compression  (left to right).}
\label{fig:three_levelsets}
\end{figure}

The ML-derived yield functions are also capable of replicating complex hardening mechanisms. In Fig.~\ref{fig:three_levelsets}, we demonstrate how the neural network can predict the yield level set and emulate the Hamilton-Jacobi extension to predict the yield surface with a hardening 
mechanism that has not been discovered in the literature. In particular, 
the yield surface is not only changing size but also \textbf{deforming} on the $\pi-$ plane. In this case, the yield surface transforms from a hexagonal shape to an oval shape.
Anticipating and then deducing the mathematically expression for the hardening law is not a simple task.

This task, if done manually, is not efficient or even feasible if 
the model is not aiming to be a surrogate of one RVE but a family of them -- in which case different microstructures could favor distinct modes of hardening and hand-crafting each one of them would become impossible. 
 
 Our framework can interpret experimental data and deduce the optimal shapes and forms of the yield surface that evolves with strain history
without the aforementioned burdens. In our implementation, the framework is able to generalize and identify multiple complicated hardening mechanisms that can be described on the $\pi$-plane. 

\subsection{Benchmark Study 3: Yield function training with Sobolev constraints}
\label{sec:sobolev_yield}

In this section, we demonstrate how we can improve the yield function neural networks predictive capacity by implementing various higher-order Sobolev training techniques described in Section~\ref{sec:framework}. These additional constraints aim to reinforce the robustness of the networks by ensuring that certain desired properties for the level set and the thermodynamic consistency are achieved.

\begin{figure}[h!]
\newcommand\siz{.40\textwidth}
\centering

\begin{tabular}{M{.40\textwidth}M{.40\textwidth}}
\hspace{-1cm}\includegraphics[width=.40\textwidth ,angle=0]{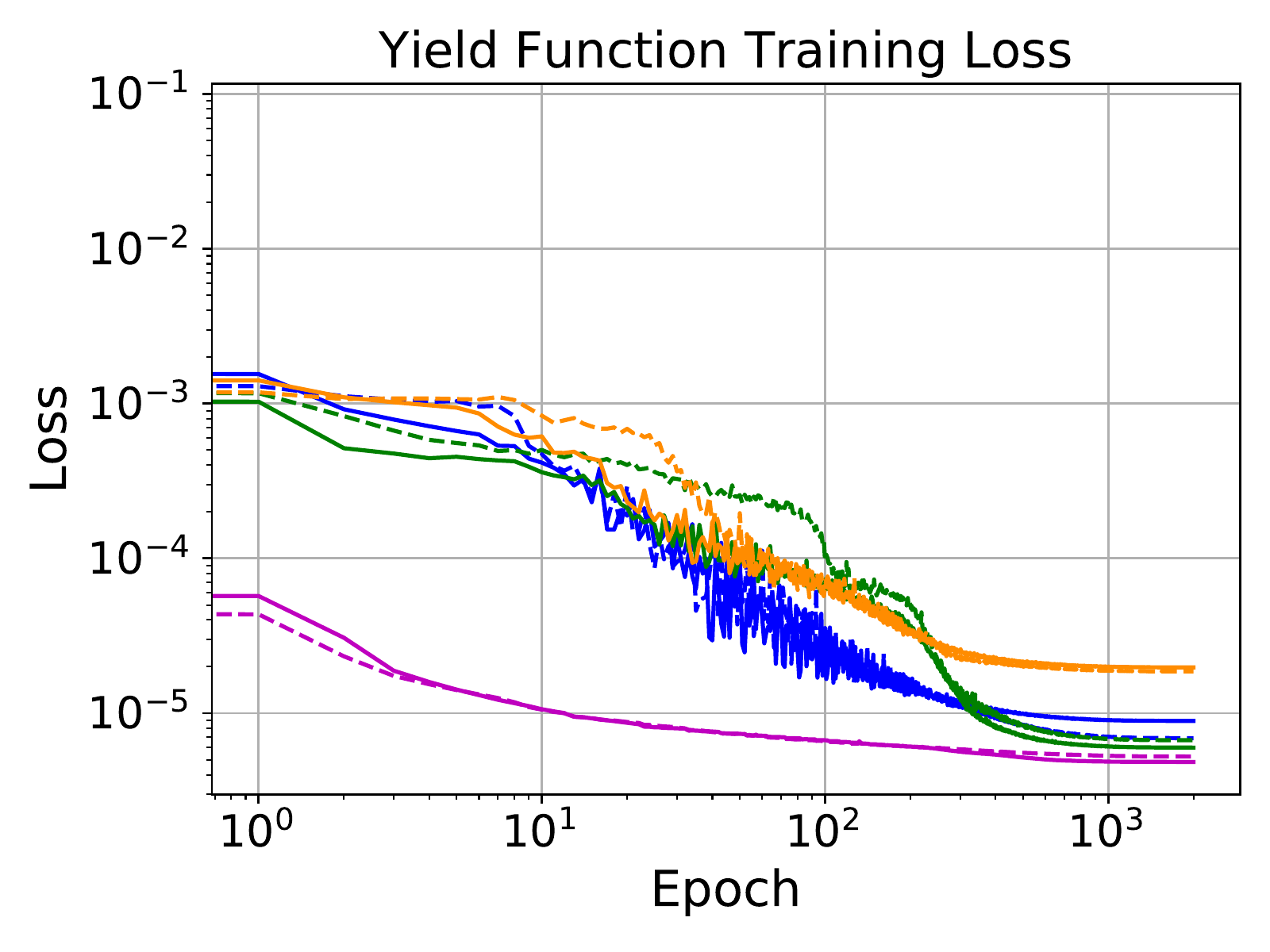} &
\hspace{-1cm}\includegraphics[width=.40\textwidth ,angle=0]{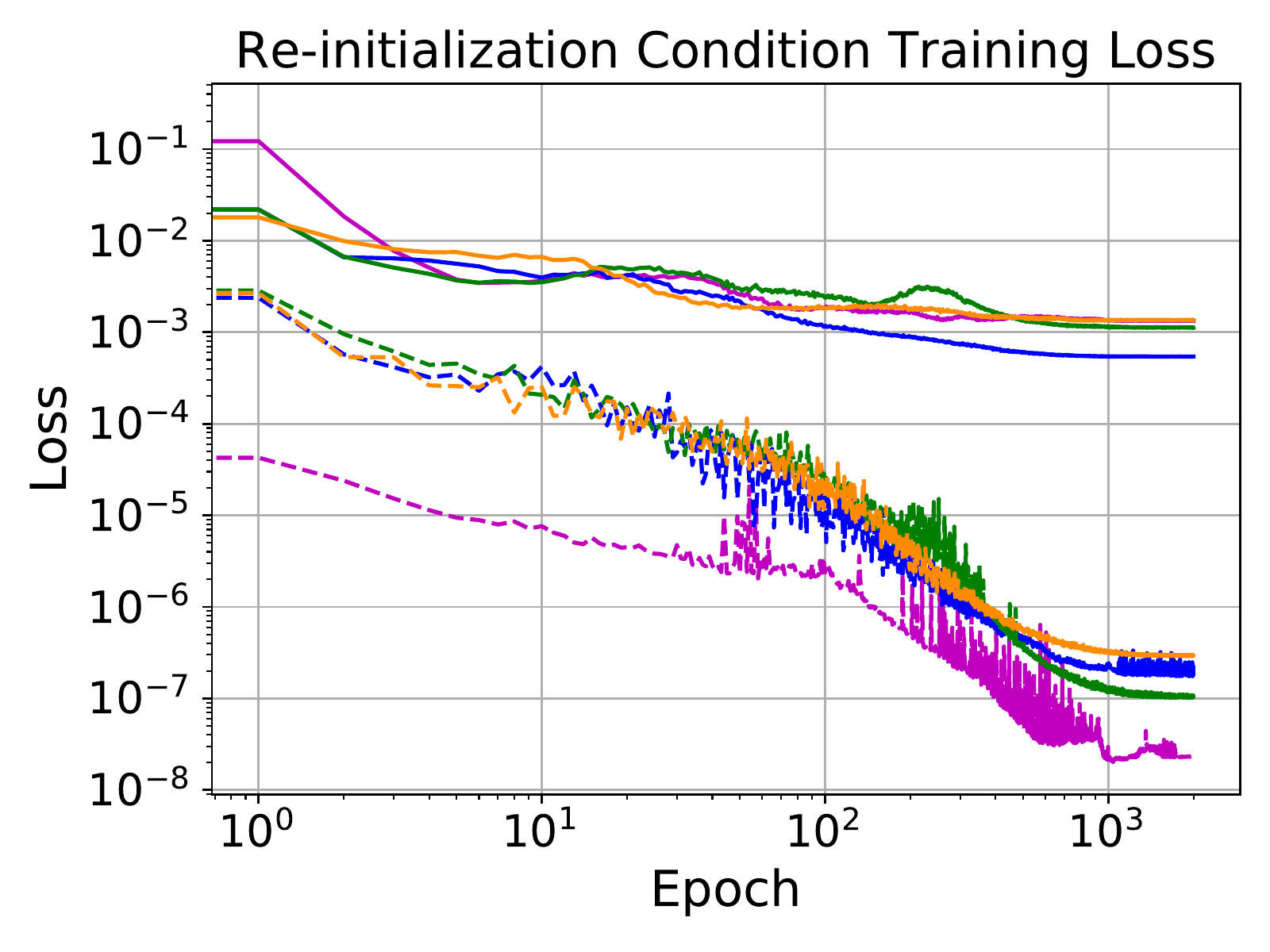} \\
 
\end{tabular}
\includegraphics[width=.70\textwidth ,angle=0]{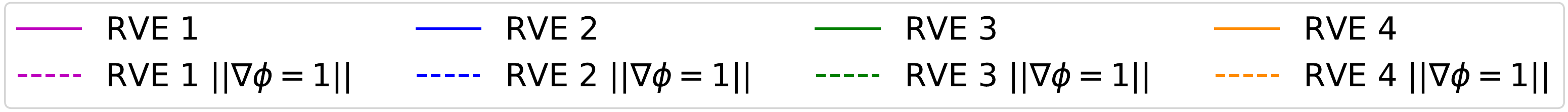} \\

\caption{Training loss function comparison for the polycrystal RVEs' yield functions with the Eikonal equation Sobolev constraint to enforce the re-initialization problem conditions.}
\label{fig:reinitialization_trainining}
\end{figure}

Our yield surface data has been pre-processed to resemble a signed distance function. Emulating the solution of the level set re-initialization problem, it would be desired that the level set at every pseudotime $t$ fulfils the Eikonal equation $|\nabla \phi|=1$. Following eq[ref eq], by applying an additional constraint in the loss function, we can enforce that the conditions of the level set re-initialization problem are met for every predicted instance of the level set function. We repeat the training of the yield function neural networks for the polycrystal RVEs with the additional re-initialization constraint. The results can be seen in Fig.~\ref{fig:reinitialization_trainining}. It is observed that the 
Sobolev constraint successfully imposes the condition $|\nabla \phi|=1$, while it does not affect the performance of the network in predicting the level set values.

\begin{figure}[h!]
\newcommand\siz{.40\textwidth}
\centering
\hspace{-1cm}\includegraphics[width=.40\textwidth ,angle=0]{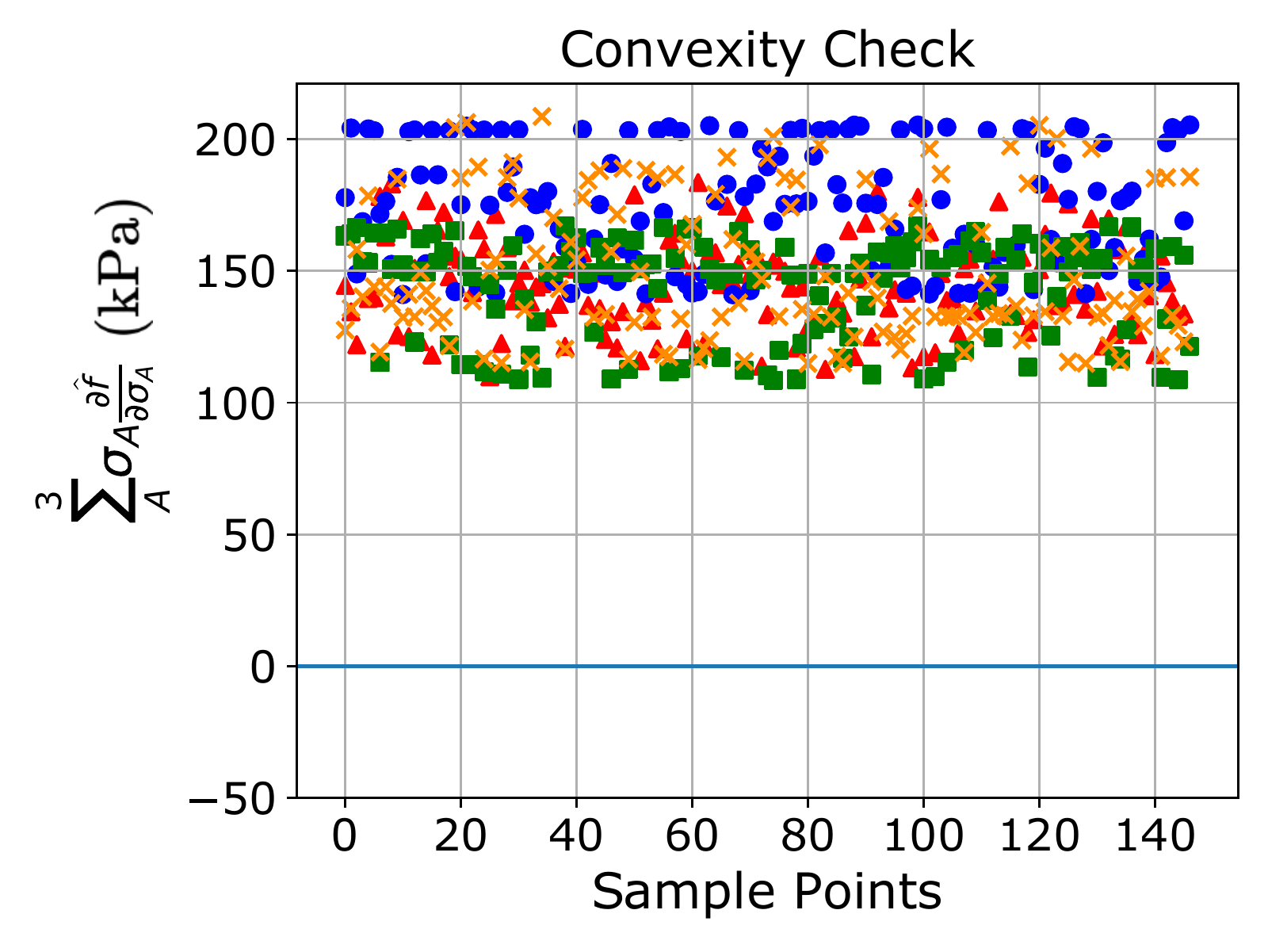} \\
\includegraphics[width=.40\textwidth ,angle=0]{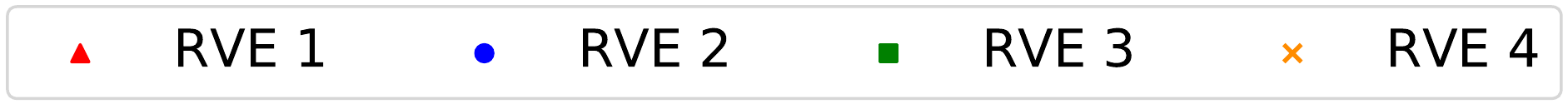} 

\caption{Convexity check for randomly sampled stress points from the polycrystal RVE datasets.}
\label{fig:convexity_check}
\end{figure}

One way to ensure that the predicted material behavior obeys the necessary thermodynamic constraints discussed in Section~\ref{sec:framework} is to ensure that the predicted yield surface is convex. 

We implement an additional inequality constraint, following Equation~\eqref{eq:convexity_penalty_yield} to penalize predictions that do not obey the convexity conditions during training. This penalty loss function term only activates when the thermodynamic inequality is violated. During the training phase of the numerical experiments presented in this paper, the penalty term did not activate. Nevertheless, the penalty term in the loss function is still 
 employed as a safeguard to prevent possible violations of the thermodynamic constraints. 
 We expect that this safeguard will be helpful in future work when 
we extend our current framework to experimental data or to anisotropic materials where the 
 visual inspection of the convexity in the principal stress or $\pi$-plane is no longer feasible. 
 A verification of the convexity is performed and the results are shown  in Fig~\ref{fig:convexity_check} where material states were randomly sampled from the polycrystal RVE database to test whether the inequality   \eqref{eq:convexity_penalty_yield} is violated. 

\begin{figure}[h!]
\newcommand\siz{.48\textwidth}
\centering
\hspace{-1cm}\includegraphics[width=.48\textwidth ,angle=0]{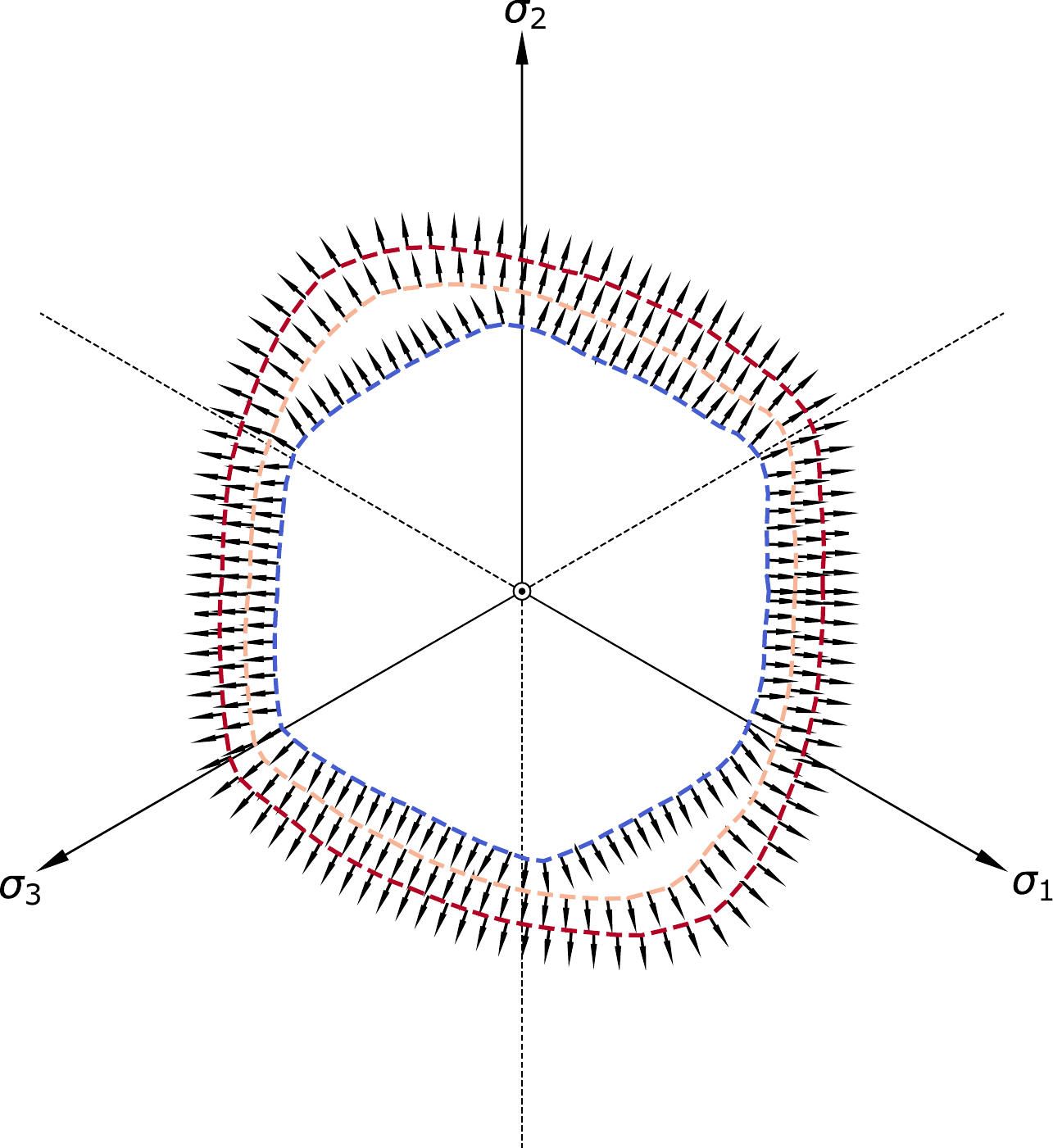} 

\caption{Predicted plastic flow of polycrystal RVE for increasing accumulated plastic strain.}
\label{fig:plastic_flow}
\end{figure}

\begin{figure}[h!]
\newcommand\siz{.48\textwidth}
\centering

\begin{tabular}{M{.48\textwidth}M{.48\textwidth}}
\hspace{-1cm}\includegraphics[width=.48\textwidth ,angle=0]{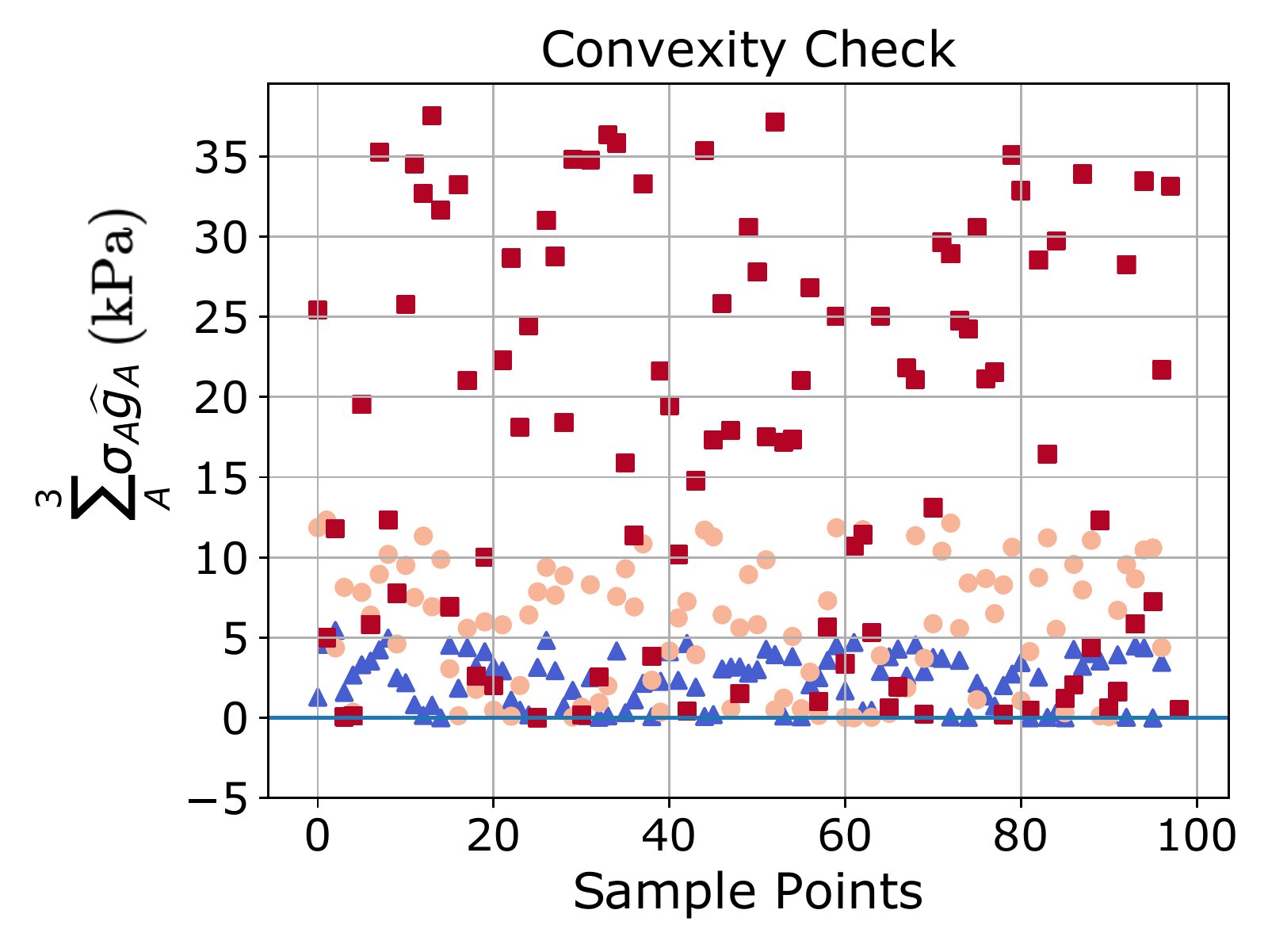} &
\hspace{-1cm}\includegraphics[width=.48\textwidth ,angle=0]{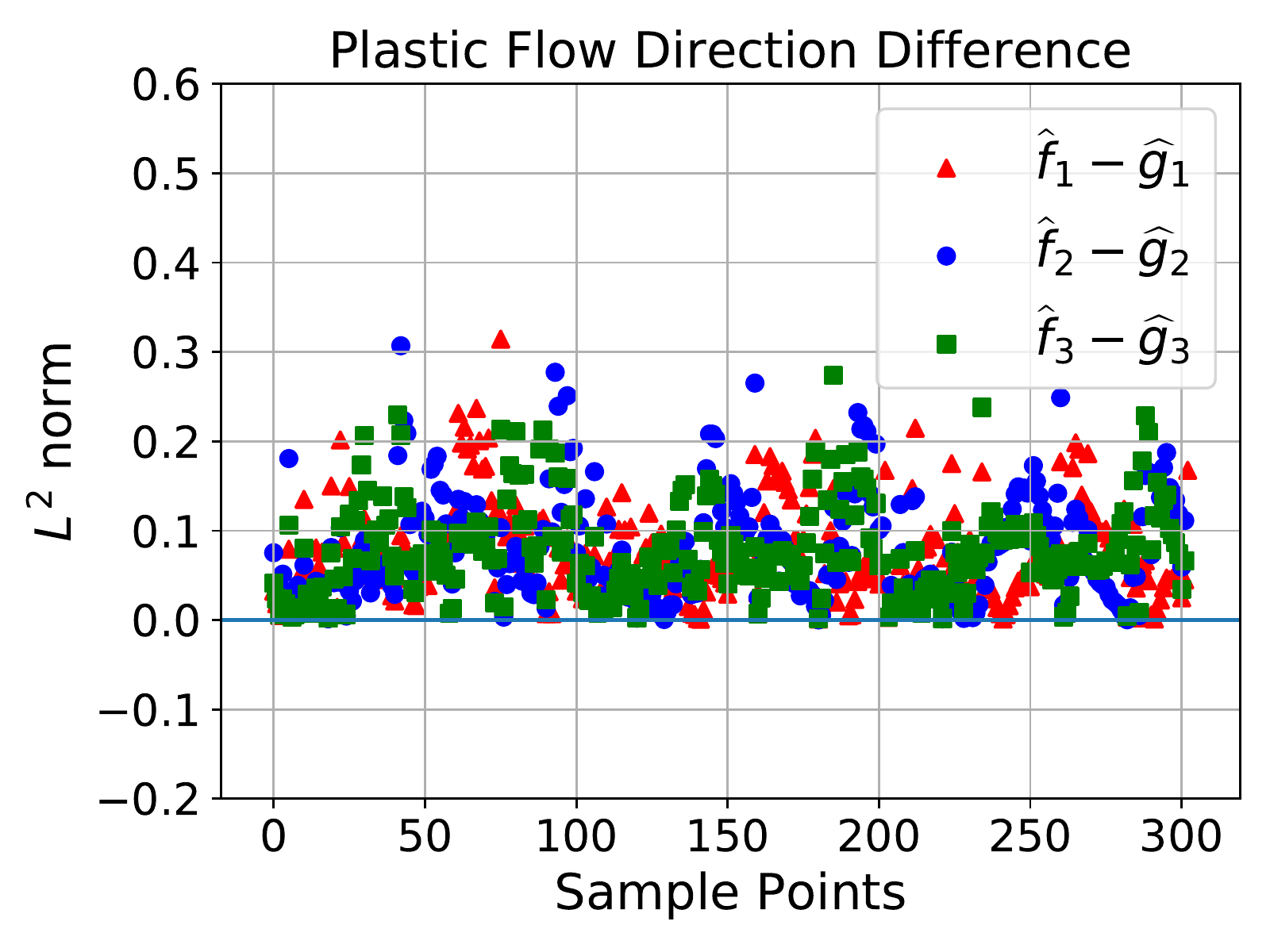} \\
 (a) & (b)
\end{tabular}

\caption{(a) The plastic flow rule is checked for convexity. (b) $L^2$ norm comparison for the predicted plastic flow direction between yield function neural network ($\widehat{f}_A$) and plastic flow neural network ($\widehat{g}_A$).}
\label{fig:plastic_flow_check}
\end{figure}

The control over the thermodynamic behavior of the material allows for the application of a thermodynamically consistent non-associative flow rule. In Fig.~\ref{fig:plastic_flow}, we train a neural network on the plastic flow information of a polycrystal RVE as described in Section~\ref{sec:nonassociative}. We have enforced the convexity of this plastic flow rule through the loss function to secure thermodynamic consistency, following Equation~\eqref{eq:convexity_penalty_flow}. To simplify the implementation of the thermodynamic constraint, we have trained the neural network with inputs the principal stresses $\sigma_1,\sigma_2,\sigma_3$ and outputs the plastic flow directions $g_1,g_2,g_3$ on the $\pi$-plane. All the other training parameters of the network are identical to the ones used for the yield function learning in Section~\ref{sec:yield_function_training}. To verify that convexity is preserved, we perform a convexity check from randomly sampled stress points on the $\pi$-plane (Fig.~\ref{fig:plastic_flow_check} (a)). For the polycrystal material modeled in this work, it is noted that the degree of non-associativity is not rather high -- the $L^2$ norm comparison for the predicted plastic flow direction between yield function neural network ($\widehat{f}_A$) and plastic flow neural network ($\widehat{g}_A$) is demonstrated in Fig.~\ref{fig:plastic_flow_check} (b). Further decoupling of the plastic flow calculation from the yield function allows for more flexibility in modeling non-associative behaviors of plasticity.

\subsection{Application 1: Surrogate model comparisons for polycrystal RVEs}
\label{sec:comparison_black_box}

This section will demonstrate how our elastoplastic Hamilton-Jacobi hardening framework (introduced in Section~\ref{sec:framework}) can compare with commonly used recurrent neural network architectures in predicting the elastoplastic response of the polycrystal material.
The data set generation for the neural network approximator $\widehat{f}$ is described in Appendix~\ref{sec:dataset_yield}. The three recurrent architectures that will be used for comparison -- a multi-step feed forward neural network, a GRU recurrent neural network, and 1D convolution architecture -- have been described in Section~\ref{sec:comparison_models}.

\begin{figure}[h!]
\newcommand\siz{.47\textwidth}
\centering

\begin{tabular}{M{.5\textwidth}M{.5\textwidth}}

\includegraphics[width=.48\textwidth]{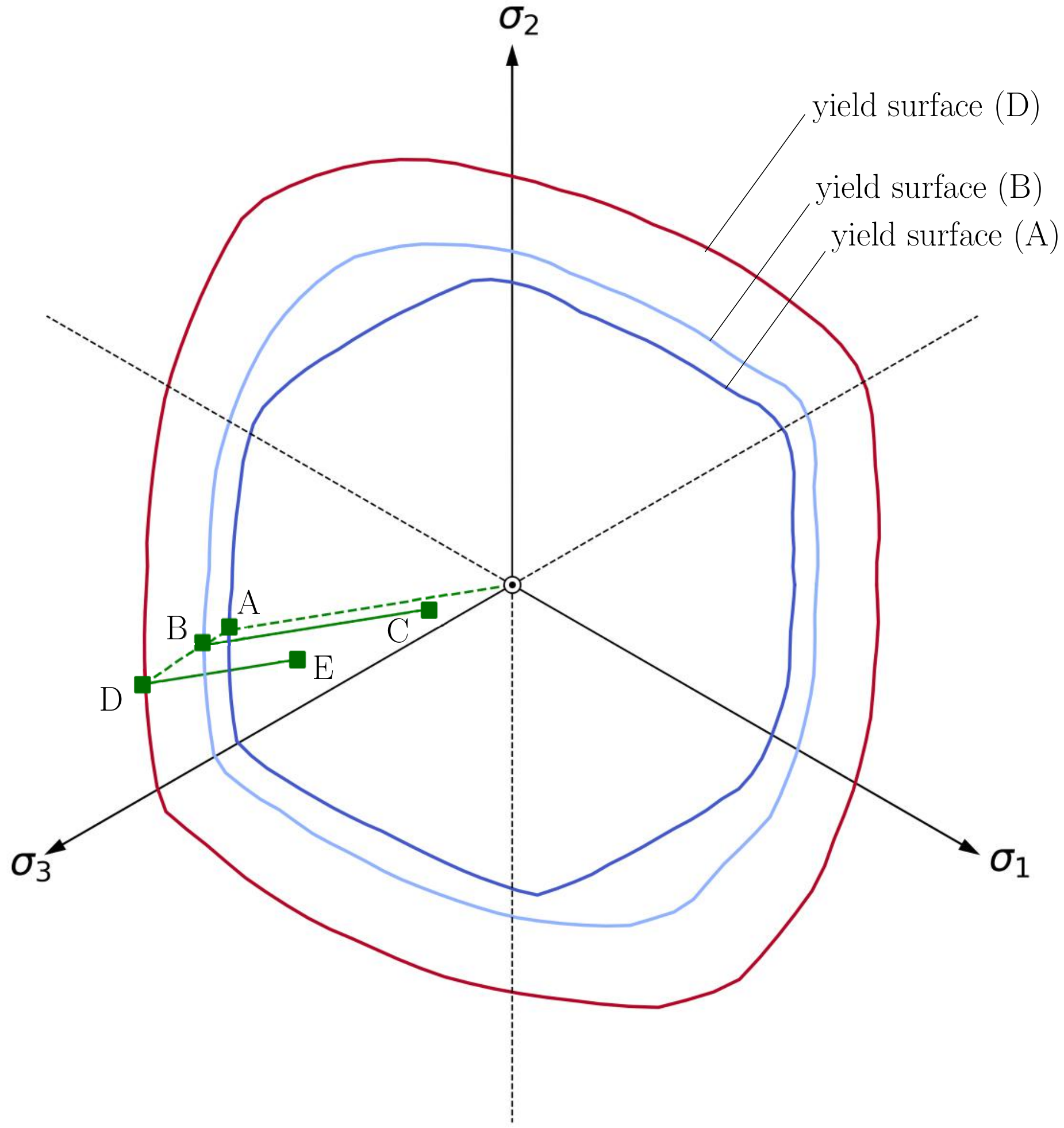} &

\includegraphics[width=.48\textwidth]{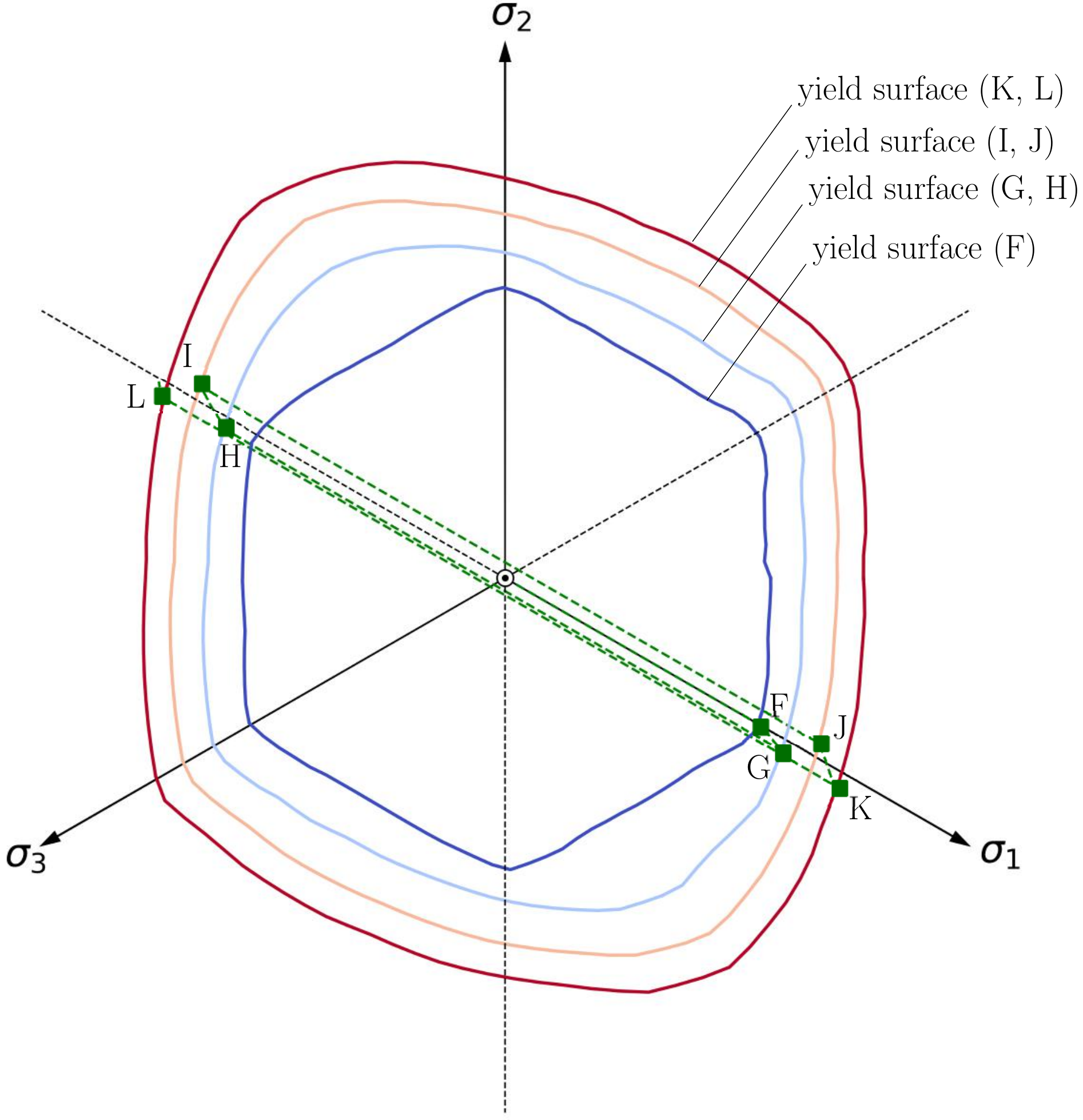} \\

(a) & (b) \\
\end{tabular}

\caption{Stress path in the $\pi$-plane for (a) a loading-unloading pattern and (b) a cyclic loading path. The yield surface neural network predicts the consecutive yield surfaces for different levels of hardening. (a) The points A, B, C, D, and E correspond to the strain-stress curve of Fig.~\ref{fig:multiunloading_networks}. (b) The points F, G, H, I, J , K, and L correspond to the strain-stress curve of Fig.~\ref{fig:cyclic_loading_comparison}. }
\label{fig:cyclic_path}
\end{figure}

To allow for a fair comparison, we have trained, tested, and compared with the recurrent models in for different amounts of data availability and for loading paths of increased difficulty. 
It is noted that the database used for the training of $\hat{f}$ will not be extended further than the 140 cases of monolithic loading cases described in Appendix~\ref{sec:dataset_yield} as it would not be necessary, even for more complex loading paths.

\begin{figure}[h!]
\newcommand\siz{.23\textwidth}
\centering
\begin{tabular}{M{.01\textwidth}M{.24\textwidth}M{.24\textwidth}M{.24\textwidth}M{.24\textwidth}}
\hspace{-2.1cm}(a) &
\hspace{-2.5cm}\includegraphics[width=.23\textwidth ,angle=0]{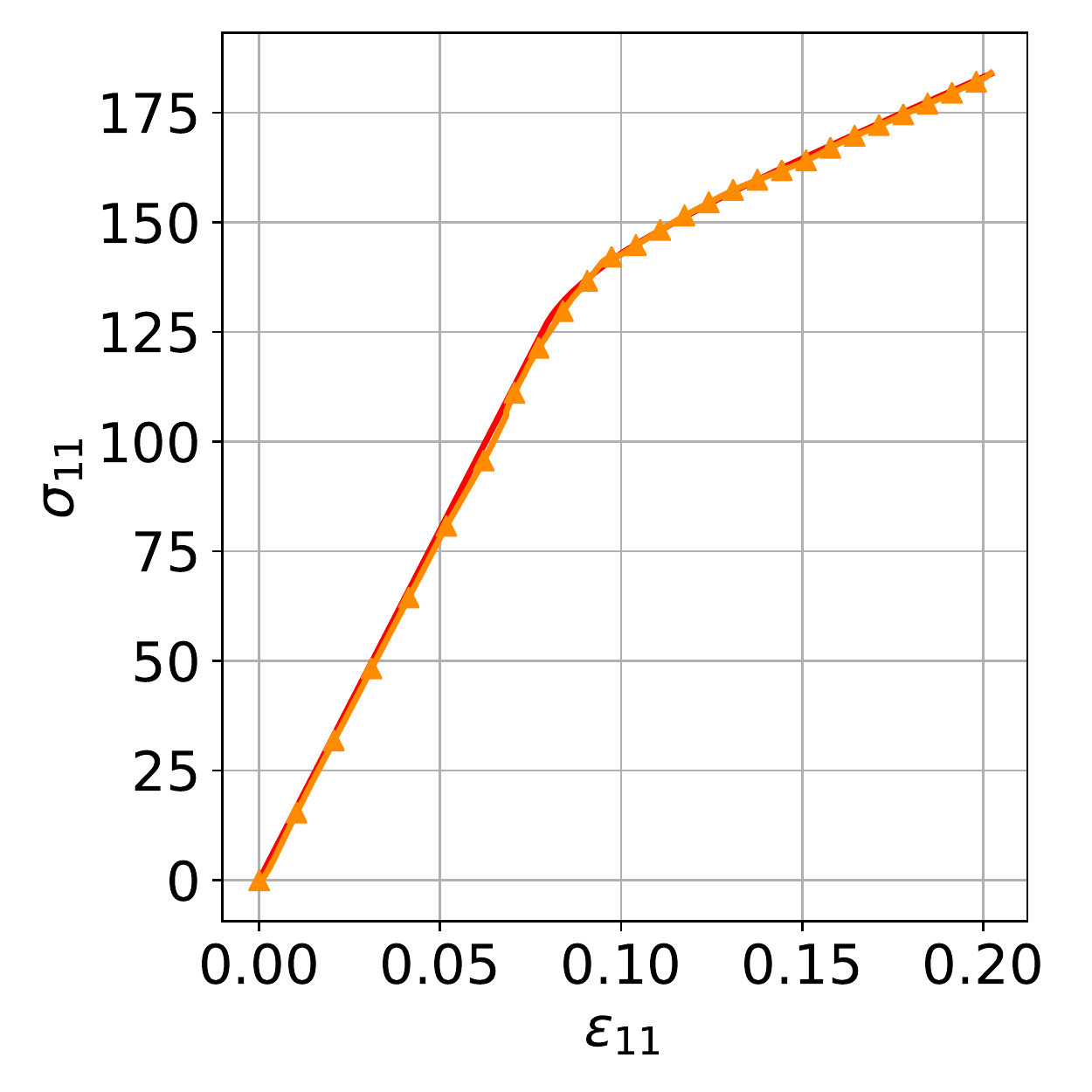} &
\hspace{-2.5cm}\includegraphics[width=.23\textwidth ,angle=0]{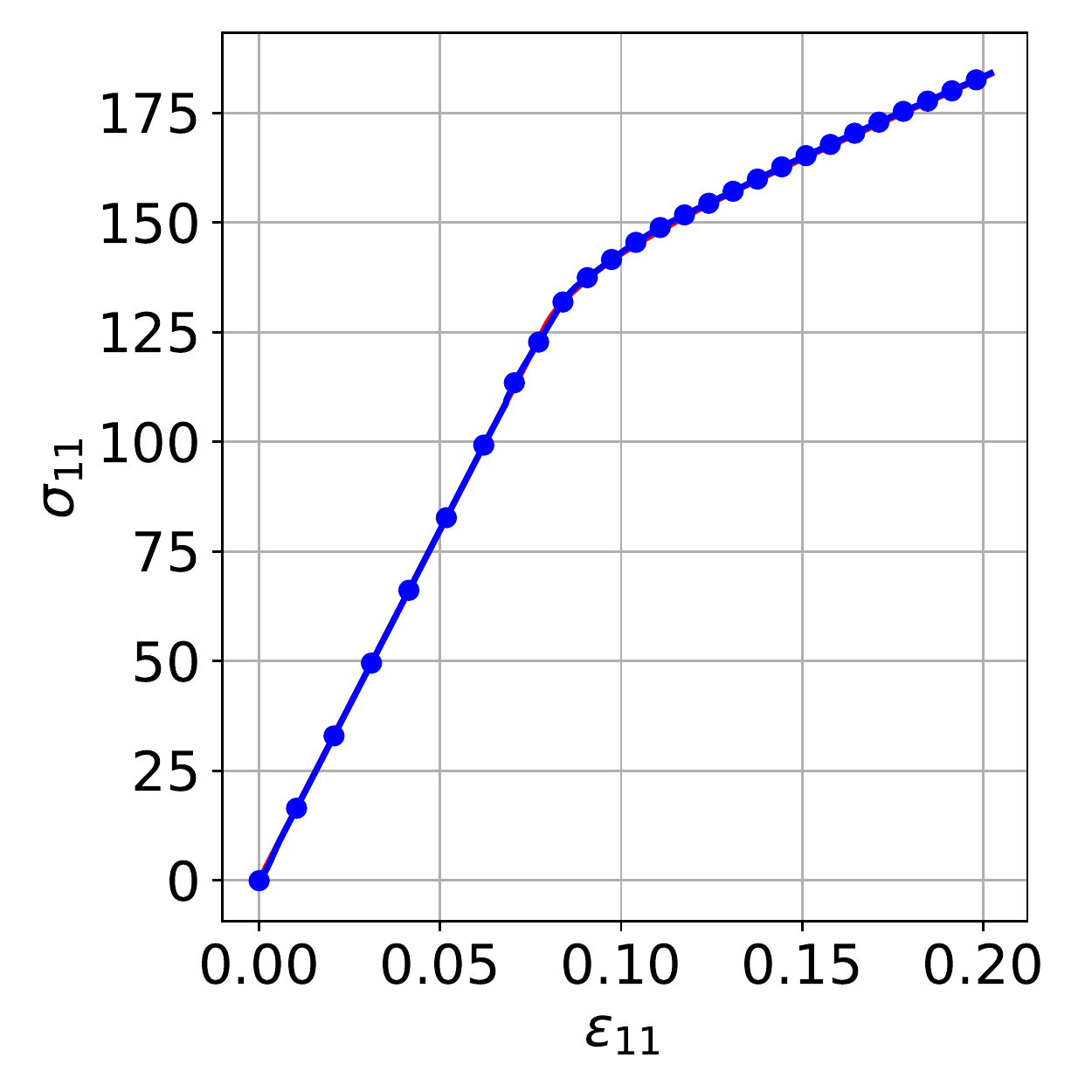} &
\hspace{-2.5cm}\includegraphics[width=.23\textwidth ,angle=0]{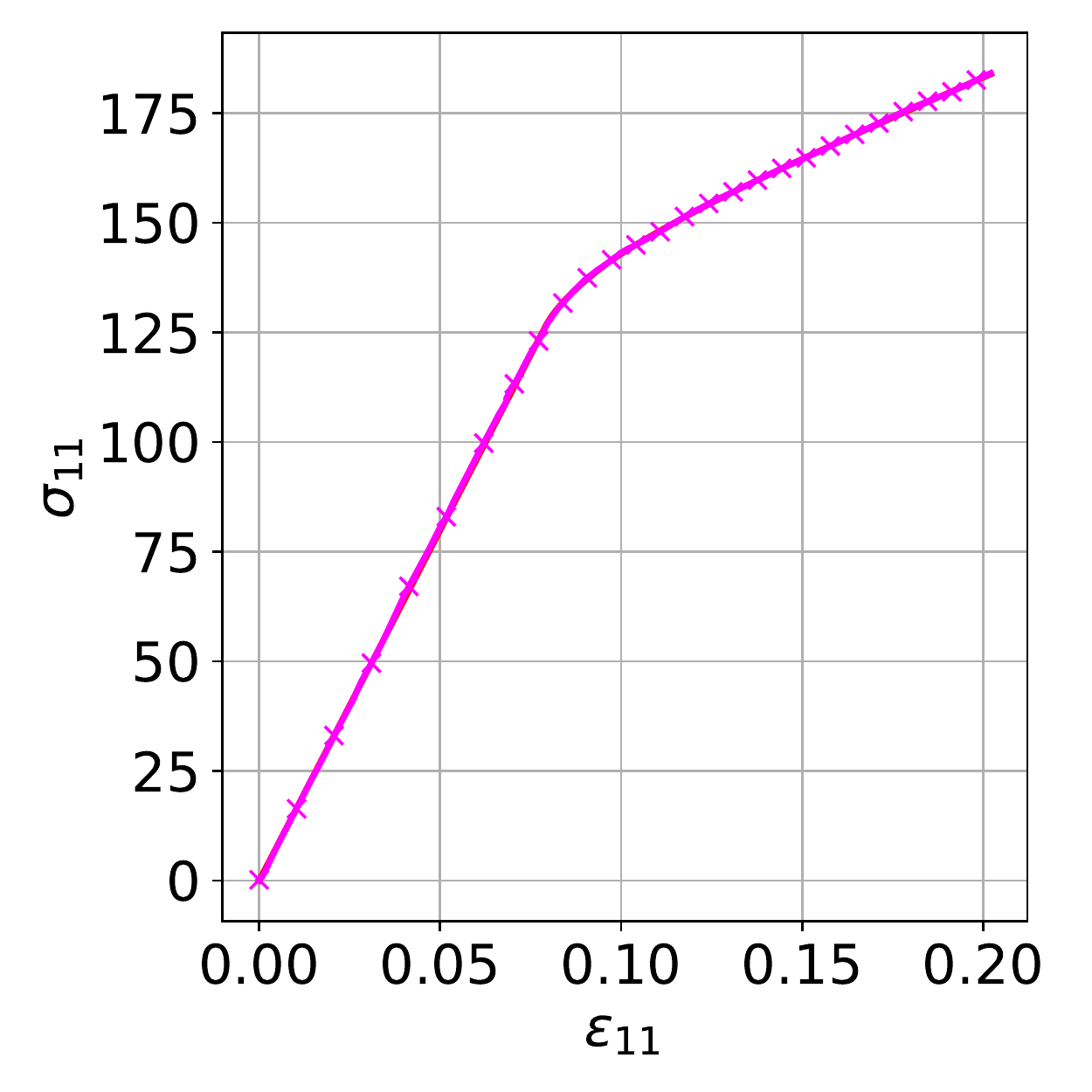} &
\vspace{0.2cm}\hspace{-2.5cm}\includegraphics[width=.23\textwidth ,angle=0]{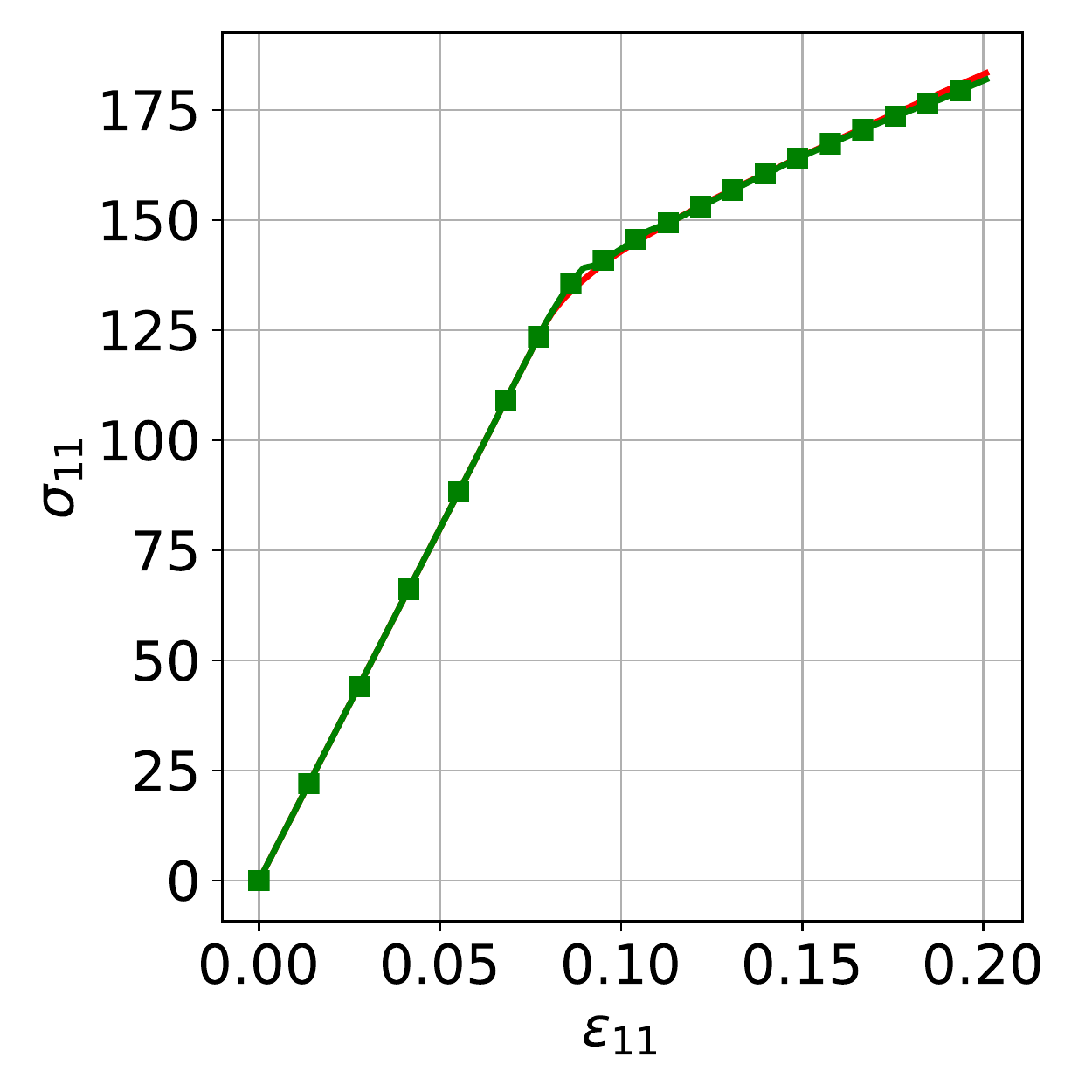} \\     

\hspace{-2.1cm}(b) &
\hspace{-2.5cm}\includegraphics[width=.23\textwidth ,angle=0]{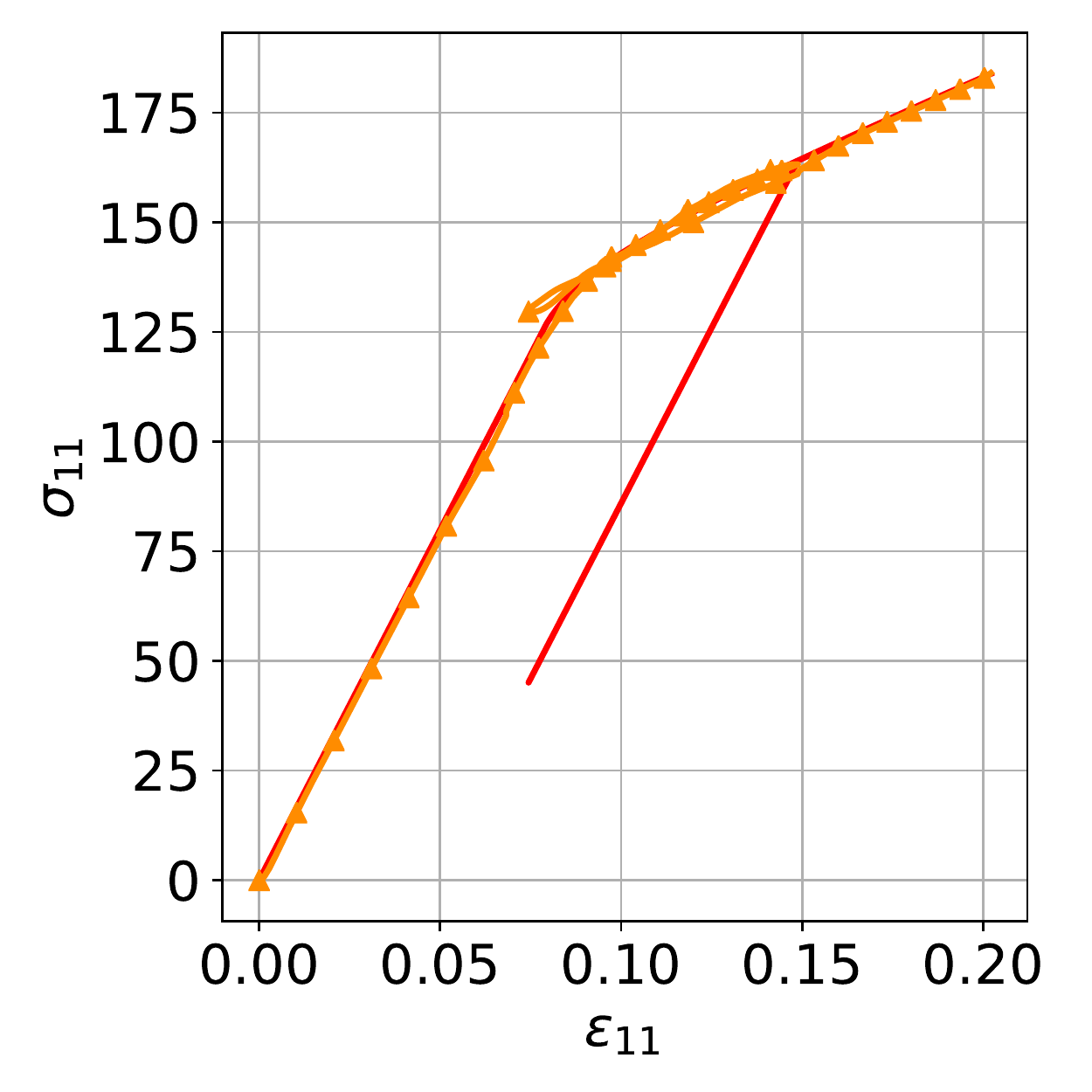} &
\hspace{-2.5cm}\includegraphics[width=.23\textwidth ,angle=0]{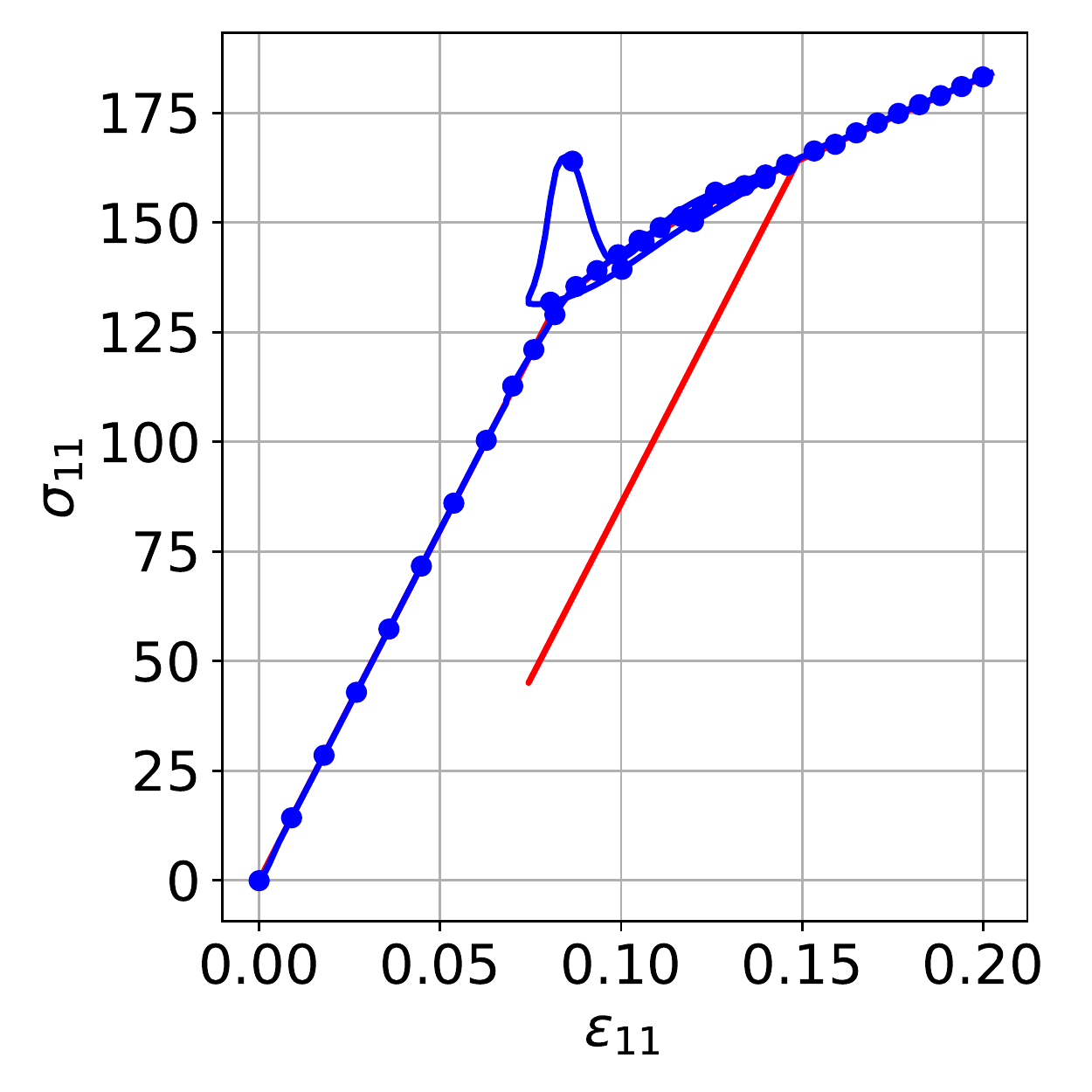} &
\hspace{-2.5cm}\includegraphics[width=.23\textwidth ,angle=0]{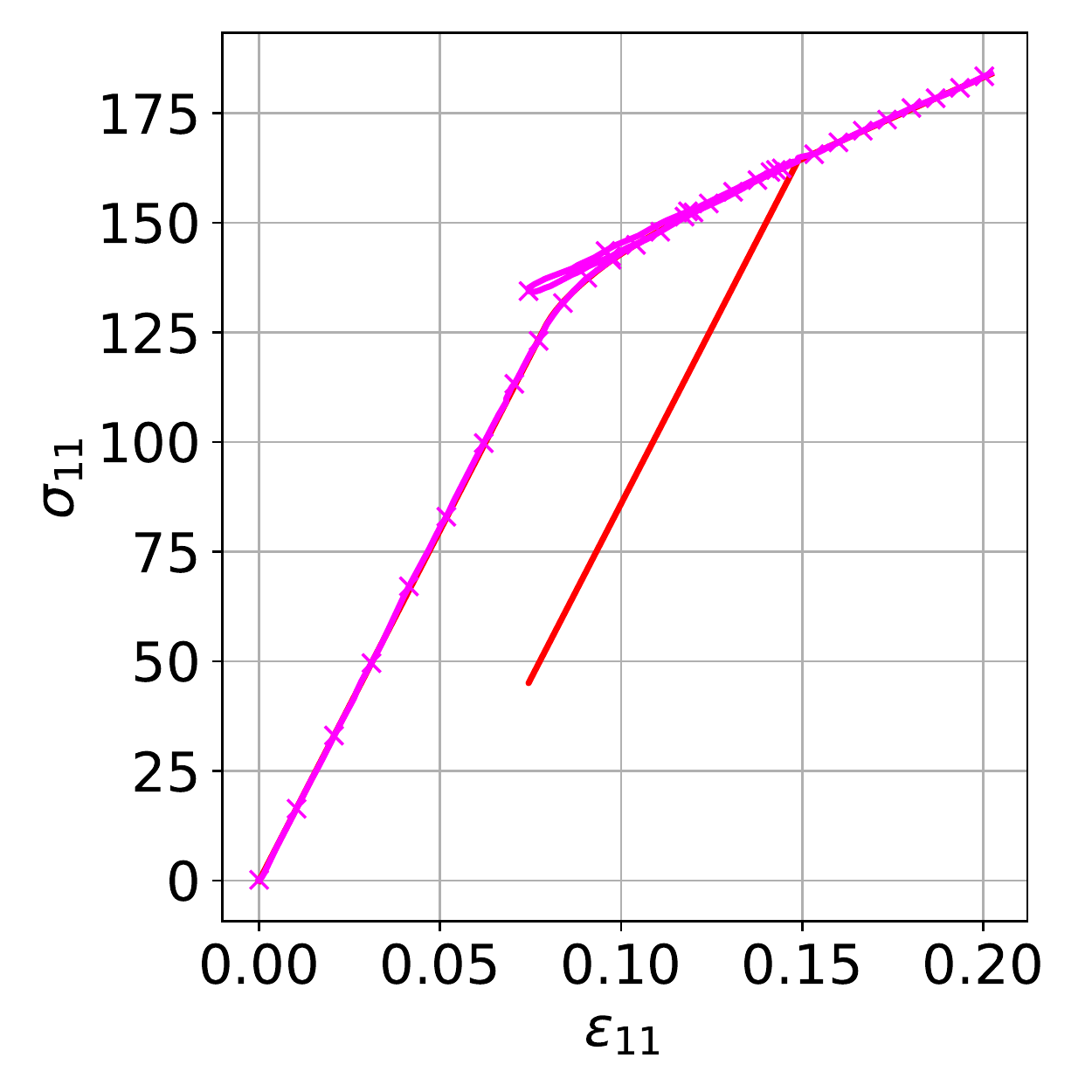} &
\vspace{0.2cm}\hspace{-2.5cm}\includegraphics[width=.23\textwidth ,angle=0]{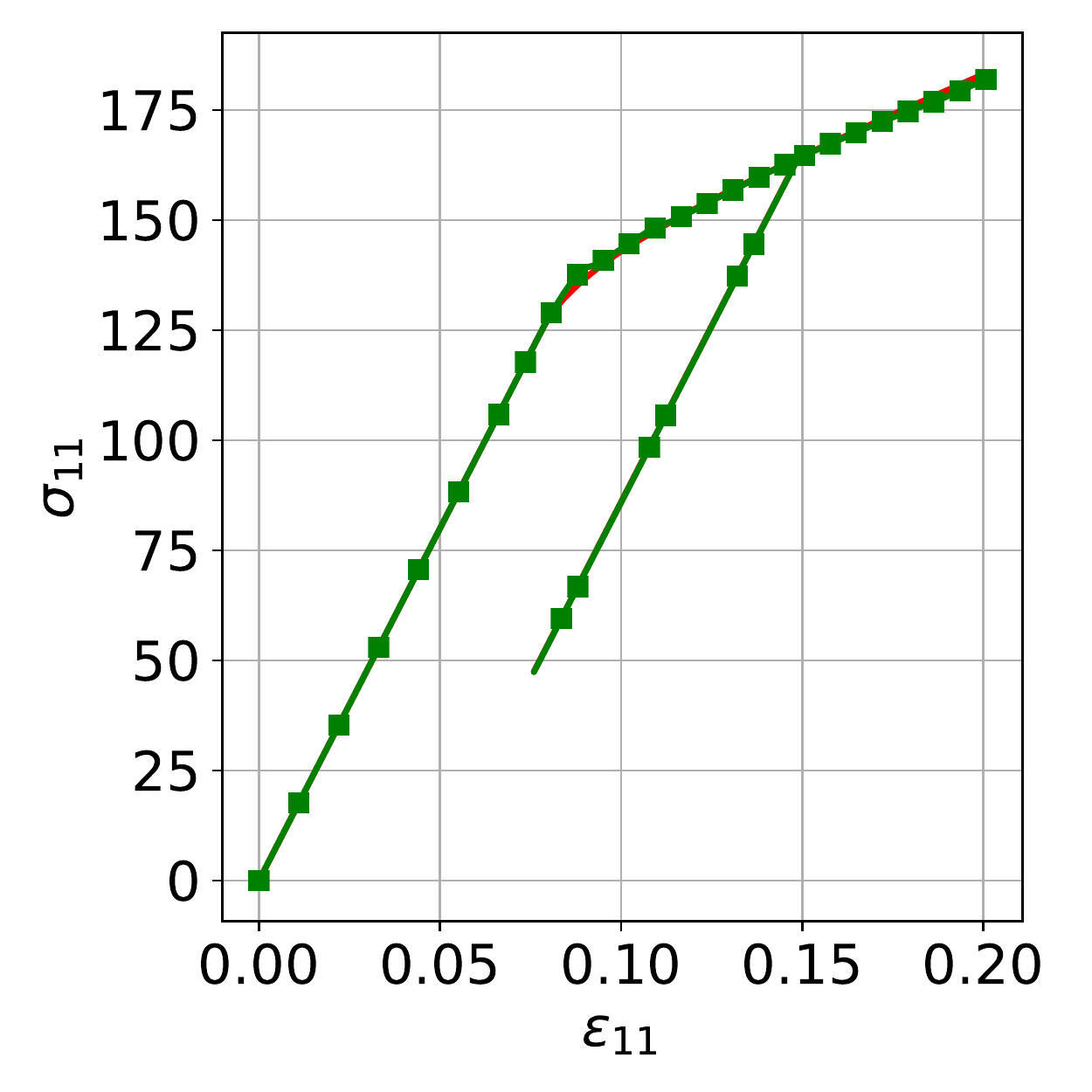}

\end{tabular}
\includegraphics[width=.7\textwidth]{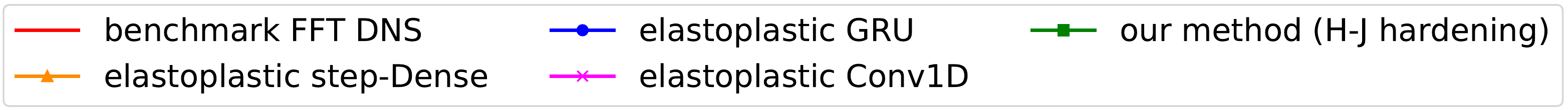} 

\caption{Comparison of black-box neural network architectures trained on monolithic data with our Hamilton-Jacobi hardening elastoplastic framework (introduced in Section~\ref{sec:framework}). The black-box models can capture the monolithic loading path (a) but cannot capture any unloading paths (b). Our framework can capture both even though it has only seen monolithic data.}
\label{fig:monolithic_networks}
\end{figure}

Initially, we train all the recurrent architectures with the 140 cases of monolithic loading (ranging from 200 to 400 deformation states per case), sampled radially from the $\pi$-plane, that were also used to train the approximator $\hat{f}$.
All models are expected to perform adequately well in blind predictions for monolithic testing cases as they have been trained for these simple patterns as seen in Fig.~\ref{fig:monolithic_networks} (a). However, the recurrent networks are expected to not be able to predict more complex loading and unloading paths as they have adequate information to recover the unseen elastic unloading paths. The black-box architectures fail to recover even a single unloading and unloading path, as seen in Fig.~\ref{fig:monolithic_networks} (b). The yield function model appears to be able to recover loading and unloading patterns well even though it was only exposed to monolithic loading paths.

\begin{figure}[h!]
\newcommand\siz{.23\textwidth}
\centering
\begin{tabular}{M{.01\textwidth}M{.24\textwidth}M{.24\textwidth}M{.24\textwidth}M{.24\textwidth}}
\hspace{-2.1cm}(a) &
\hspace{-2.5cm}\includegraphics[width=.23\textwidth ,angle=0]{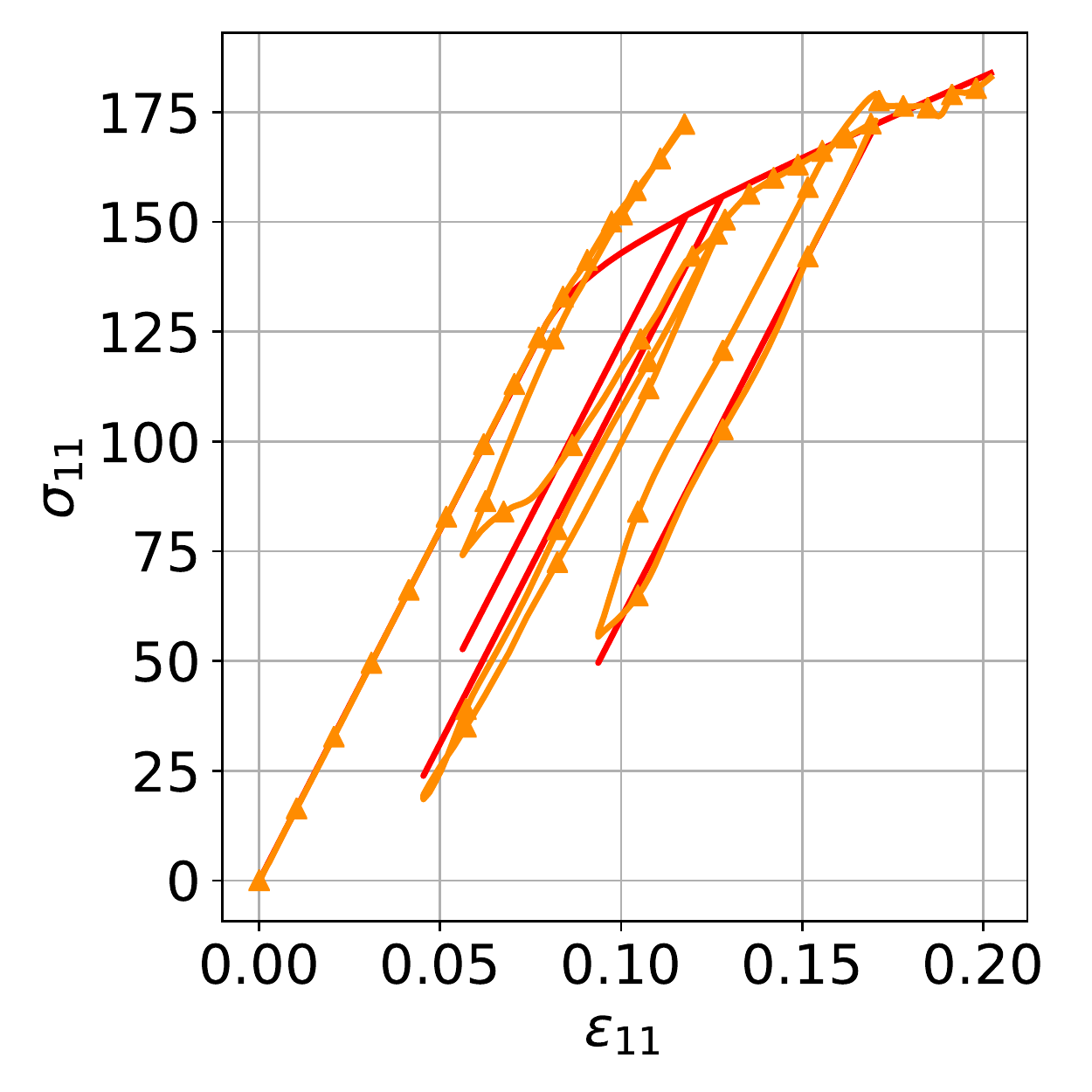} &
\hspace{-2.5cm}\includegraphics[width=.23\textwidth ,angle=0]{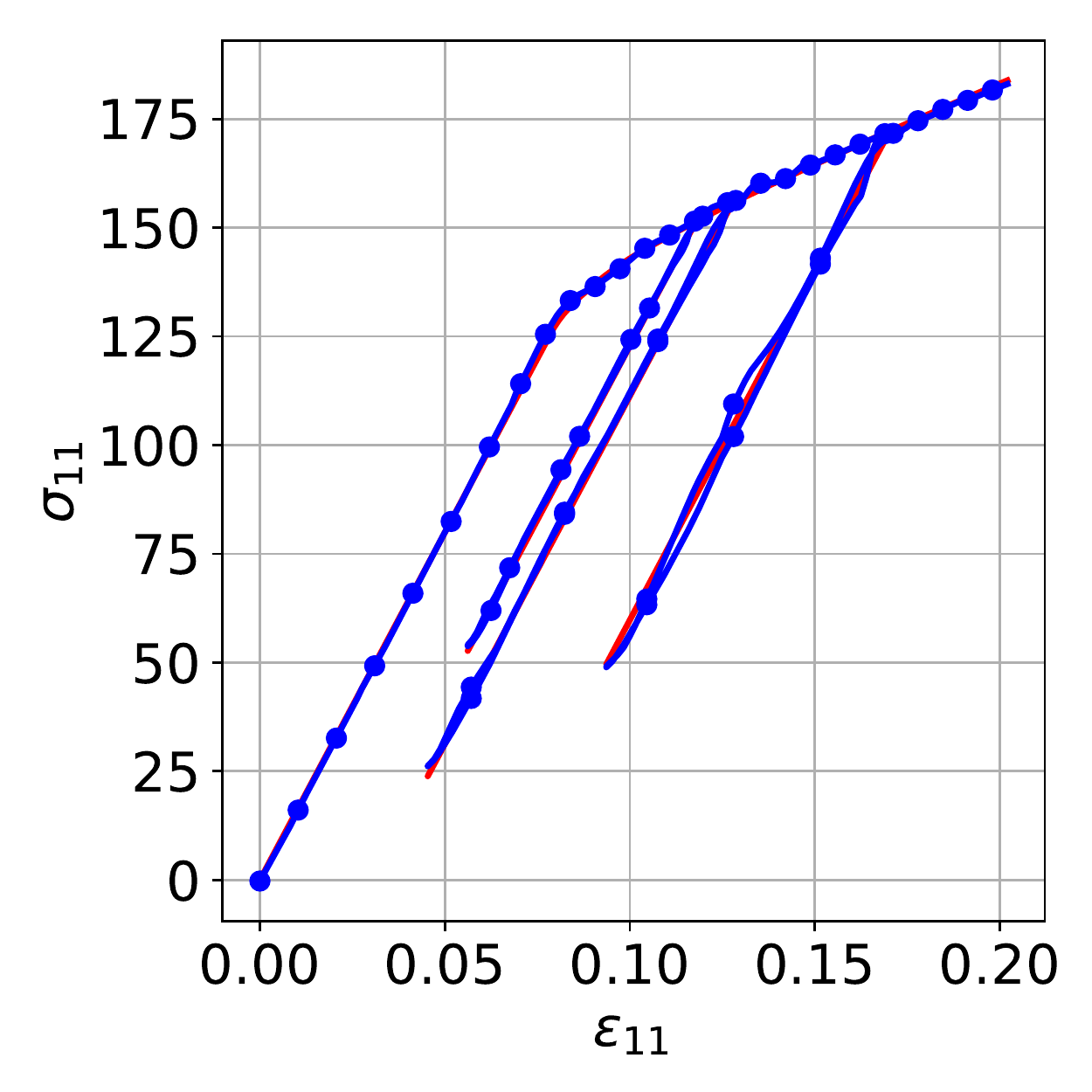} &
\hspace{-2.5cm}\includegraphics[width=.23\textwidth ,angle=0]{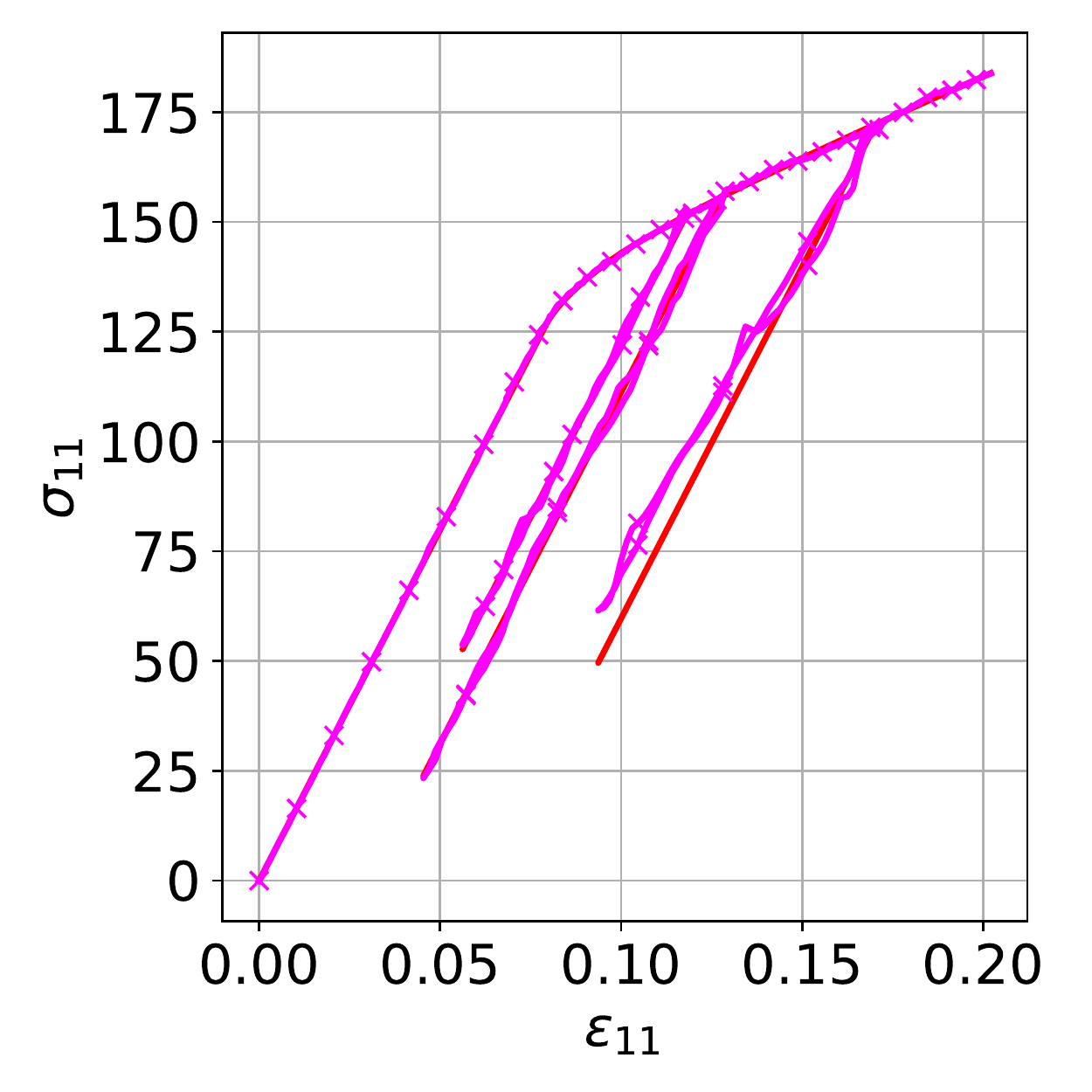} &
\vspace{0.2cm}\hspace{-2.5cm}\includegraphics[width=.23\textwidth ,angle=0]{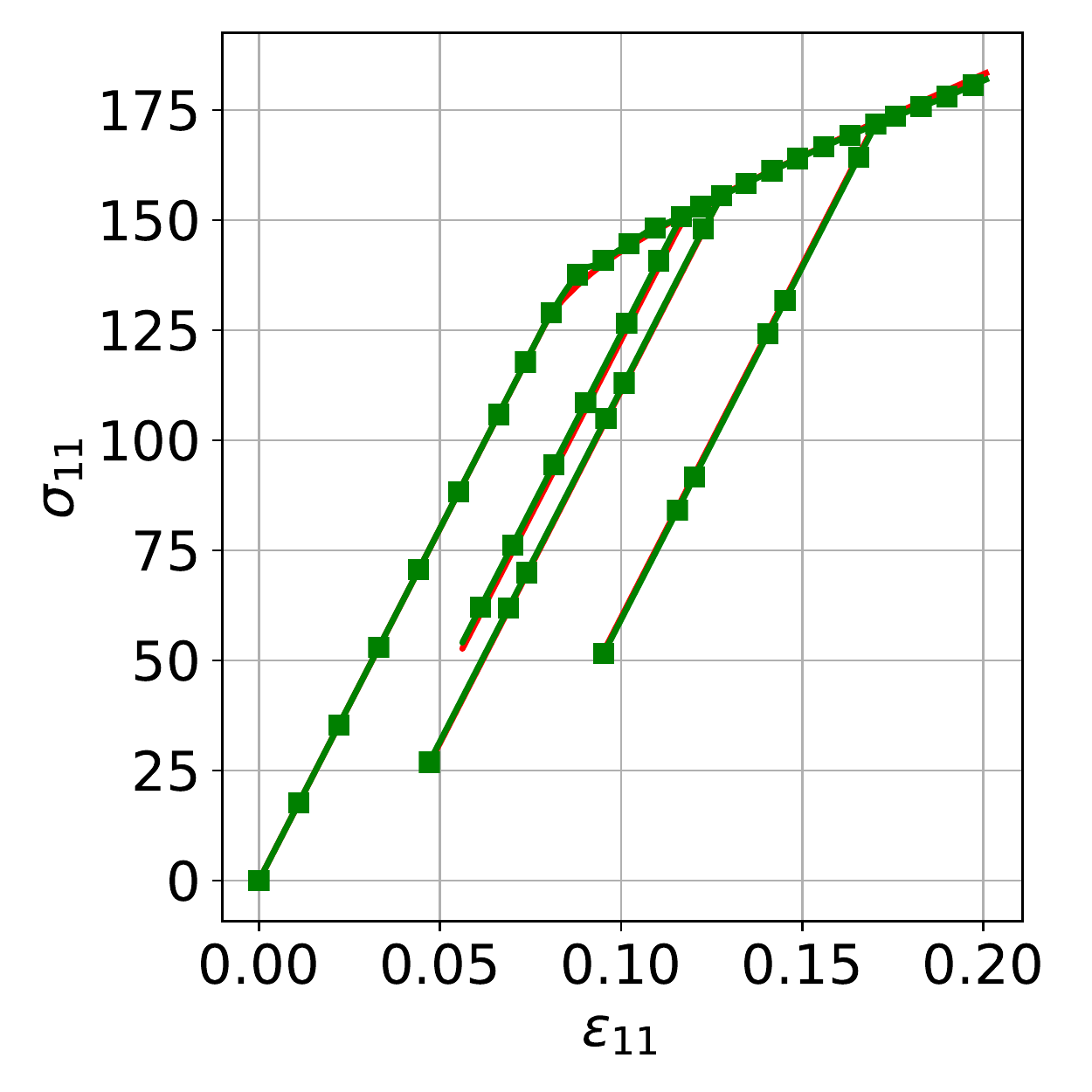} \\

\hspace{-2.1cm}(b) &
\hspace{-2.5cm}\includegraphics[width=.23\textwidth ,angle=0]{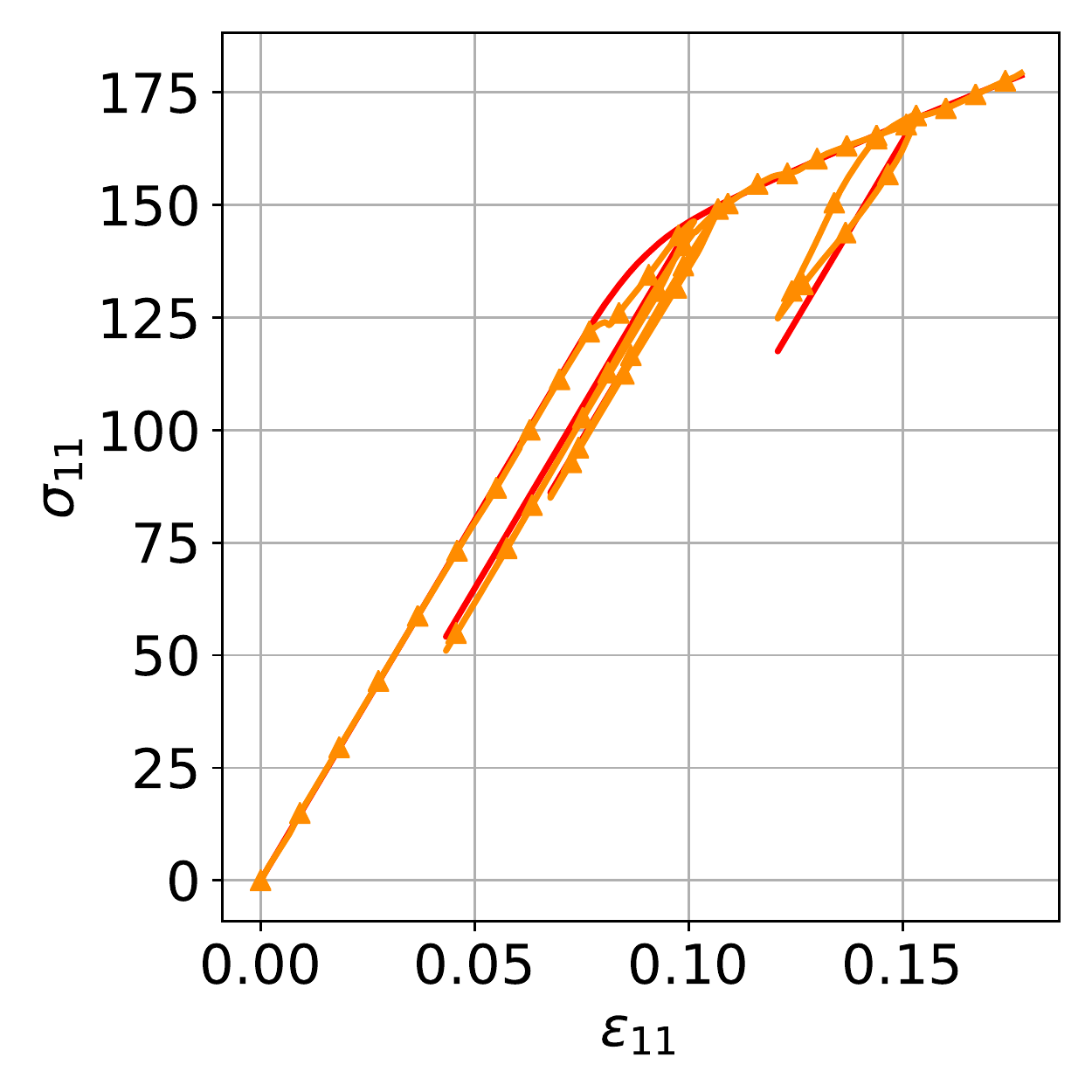} &
\hspace{-2.5cm}\includegraphics[width=.23\textwidth ,angle=0]{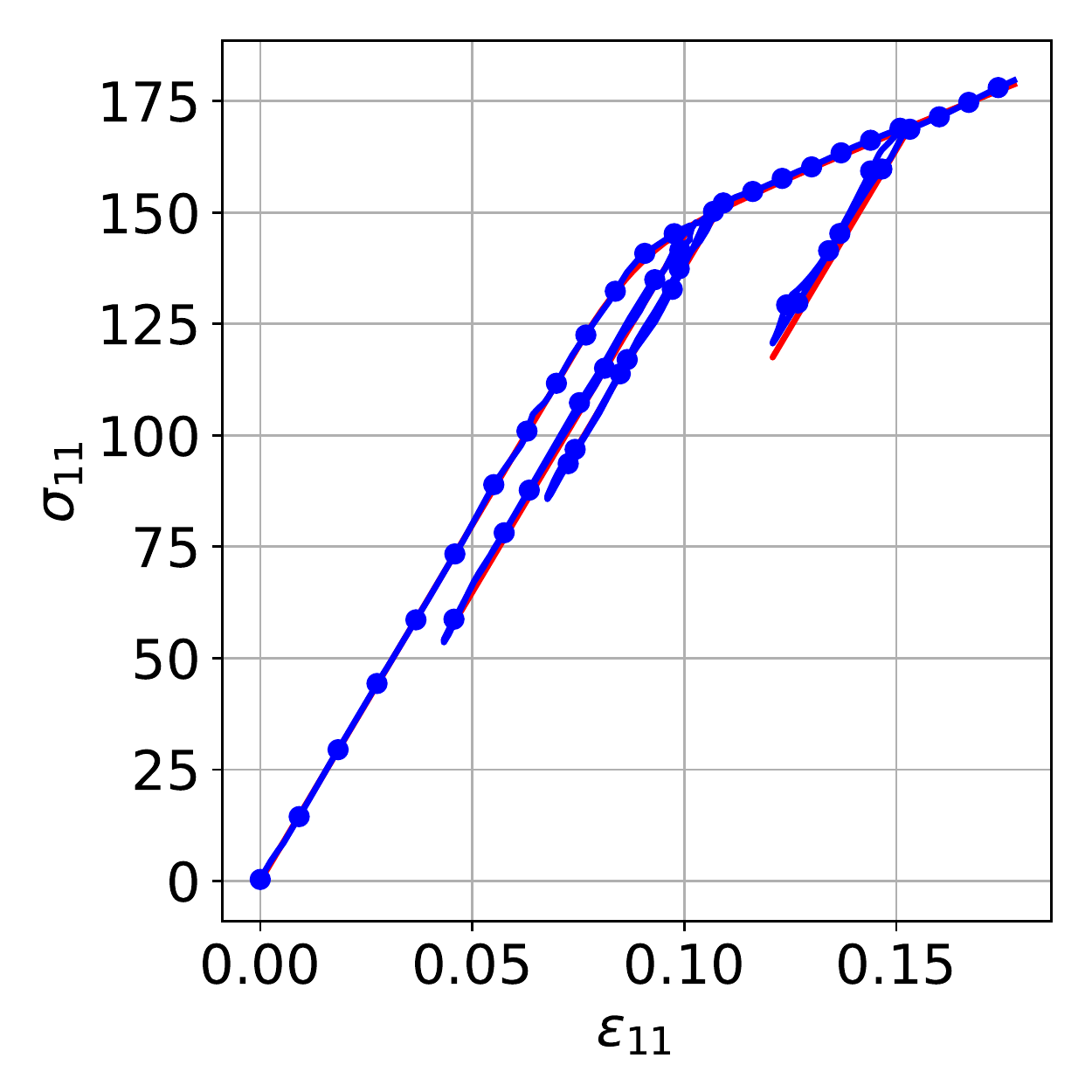} &
\hspace{-2.5cm}\includegraphics[width=.23\textwidth ,angle=0]{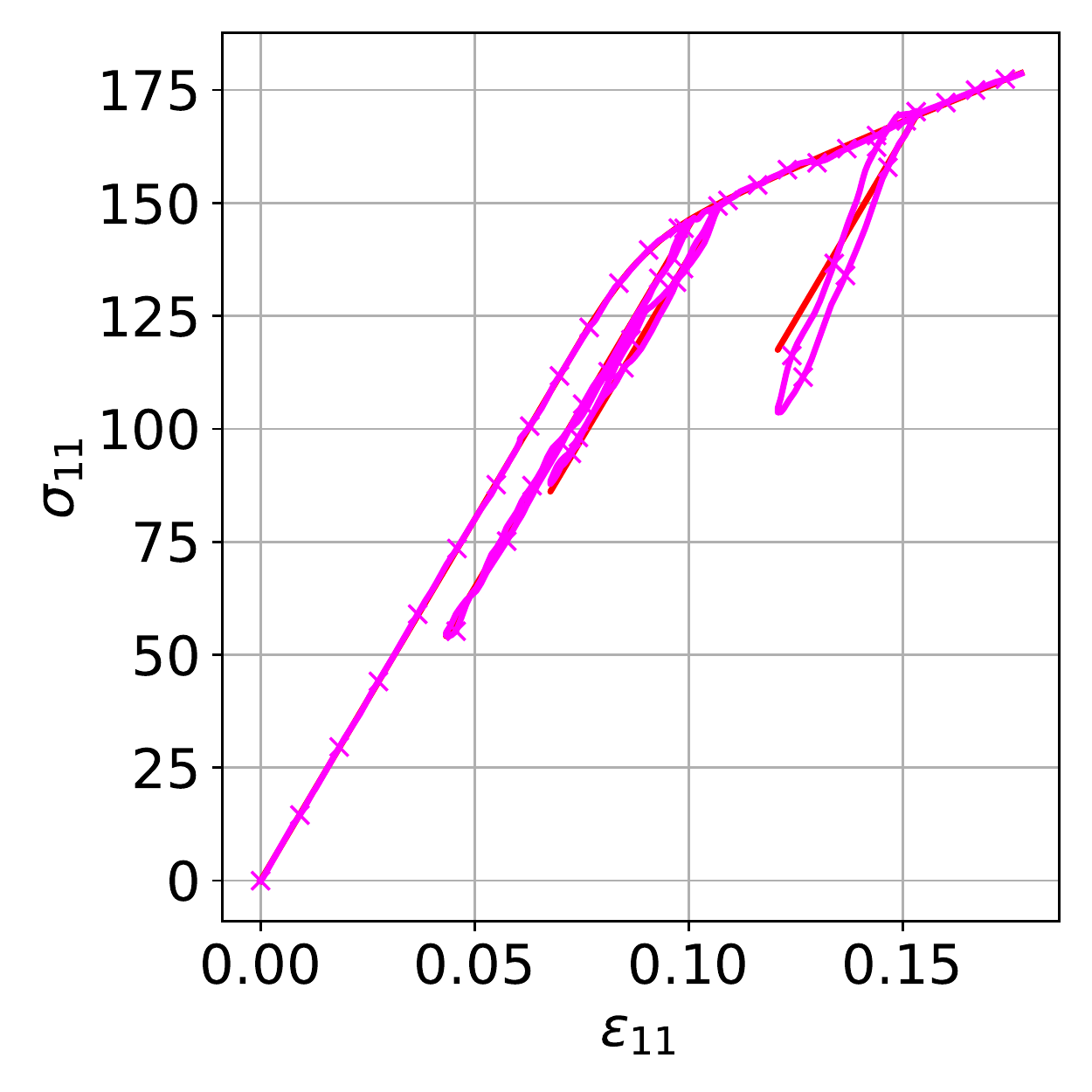} &
\vspace{0.2cm}\hspace{-2.5cm}\includegraphics[width=.23\textwidth ,angle=0]{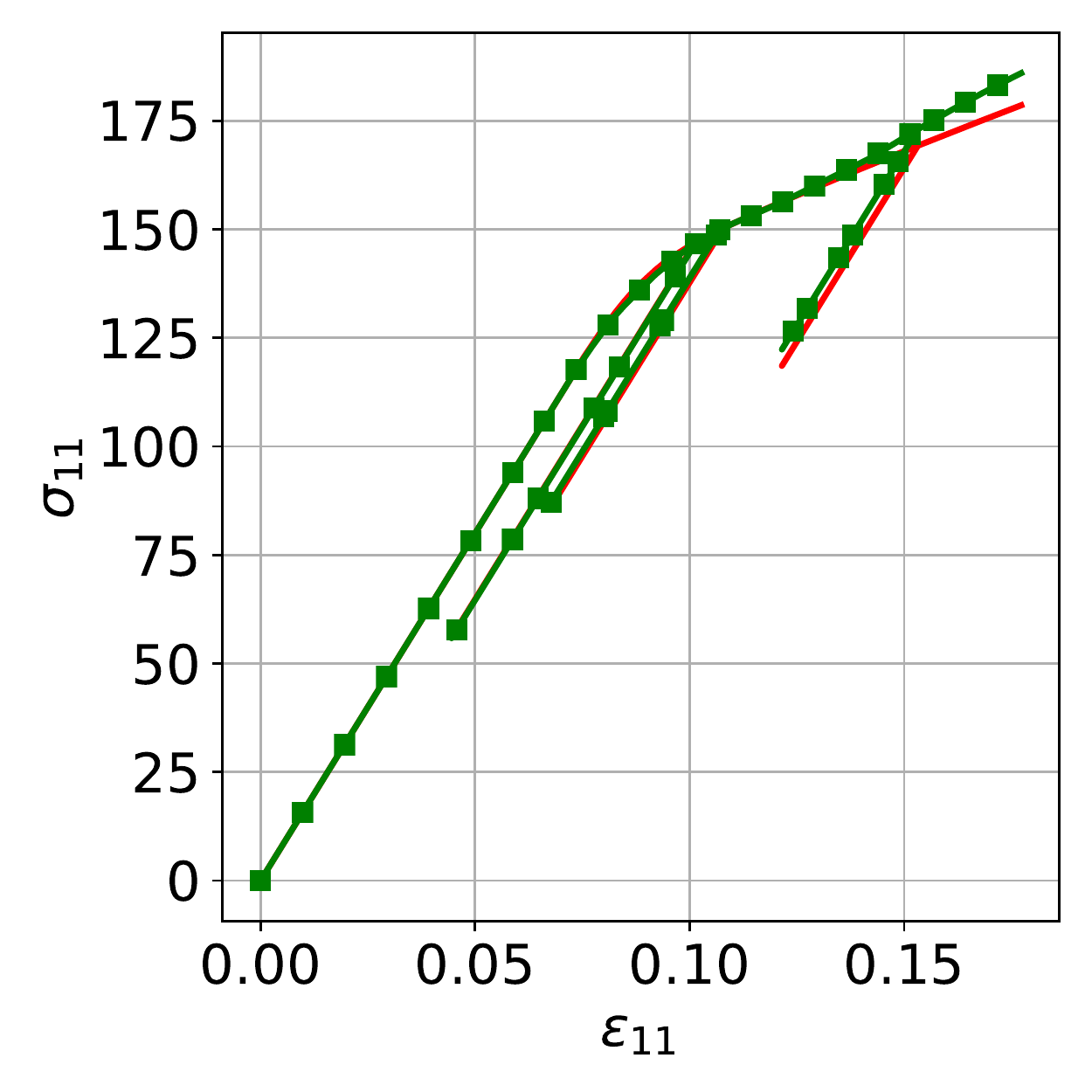} \\

\hspace{-2.1cm}(c) &
\hspace{-2.5cm}\includegraphics[width=.23\textwidth ,angle=0]{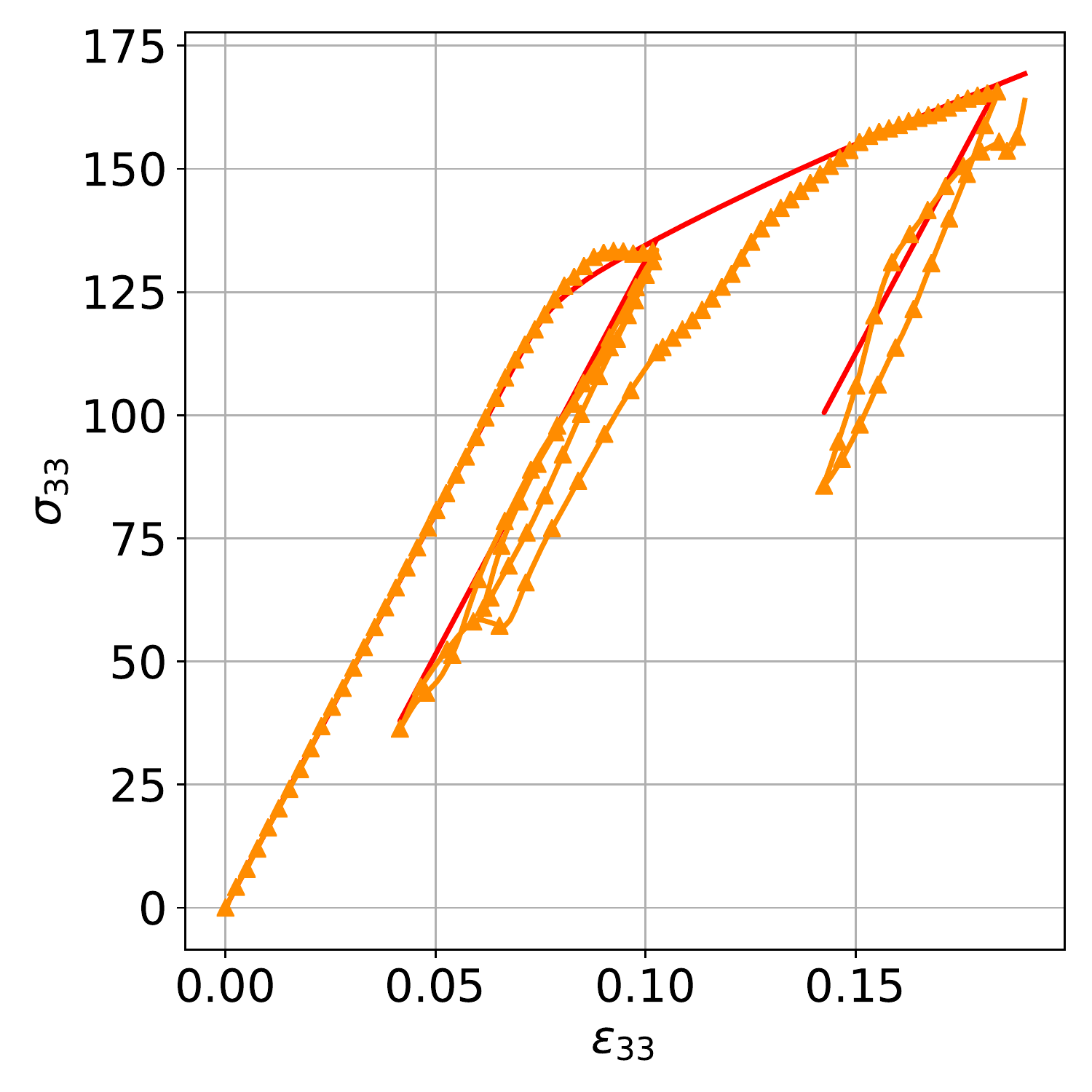} &
\hspace{-2.5cm}\includegraphics[width=.23\textwidth ,angle=0]{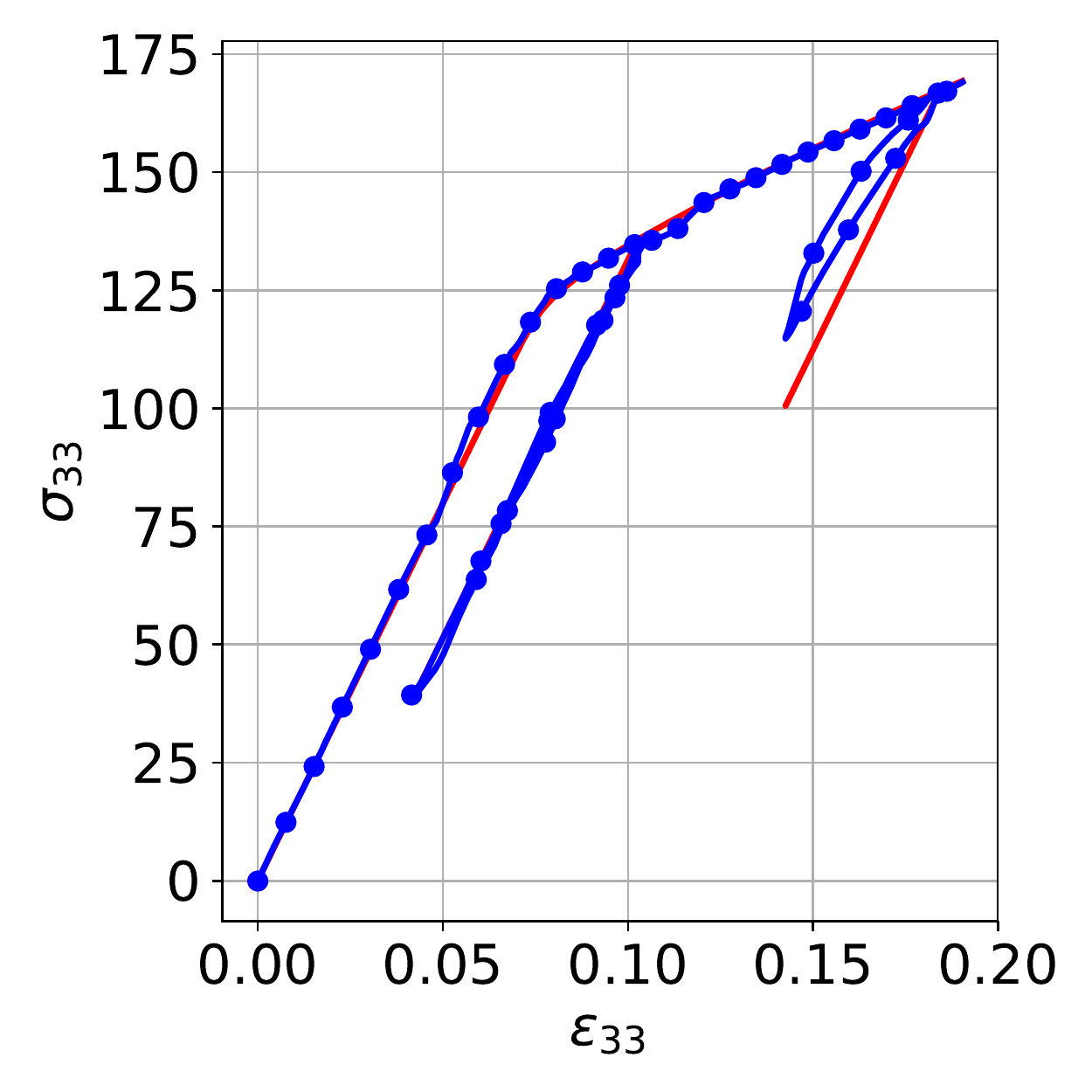} &
\hspace{-2.5cm}\includegraphics[width=.23\textwidth ,angle=0]{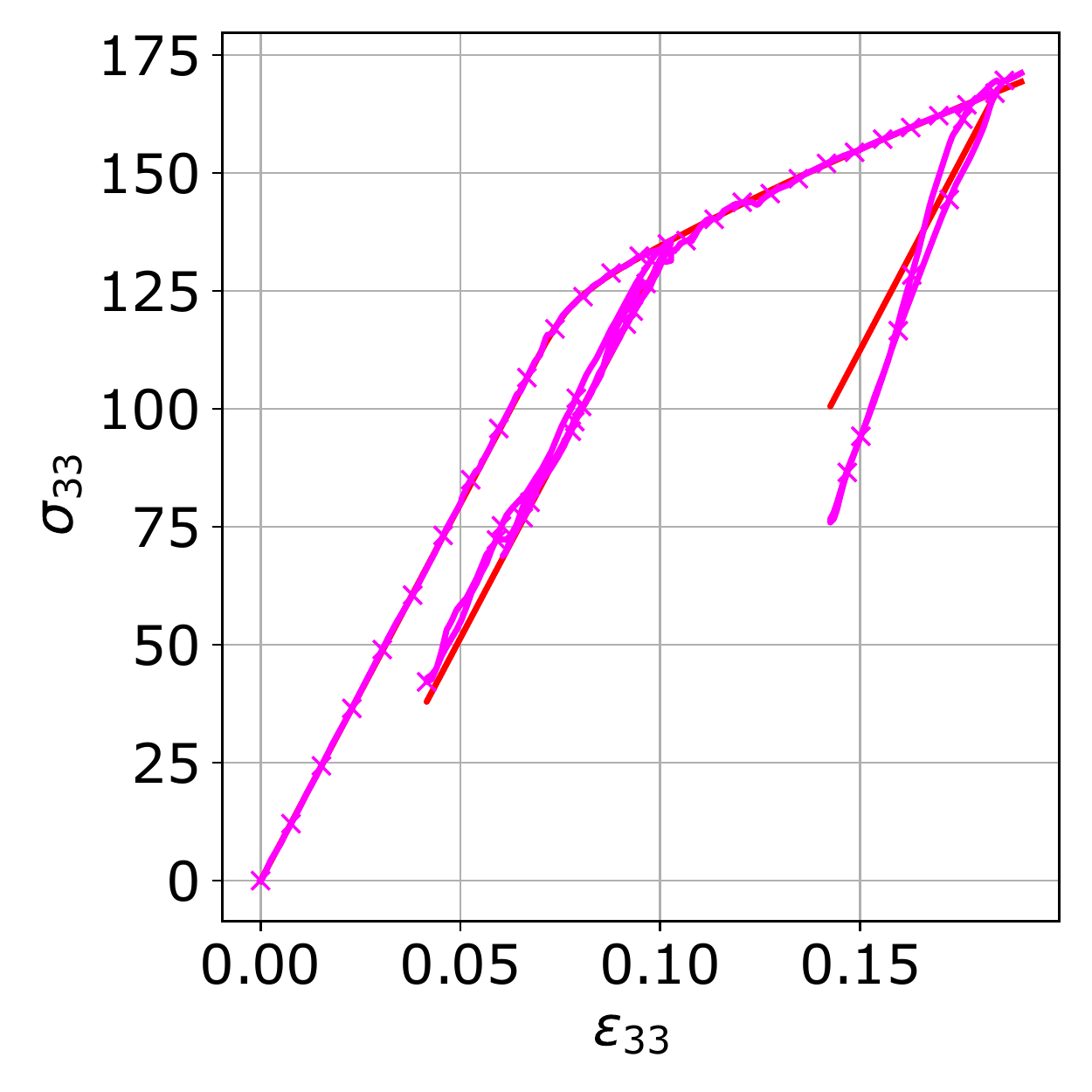} &
\vspace{0.2cm}\hspace{-2.5cm}\includegraphics[width=.23\textwidth ,angle=0]{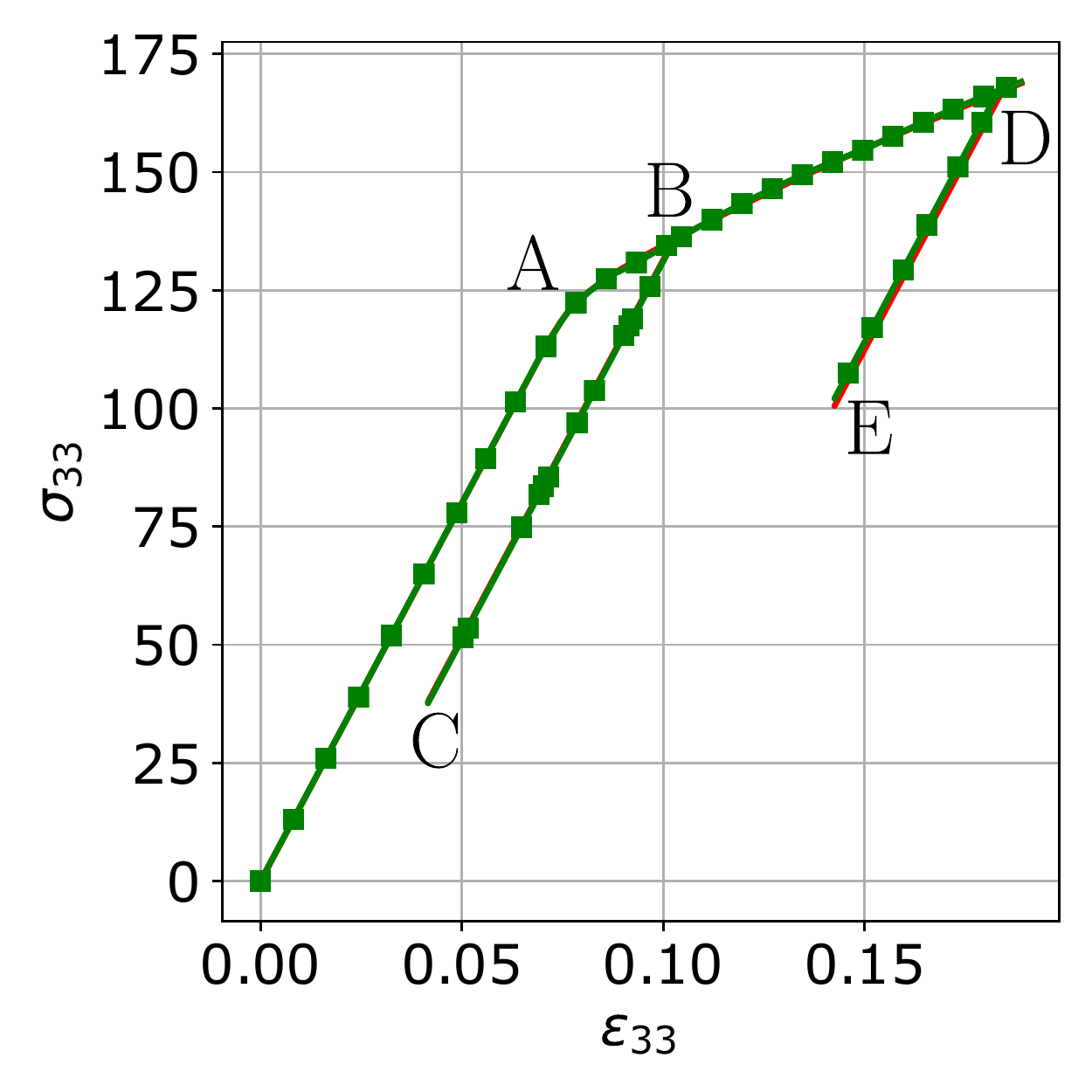} \\     

\end{tabular}
\includegraphics[width=.7\textwidth]{./figure/nn_comparison_legend.pdf} 

\caption{Comparison of black-box neural network architectures trained on random loading-unloading data with our Hamilton-Jacobi hardening elastoplastic framework. Three different cases of loading-unloading are demonstrated (a and b). The black-box models can capture loading-unloading behaviors better than the monolithic data trained ones but still show difficulty capturing some unseen unloading paths. Our framework appears to be more robust in loading-unloading path predictions -- even though it is only trained on monolithic data.}
\label{fig:multiunloading_networks}
\end{figure}

In the second numerical experiment, we increase the complexity of the database that the recurrent neural networks are trained on. Following the same loading path angles as a basis on the $\pi$-plane, we generate cases that now include complex unloading and reloading paths. We, thus, allow the recurrent architectures to be exposed to the previously missing elastic unloading paths. We randomly assign the unloading and loading paths randomly for every loading direction. At every direction, we randomly assign from 1 to 3 unloading and reloading paths with the unloading target strain also randomly chosen each time. Using this method, we generate we double the number of the sample points of the initial cases by adding random unloading and reloading patterns to retrain the recurrent architectures on. The performance of all the models is again compared against complex unseen loading and unloading  cases and the results for three testing cases can be seen in Fig.~\ref{fig:multiunloading_networks}.

\begin{figure}[h!]
\newcommand\siz{.25\textwidth}
\centering
\begin{tabular}{M{.25\textwidth}M{.25\textwidth}M{.25\textwidth}M{.25\textwidth}}

\hspace{-2.5cm}\includegraphics[width=.25\textwidth ,angle=0]{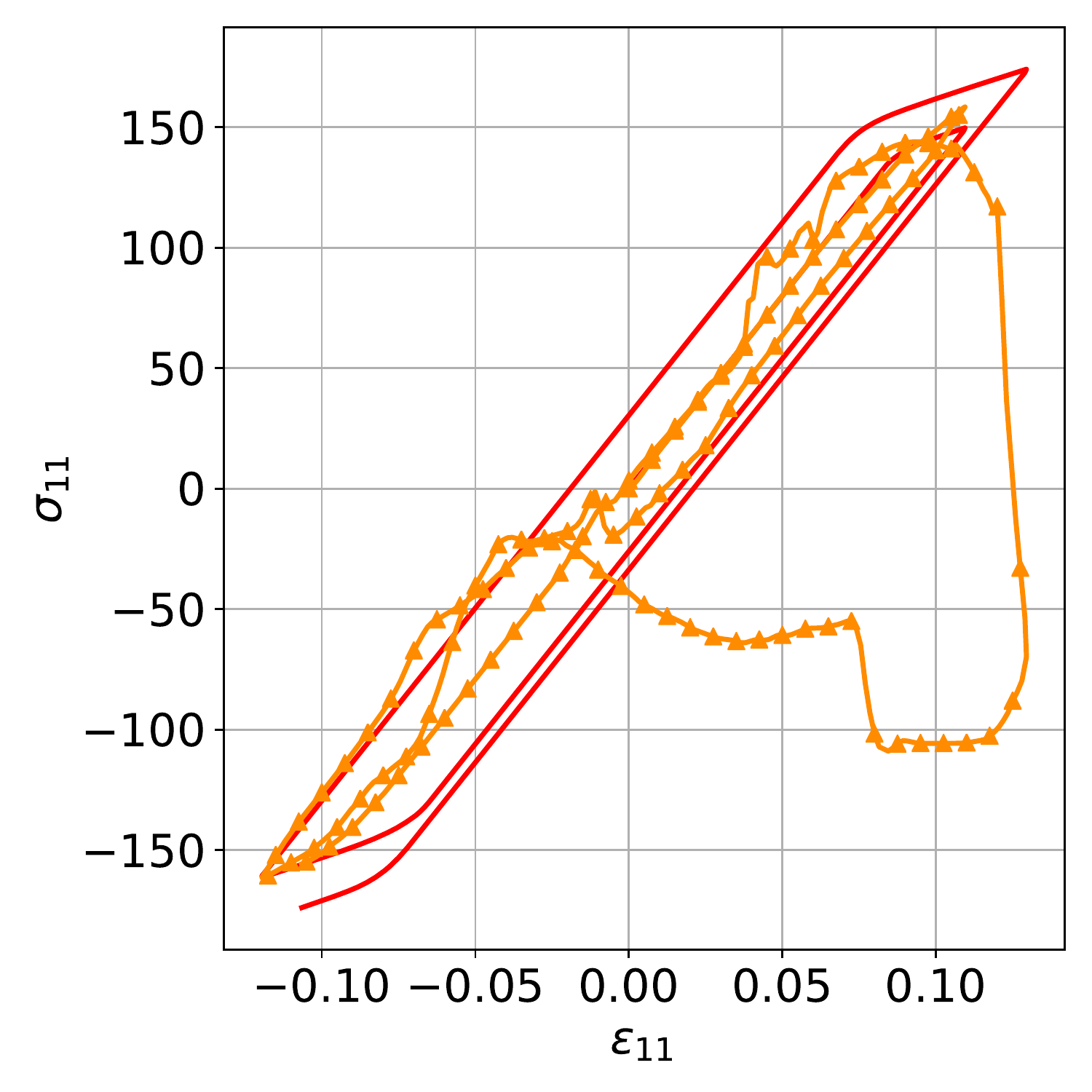} &
\hspace{-2.5cm}\includegraphics[width=.25\textwidth ,angle=0]{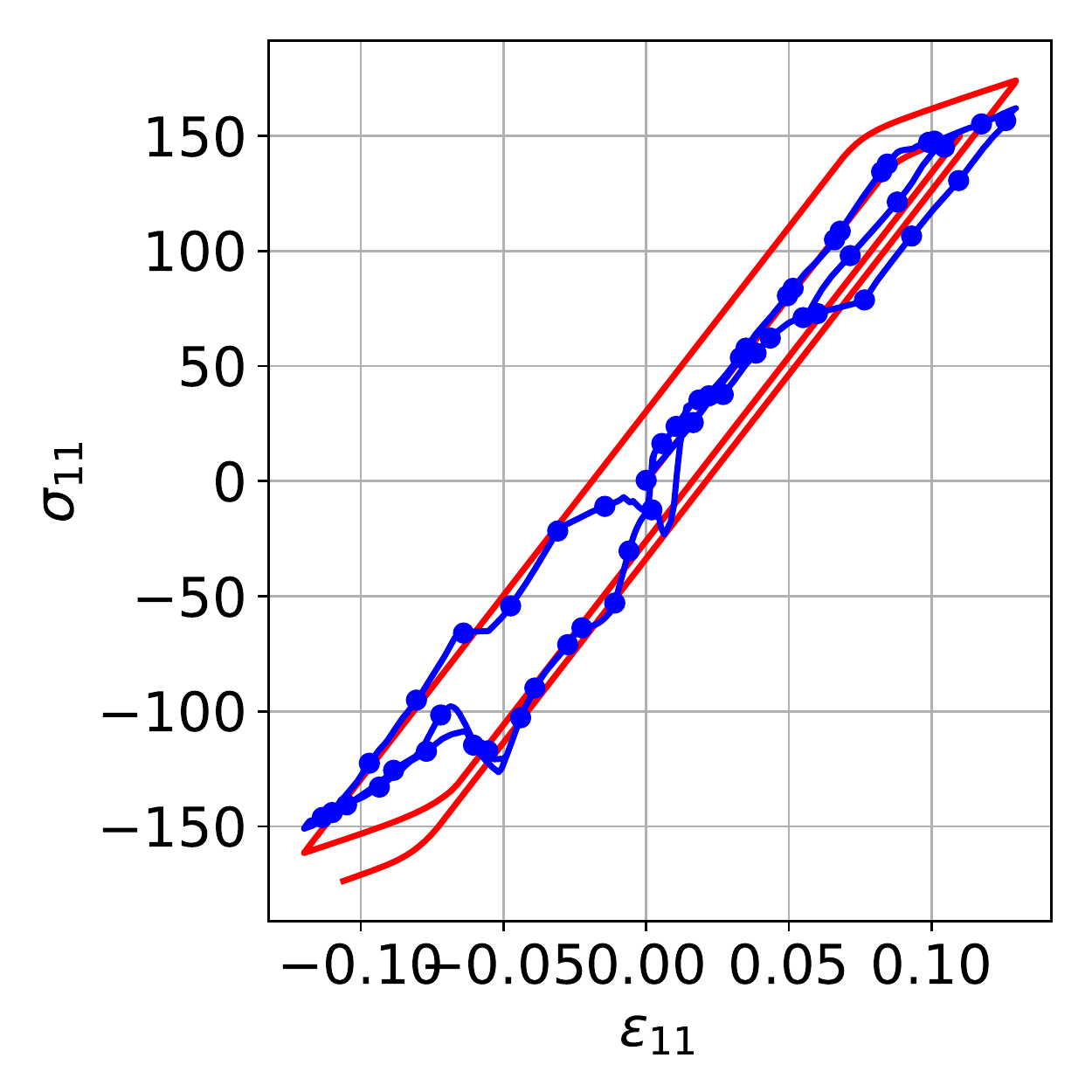} &
\hspace{-2.5cm}\includegraphics[width=.25\textwidth ,angle=0]{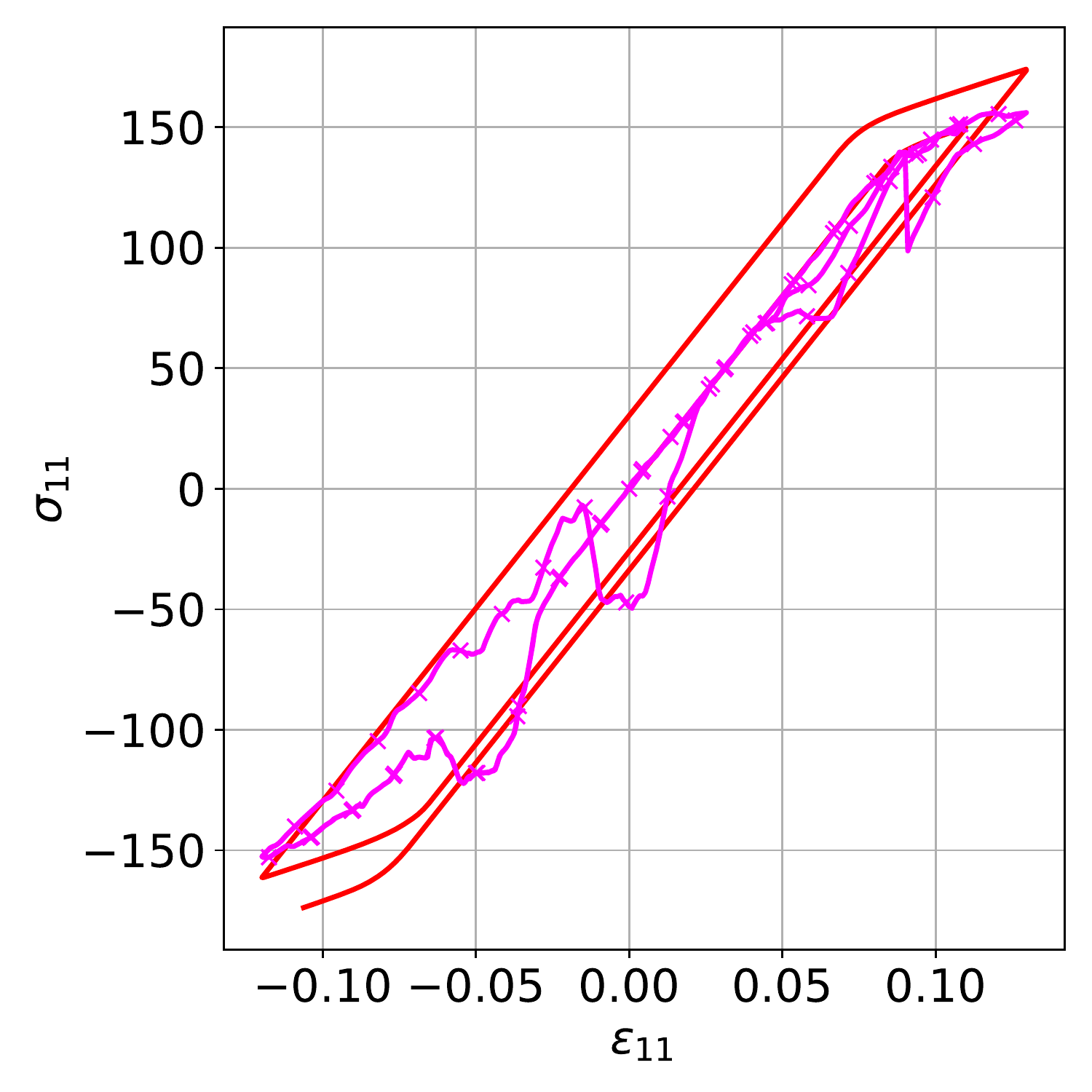} &
\hspace{-2.5cm}\includegraphics[width=.25\textwidth ,angle=0]{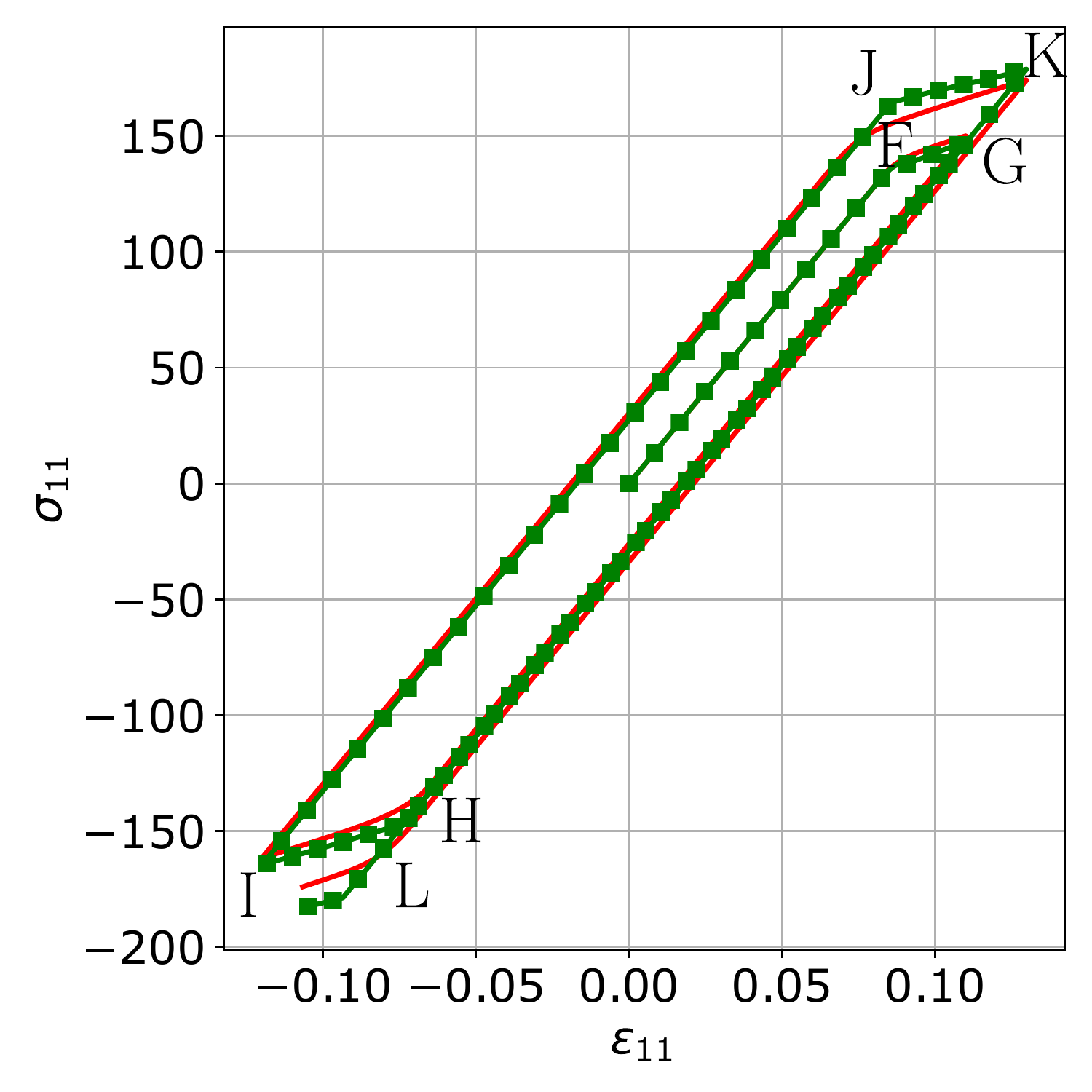} \\     

\end{tabular}
\includegraphics[width=.7\textwidth]{./figure/nn_comparison_legend.pdf} 

\caption{Comparison of black-box neural network architectures trained on random loading-unloading data with our Hamilton-Jacobi hardening elastoplastic framework for a cyclic loading path. }
\label{fig:cyclic_loading_comparison}
\end{figure}

As third comparison experiment, we test the models capabilities to predict cyclic loading and unloading paths. The results for the cyclic testing can be seen in Fig.~\ref{fig:cyclic_loading_comparison}. As expected, the black-box models -- even with an extended training data set -- fail to capture the cyclic behaviors. Our elastoplastic framework, while only trained with monolithic data, shows great capacity to capture cyclic behaviors.

\subsection{Application 2: Finite element simulations with machine learning derived polycrystal plasticity models}
\label{sec:computational_examples}

The return mapping algorithm of our elastoplastic neural network framework, described in Section~\ref{sec:return_mapping_algorithm}, is implemented in series of benchmark finite element simulations. The aim of these computational examples is to demonstrate the framework's ability to be integrated in multi-scale simulations by predicting the homogenized elastoplastic response of the microstructure -- completely replacing the local elastic and plastic constitutive laws with their data-driven counterparts. The return mapping algorithm demonstrated is fully generalized for any isotropic energy functional and isotropic yield function. 

\begin{figure}[h!]
\newcommand\siz{.95\textwidth}
\centering

\includegraphics[width=.95\textwidth ,angle=0]{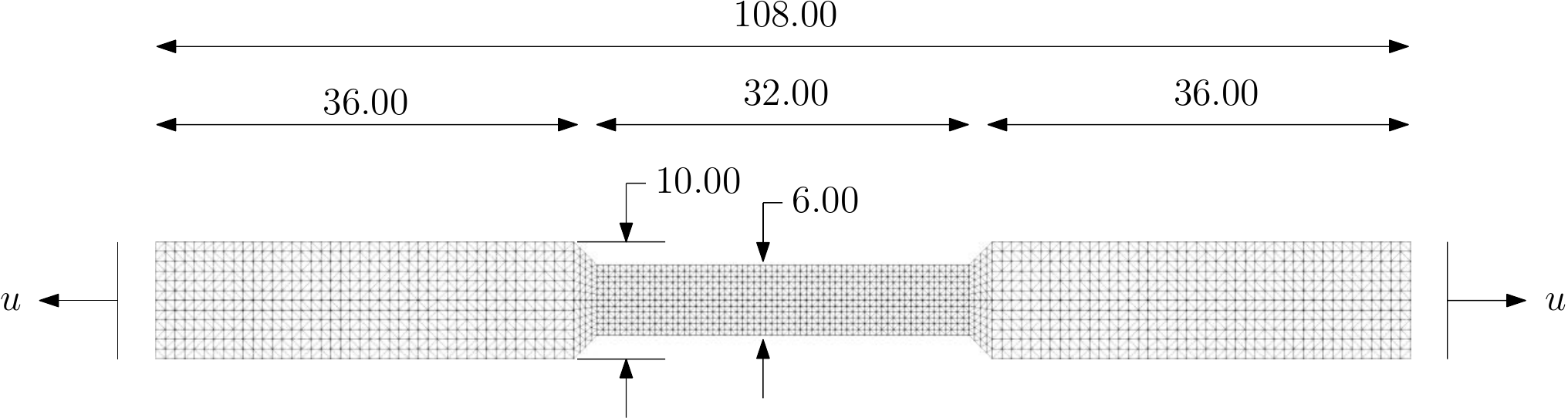}

\caption{Macroscopic structure and boundary conditions used in finite element simulations. The domain is symmetric along the horizontal and vertical axes so only one quarter of the domain is modeled. The units are in mm.}
\label{fig:fem_mesh}
\end{figure}

We perform the finite element quasi-static simulation of macroscopic monotonic uniaxial displacement of a bar depicted in Fig.~\ref{fig:fem_mesh}. The domain is symmetric along the horizontal and vertical axes and the elasticity and plasticity model used are isotropic, thus, we are modeling one quarter of the domain to predict the symmetric behavior. The domain is meshed with 3800 triangular elements with an average side length of $6.75 \times 10^{-4}$ meters. The displacement $u$ is applied at the boundaries as shown in Fig.~\ref{fig:fem_mesh} in increments of $\Delta u = 5 \times 10^{-5}$ meters. The microscopic elastoplastic behavior of every material point in the mesh is predicted by an elastic energy functional neural and yield function level set neural network, integrated by the return mapping algorithm of Section~\ref{sec:return_mapping_algorithm}.

\begin{figure}[h!]
\newcommand\siz{7cm}
\centering
\begin{tabular}{M{7cm}M{7cm}M{2cm}}
\includegraphics[width=7cm ,angle=0]{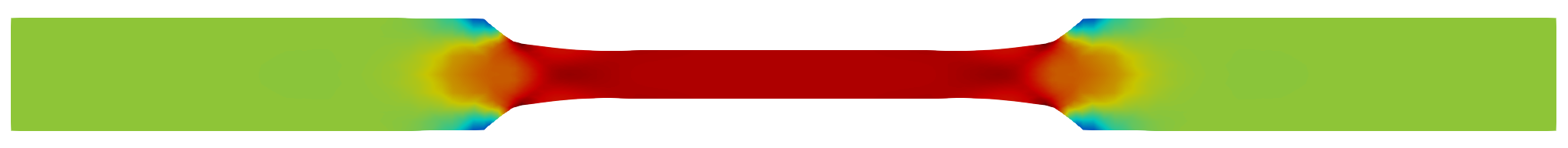} &
\includegraphics[width=7cm ,angle=0]{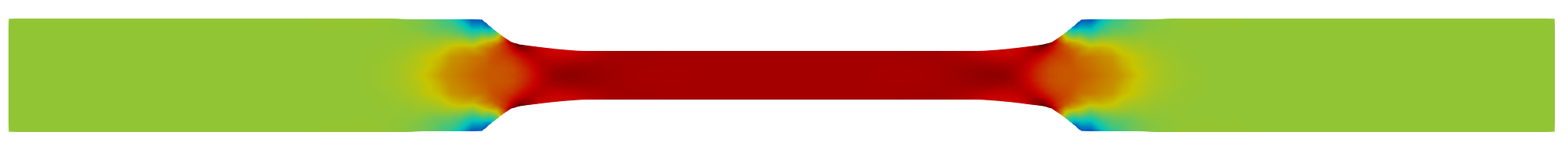} &   
\includegraphics[height=3cm,angle=0]{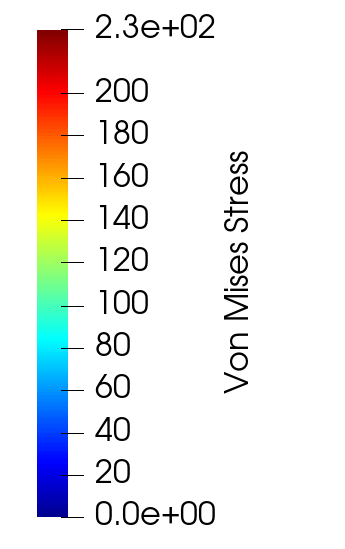} \\

\includegraphics[width=7cm ,angle=0]{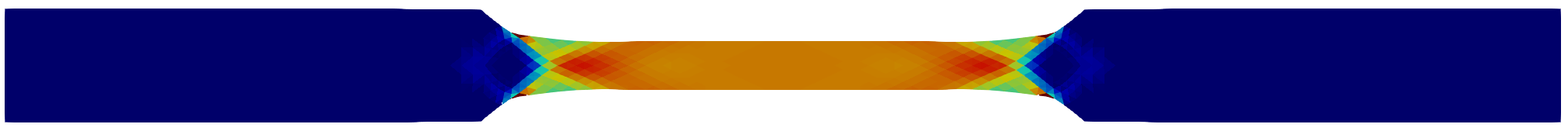} &
\includegraphics[width=7cm ,angle=0]{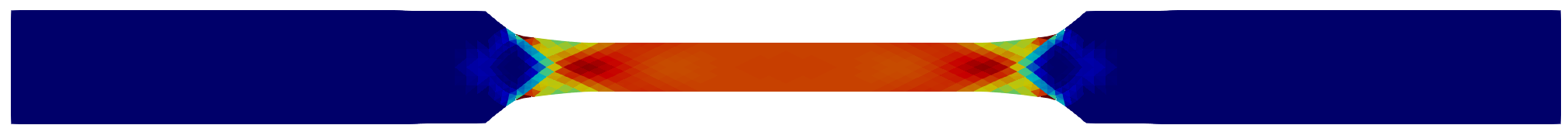} &   
\hspace{-0.2cm}\includegraphics[height=3cm,angle=0]{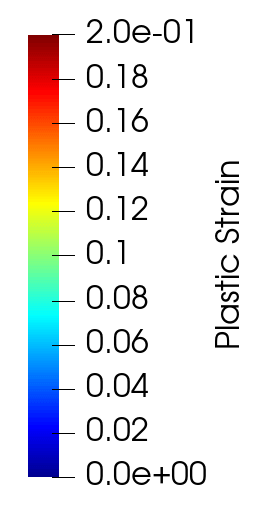} \\

\end{tabular}
\caption{Von Mises stress (top) and accumulated plastic strain (bottom) for the benchmark J2 plasticity (left) simulation and neural network J2 yield function (right) FEM simulations.}
\label{fig:j2_plasticity_simulations}

\end{figure}

As a first numerical verification exercise, we combine a quadratic energy functional of linear elasticity and J2 plasticity yield function with isotropic hardening. The neural networks and their training for the elastic response has been described in Section~\ref{sec:small_strain_sobolev} and, for the plastic response, in Sections~\ref{sec:yield_function_training} and \ref{sec:verification_j2}. The goal displacement for the uniaxial loading simulation is $u_{\text{goal}}=5.5 \times 10^{-3}$ meters. The results at the goal displacement for the benchmark solution and our elastoplastic Hamilton-Jacobi hardening framework are demonstrated in Fig.~\ref{fig:j2_plasticity_simulations} and appear to be in close agreement. 

\begin{figure}[h!]
\newcommand\siz{0.8\textwidth}
\centering
\vspace{1.5cm}
\begin{tabular}{M{0.8\textwidth}M{0.2\textwidth}}
\vspace{-1.5cm}\includegraphics[width=0.8\textwidth ,angle=0]{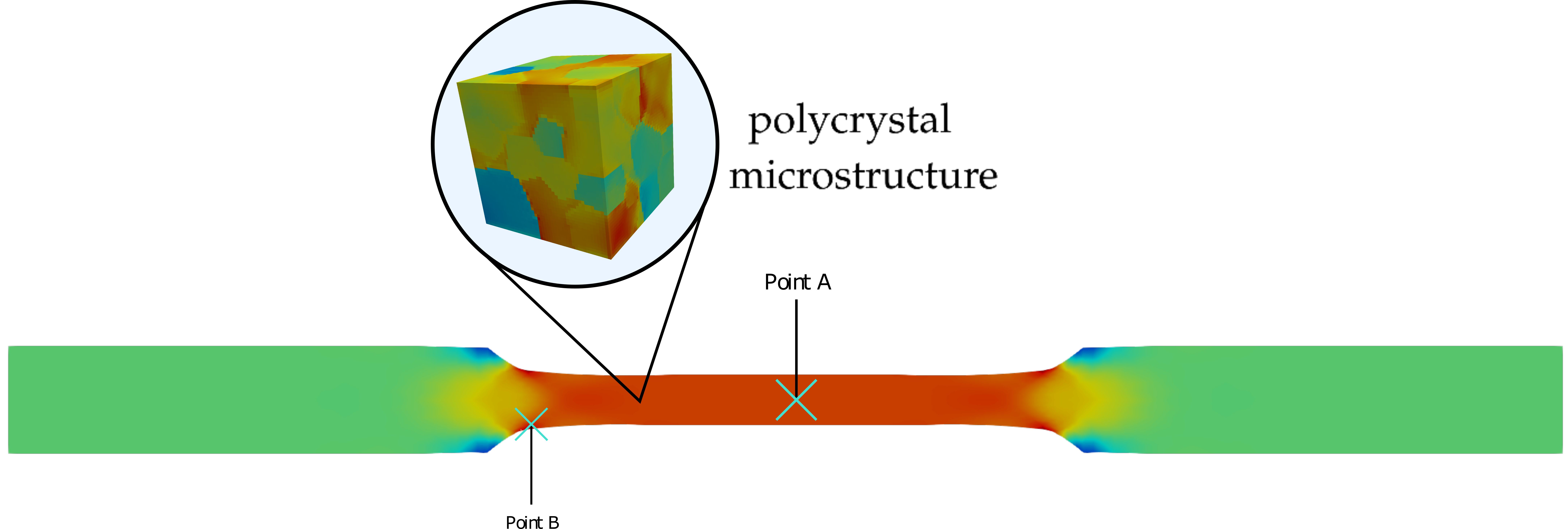} &
\includegraphics[height=3cm,angle=0]{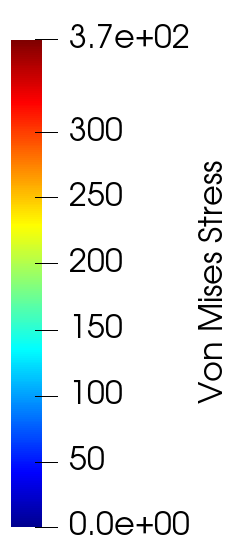} \\

\includegraphics[width=0.8\textwidth ,angle=0]{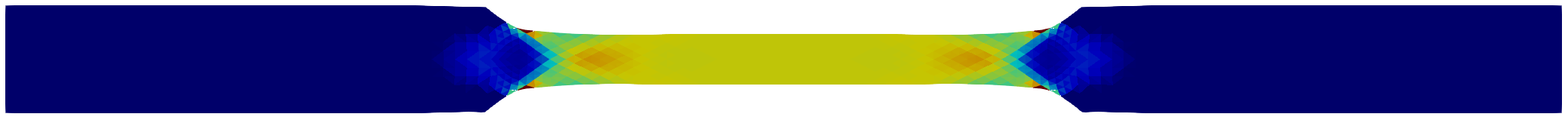} &
\includegraphics[height=3cm,angle=0]{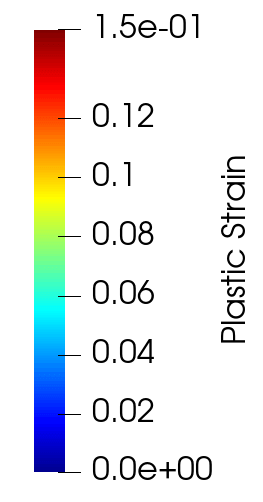} 

\end{tabular}

\caption{Von Mises stress (top) and accumulated plastic strain (bottom) for the neural network polycrystal yield function FEM simulations.}
\label{fig:polycrystal_plasticity_simulations}
\end{figure}

In a second numerical experiment, we are simulating the behavior under uniaxial loading of the domain in which the material points represent a polycrystalline microstructure. The elastoplastic framework in this simulation consists of a quadratic hyperelastic energy functional neural network and a polycrystal yield function the training of which is described in Sections~\ref{sec:small_strain_sobolev} and \ref{sec:yield_function_training} respectively. Both networks were trained on FFT simulation data as described in Appendix~\ref{sec:dataset_yield} to predict the homogenized elastoplastic behavior of the polycrystal.  In prevous work, the polycrystal incremental constitutive behavior has been calculated through a coupling of the FFT and FEM method (e.g. \citet{kochmann2016two,kochmann2018efficient}). 

\begin{figure}[h!]
\newcommand\siz{0.3\textwidth}
\centering
\begin{tabular}{M{0.3\textwidth}M{0.3\textwidth}M{0.4\textwidth}}

\hspace{-1.5cm}\includegraphics[width=0.3\textwidth ,angle=0]{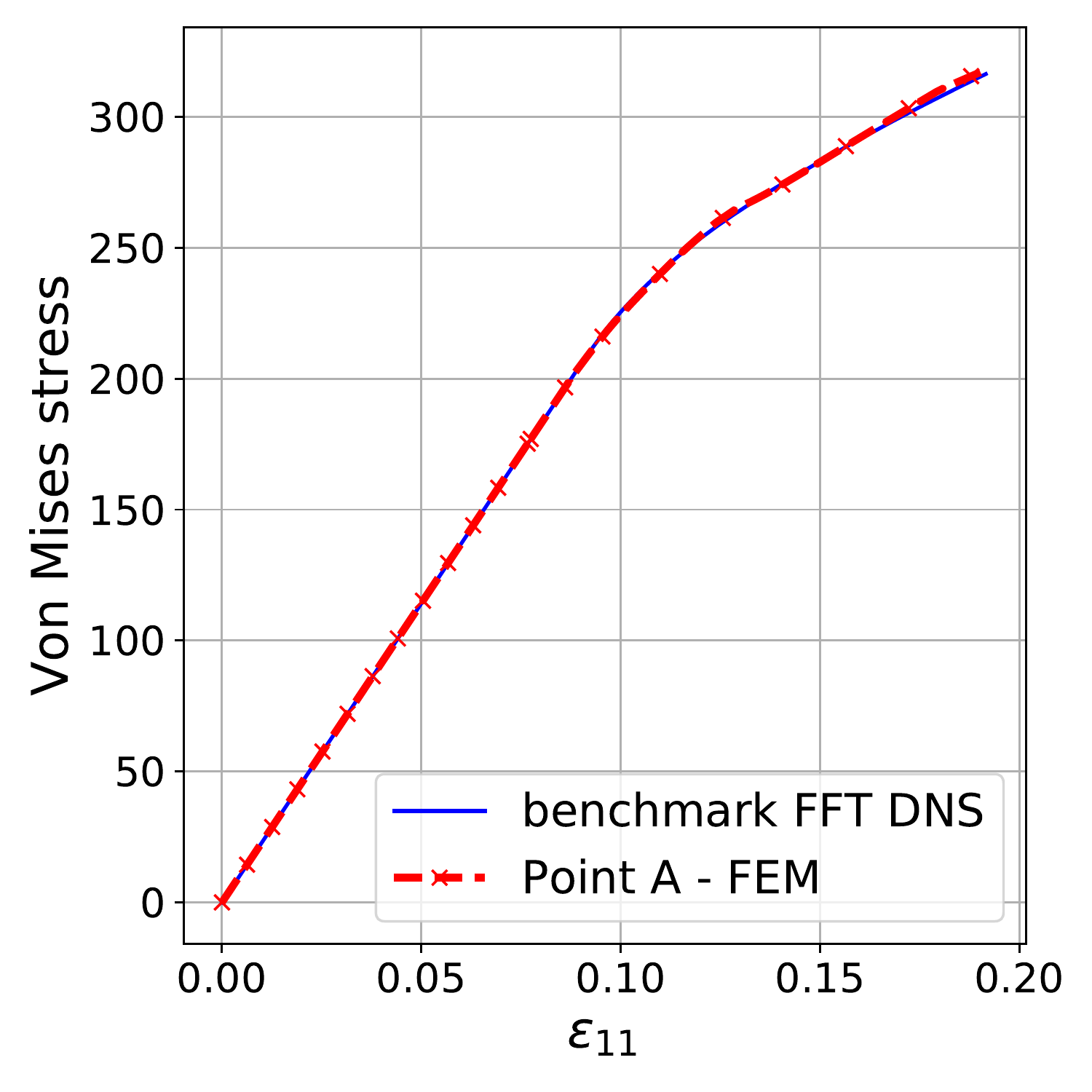} &
\hspace{-1.5cm}\includegraphics[height=0.3\textwidth ,angle=0]{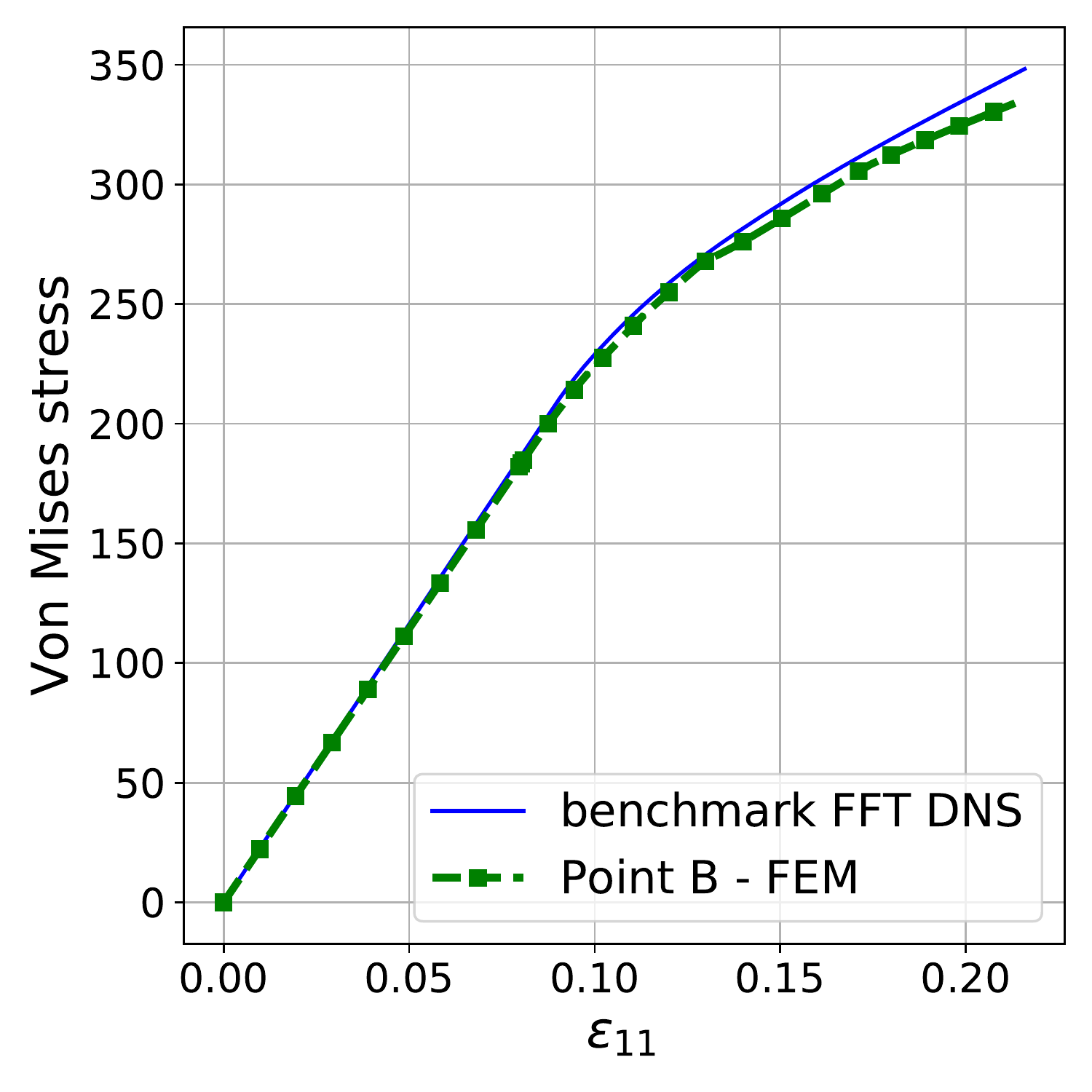} &
\hspace{-1.5cm}\includegraphics[height=0.4\textwidth ,angle=0]{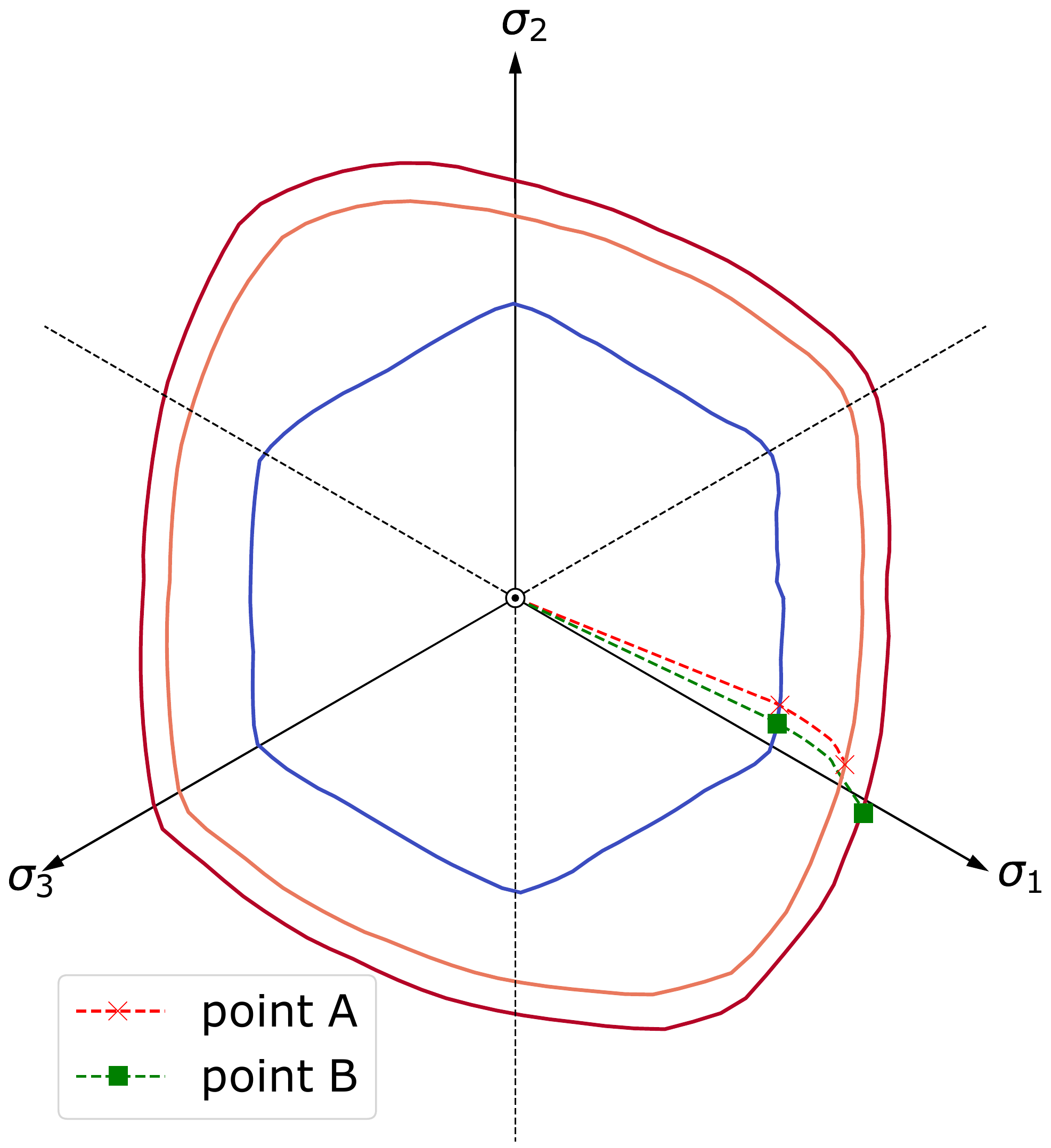} 
\end{tabular}
\caption{Von Mises stress curves and stress paths on the $\pi$-plane for Points A and B of the domain of Fig.~\ref{fig:polycrystal_plasticity_simulations}.}
\label{fig:pick_points_FEM}
\end{figure}

However, the efficiency of these methods depends on the heterogeneity of the polycrystal as it can affect the computational cost of the simulations for a large number of crystals and the stability when there are sharp material property differences. In the current work, there is no need for online FFT simulations to be run in parallel with the FEM simulations. The neural network training database for the elasticity and the plasticity are built separately offline for a discrete number of FFT simulations and the trained networks will be interpolating the behaviors and making blind predictions during the FEM simulation. The results of the simulation for the Von Mises stress and the accumulated plastic strain for the simulation at the displacement goal of $u_{\text{goal}}=6.5 \times 10^{-3}$ meters is demonstrated in Fig.~\ref{fig:polycrystal_plasticity_simulations}. The stress curves and stress paths on the planes for two points of the domain are also demonstrated in Fig.~\ref{fig:pick_points_FEM}.

\section{Conclusions}
The history of plasticity theory is influenced by the geometrical interpretations of mechanics concepts in different parametric spaces
 \citep{de1870memoire, lode1926versuche, hill1998mathematical, rice1971inelastic}. Forming a vector space that uses different invariants 
 or measures of stress as orthogonal bases had helped us understand 
yielding and the subsequent hardening and softening through visualization. 
However,  these new mechanisms often take decades to be discovered and 
adopted by the mechanics community. 
In this work, our contributions are twofold. First, we leverage the geometrical interpretation of plasticity theory to establish a connection 
between elastoplasticity and level set theories. Second, we introduce a 
new variety of deep machine learning that is designed to train functionals with sufficient smoothness.  
By using higher-order training 
to regularize the continuity and smoothness of the energy functional, 
the yield function, the flow rules and the hardening mechanisms, 
we create a framework that retains the simplicity afforded by the geometrical interpretation of the models without limiting our choices of elasticity, yield function and hardening mechanisms. 
Thermodynamic constraints can be easily checked and introduced, as the machine learning generated models are now 
geometrically interpretable. Finally, the most significant part of this research is that it provides a generalized framework where the yield function may form in any arbitrary shape and evolve in any generic way
that optimizes the quality of the predictions. As shown in the paper, the level set framework may manifest many classical plasticity models when given the corresponding data but it may also introduce new yield surface and hardening laws that are difficult to hand-craft. 
 
Comparisons against the black-box predictions commonly used to generate elasto-plasticity responses show that the approach in this paper is 
not only provide more robust and accurate responses for forward predictions, 
but are also more effective in training given the same set of data. 

\begin{appendix}
\normalsize{

\section{Appendix: Data generation for the hyperelasticity benchmark}
\label{sec:dataset_hyperelastic}

In this work, the numerical experiments (Section~\ref{sec:numerical_experiments}) are performed on synthetic data sets generated for two small strain hyperelastic laws. One of them is isotropic linear elasticity. The second is a small-strain hyperelastic law designed for the Modified Cam-Clay plasticity model \citep{roscoe1968generalized,houlsby1985use,borja2001cam}. The hyperelastic energy functional allows full coupling between the elastic volumetric and deviatoric responses and is described as:

\begin{equation}
\psi \left(\epsilon_{v}^{\mathrm{e}}, \epsilon_{s}^{\mathrm{e}}\right)=-p_{0} c_{r} \exp \left(\frac{\epsilon_{v 0}-\epsilon_{v}^{\mathrm{e}}}{\xi}\right)-\frac{3}{2} c_{\mu} p_{0} \exp \left(\frac{\epsilon_{v 0}-\epsilon_{v}^{\mathrm{e}}}{\xi}\right)\left(\epsilon_{s}^{\mathrm{e}}\right)^{2},
\label{eq:borja_hyperelastic_psi}
\end{equation}
where $\epsilon_{v0}$ is the initial volumetric strain, $p_0$ is the initial mean pressure when $\epsilon_v = \epsilon_{v0}$, $\xi > 0$ is the elastic compressibility index, and $c_\mu>0$ is a constant. The hyperelastic energy functional is designed to describe an elastic compression law where the equivalent elastic bulk modulus and the equivalent  shear modulus vary linearly with $-p$, while the mean pressure $p$ varies exponentially with the change of the volumetric strain $\Delta\epsilon_v = \epsilon_{v0} - \epsilon_v$. The specifics and the utility of this hyperelastic law is outside the scope of this current work and will be omitted. The numerical parameters of this model where chosen as $\epsilon_{v0} = 0$, $p_0= -100 $ KPa, $c_\mu = 5.4 $, and $\xi = 0.018$.
Taking the partial derivatives of the energy functional with respect to the strain invariants, the stress invariants are derived as:

\begin{equation}
p=\frac{\partial \psi }{\partial \epsilon_{v}^{\mathrm{e}}}=p_{0}\left(1+\frac{3 c_{\mu}}{2 \xi}\left(\epsilon_{s}^{\mathrm{e}}\right)^{2}\right) \exp \left(\frac{\epsilon_{v 0}-\epsilon_{v}^{\mathrm{e}}}{\xi}\right),
\end{equation}

\begin{equation}
q=\frac{\partial \psi}{\partial \epsilon_{s}^{\mathrm{e}}}=-3 c_{\mu} p_{0} \exp \left(\frac{\epsilon_{v 0}-\epsilon_{v}^{\mathrm{e}}}{\xi}\right) \epsilon_{s}^{\mathrm{e}}.
\end{equation}

The components of the symmetric stiffness Hessian matrix $\tensor{D}^e$ are derived by taking the second-order partial derivative of the energy functional with respect to the two strain invariants:

\begin{equation}
\begin{aligned}
D_{11}^{\mathrm{e}} &=\frac{\partial ^2 \psi }{\partial \epsilon_{v}^{\mathrm{e} \, 2}}=-\frac{p_{0}}{c_{r}}\left(1+\frac{3 c_{\mu}}{2 c_{r}}\left(\epsilon_{s}^{\mathrm{e}}\right)^{2}\right) \exp \left(\frac{\epsilon_{v 0}-\epsilon_{v}^{\mathrm{e}}}{c_{r}}\right), \\
D_{22}^{\mathrm{e}} &=\frac{\partial ^2 \psi }{\partial \epsilon_{s}^{\mathrm{e} \, 2}}=-3 c_{\mu} p_{0} \exp \left(\frac{\epsilon_{v 0}-\epsilon_{v}^{\mathrm{e}}}{c_{r}}\right), \\
D_{12}^{\mathrm{e}} &=D_{21}^{\mathrm{e}}=\frac{\partial ^2 \psi }{\partial \epsilon_{v}^{\mathrm{e}}\partial \epsilon_{s}^{\mathrm{e}}}=\frac{3 p_{0} c_{\mu} \epsilon_{s}^{\mathrm{e}}}{c_{r}} \exp \left(\frac{\epsilon_{v 0}-\epsilon_{v}^{\mathrm{e}}}{c_{r}}\right).
\end{aligned}
\label{eq:borja_hyperelastic_D}
\end{equation}

\section{Appendix: Data generation for polycrystal yield function}
\label{sec:dataset_yield}

The data set for the polycrystal yield function neural networks is generated by exploration of the stress space using the $\pi$-plane. Visualization of the data generation on the $\pi$-plane greatly facilitates the geometric interpretation of the yield function and the illustrates the amount of data necessary to span the stress space of the yield function. This allows for an effective planning of the data acquisition and insight on the required experimental set up. A demonstration of a yield surface data set generation is shown in Fig.~\ref{fig:data_generation_scheme}. Sample points are collected radially. The stress space is then partitioned by the Lode's angle where each angle is assigned a stress path that moves toward the radial direction on the $\pi-$plane. To accelerate the plasticity data acquisition, we can initially perform three experiments in the $\sigma_1$, $\sigma_2$, and $\sigma_3$ principal directions and, assuming the convexity of the yield surface, we can define an initial path-independent elastic region. The plasticity data generation is the focused outside the elastic region, effectively reducing the exploration space.

The yield function data sets are generated offline by fast Fourrier transform (FFT) method based simulations on polycrystal microstructures. The homogenized mesoscale polycrystal material responses are calculated on a 3d periodic domain by solving a Lippman-Schwinger equation using the FFT spectral method \citep{ma_fft_2019}. The underlying elasticity model of the polycrystals is linear elasticity with a Young's Modulus of $E = 2.0799 \text{MPa}$ and a Poisson ratio of $\nu = 0.3$. The material's plastic behavior was calculated using the ultimate algorithm for crystal plasticity \citep{borja1993discrete}. The model has 12 linearly independent slip systems with a yield stress of $100 \text{kPa}$ and a hardening modulus of $100 \text{kPa}$. An FFT elastoplastic simulation is performed radially for each of 140 different Lode's angles spanning the $\pi$-plane.

Each yield function stress data point that is generated by the FFT simulations is described by a radius $\rho$, an angle $\theta$, and an accumulated plastic strain $\bar{\epsilon}_p$. As stated in Section~\ref{sec:level_set}, based on the level-set re-initialization problem, every yield surface point is pre-processed to construct a level set. For every generated sample point $(\rho_o,\theta_o)$ on the $\pi$-plane, we construct 14 level set training points using a signed distance function, distributed uniformly on the radial direction with a distance range of $\pm \rho_o$ from point  $(\rho_o,\theta_o)$.

After generating the points of the level set, every point has a corresponding output value equal to the signed distance function $\phi(\rho_o,\theta_o, \bar{\epsilon}_{p,o})$ for that point. In this way, all the level set points on an isocontour will have the same output value. This proven to be an obstacle in the back-propagation during the neural network training -- many input combinations correspond to the same output value. To increase the variation of the output values of each sample during training, we introduce a helper transformation function $\zeta(\rho,\theta)$ of the output values in the data pre-processing step. Thus, during training, every level set input sample point $(\rho_o,\theta_o, \bar{\epsilon}_{p,o})$ is mapped to an output value:

\begin{equation}
\phi_\zeta (\rho_o,\theta_o, \bar{\epsilon}_{p,o}) = \phi (\rho_o,\theta_o, \bar{\epsilon}_{p,o}) + \zeta(\rho_o,\theta_o).
\label{eq:helper_function}
\end{equation}

During the prediction step, the true value of the level set can recovered by subtracting the know value of $\zeta(\rho_o,\theta_o)$ from the prediction output. The helper function in this work was chosen as $\zeta(\rho,\theta) = 2\bar{\rho}\cos(\theta/3)$, where $\bar{\rho}$ is the mean value of the radii in the yield function data set.

\section{Appendix: Verification exercise with custom hardening} 
\label{sec:custom_hardening}

The plasticity components of the neural network elastoplasticity framework can further be decomposed by separating the initial yield surface and its evolution -- the hardening law. We are introducing a method to apply custom hardening laws to the neural network approximated yield functions. The initial yield surface is controlled by a neural network of the form $\tilde{f}(\rho,\theta)$ with only the Lode's coordinates as inputs. The hardening is handled by a separate hardening law. In plasticity literature, hardening is usually implemented by transforming the yield surface -- changing the yield stress value. However, in the case of our neural network yield function approximation, the yield stress in not explicitly defined and cannot be immediately modified. To overcome this obstacle, we define the desired hardening laws to the neural network input instead of the assumed yield stress. Specifically, we define a hardening law as transformation $\tensor{L}$ of the original Lode's coordinates $\rho$ and $\theta$, such that:

 \begin{equation}
 \tensor{L}(\rho,\theta,\xi) = \langle L_\rho (\rho,\theta,\xi) , L_\theta (\rho,\theta,\xi) \rangle =\langle \rho_L , \theta_L \rangle, 
\label{eq:custom_hardening_definition}
\end{equation} 
where $L_r (\rho,\theta,\xi)$ and $L_\theta (\rho,\theta,\xi) $ are the parametric equations that transform $\rho$ and $\theta$ into the input variables $\rho_L$, $\theta_L$ respectively after hardening, and $\xi$ is an internal hardening variable. 

Common literature hardening laws can be translated into input transformations of this type and applied to the neural network yield functions through geometric interpretation. For example, in the simple case of isotropic hardening of the Von Mises plasticity model, hardening in the $\pi$-plane can be interpreted as the dilation of the circular yield surface -- i.e. increase of the radius $\rho_y$ where there is yielding. In the case of a neural network approximating the Von Mises yield function, the value of the current $r_y$ would not be readily available to modify. For that reason, instead of increasing the yield radius $\rho_y$, we opt for decreasing the input radius $\rho$ of the neural network an equivalent amount. The transformed radius $\bar{\rho}$ is defined as:

\begin{equation}
\bar{\rho} = \bar{L_\rho}(\rho,\bar{\epsilon}_p) = \rho - \sqrt{\frac{2}{3}} H \bar{\epsilon}_p,
\label{eq:isotropic_hardening_law}
\end{equation}
where $H$ is the material's identified hardening modulus and $ \bar{\epsilon}_p$ is the accumulated plastic strain. Any custom hardening model can be applied to with the right conversion to an input transformation. This enables for even more flexibility when assembling the theoretical components of the elastoplastic framework system. The hyperelastic energy functional, the initial yield surface, and the hardening law are independent of each other and can separately replaced. Furthermore, being able to assign a hardening law as a separate process in the data-driven yield function could prove valuable when only information of the initial yield surface is available in the data. This facilitates more flexibility in modeling the material by assigning hand-derived hardening laws to duplicate the elastoplastic response, using domain expertise.
 
\begin{figure}[h!]
\newcommand\siz{.43\textwidth}
\centering

\begin{tabular}{M{.45\textwidth}M{.45\textwidth}}
\includegraphics[width=.43\textwidth ,angle=0]{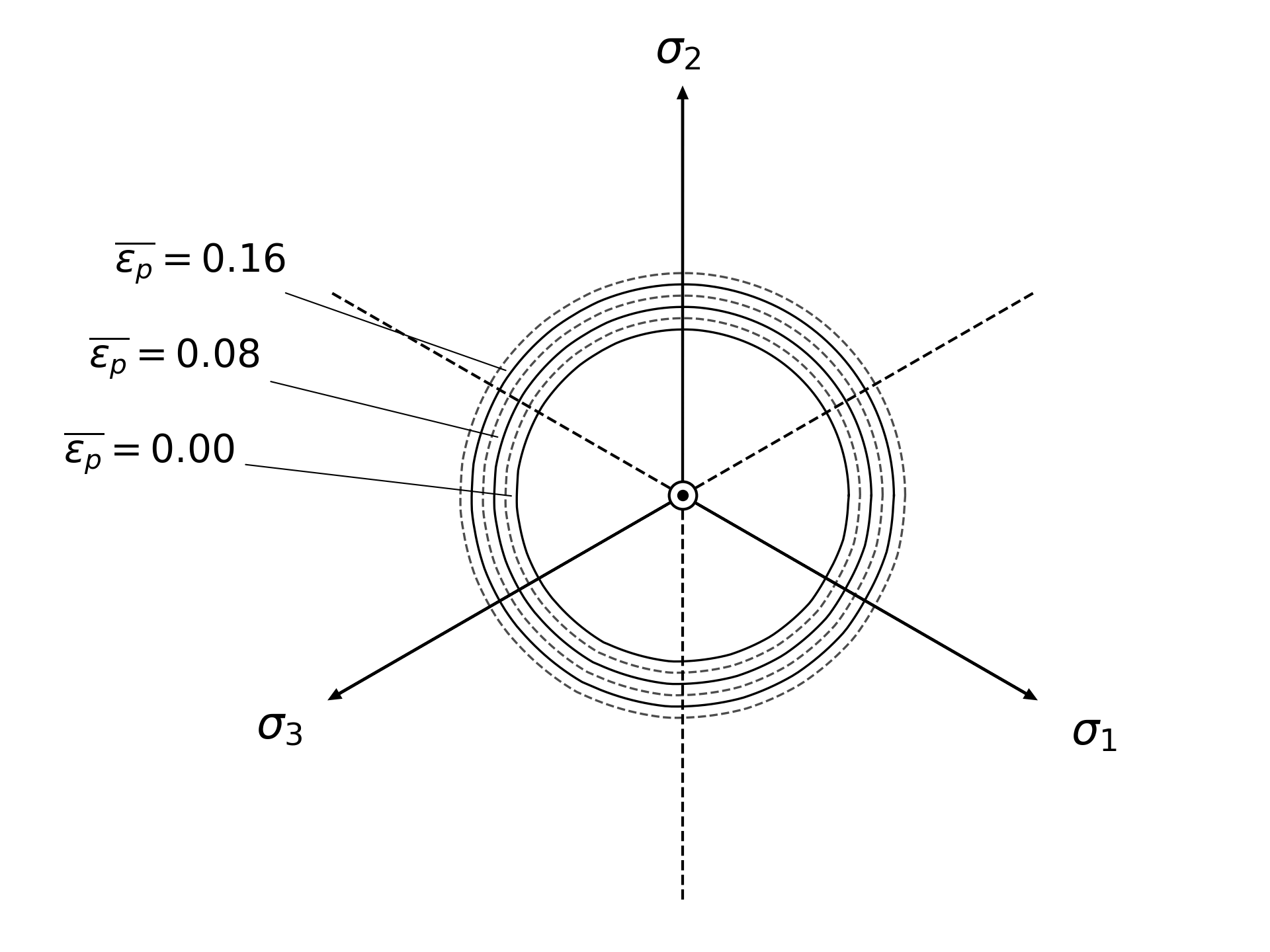} &
\includegraphics[width=.43\textwidth ,angle=0]{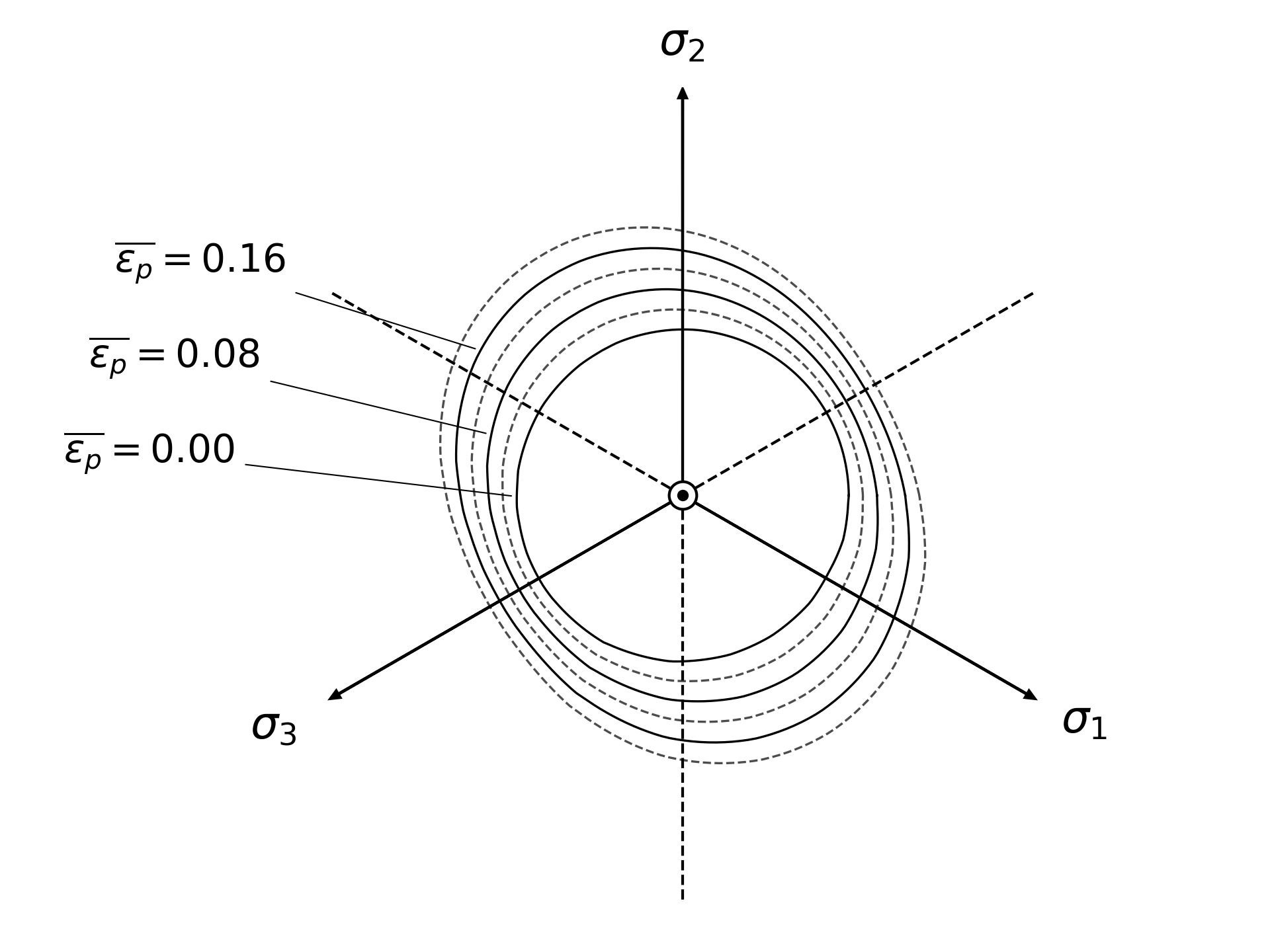} \\     

(a) $\bar{\rho}  = \rho - \sqrt{\frac{2}{3}} H \bar{\epsilon}_p$ & (b) $\bar{\rho} = \rho - \sqrt{\frac{2}{3}} H \bar{\epsilon}_p ( 1 + \cos^2(\theta - \frac{\pi}{ 6}) )$ \\

\includegraphics[width=.43\textwidth ,angle=0]{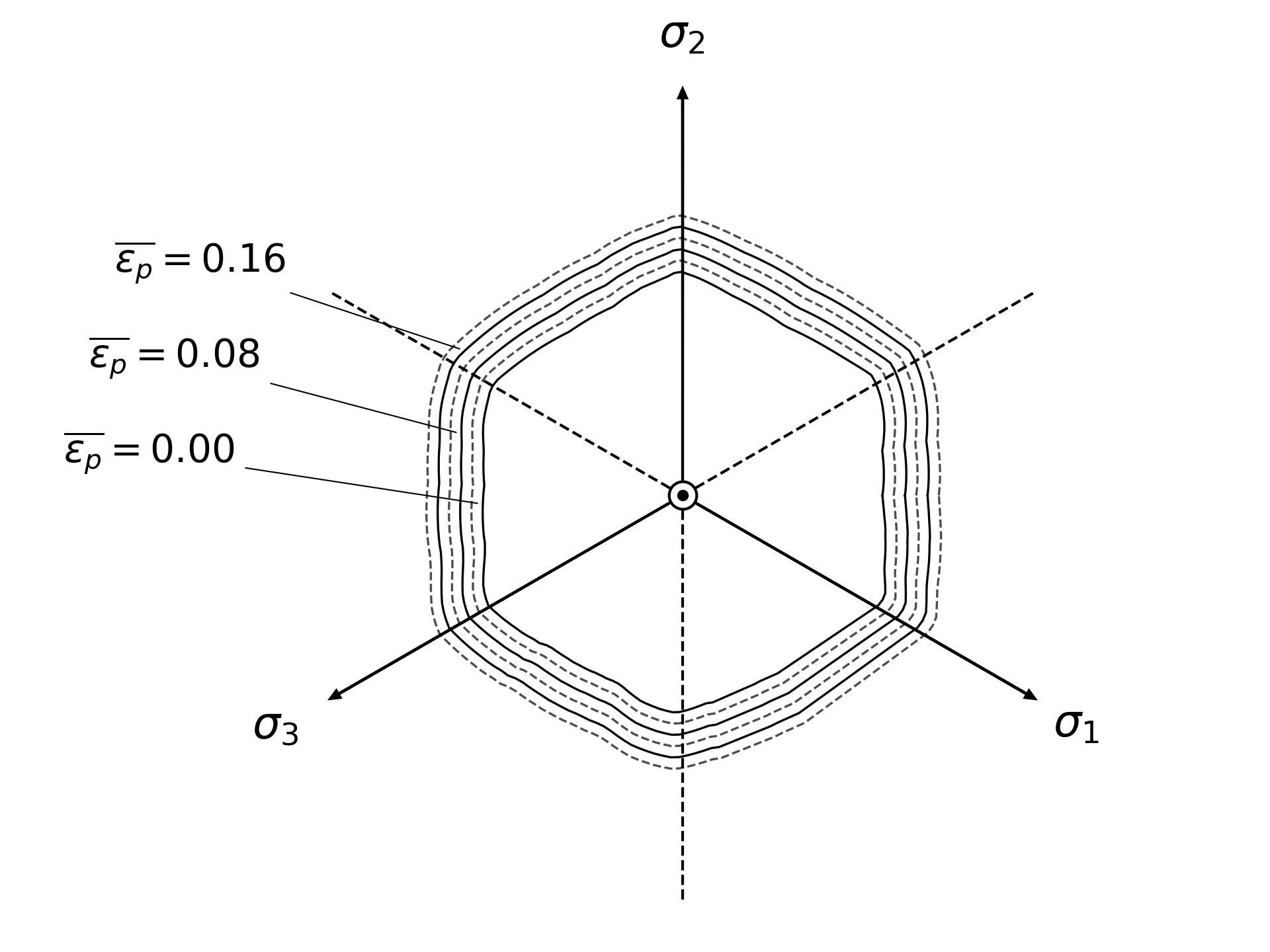} &
\includegraphics[width=.43\textwidth ,angle=0]{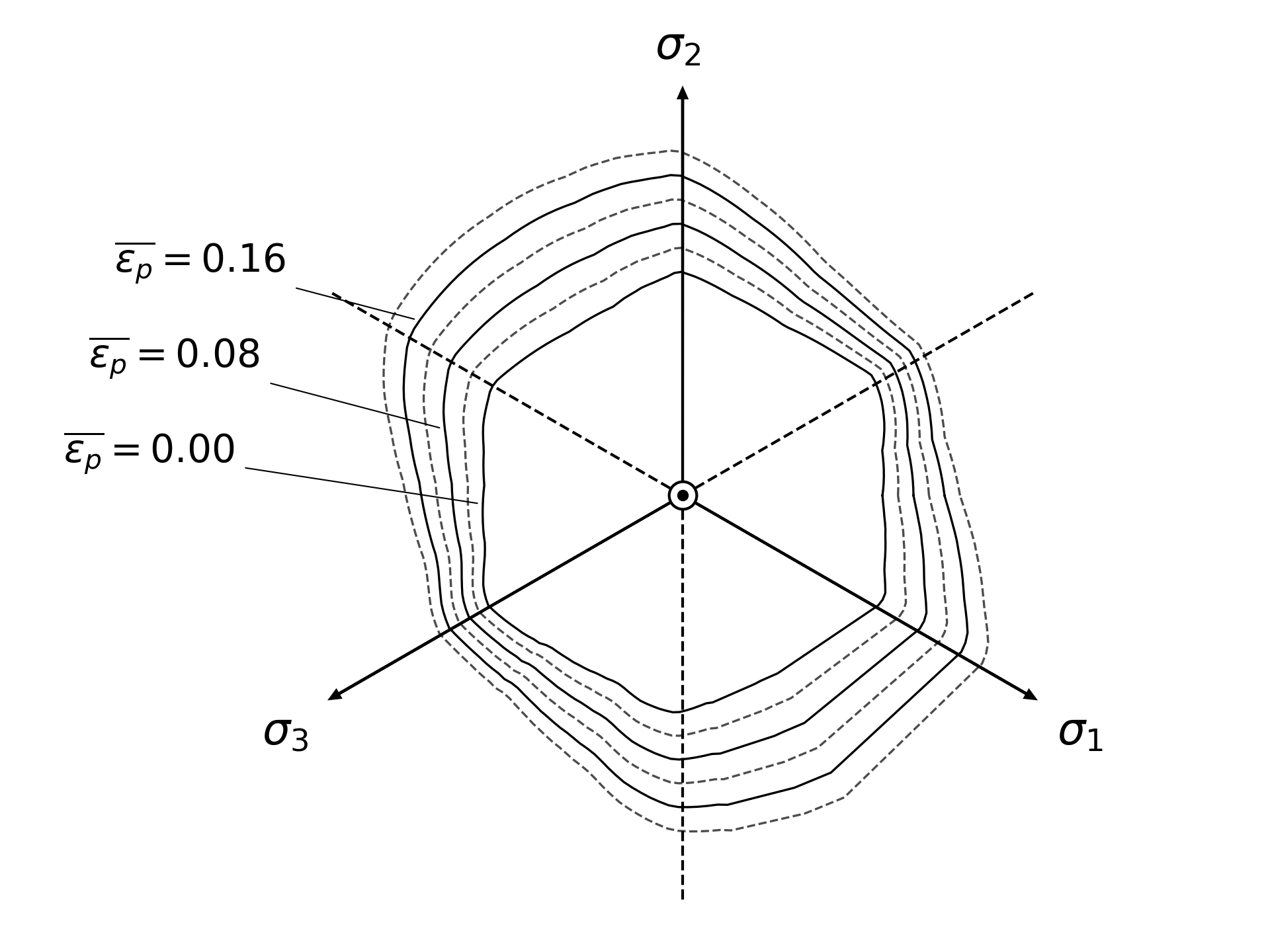} \\

(c) $\bar{\rho} = \rho - \sqrt{\frac{2}{3}} H \bar{\epsilon}_p$ & (d) $\bar{\rho} = \rho - \sqrt{\frac{2}{3}} H \bar{\epsilon}_p ( 1 + \cos^2(\theta - \frac{\pi}{ 6}) )$ 

\end{tabular}

\caption{Custom hardening transformations of initial neural network yield surfaces. The transformations are implemented by modifying the neural network input radius $\rho$. Transformations (a) and (c) emulate simple isotropic hardening (dilation of yield surface). Transformations (b) and (d) emulate a mixed mode hardening mechanism (dilation and change of shape).}
\label{fig:custom_hardening_modes}
\end{figure} 
 
 A few different cases of custom hardening transformations are demonstrated in Fig.~\ref{fig:custom_hardening_modes}. The initial yield surfaces are predicted from a neural network approximator -- all the points approximated have an accumulated plastic strain  $ \bar{\epsilon}_p = 0$. Fig.~\ref{fig:custom_hardening_modes} (a) and (c) showcase a simple isotropic hardening cases emulated by reducing the neural network input radius $\rho$ uniformly for all the Lode's angles $\theta$ on the $\pi$-plane. The hardening mechanism can be geometrically interpreted as a dilation of the initial yield surface. Fig.~\ref{fig:custom_hardening_modes} (b) and (d) showcase two modes of hardening acting simultaneously -- a dilation and an elongation towards a preferred direction of the initial yield surface. The elastoplastic framework implemented in this work allows for the integration of any isotropic hardening mechanism that transforms the size and shape of the initial yield surface.

In the current formulation, the neural network elastoplastic framework can consist of any isotropic hyperelastic energy functional and isotropic yield function. To demonstrate the framework's capability to capture non-linear behaviors, we have implemented a fictitious highly non-linear and a fictitious non-linear custom hardening law. The energy functional neural network is trained on data set based a modification on the linear elastic energy functional with the shear part replaced with a highly non-linear term:

\begin{equation}
\breve{\psi}(\epsilon_v^{\mathrm{e}},\epsilon_s^{\mathrm{e}}  ) = \frac{1}{2} K \epsilon_v^{\mathrm{e} \, 2} + \frac{3}{2} G \epsilon_s^{\mathrm{e} \, 4}.
\label{eq:fictitious_energy_functional}
\end{equation}

The non-linear hardening law is implemented by applying a transformation on the Lode's radius input of the Von Mises yield function neural network. The hardening law $\breve{L}$ provides a transformed radius:

\begin{equation}
\breve{\rho}=\breve{\rho}(\rho,\bar{\epsilon}_p) = \rho \cdot (1 - \bar{\epsilon}_p^2)^6.
\label{eq:fictitious_hardening_law}
\end{equation}

The prediction of the framework is demonstrated in Fig.~\ref{fig:non_lin_nets}. The framework provides great flexibility to decompose the material behavior for the elasticity, yield surface and hardening law -- all of which can be individually replace. This also allows for a combination of data-driven and handcrafted laws that can be tuned to closely replicate observed material behaviors.
 
\begin{figure}[h!]
\newcommand\siz{.32\textwidth}
\centering
\begin{tabular}{M{.33\textwidth}M{.33\textwidth}M{.33\textwidth}}
\hspace{-2cm}\includegraphics[width=.32\textwidth ,angle=0]{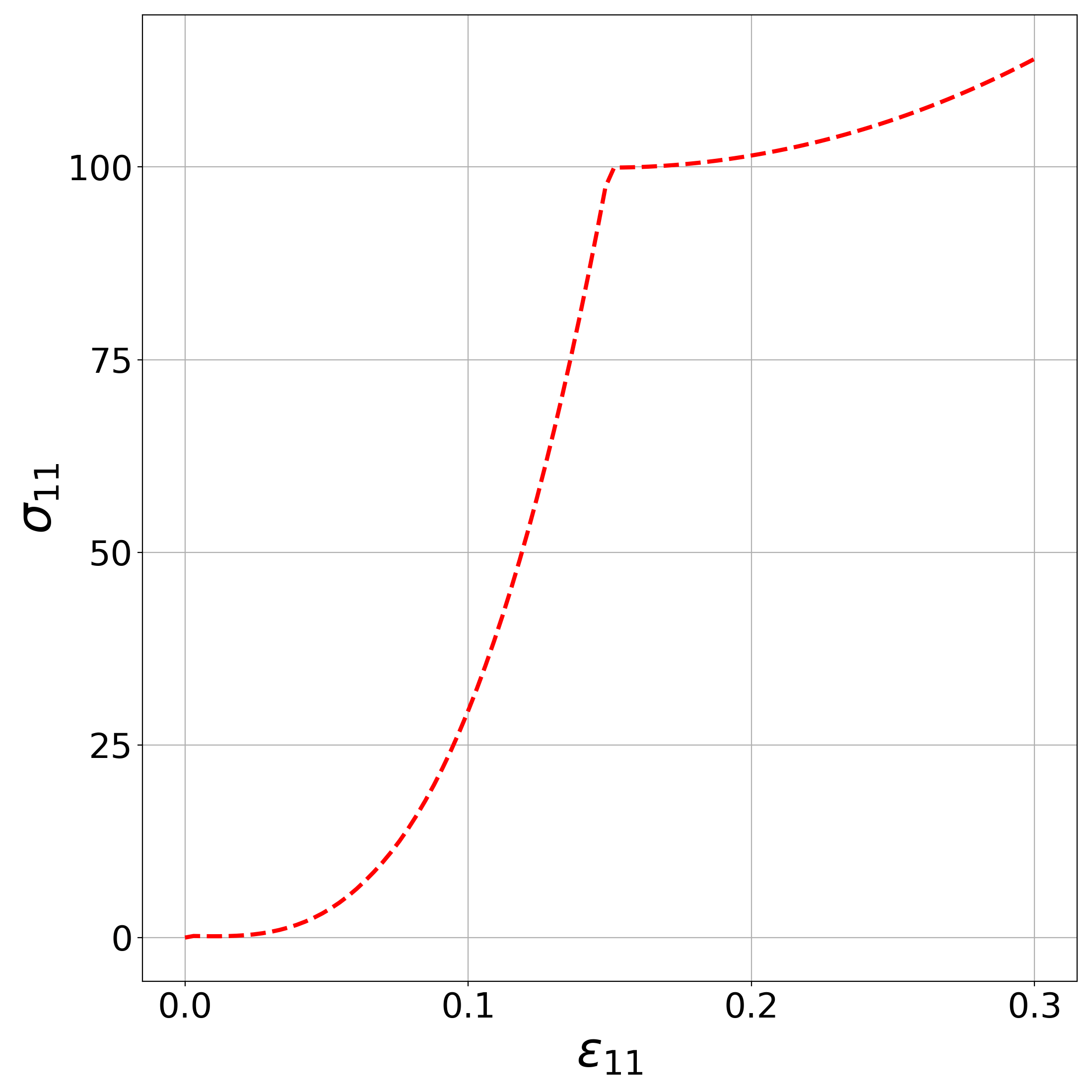} &
\hspace{-2cm}\includegraphics[width=.32\textwidth ,angle=0]{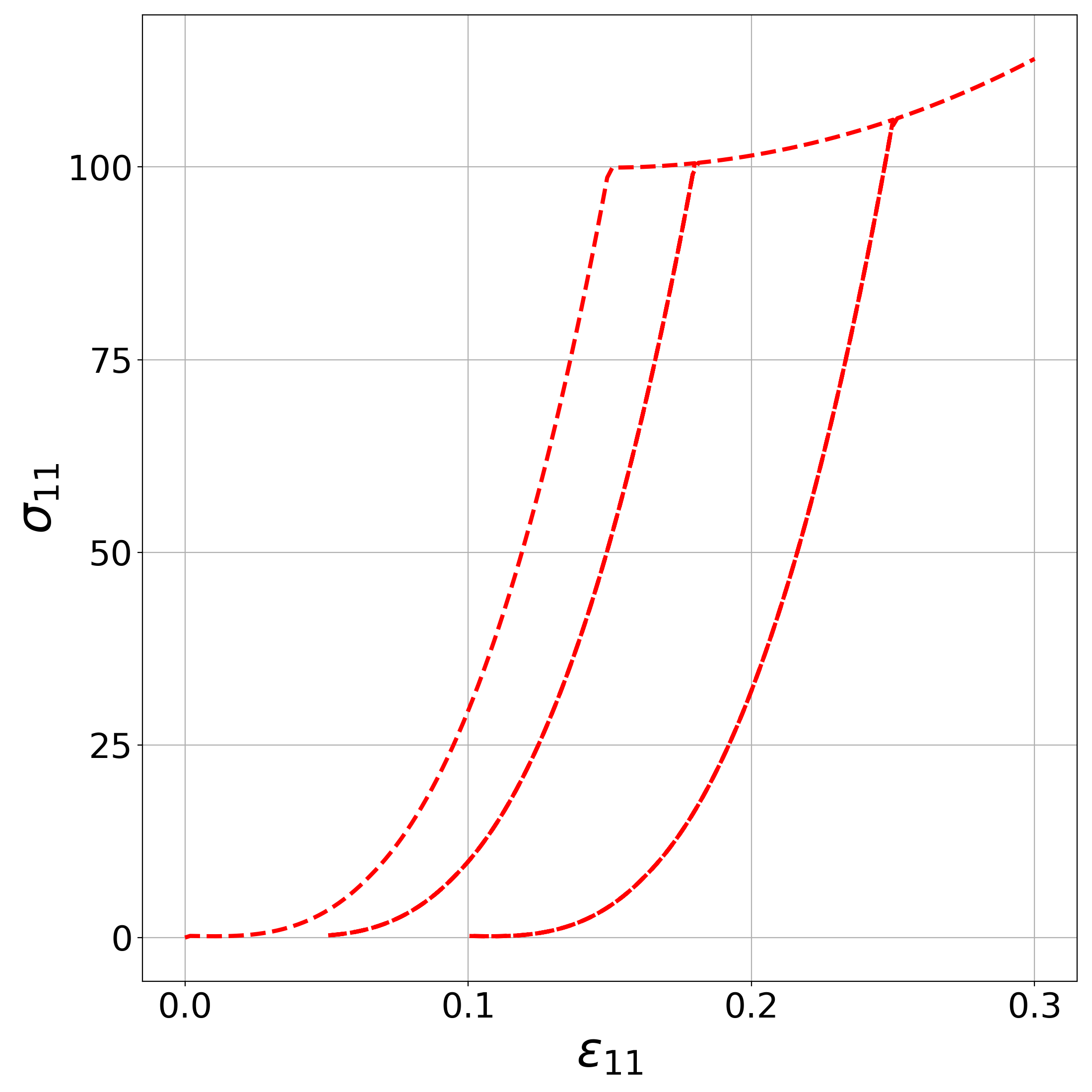} &
\hspace{-2cm}\includegraphics[width=.32\textwidth ,angle=0]{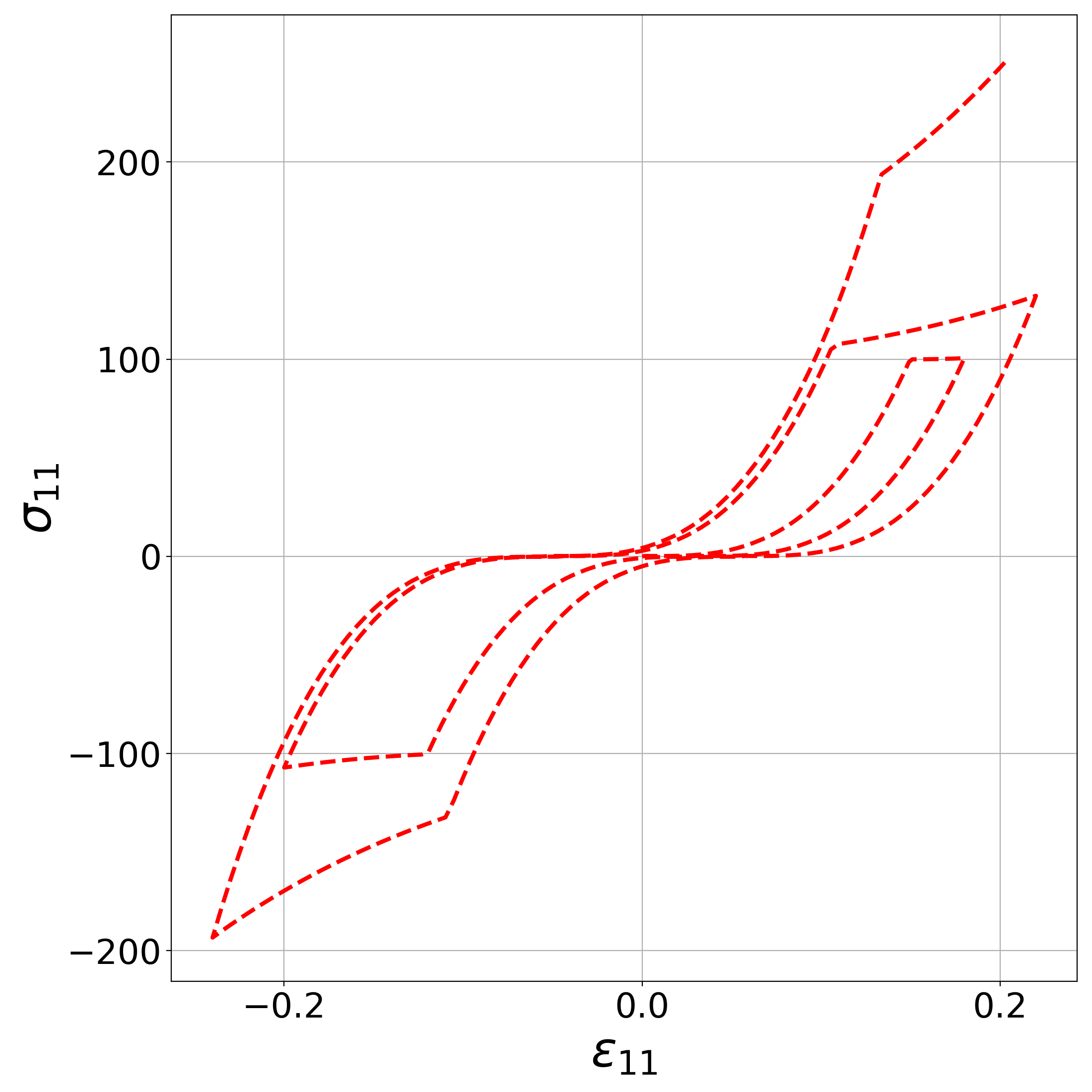} \\     

\end{tabular}

\caption{The ANN elastoplastic framework can handle highly non-linear hyperelastic energy functionals and custom hardening laws. Three loading paths are demonstrated for a fictitious non-linear energy functional and hardening law. The initial yield surface is predicted by the Von Mises yield surface neural network.}
\label{fig:non_lin_nets}
\end{figure}

\section{Appendix: Verification exercise on learning classical J2 plasticity with isotropic hardening}
\label{sec:verification_j2}
As a part of the verification exercise, we also test whether the proposed framework is able to deduce an elasto-plasticity model with
linear elasticity and Von Mises plasticity with isotropic hardening solely by learning from limited data. The elastoplastic ANN framework consists of a neural network approximating the linear elastic energy functional and a yield function neural network that approximates a Von Mises yield surface. The hardening law of the system is implemented in two different ways to demonstrate the flexibility of the framework. The hardening law can be directly enforced in the yield function neural network by utilizing the accumulated plastic strain as an input of the architecture and allowing to define the evolution of the yield surface. The hardening law can also be separately defined in the framework. In this case, the yield surface can evolve following a hardening mode as described in Section~\ref{sec:custom_hardening}, identified to match the material's plastic behavior. To emulate isotropic hardening, a transformation law $\bar{L}$ is applied on the Lode's radius input of the yield function neural network, following Eq.~\eqref{eq:isotropic_hardening_law}. The material has a Young's Modulus of $E = 2.0799 \text{MPa}$, a Poisson ratio of $\nu = 0.3$, an initial yield stress of $100 \text{kPa}$, and a hardening modulus of $H = 0.1 E$.

The comparison of the neural network elastoplastic framework with three benchmark simulations is shown in Fig.~\ref{fig:j2_benchmarks}. The framework is tested against a monolithic loading path, a loading path with multiple unloading patterns, and a cyclic loading path. It is noted that both of the networks of the framework are feed-forward, do not retain any loading history information and the data sets they were trained on do not provide any loading and unloading strain history information. However, by integrating the two network predictions through the return mapping algorithm described in Section~\ref{sec:return_mapping_algorithm}, the framework can adequately capture loading and unloading patterns it has not been explicitly trained on.

\begin{figure}[h!]
\newcommand\siz{.32\textwidth}
\centering
\begin{tabular}{M{.01\textwidth}M{.33\textwidth}M{.33\textwidth}M{.33\textwidth}}
\hspace{-2.1cm}(a) &
\hspace{-2.5cm}\includegraphics[width=.32\textwidth ,angle=0]{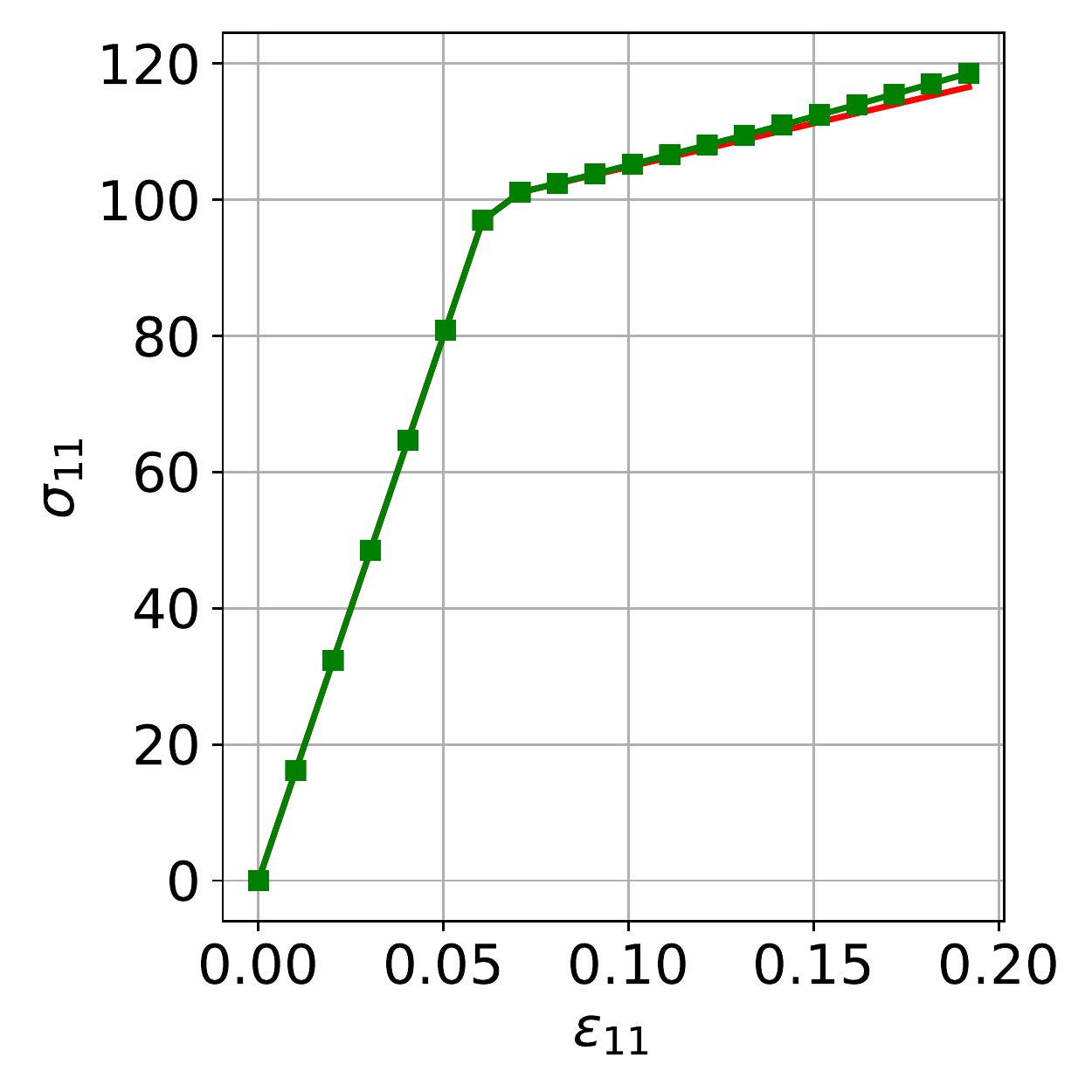} &
\hspace{-2.5cm}\includegraphics[width=.32\textwidth ,angle=0]{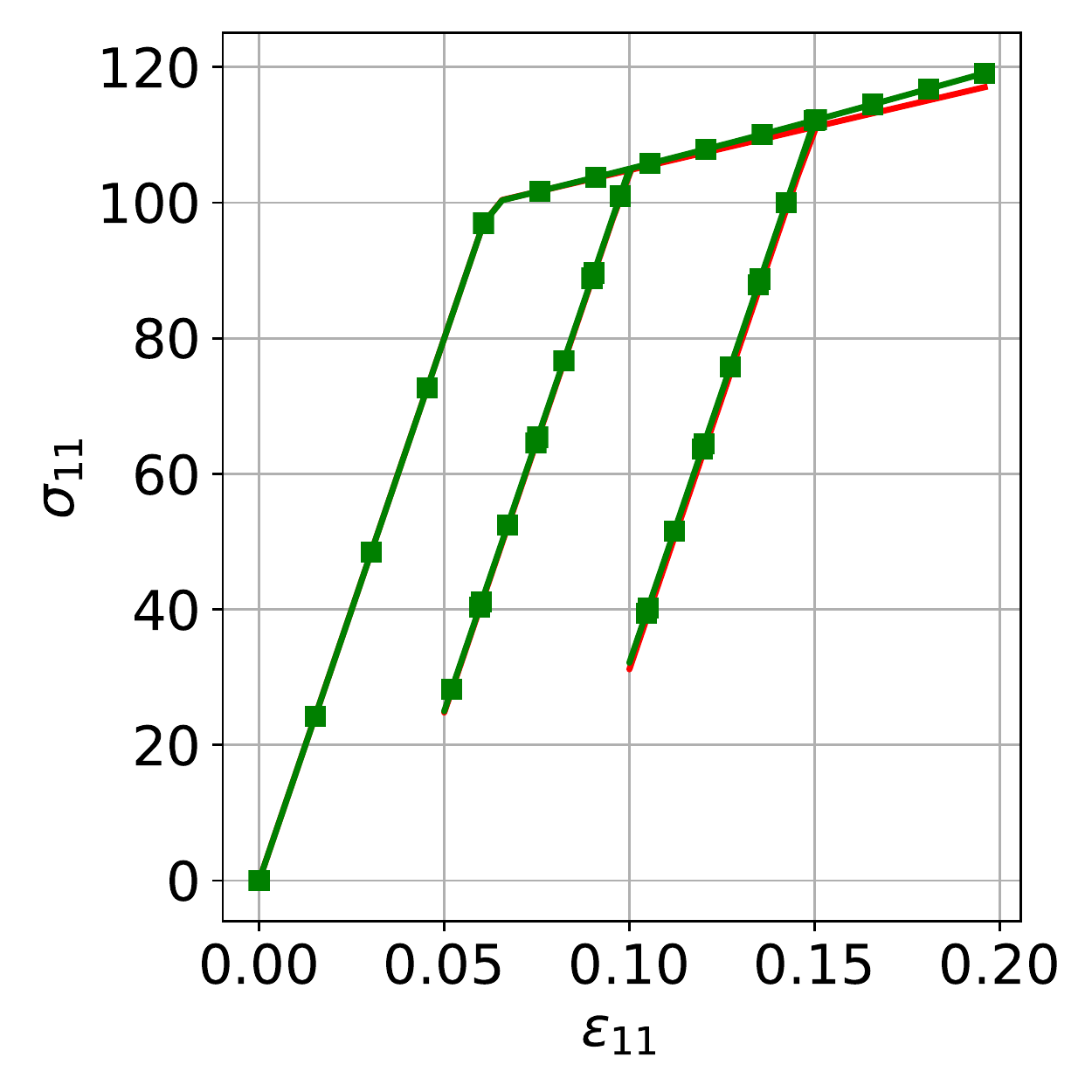} &
\hspace{-2.5cm}\includegraphics[width=.32\textwidth ,angle=0]{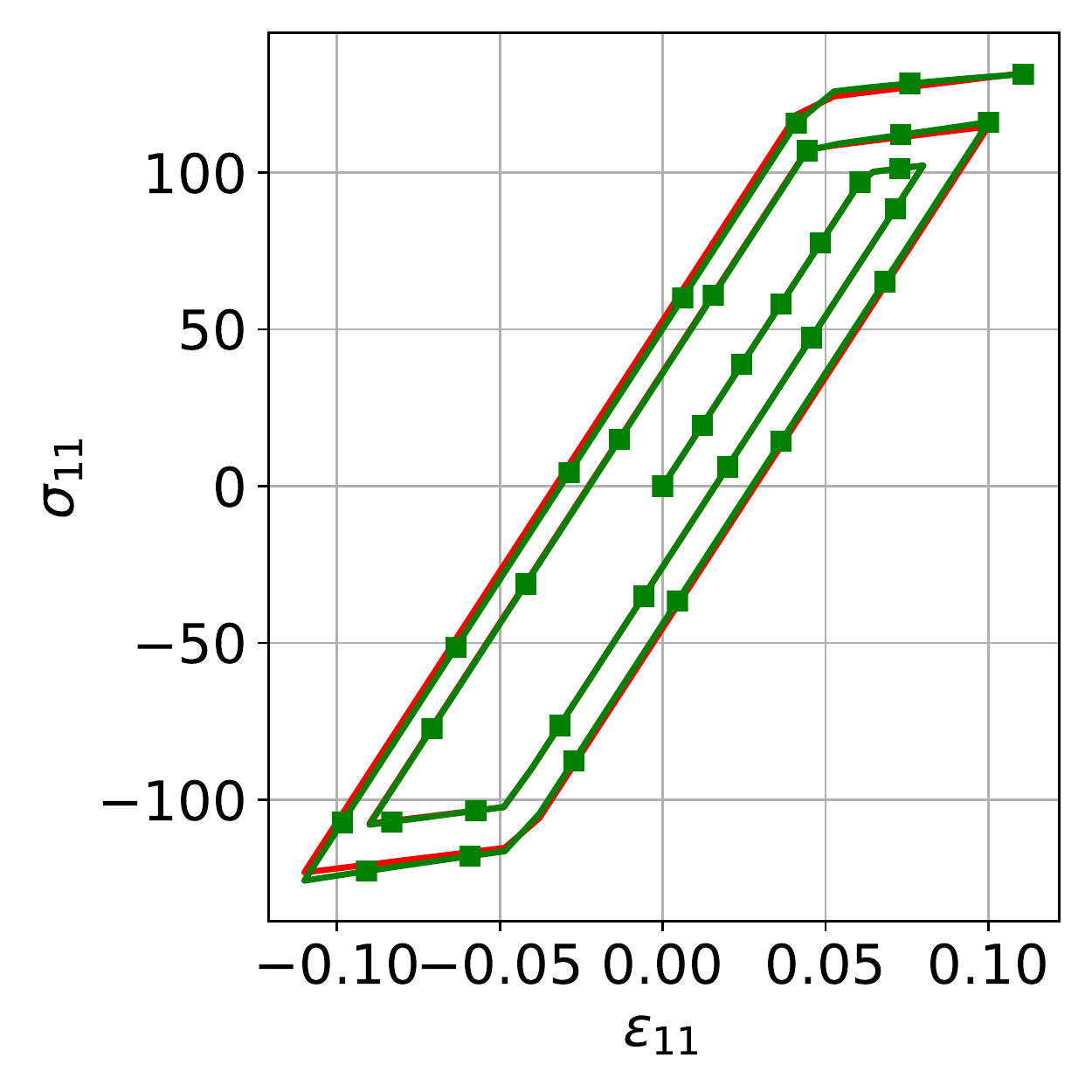} \\     

\hspace{-2.1cm}(b) &
\hspace{-2.5cm}\includegraphics[width=.32\textwidth ,angle=0]{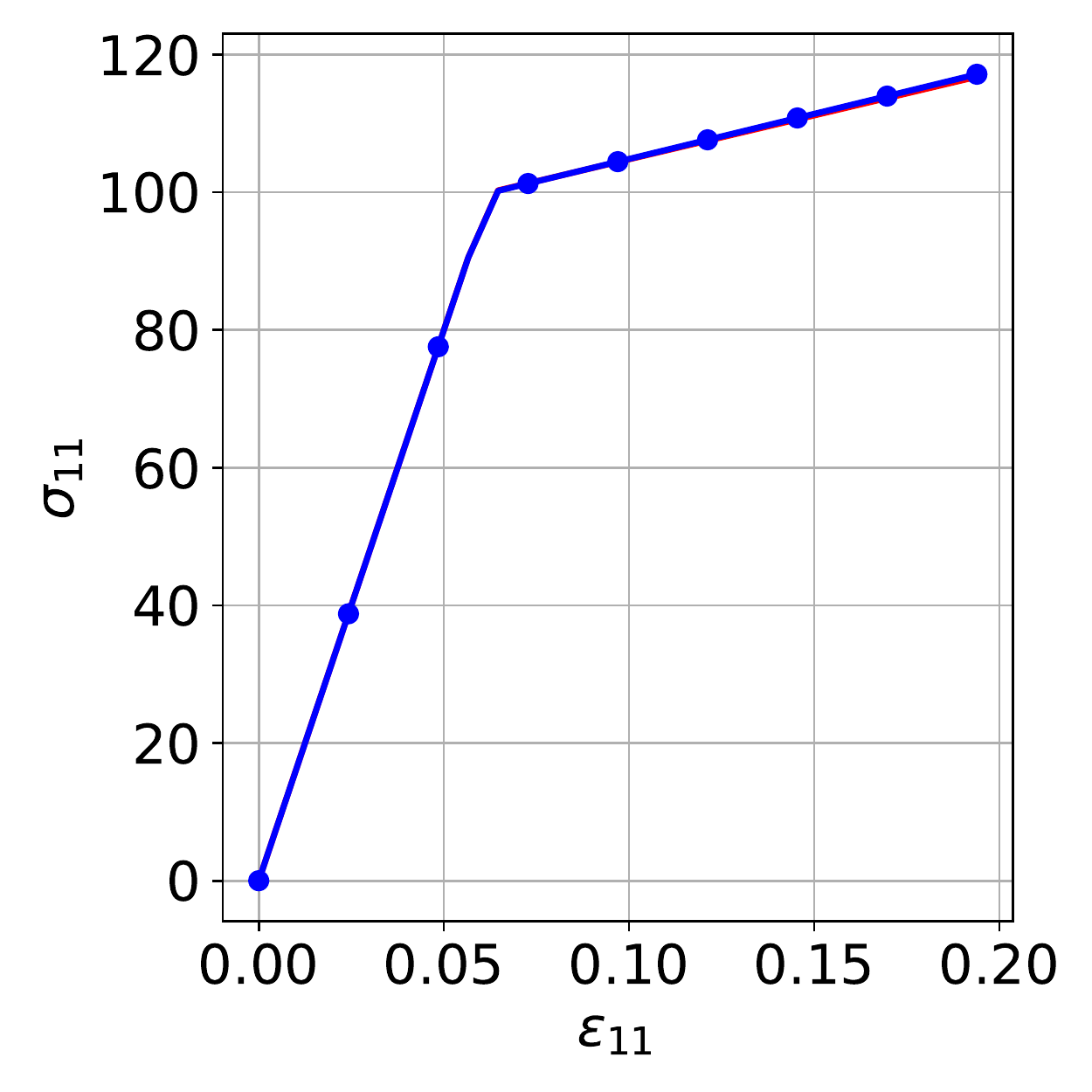} &
\hspace{-2.5cm}\includegraphics[width=.32\textwidth ,angle=0]{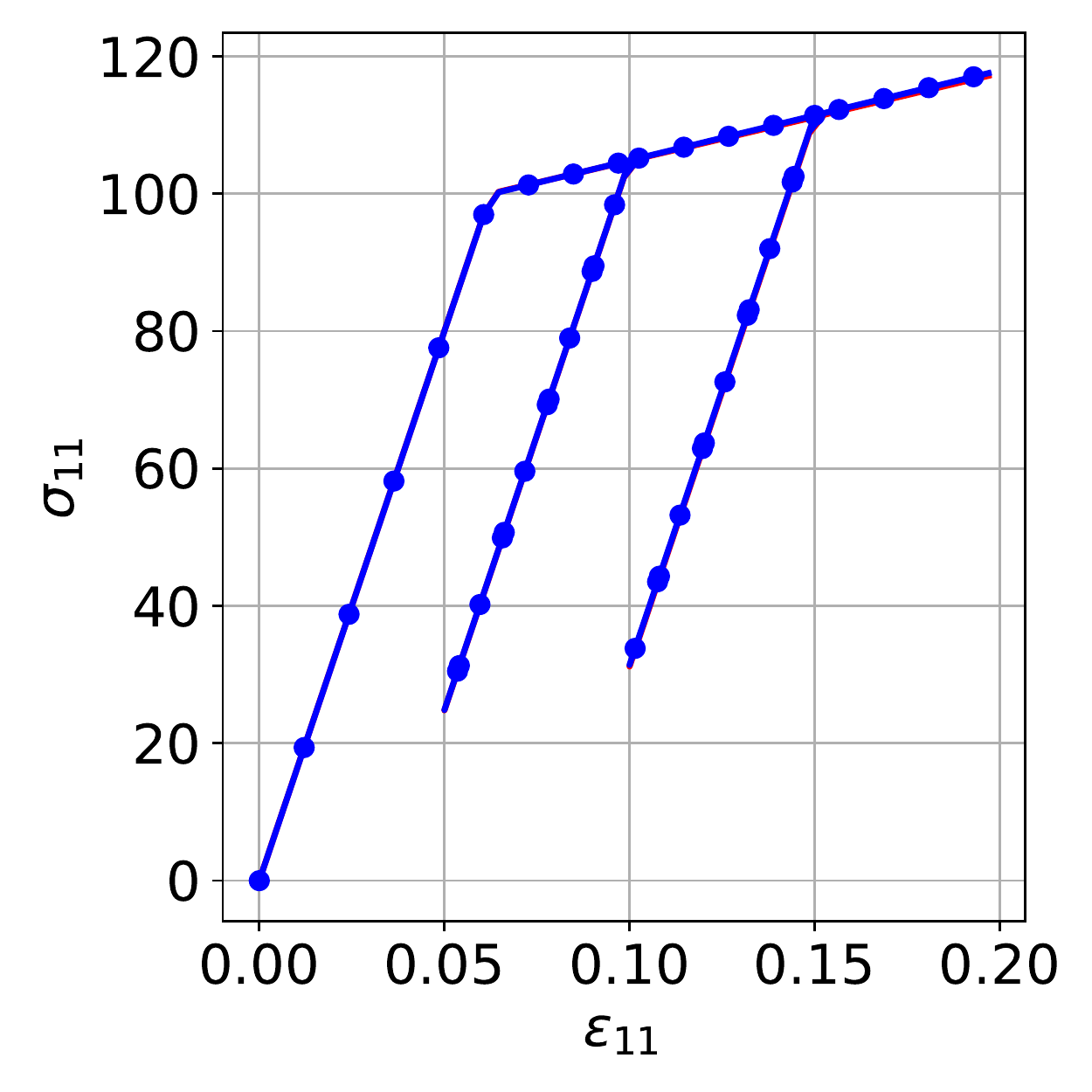} &
\hspace{-2.5cm}\includegraphics[width=.32\textwidth ,angle=0]{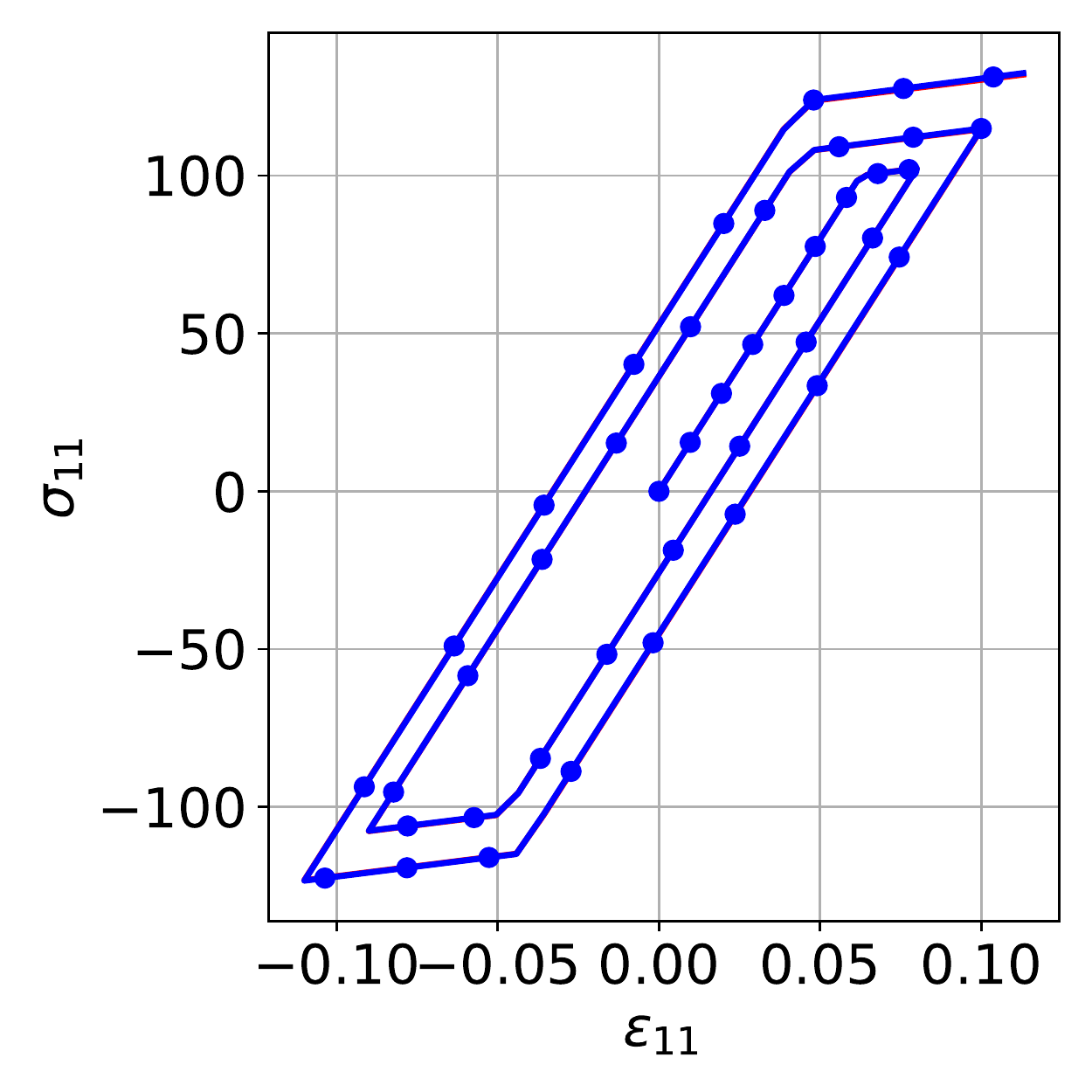} \\     

\end{tabular}
\includegraphics[width=.7\textwidth]{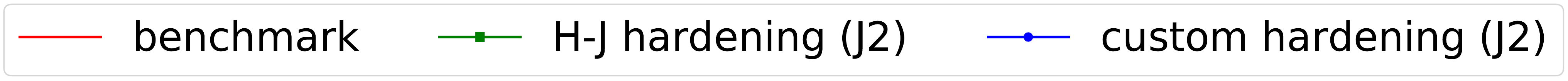} 

\caption{Comparison of the neural network elastoplastic framework (linear elasticity and Von Mises plasticity) with benchmark simulation data. (a) The yield function NN replaces the yield function and the hardening law. (b) The yield function NN predicts only the initial yield surface and a custom identified hardening law is applied.}
\label{fig:j2_benchmarks}
\end{figure}

}

\end{appendix}

\section{Acknowledgments}
The authors are supported by 
by the NSF CAREER grant from Mechanics of Materials and Structures program
at National Science Foundation under grant contracts CMMI-1846875 and OAC-1940203, 
the Dynamic Materials and Interactions Program from the Air Force Office of Scientific 
Research under grant contracts FA9550-17-1-0169 and FA9550-19-1-0318.
These supports are gratefully acknowledged. 
The views and conclusions contained in this document are those of the authors, 
and should not be interpreted as representing the official policies, either expressed or implied, 
of the sponsors, including the Army Research Laboratory or the U.S. Government. 
The U.S. Government is authorized to reproduce and distribute reprints for 
Government purposes notwithstanding any copyright notation herein.

\bibliographystyle{plainnat}
\bibliography{main}

\end{document}